\documentclass[twoside,11pt]{article}

\usepackage[abbrvbib, preprint]{jmlr2e}

\usepackage{blindtext}
\usepackage{float}
\usepackage{wrapfig}
\usepackage{enumitem}
\usepackage{graphicx}
\usepackage{subfigure}
\usepackage{multirow}
\usepackage{subcaption}
\usepackage{pifont}
\usepackage{amsmath}
\usepackage{xcolor}
\newcommand{\cmark}{\ding{51}}%
\newcommand{\xmark}{\ding{55}}%

\usepackage{lastpage}
\jmlrheading{27}{2026}{1-\pageref{LastPage}}{03/16; Revised TBD/TBD}{TBD/TBD}{21-0000}{Israel Mason-Williams, Gabryel Mason-Williams and Helen Yannakoudakis}

\ShortHeadings{Mason-Williams, Mason-Williams and Yannakoudakis}{ Understanding Knowledge Distillation}
\firstpageno{1}

\begin{document}

\title{A Functional Perspective on Knowledge Distillation in Neural Networks}

\author{\name Israel Mason-Williams \email israel.mason-williams@kcl.ac.uk \\
       \addr Department of Informatics\\
       UKRI Safe and Trusted AI\\
       London, United Kingdom
       \AND
       \name Gabryel Mason-Williams \email  g.t.mason-williams@qmul.ac.uk \\
       \addr Department of Informatics\\
       Queen Mary University of London\\
       London, United Kingdom
        \AND
       \name Helen Yannakoudakis \email  helen.yannakoudakis@kcl.ac.uk \\
       \addr Department of Informatics\\
       King's College London \\
       London, United Kingdom}

\editor{My editor}

\maketitle

\begin{abstract}
Knowledge distillation is considered a compression mechanism when judged on the resulting student's accuracy and loss, yet its functional impact is poorly understood. We quantify the compression capacity of knowledge distillation and the resulting knowledge transfer from a functional perspective, decoupling compression from architectural reduction to provide an improved understanding of knowledge distillation. We employ a control-driven experimental protocol with hypothesis testing and random control distillation to isolate and understand knowledge transfer mechanisms across data modalities. To test the breadth and limits of our analyses, we study self-distillation, standard distillation, feature-map matching variants, distillation scaling laws across model sizes, and the impact of temperature on knowledge transfer. We find statistically supported knowledge transfer in some modalities and architectures; however, the extent of this transfer is less pronounced than anticipated, even under conditions that maximise knowledge sharing. Notably, in cases of significant functional transfer, we identify a consistent and severe asymmetric transfer of negative knowledge to the student, raising safety concerns for knowledge distillation. Across 22 experimental setups, 9 architectures, and 7 datasets, our results suggest that knowledge distillation functions less as a robust compression-by-transfer mechanism and more as a data-dependent regulariser whose transfer component is biased towards negative asymmetric transfer. 
\end{abstract}

\begin{keywords}
Knowledge Distillation, Functional Similarity, Asymmetric Transfer, Distillation Scaling Laws, Adversarial Transfer
\end{keywords}

\section{Introduction}

Large neural networks have achieved remarkable results across domains \citep{gpt3, dosovitskiy2020image, kirillov2023segment}, but at significant computational cost. This has motivated techniques that reduce model size while maintaining performance. Knowledge distillation (KD) has emerged as a widely adopted method to compress models by training a student model to mimic a larger teacher~\citep{model_compression, hinton2015distilling,gu2024minillm,muralidharan2024compact}. While KD can be applied across architectures and modalities, including in self-distillation regimes where the teacher and student share the same architecture  \citep{allen2020towards, zhang2019your}, the mechanism by which KD improves student performance remains contested~\citep{busbridge2025distillation}. Recent studies have challenged the assumption that KD works through meaningful knowledge transfer, showing that performance gains have been observed even with randomly initialised teachers~\citep{does_kd_really_work}. This motivates a rigorous examination of KD's functional impact. 

In this work, we move beyond the question of whether knowledge is transferred: we re-examine the framing of knowledge distillation as a robust mechanism of knowledge transfer. Our contribution is not only to examine the regularisation view of KD, but to make it testable: we introduce a functional, control-driven evaluation protocol that measures teacher-student functional similarity relative to counterfactual baselines and decomposes transferred agreement into teacher-correct and teacher-incorrect regions. This lets us distinguish apparent KD gains that are consistent with data-dependent regularisation from gains accompanied by measurable teacher-function transfer. 
\textit{Throughout the paper, `knowledge transfer' refers to our operational notion of teacher-function transfer, measured relative to counterfactual controls.}

Our study is grounded in two research questions: 1) Does knowledge distillation produce statistically supported teacher-student functional similarity, relative to control conditions, across architectures and data modalities? 2) When functional transfer occurs, is the resulting teacher-student alignment symmetric across teacher-correct and teacher-incorrect behaviour, or is it systematically asymmetric? Across our experiments, we find that improvements under KD do not necessarily arise from meaningful transfer of the teacher's knowledge, but are often consistent with a substantial data-dependent regularisation effect~\citep{does_kd_really_work,yun2020regularizing,ge2021self,yuan2020revisiting}.Crucially, when functional transfer is statistically supported, we identify a systematic \textit{negative asymmetric payoff}: KD disproportionately increases agreement on the teacher's incorrect predictions relative to agreement on its correct predictions.

We first focus on self-distillation, where the student has the capacity to match the teacher's functional representation perfectly, ensuring that any observed differences are solely due to the distillation signal. We then verify our findings in the standard distillation setting with smaller student models (Section~\ref{sec:standard-kd-ss}), as well as with different KD variants   (Section~\ref{sec:feature_map_matching}). 
Our methodological framework isolates the core mechanics of knowledge distillation through: 1) a controlled training setup where all models share initialisations and data orders, enabling precise functional comparison; 2) two counterfactual controls: independent models with the same initialisation seed and different data order (SIDDO) as the teacher (using the same architecture in the self-distillation case), and a Random Control Distillation (RCD) where the distillation target is replaced with a class-uniform distribution,
all functionally compared to the teacher model used in the standard distillation process; 3) a broad array of functional similarity metrics including Activation Distance, Rank Disagreement, Prediction Disagreement, JS Divergence and Prediction Agreement to quantify teacher-student functional alignment and functional transfer between each control and the teacher. We conduct experiments across 7 datasets, 3 data modalities (image, audio, and language), and 9 architectures, training over 4,000 models. Our findings show that: 
\begin{itemize}
\itemsep0pt
  \item The strongest improvements in accuracy and loss frequently arise under Random Control Distillation, challenging the assumption that KD performance gains necessarily reflect successful knowledge transfer, and supporting a data-dependent regularisation perspective of KD.
  \item KD can lead to statistically supported functional similarity between teacher and student, but this similarity is often marginal and inconsistent across datasets and modalities. Furthermore, increasing the distillation temperature significantly reduces the degree of functional transfer.
  \item When functional transfer is supported and not marginal, the resulting teacher-student alignment is often asymmetric, with disproportionate agreement on the teacher's incorrect predictions. 
  This asymmetry becomes more pronounced as dependence on the teacher signal increases. 
\end{itemize}
In the regimes we study, our findings motivate a re-characterisation of KD, not as a uniformly robust knowledge transfer mechanism, but as a data-dependent regulariser with inconsistent and negative asymmetric knowledge-sharing capacity. 
This perspective raises important, practical safety concerns: when functional transfer is significant, KD may amplify incorrect or harmful behaviour encoded in the teacher. We further support this claim via a concrete case of adversarial transfer facilitated by KD. Concretely, our contributions are as follows:
\begin{itemize}
\itemsep0pt
  \item We provide a control-driven agnostic experimental framework that isolates the contribution of the teacher signal and evaluates KD gains against counterfactual baselines rather than performance metrics alone, something that prior work has not quantitatively disentangled to this level. 
  \item We identify and characterise a novel phenomenon across conditions, modalities and architectures: when statistically supported functional transfer occurs, KD disproportionately favours the teacher's incorrect predictions, revealing a systematic error amplification effect with safety implications.
  \item We demonstrate the diagnostic utility of RCD as a crucial counterfactual, showing it frequently outperforms KD, concretely weakening the assumption that KD gains primarily arise from knowledge transfer. 
  \item To our knowledge, we conduct the largest functional study of KD to date spanning over 4,000 trained models across 9 architectures, 7 datasets, and 3 modalities (vision, audio, and language), supporting the generality of our empirical observations and reproducibility of our evaluation framework.
  \item We show targeted and scalable negative transfer via adversarial and capacity-scaling experiments, demonstrating that KD can reliably copy specific erroneous behaviours, and that negative asymmetric transfer can increase with student capacity -- underscoring the safety implications of KD applications in high-stakes settings.  
\end{itemize} 
Our study contributes a timely and nuanced perspective on knowledge distillation by turning competing explanations of KD into an empirically testable question, helping to adjudicate between data-dependent regularisation and reliable knowledge-transfer accounts, while uncovering a systematic error-asymmetric transfer dynamic with important safety implications across modalities. 
\section{Related Work}

\paragraph{Knowledge Distillation (KD):}

KD transfers behaviour from a teacher (or ensemble) to a student \citep{model_compression,hinton2015distilling}, with strong empirical results across modalities \citep{beyer2022knowledge,jung2020knowledge,sanh2019distilbert,aghli2021combining,li2020few,fang2021compressing,wang2022efficient} and architectures \citep{touvron2021training,miles2024vkd}. However, 
the mechanism behind KD improvements remains debated \citep{mason-williams2024neural,does_kd_really_work,ojha2023knowledge,pmlr-v139-menon21a}. A line of work interprets KD as a form of data-dependent regularisation \citep{yun2020regularizing,ge2021self,yuan2020revisiting}, arguing that limited or no knowledge is transferred under its optimisation objective. Other literature, on the other hand, strongly argues that KD can effectively transfer meaningful  knowledge from teacher to student~\citep{shen2021label,sultan2023knowledge}. In this paper, we advance the discussion of KD as a regulariser by introducing a functional perspective that spans vision, audio, and language. There remains substantial contention in the community about how knowledge distillation fundamentally operates~\citep{busbridge2025distillation}, with misaligned interpretations across data modalities. This motivates our multi-modal, multi-architecture, and multi-dataset study of KD. We present a control-driven functional protocol that decouples compression effects from architectural size reduction, and measures teacher-student alignment for statistical significance beyond only analysing accuracy and loss. Using this framework, we provide evidence that KD often acts as a data-dependent regulariser, while also exposing a new dimension of this regularisation: when substantial and statistically supported knowledge transfer occurs, there is a strong negative asymmetric transfer to the student.

\paragraph{Functional Similarity Metrics:}

Functional similarity methods compare models by their outputs rather than their accuracy alone \citep{klabunde2023similarity}. Such analyses have been used in unlearning \citep{golatkar2021mixed,chundawat2023can}, understanding ensemble dynamics \citep{fort2019deep}, and compression/pruning \citep{mason2024makes,mason-williams2024neural}.
Metrics such as Activation Distance, Prediction Dissimilarity and JS Divergence have been used for functional analysis. Activation Distance represents the $\mathcal{L}_{2}$ distance between the softmax output distribution of two models, enabling output-space functional comparison. JS Divergence represents the Jensen-Shannon information-theoretic divergence that gives a directed measure of divergence between non-continuous probability distributions by employing a weighted average of the KL divergences from each distribution ~\citep{lin1991divergence}. 
Prediction Disagreement measures the proportion of inputs for which two models produce different top-1 predictions, providing a complementary view of functional alignment at the label level \citep{fort2019deep}. 
We employ all of the above to conduct a functional analysis of KD.  

\section{Experimental Setup}
We begin with self-distillation, where the student model has the same architecture and initialisation as the teacher. In theory, this setting gives the student the exact capacity to entirely recover the teacher's function, allowing isolation of the effects of the distillation signal. 
Our core experimental findings are derived from this controlled self-distillation setup. We then verify the generality of our findings in standard KD (smaller student; Section~\ref{sec:standard-kd-ss}), feature-map matching distillation (Section~\ref{sec:feature_map_matching}) and temperature variants (Section~\ref{sec:temperature-kd}). 

Let $M_T$ denote the trained teacher model, trained from initialisation $M_0$. All comparative models (including students and controls) share architecture and initialisation $M_0$, ensuring they begin from the same point in the loss landscape. Thus any observed differences in functional behaviour arise purely from the distillation signal rather than confounds from architecture and initialisation. In self-distillation, students start from $M_0$ and are trained to match the trained teacher $M_T$ with the standard logit-matching objective:

\begin{equation}\label{eq:kd}
  \begin{aligned}
  \mathcal{L}(x;M_S) = (1-\alpha) \, \mathcal{H}(y, \sigma(z_s; T=1)) \\
  +\alpha \ \, \mathcal{KL}(\sigma(z_t/T), \sigma(z_s/T))
  \end{aligned}
\end{equation}
where $x$ is the input, $M_S$ is the student model parameters, $\alpha$ is the teacher-weighting coefficient, $\mathcal{H}$ is the cross-entropy loss function, $\mathcal{KL}$ is the kullback-leibler divergence loss function, $y$ is the ground-truth label, $\sigma$ is the softmax function parameterised by the temperature $T$, and $z_s$ and $z_t$ are the student and teacher logits, respectively. Unless otherwise stated, we keep all training hyperparameters fixed across conditions: optimiser, learning-rate schedule, batch size, data augmentations/preprocessing, number of epochs, and evaluation protocol.
\begin{figure}[H]
\centering
 \subfigure[KD with a Teacher Model.] {\includegraphics[width=0.35\textwidth]{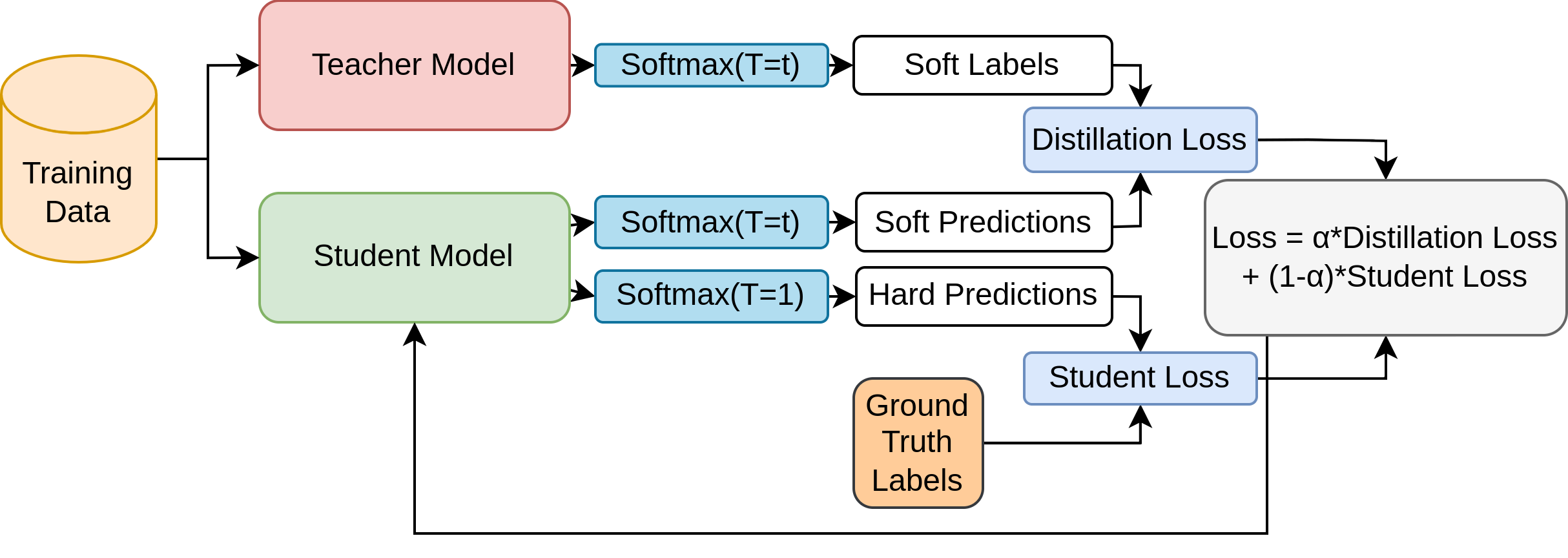}} \label{fig:kd-setup}
 \subfigure[RCD] {\includegraphics[width=0.35\textwidth]{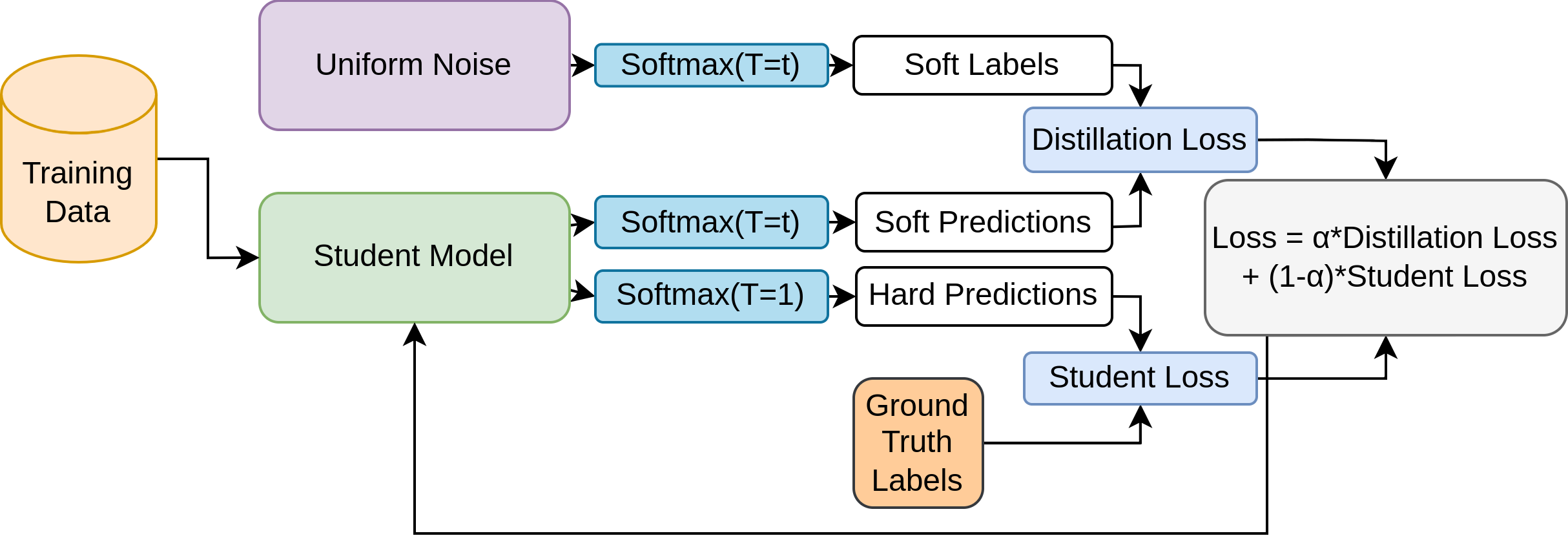}} \label{fig:kd-setup-rand}
 \caption{Knowledge Distillation (KD) and Random Control Distillation (RCD) Setups.}
 \label{fig:kd-setups}
\end{figure}
To isolate the effect of the teacher signal, we compare KD against two controls: SIDDO (same initialisation, different data order), using the same architecture and initialisation as the teacher, but trained independently with different training-order randomness; RCD (random control distillation; analogous to a randomised control trial~\citep{hariton2018randomised}), using the same optimisation as KD (Eq.~\ref{eq:kd}), but where teacher distribution is replaced with a class-uniform target $p^{(\textrm{RCD})}(y\mid x)=\frac{1}{K}$ for $K$ classes (i.e., uniform teacher logits). The setup is visualised in Figure~\ref{fig:kd-setups}. 
RCD is intended as a counterfactual control that preserves the distillation training form while removing teacher information content. 

We vary the distillation coefficient $\alpha \in \{0.1, 0.5, 0.9\}$ to modulate reliance on the teacher. At 0.1, the teacher signal contributes minimally; at 0.5, there is an equal weighting of label and teacher supervision; at 0.9, training is predominantly guided by the teacher. If KD induces meaningful knowledge transfer, functional similarity should monotonically increase with an increase of the $\alpha$ (although optimiser dynamics and task effects may introduce deviations). All experiments use temperature $T = 1$ to preserve the original teacher distribution allowing for full representation matching. However, in Section~\ref{sec:temperature-kd}, we vary temperature to analyse its impact on knowledge transfer. For each architecture--dataset pair across modalities, we train 3 teacher models (seeds 0-2). For each teacher, we then train 10 student models for each distillation setup (KD, RCD, SIDDO) over 3 $\alpha$ values (0.1, 0.5, 0.9) for KD  and RCD - we use 10 training seeds (10-19). This results in 71 models per architecture-teacher combination, and 2,556 models in the self-distillation experiments alone. For smaller-student KD we train 790 models, while for temperature experiments we train 490 models, for a total of \textbf{4,189 models} across the paper when including adversarial and feature map matching experiments. A full breakdown of models trained per experimental setup is provided in Appendix Section~\ref{app:compute_usage}. Table~\ref{fig:modality-dataset-architecture} provides a list of architectures and datasets used, spanning settings from SVHN to ImageNet. Results are reported using the Standard Error of the Mean (SEM)~\citep{belia2005researchers}, which better reflects estimation uncertainty across independent runs.

\begin{table}[H]
\centering
\caption{Modalities used in our experiments, along with their respective datasets and architectures.}
\label{fig:modality-dataset-architecture}
\resizebox{\linewidth}{!}{
\begin{tabular}{|l|l|l|}
\hline
\textbf{Modality} & \textbf{Datasets} & \textbf{Architectures} \\ \hline
\textbf{Image} & \begin{tabular}[c]{@{}l@{}} ImageNet~\citep{deng2009imagenet} \& \\
TinyImageNet~\citep{le2015tiny},\\  SVHN~\citep{netzer2011reading}\end{tabular} &
\begin{tabular}[c]{@{}l@{}}ResNet-50, ResNet-18~\citep{resnet}, VGG19BN\\ VGG19~\citep{simonyan2014very},\\  Vision Transformer (ViT)~\citep{dosovitskiy2020image} \end{tabular} \\ \hline 
\textbf{Audio} &
\begin{tabular}[c]{@{}l@{}}SpeechCommandsV2~\citep{warden2018speech}, \\ UrbanSound8K~\citep{10.1145/2647868.2655045}\end{tabular} &
\begin{tabular}[c]{@{}l@{}} 
VGGish~\citep{hershey2017cnn} AST~\citep{gong2021ast} \\ (Audio Spectrogram Transformer)\end{tabular}\\ \hline
\textbf{Language} &  \begin{tabular}[c]{@{}l@{}} 
TinyShakespeare~\citep{blog2015unreasonable}, \\Adversarial TinyShakespeare `th\_' \end{tabular} &
Nano-GPT, Pico-GPT~\citep{Karpathy2022} \\ \hline 
\end{tabular}
}
\end{table}

\subsection{Functional Similarity Metrics}
We evaluate student--teacher alignment using functional similarity metrics computed on the test set, $\mathcal{D}_{\text{test}}$. This enables an analysis of how similar teacher and student representations are on unseen data. Let ${M}_{T}$ denote the teacher and $M_{C}$ a comparative model (KD student, SIDDO, RCD). For the metrics described below, with the exception of Prediction Agreement, \textbf{lower ($\downarrow$) values indicate greater functional similarity}. The functional similarity metrics examined in this paper are described as follows: 

\begin{itemize}
\itemsep0pt
\item  \textbf{Activation distance:} The mean $\mathcal{L}_{2}$ distance between the softmax output distributions of ${M}_{T}$ and $M_{C}$ on $\mathcal{D}_{test}$, formally expressed as: $1-\frac{1}{|\mathcal{D}_{test}|}\sum\limits^{|\mathcal{D}_{test}|}_{i=1}\Big(\sum\limits^{K}_{k=1}\big(\sigma(M_{T_k}^{(i)}) - \sigma(M_{C_k}^{(i)})\big)^2\Big)^\frac{1}{2}$.

\item  \textbf{Jensen-Shannon Divergence (JS Divergence):} The mean Jensen--Shannon divergence (weighted average of KL divergence~\citep{lin1991divergence}) between the softmax output distributions of ${M}_{T}$ and $M_{C}$ on $\mathcal{D}_{test}$, formally expressed as: \\ $\frac{1}{|\mathcal{D}_{test}|}\sum\limits^{|\mathcal{D}_{test}|}_{i=1} \frac{1}{2}\bigg{(}\mathcal{KL}\big(\sigma(M^{(i)}_{T}) || N^{(i)} \big) + \mathcal{KL}\big(\sigma(M^{(i)}_{C}) || N^{(i)}\big)\bigg)$, where $N^{(i)}=\frac{1}{2}\big(\sigma(M^{(i)}_{T}) + \sigma(M^{(i)}_{C})\big)$.

\item \textbf{Rank Disagreement:} The percentage of disagreement in the hierarchical order of output predictions between ${M}_{T}$ and $M_{C}$ on $\mathcal{D}_{test}$ formally expressed as: \\$\frac{1}{|\mathcal{D}_{test}|}\sum\limits^{|\mathcal{D}_{test}|}_{i=1}\sum\limits^{K}_{k=1}(\boldsymbol{1}(M_{T_k}^{(i)}\neq M_{C_k}^{(i)}))$.

\item \textbf{Prediction Disagreement:} The proportion of inputs on which ${M}_{T}$ and $M_{C}$ produce different top-1 predictions on $\mathcal{D}_{test}$, formally expressed as: $\frac{1}{|\mathcal{D}_{test}|}\sum\limits^{|\mathcal{D}_{test}|}_{i=1}(\boldsymbol{1}(arg\space max\space(M_{T}^{(i)})\neq arg\space max (M_{C_k}^{(i)})))$. We also use Prediction Agreement, the complement of Prediction Disagreement, to analyse prediction transfer, for predication agreement increased values indicate more functional similarity. 
\end{itemize}
These metrics move beyond accuracy and loss by quantifying the extent to which students reproduce the teacher's output function on a given task, which is essential for understanding student--teacher alignment in practice. In Appendix Section~\ref{app:extended_func_analysis}, we show how the functional analysis used in this paper relates to information-theoretic and geometric perspectives. 

\subsection{Knowledge Transfer Definitions}
In this section, we define what, under the experimental conditions explored in this paper, constitutes meaningful knowledge transfer, how it is expected to manifest in the student model, and how different transfer payoffs should be interpreted. We quantify knowledge transfer through functional similarity between teacher and student, since this directly measures the extent to which the student reproduces the teacher's output function on unseen data. This enables us to understand how knowledge transfer affects the representation of the student model. Accordingly, when knowledge transfer occurs, we expect increased functional alignment between teacher and student, especially in self-distillation settings where the student has the capacity to match the teacher's function. 

\begin{figure}[H]
    \centering
    \includegraphics[width=0.32\linewidth]{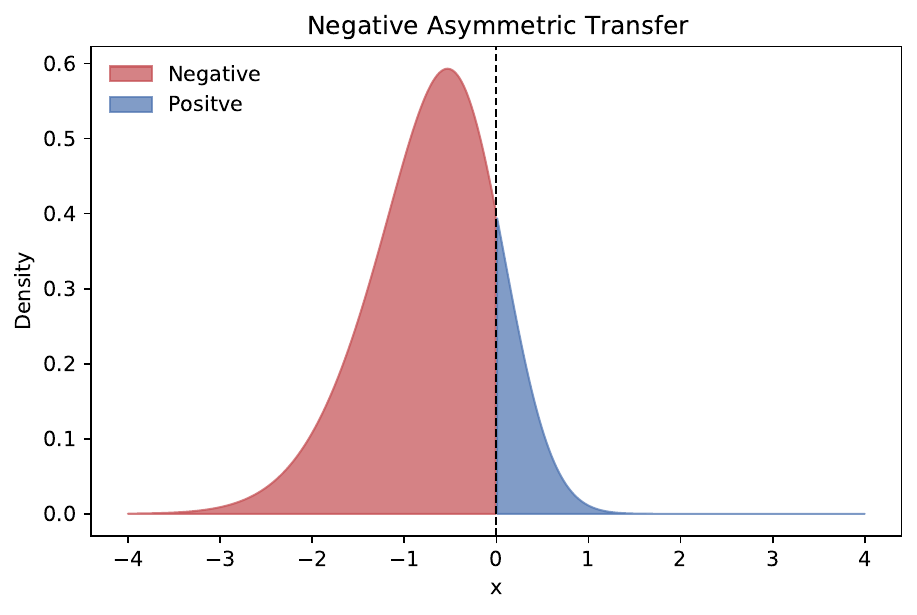}
    \includegraphics[width=0.32\linewidth]{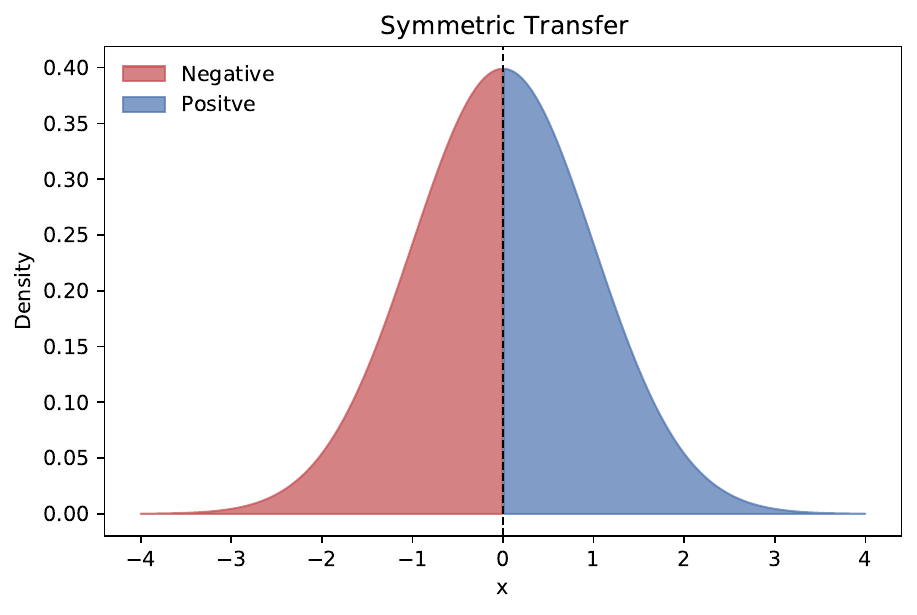}
    \includegraphics[width=0.32\linewidth]{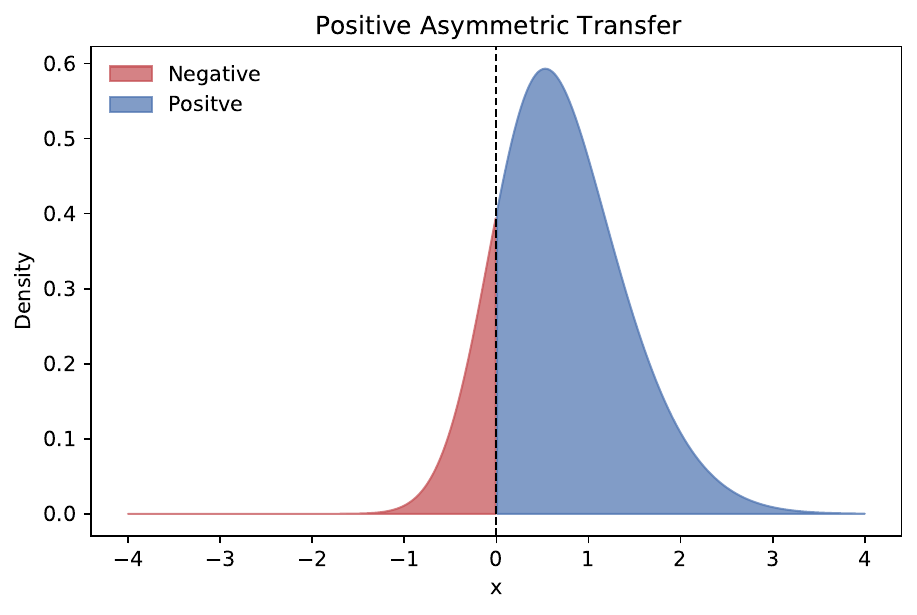}
    \caption{Types of transfer that may arise under knowledge distillation. Left: negative asymmetric transfer; middle: symmetric transfer; right: positive asymmetric transfer.}
    \label{fig:payoffs}
\end{figure}

\paragraph{Knowledge transfer:}We operationally define knowledge transfer as a statistically significant increase in teacher--student functional similarity relative to both RCD and SIDDO controls. Concretely, we say that knowledge transfer has occurred when most functional similarity metrics (e.g., Activation Distance, Rank Disagreement, Prediction Disagreement, and JS Divergence) improve significantly when comparing the student to the teacher against these baselines. A decrease in these metrics indicates increased functional similarity between teacher and student under KD. If this criterion is met, the resulting agreement between student and teacher, relative to the baselines, can fall into one of three cases detailed below (see Figure~\ref{fig:payoffs} for a diagrammatic example). 

Let $A_{\text{correct}}(M_S, M_T)$ denote the agreement in top-1 label rate between student and teacher restricted to inputs on which the teacher is correct, and let $A_{\text{incorrect}}(M_S, M_T)$ denote the agreement rate restricted to inputs on which the teacher is incorrect. Let $A_{\text{correct}}(B, M_T)$ and $A_{\text{incorrect}}(B, M_T)$ denote the corresponding agreement rates for a baseline model $B$ (e.g., SIDDO or RCD). We then define:

\begin{equation}
\Delta_{\text{correct\_agreement}}
=
A_{\text{correct}}(M_S, M_T) - A_{\text{correct}}(B, M_T)
\end{equation}
and
\begin{equation}
\Delta_{\text{incorrect\_agreement}}
=
A_{\text{incorrect}}(M_S, M_T) - A_{\text{incorrect}}(B, M_T).
\end{equation}

These quantities measure how much KD changes teacher--student agreement relative to the control baseline on teacher-correct and teacher-incorrect predictions, respectively. Based on these quantities, transfer can take one of three forms:

\begin{equation}
     \mbox{Negative asymmetric transfer: } 
\Delta_{\text{correct\_agreement}} < \Delta_{\text{incorrect\_agreement}}
\label{eq:negative_assymetric_transfer}
\end{equation}
\begin{equation}
     \mbox{Symmetric transfer: } 
\Delta_{\text{correct\_agreement}} = \Delta_{\text{incorrect\_agreement}}
\end{equation}
\begin{equation}
     \mbox{Positive asymmetric transfer: } 
\Delta_{\text{correct\_agreement}} > \Delta_{\text{incorrect\_agreement}}
\end{equation}

\paragraph{Asymmetric payoff:} An asymmetric payoff occurs when the increase in teacher--student prediction agreement, relative to controls, differs between teacher-correct and teacher-incorrect predictions. We report the separate changes in correct-agreement, $\Delta_{\text{correct\_agreement}}$, and incorrect-agreement, $\Delta_{\text{incorrect\_agreement}}$, in order to characterise the structure of transferred agreement. 

\paragraph{Negative asymmetric transfer:} We define negative transfer as the regime in which two properties hold simultaneously: (i) functional similarity improves relative to controls, but (ii) the increase in incorrect-agreement exceeds the increase in correct-agreement (negative asymmetry). In other words, the student becomes more functionally similar to the teacher while absorbing proportionally more of the teacher's mistakes than its correct behaviour. 

\subsection{Hypothesis Testing} \label{sec:hypo}

To evaluate whether KD increases teacher--student functional similarity beyond controls, we compare KD to SIDDO and RCD on each metric and condition.
We use paired tests by matching runs across conditions: for each architecture--dataset--teacher-seed setting and student seed, models share the same initialisation and the same data order, isolating the effect of the distillation signal from stochastic factors. 
For a given architecture--dataset--teacher-seed setting and KD coefficient $\alpha$, our primary hypotheses against controls are:

\begin{itemize}
\setlength{\parskip}{0pt}
  \setlength{\itemsep}{0pt}
  \item[$H_0$:] KD students, on average, are not more functionally similar to the teacher than control models.
  \item[$H_a$:] KD students, on average, are more functionally similar to the teacher than control models. 
\end{itemize}

\noindent For each metric, we compare the 10 KD runs for each $\alpha$ value to the corresponding 
control values offered by \textit{both} SIDDO and RCD using one-sided Wilcoxon signed-rank tests~\citep{woolson2007wilcoxon}. Our scientific question is directional:
functional similarity corresponds to improved values in the preferred direction for each metric (lower for distance/disagreement metrics; higher for agreement/accuracy; lower for loss). 

To control the expected false discovery rate across the functional similarity metrics within each setting, we apply Benjamini--Hochberg False Discovery Rate (FDR) correction ~\citep{yekutieli1999resampling} for each $\alpha$ in \{0.1,0.5 and 0.9\} and teacher seed, and report significance after FDR. For accuracy and loss, we keep the significance threshold at $0.05$. We classify transfer as \textbf{statistically supported} only when i) the corrected significance threshold is satisfied, and ii) the effect is in the expected direction for a \textit{majority} of the functional similarity metrics under both control comparisons. The control conditions are SIDDO and RCD, and variable of interest is standard KD. 
We use \textit{marginal statistically supported transfer} to describe cases where transfer is statistically supported, but the raw transfer metrics remain close to control. This indicates limited shared function, with only weak student--teacher agreement on both correct and incorrect teacher predictions. 

\section{Results and Discussion}
We examine functional transfer facilitated by knowledge distillation across three modalities: computer vision, audio, and language, presenting results in this order. In Section~\ref{sec:self-distillation} we study self-distillation, where the student has the capacity to match the teacher's function, and in the main body we analyse teacher seed 0. Concretely, we report SVHN (ResNet18) and TinyImageNet (ResNet50) for vision, SpeechCommands and UrbanSound8K (VGGish) for audio, and TinyShakespeare (Nano-GPT/Pico-GPT) for language. Full results for remaining teacher seeds (1--2), additional datasets, and architectures are provided in the appendix (SVHN: Appendix~\ref{SHVN_results}; TinyImageNet: Appendix~\ref{app:tinyImageNet}; audio: Appendix~\ref{audio_results}; language: Appendix~\ref{Language_results}). We choose widely used benchmarks across vision, audio, and language to test the generality of our findings across modalities. 
  
Following self-distillation, Section~\ref{sec:standard-kd-ss} studies standard distillation with smaller students and examines how negative asymmetric transfer changes with student capacity, including a scaling-law analysis~\citep{busbridge2025distillation} (additional seeds in Appendix~\ref{app:standard-kd-ss}). We then test KD variants via feature-map matching (Section~\ref{sec:feature_map_matching}), and probe how transfer depends on KD hyperparameters by studying the impact of temperature (Section~\ref{sec:temperature-kd}). Our final experiments illustrate downstream implications of negative asymmetric transfer via a controlled targeted error-transfer setting, showing that KD can copy a specific erroneous pattern from an adversarially modified teacher (Section~\ref{sec:adv-transfer}). Section~\ref{sec:theory} summarises our main empirical findings with a gradient-level mechanistic explanation, showing that negative asymmetric behaviour arises naturally from the standard KD objective under teacher error mass and motivating safety-relevant auditing of teacher error modes. In Appendix~\ref{app:extended_func_analysis}, we show that these conclusions are consistent under information-theoretic and geometric measures in addition to our functional similarity metrics, and Appendix~\ref{app:label_smoothing} shows that RCD is equivalent to label smoothing while preserving the KD optimisation shown in equation~\eqref{eq:kd}.

Training details for all settings are also provided in the appendix. Unless specified otherwise,~\textbf{we report means and $\pm$1~SEM over 10 runs per teacher seed and condition in tables and figures.} We use the terms \textit{statistically supported} and \textit{marginal} as defined in Section~\ref{sec:hypo}. 

\subsection{Self-Distillation}
\label{sec:self-distillation}
We begin with the computer vision domain (Section~\ref{subsec:CV}). In small-scale settings, we observe that when functional transfer is non-marginal and statistically supported, transferred agreement is consistently \textit{asymmetric} toward the teacher's errors. We then validate these findings at larger scale on TinyImageNet. In the base setting, transfer is marginal; however, increasing teacher generalisation performance (via augmentation) amplifies both functional transfer and the negative asymmetry of transfer. Finally, we evaluate audio and language (Sections~\ref{subsec:audio} and~\ref{subsec:language}), where we observe stronger functional transfer coupled with increased negative asymmetry.

\subsubsection{Computer Vision}
\label{subsec:CV}
For the small-scale experiments, we use SVHN with ResNet18. Here, we observe that KD yields statistically supported functional similarity at high $\alpha$ values, but the magnitude and asymmetry of transfer vary across teacher seeds. When transfer is non-marginal, we observe a systematic increase in student--teacher agreement on incorrect predictions relative to correct ones. Table~\ref{tab:resnet-svhn-teacher-main} shows teacher variability: train losses of $6.46\times10^{-4}$, $6.1\times10^{-5}$, and $4.66\times10^{-3}$ with a generalisation gap of $\approx 0.04$ for seeds 0, 1, and 2, respectively. 

\begin{figure}[H]
    \centering 
    \includegraphics[width=\linewidth]{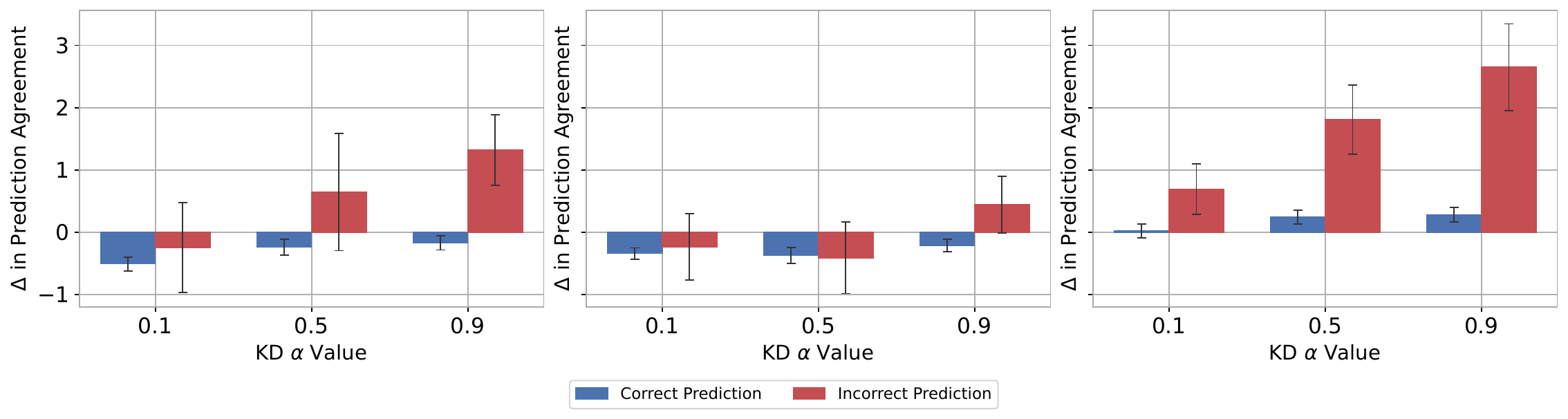}
    \caption{Difference in prediction agreement between KD students and the best control baseline on correct (blue) vs. incorrect (red) predictions; for the SVHN ResNet18 (seeds 0 to 2, left to right).}
    \label{fig:resnet-svhn-prediction-main}
\end{figure}

\begin{table}[H]
\caption{SVHN ResNet18 Teacher Performance on Train and Test Sets.}
\scriptsize
\label{tab:resnet-svhn-teacher-main}
\centering
\begin{tabular}{|c|c|c|c|c|}
\hline
\textbf{Teacher Seed} & \textbf{Train Loss} & \textbf{Train Accuracy} & \textbf{Test Loss} & \textbf{Test Accuracy} \\ \hline
0 & 0.000646 & 0.999850 & 0.381410 & 0.951829 \\ \hline
1 & 0.000061 & 0.999973 & 0.331054 & 0.952251 \\ \hline
2 & 0.004657 & 0.998580 & 0.309702 & 0.947104 \\ \hline
\end{tabular}
\end{table}
When analysing the functional transfer for teacher seed 0 (Table \ref{tab:svhn-resnet18-ts-0-main}), we observe that KD across $\alpha$ values yields only marginal transfer relative to the SIDDO and RCD baselines. Notably, the best test loss and accuracy (Table \ref{tab:svhn-resnet18-ts-0-main}) are achieved by random control distillation, reducing confidence that KD's performance gains arise from meaningful knowledge transfer and instead supporting the view of KD as a data-dependent regulariser.
For the highest-train-loss teacher (seed 2), KD produces statistically supported functional transfer across metrics at most $\alpha$ values (Table~\ref{tab:resnet-svhn-significance}), with the exception of Prediction Disagreement at $\alpha=0.1$. This transfer coincides with a large asymmetric payoff in prediction agreement toward the teacher's incorrect predictions (Figure~\ref{fig:resnet-svhn-prediction-main}). The lowest-train-loss teacher (seed 1) shows no statistically supported transfer at $\alpha \in \{0.1, 0.5\}$ and only partial transfer at $\alpha=0.9$ (again, excluding Prediction Disagreement). For seed 0 (intermediate train loss), we observe statistically supported transfer at $\alpha=0.5$ and $0.9$, accompanied by asymmetric incorrect agreement (Figure~\ref{fig:resnet-svhn-prediction-main}).

\begin{table}[H]
\centering

\caption{SVHN ResNet18 (teacher seed 0): mean $\pm$ 1 SEM over 10 runs. \textbf{Bold} indicates the best mean per metric. Arrows ($\uparrow$/$\downarrow$) denote the preferred direction for each metric.}
\label{tab:svhn-resnet18-ts-0-main}
\resizebox{\textwidth}{!}{%
\begin{tabular}{|l|c|ccc|ccc|}
\hline
\multicolumn{1}{|c|}{\multirow{2}{*}{\textbf{Metrics}}} & \textbf{Control} & \multicolumn{3}{c|}{\textbf{Knowledge Distillation}} & \multicolumn{3}{c|}{\textbf{Random Control Distillation}} \\ \cline{2-8} 
\multicolumn{1}{|c|}{} & \textbf{SIDDO} & \multicolumn{1}{c|}{\textbf{0.1}} & \multicolumn{1}{c|}{\textbf{0.5}} & \textbf{0.9} & \multicolumn{1}{c|}{\textbf{0.1}} & \multicolumn{1}{c|}{\textbf{0.5}} & \textbf{0.9} \\ \hline
Activation Distance \textbf{($\mathbf{\downarrow}$)} & 0.063$\pm{0.002}$ & \multicolumn{1}{c|}{0.064$\pm{0.001}$} & \multicolumn{1}{c|}{0.060$\pm{0.001}$} & \textbf{0.059$\pm{0.001}$} & \multicolumn{1}{c|}{0.144$\pm{0.001}$} & \multicolumn{1}{c|}{0.493$\pm{0.000}$} & 0.849$\pm{0.000}$ \\ \hline
Rank Disagreement  \textbf{($\mathbf{\downarrow}$)} & 0.696$\pm{0.003}$ & \multicolumn{1}{c|}{0.688$\pm{0.004}$} & \multicolumn{1}{c|}{0.684$\pm{0.003}$} & \textbf{0.681$\pm{0.003}$} & \multicolumn{1}{c|}{0.800$\pm{0.002}$} & \multicolumn{1}{c|}{0.798$\pm{0.002}$} & 0.802$\pm{0.003}$ \\ \hline
Prediction Disagreement  \textbf{($\mathbf{\downarrow}$)} & 0.045$\pm{0.001}$ & \multicolumn{1}{c|}{0.046$\pm{0.001}$} & \multicolumn{1}{c|}{0.043$\pm{0.001}$} & \textbf{0.042$\pm{0.001}$} & \multicolumn{1}{c|}{\textbf{0.042$\pm{0.001}$}} & \multicolumn{1}{c|}{0.043$\pm{0.001}$} & 0.046$\pm{0.001}$ \\ \hline
JS Divergence  \textbf{($\mathbf{\downarrow}$)} & 0.025$\pm{0.001}$ & \multicolumn{1}{c|}{0.025$\pm{0.001}$} & \multicolumn{1}{c|}{0.023$\pm{0.001}$} & \textbf{0.022$\pm{0.000}$} & \multicolumn{1}{c|}{0.053$\pm{0.000}$} & \multicolumn{1}{c|}{0.201$\pm{0.000}$} & 0.431$\pm{0.000}$ \\ \hline
Accuracy  \textbf{($\mathbf{\uparrow}$)} & 0.952$\pm{0.001}$ & \multicolumn{1}{c|}{0.951$\pm{0.001}$} & \multicolumn{1}{c|}{0.954$\pm{0.001}$} & 0.954$\pm{0.001}$ & \multicolumn{1}{c|}{\textbf{0.957$\pm{0.001}$}} & \multicolumn{1}{c|}{\textbf{0.957$\pm{0.001}$}} & 0.955$\pm{0.001}$ \\ \hline
Loss  \textbf{($\mathbf{\downarrow}$)} & 0.385$\pm{0.011}$ & \multicolumn{1}{c|}{0.344$\pm{0.008}$} & \multicolumn{1}{c|}{0.310$\pm{0.006}$} & 0.293$\pm{0.004}$ & \multicolumn{1}{c|}{\textbf{0.236$\pm{0.003}$}} & \multicolumn{1}{c|}{0.692$\pm{0.001}$} & 1.698$\pm{0.001}$ \\ \hline
\end{tabular}%
}
\end{table}
\begin{table}[H]
    \centering
    \caption{SVHN ResNet18 (significance testing). \cmark~indicates significant transfer compared to controls; \xmark~indicates no significance. Each triplet corresponds to teacher seeds 0-2 (left to right).}
    \label{tab:resnet-svhn-significance}
\resizebox{\textwidth}{!}{
\begin{tabular}{|c|c|c|c|c|c|c|}
\hline
\textbf{} & \textbf{Activation Distance}                                      & \textbf{Rank Disagreement}                                        & \textbf{Prediction Disagreement}                                  & \textbf{JS Divergence}                                            & \textbf{Accuracy}                                                 & \textbf{Loss}                                                     \\ \hline
KD 0.1    & \xmark \xmark \cmark & \xmark \xmark \cmark & \xmark \xmark \xmark & \xmark \xmark \cmark & \xmark \xmark \xmark & \xmark \xmark \xmark \\ \hline
KD 0.5    & \xmark \xmark \cmark & \xmark \xmark \cmark & \xmark \xmark \cmark & \xmark \xmark \cmark & \xmark \xmark \xmark & \xmark \xmark \cmark \\ \hline
KD 0.9    & \cmark \xmark \cmark & \cmark \xmark \cmark & \xmark \xmark \cmark & \cmark \xmark \cmark & \xmark \xmark \xmark & \xmark \xmark \cmark \\ \hline
\end{tabular}}
\end{table}

Across seeds, higher teacher train loss is associated with stronger, more asymmetric, functional transfer, consistent with a teacher that deviates more from ground-truth labels, thereby exposing students to incorrect structure that is preferentially transferred under KD.

\begin{figure}[H]
    \centering
    \includegraphics[width=\linewidth]{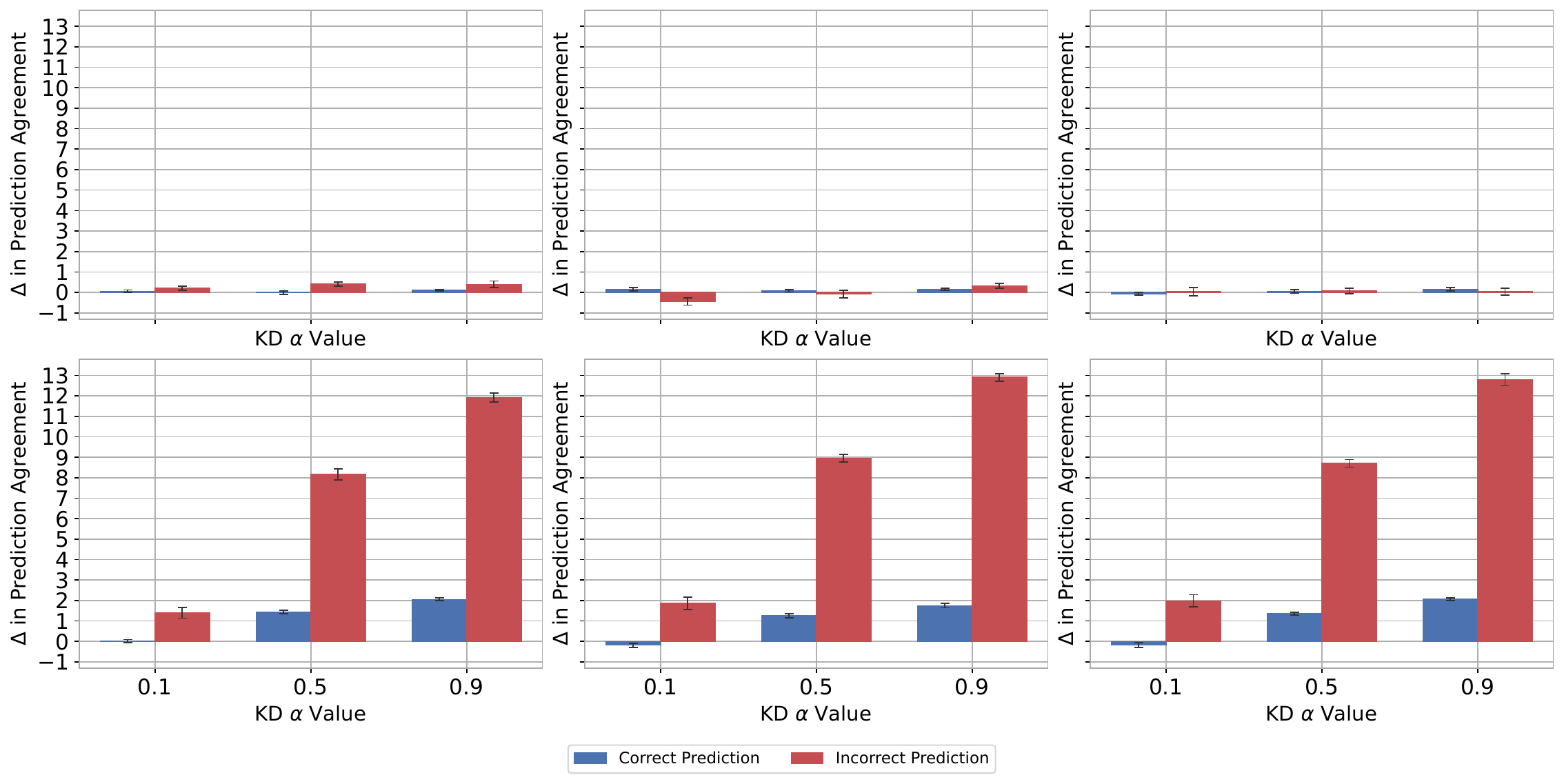}
     \caption{Difference in prediction agreement between KD students and the best control baseline on correct (blue) vs.\ incorrect (red) predictions; on TinyImageNet, ResNet-50. Top: base teachers. Bottom: augmented teachers  (seeds 0 to 2, left to right).}
    \label{fig:resnet-TiN-prediction-main}
\end{figure}

\paragraph{Function Transfer in Larger-Scale Settings}
To understand how these results scale, we next study TinyImageNet with ResNet50. In the base setting, KD produces statistically supported but marginal functional gains relative to SIDDO in Table~\ref{tab:tin-resnet50-ts-0-main}; the corresponding prediction agreement shows no clear preference toward correct or incorrect agreement in Figure~\ref{fig:resnet-TiN-prediction-main}. Motivated by the SVHN analysis, we increase the teacher train loss via data augmentation using RandAugment \citep{cubuk2020randaugment} with the default settings, and examine the consequences for functional transfer and asymmetry. Here, we increase generalisation performance while increasing the teacher's loss (Table~\ref{TinyImageNet-ResNet50-Teachers-Base-Main-Aug}) to study the nature of knowledge transfer at increased dataset and architecture complexity.

\begin{table}[H]
\caption{TinyImageNet ResNet50 Teacher Performance: Base vs. RandAugment.}
\label{TinyImageNet-ResNet50-Teachers-Base-Main-Aug}
\centering
\scriptsize
\begin{tabular}{|c|c|c|c|c|}
\hline
\textbf{Teacher Seed} & \textbf{Train Loss} & \textbf{Train Accuracy} & \textbf{Test Loss} & \textbf{Test Accuracy} \\ \hline
\multicolumn{5}{|c|}{\textbf{Base}} \\ \hline
0 & 0.001426 & 0.999800 & 2.070590 & 0.605300 \\ \hline
1 & 0.001393 & 0.999800 & 2.051494 & 0.607900 \\ \hline
2 & 0.001436 & 0.999800 & 2.051024 & 0.610600 \\ \hline
\hline
\multicolumn{5}{|c|}{\textbf{RandAugment}} \\ \hline
0 & 0.672748 & 0.840410 & 1.620552 & 0.638800 \\ \hline
1 & 0.678245 & 0.839200 & 1.629393 & 0.641800 \\ \hline
2 & 0.667570 & 0.840750 & 1.624969 & 0.641100 \\ \hline
\end{tabular}
\end{table}

In the base setting (Table~\ref{TinyImageNet-ResNet50-Teachers-Base-Main-Aug}), teachers have very low train loss and moderate test accuracy; with augmentation, 
train loss increases while test accuracy improves, as expected.
\begin{table}[H]

\caption{TinyImageNet (base): ResNet50 mean $\pm$ SEM over 10 runs (teacher seed 0). \textbf{Bold} indicates best mean.}
\label{tab:tin-resnet50-ts-0-main}
\resizebox{\textwidth}{!}{%
\begin{tabular}{|c|c|ccc|ccc|}
\hline
\multicolumn{1}{|c|}{\multirow{2}{*}{\textbf{Metrics}}} & \multicolumn{1}{c|}{\textbf{Control}} & \multicolumn{3}{c|}{\textbf{Knowledge Distillation}} & \multicolumn{3}{c|}{\textbf{Random Control Distillation}} \\ \cline{2-8} 
\multicolumn{1}{|c|}{} & \multicolumn{1}{c|}{\textbf{SIDDO}} & \multicolumn{1}{c|}{\textbf{0.1}} & \multicolumn{1}{c|}{\textbf{0.5}} & \multicolumn{1}{c|}{\textbf{0.9}} & \multicolumn{1}{c|}{\textbf{0.1}} & \multicolumn{1}{c|}{\textbf{0.5}} & \multicolumn{1}{c|}{\textbf{0.9}} \\ \hline
Activation Distance & 0.157 $\pm{}$ 0.001 & \multicolumn{1}{c|}{0.157 $\pm{}$ 0.001} & \multicolumn{1}{c|}{0.156 $\pm{}$ 0.001} & \textbf{0.155 $\pm{}$ 0.000}  & \multicolumn{1}{c|}{0.343 $\pm{}$ 0.000}  & \multicolumn{1}{c|}{0.581 $\pm{}$ 0.000}  & 0.791 $\pm{}$ 0.000 \\ \hline
Rank Disagreement & \textbf{0.939 $\pm{}$ 0.000}  & \multicolumn{1}{c|}{\textbf{0.939 $\pm{}$ 0.000} } & \multicolumn{1}{c|}{\textbf{0.939 $\pm{}$ 0.000} } & \textbf{0.939 $\pm{}$ 0.000}  & \multicolumn{1}{c|}{0.980 $\pm{}$ 0.000}  & \multicolumn{1}{c|}{0.984 $\pm{}$ 0.000}  & 0.984 $\pm{}$ 0.000 \\ \hline
Prediction Disagreement & 0.153 $\pm{}$ 0.001 & \multicolumn{1}{c|}{0.152 $\pm{}$ 0.001} & \multicolumn{1}{c|}{\textbf{0.151 $\pm{}$ 0.001}} & \textbf{0.151 $\pm{}$ 0.001} & \multicolumn{1}{c|}{0.190 $\pm{}$ 0.001} & \multicolumn{1}{c|}{0.214 $\pm{}$ 0.000}  & 0.324 $\pm{}$ 0.000 \\ \hline
JS Divergence & 0.040 $\pm{}$ 0.000 & \multicolumn{1}{c|}{0.040 $\pm{}$ 0.000}  & \multicolumn{1}{c|}{\textbf{0.039 $\pm{}$ 0.000} } & \textbf{0.039 $\pm{}$ 0.000}  & \multicolumn{1}{c|}{0.171 $\pm{}$ 0.000}  & \multicolumn{1}{c|}{0.333 $\pm{}$ 0.000}  & 0.533 $\pm{}$ 0.000 \\ \hline
Accuracy & 0.605 $\pm{}$ 0.001 & \multicolumn{1}{c|}{0.605 $\pm{}$ 0.000}  & \multicolumn{1}{c|}{0.604 $\pm{}$ 0.001} & 0.605 $\pm{}$ 0.001 & \multicolumn{1}{c|}{\textbf{0.607 $\pm{}$ 0.000} } & \multicolumn{1}{c|}{0.606 $\pm{}$ 0.001} & 0.580 $\pm{}$ 0.000 \\ \hline
Loss & 2.068 $\pm{}$ 0.001 & \multicolumn{1}{c|}{2.065 $\pm{}$ 0.002} & \multicolumn{1}{c|}{2.055 $\pm{}$ 0.001} & 2.043 $\pm{}$ 0.002 & \multicolumn{1}{c|}{\textbf{1.977 $\pm{}$ 0.001}} & \multicolumn{1}{c|}{2.497 $\pm{}$ 0.001} & 3.612 $\pm{}$ 0.002 \\ \hline
\end{tabular}%
}
\end{table}
\begin{table}[H]
\caption{ResNet50 on TinyImageNet (significance testing). \cmark~indicates significant results compared to controls; \xmark~indicates insignificant results. Each tick represents a teacher (seeds 0 to 2, left to right).}
\label{tab:resnet50-tin-significance}
\resizebox{\textwidth}{!}{
\begin{tabular}{|c|c|c|c|c|c|c|}
\hline
\textbf{} & \textbf{Activation Distance}                                      & \textbf{Rank Disagreement}                                        & \textbf{Prediction Disagreement}                                  & \textbf{JS Divergence}                                            & \textbf{Accuracy}                                                 & \textbf{Loss}                                                     \\ \hline
KD 0.1    & \xmark \xmark \xmark & \xmark \xmark \xmark & \xmark \xmark \xmark & \xmark \xmark \xmark & \xmark \cmark \xmark & \xmark \xmark \xmark \\ \hline
KD 0.5    & \cmark \xmark \cmark & \cmark \cmark \cmark & \xmark \xmark \xmark & \cmark \cmark \cmark & \xmark \cmark \xmark & \xmark \xmark \xmark \\ \hline
KD 0.9    & \cmark \cmark \cmark & \cmark \cmark \cmark & \cmark \cmark \xmark & \cmark \cmark \cmark & \xmark \xmark \xmark & \xmark \xmark \xmark \\ \hline
\end{tabular}}
\end{table}

Having established how augmentation changes the teacher regime, we now examine how this impacts functional similarity between students and teachers (teacher seed 0). In the base case (Table~\ref{tab:tin-resnet50-ts-0-main}), KD with $\alpha=0.9$ improves functional similarity over SIDDO by at most $0.002$ (Activation Distance), $0.000$ (Rank Disagreement), $0.002$ (Prediction Disagreement), and $0.001$ (JS Divergence). Table~\ref{tab:resnet50-tin-significance} shows that this qualifies as statistically supported transfer; however, it is marginal in magnitude. Under augmentation (Table~\ref{tab:tina-resnet-ts-0-main}), we observe substantially larger functional similarity gains: KD with $\alpha=0.9$ improves by $0.062$ (Activation Distance), $0.016$ (Rank Disagreement), $0.060$ (Prediction Disagreement), and $0.030$ (JS Divergence), 
corroborating the hypothesis that increased teacher error mass is associated with stronger functional transfer. Relative to the base teachers, augmentation increases train loss while improving test accuracy, and we observe larger functional alignment under KD in this regime. 
Notably, in both the base and augmented settings, the best test loss/accuracy occurs under random control distillation, indicating that improved performance does not require a meaningful teacher signal.

\begin{table}[H]

\caption{TinyImageNet (RandAugment): ResNet50 mean $\pm$ SEM over 10 runs (teacher seed 0). \textbf{Bold} indicates best mean.}
\label{tab:tina-resnet-ts-0-main}
\resizebox{\textwidth}{!}{%
\begin{tabular}{|c|c|ccc|ccc|}
\hline
\multicolumn{1}{|c|}{\multirow{2}{*}{\textbf{Metrics}}} & \multicolumn{1}{c|}{\textbf{Control}} & \multicolumn{3}{c|}{\textbf{Knowledge Distillation}} & \multicolumn{3}{c|}{\textbf{Random Control Distillation}} \\ \cline{2-8} 
\multicolumn{1}{|c|}{} & \multicolumn{1}{c|}{\textbf{SIDDO}} & \multicolumn{1}{c|}{\textbf{0.1}} & \multicolumn{1}{c|}{\textbf{0.5}} & \multicolumn{1}{c|}{\textbf{0.9}} & \multicolumn{1}{c|}{\textbf{0.1}} & \multicolumn{1}{c|}{\textbf{0.5}} & \multicolumn{1}{c|}{\textbf{0.9}} \\ \hline
Activation Distance & 0.193 $\pm{}$ 0.000 & \multicolumn{1}{c|}{0.183 $\pm{}$ 0.000}  & \multicolumn{1}{c|}{0.150 $\pm{}$ 0.000}  & \textbf{0.131 $\pm{}$ 0.000}  & \multicolumn{1}{c|}{0.245 $\pm{}$ 0.001} & \multicolumn{1}{c|}{0.501 $\pm{}$ 0.001} & 0.781 $\pm{}$ 0.000 \\ \hline
Rank Disagreement & 0.959 $\pm{}$ 0.000 & \multicolumn{1}{c|}{0.957 $\pm{}$ 0.000}  & \multicolumn{1}{c|}{0.948 $\pm{}$ 0.000}  & \textbf{0.943 $\pm{}$ 0.000}  & \multicolumn{1}{c|}{0.975 $\pm{}$ 0.000}  & \multicolumn{1}{c|}{0.981 $\pm{}$ 0.000}  & 0.987 $\pm{}$ 0.000 \\ \hline
Prediction Disagreement & 0.196 $\pm{}$ 0.001 & \multicolumn{1}{c|}{0.188 $\pm{}$ 0.001} & \multicolumn{1}{c|}{0.154 $\pm{}$ 0.001} & \textbf{0.136 $\pm{}$ 0.001} & \multicolumn{1}{c|}{0.195 $\pm{}$ 0.001} & \multicolumn{1}{c|}{0.240 $\pm{}$ 0.001} & 0.572 $\pm{}$ 0.001 \\ \hline
JS Divergence & 0.058 $\pm{}$ 0.000 & \multicolumn{1}{c|}{0.052 $\pm{}$ 0.000}  & \multicolumn{1}{c|}{0.036 $\pm{}$ 0.000}  & \textbf{0.028 $\pm{}$ 0.000}  & \multicolumn{1}{c|}{0.094 $\pm{}$ 0.000}  & \multicolumn{1}{c|}{0.266 $\pm{}$ 0.000}  & 0.563 $\pm{}$ 0.000 \\ \hline
Accuracy & 0.640 $\pm{}$ 0.000 & \multicolumn{1}{c|}{0.643 $\pm{}$ 0.001} & \multicolumn{1}{c|}{0.644 $\pm{}$ 0.000}  & 0.642 $\pm{}$ 0.000 & \multicolumn{1}{c|}{0.646 $\pm{}$ 0.001} & \multicolumn{1}{c|}{\textbf{0.657 $\pm{}$ 0.001}} & 0.400 $\pm{}$ 0.001 \\ \hline
Loss & 1.619 $\pm{}$ 0.003 & \multicolumn{1}{c|}{1.600 $\pm{}$ 0.001} & \multicolumn{1}{c|}{1.578 $\pm{}$ 0.001} & 1.577 $\pm{}$ 0.001 & \multicolumn{1}{c|}{\textbf{1.551 $\pm{}$ 0.001}} & \multicolumn{1}{c|}{1.984 $\pm{}$ 0.002} & 4.211 $\pm{}$ 0.001 \\ \hline
\end{tabular}%
}
\end{table}
\begin{table}[H]
\caption{ResNet50 on TinyImageNet with RandAugment (significance testing). \cmark~indicates significant results compared to controls; \xmark~indicates insignificant results. Each tick represents a teacher (seeds 0 to 2, left to right).}
\label{tab:resnet-tina-significance}
\resizebox{\textwidth}{!}{
\begin{tabular}{|c|c|c|c|c|c|c|}
\hline
\textbf{} & \textbf{Activation Distance}                                      & \textbf{Rank Disagreement}                                        & \textbf{Prediction Disagreement}                                  & \textbf{JS Divergence}                                            & \textbf{Accuracy}                                                 & \textbf{Loss}                                                     \\ \hline
KD 0.1    & \cmark \cmark \cmark & \cmark \cmark \cmark & \cmark \cmark \cmark & \cmark \cmark \cmark & \xmark \xmark \xmark & \xmark \xmark \xmark \\ \hline
KD 0.5    & \cmark \cmark \cmark & \cmark \cmark \cmark & \cmark \cmark \cmark & \cmark \cmark \cmark & \xmark \xmark \xmark & \xmark \xmark \xmark \\ \hline
KD 0.9    & \cmark \cmark \cmark & \cmark \cmark \cmark & \cmark \cmark \cmark & \cmark \cmark \cmark & \xmark \xmark \xmark & \xmark \xmark \xmark \\ \hline
\end{tabular}}
\end{table}

Figure~\ref{fig:resnet-TiN-prediction-main} shows the corresponding prediction agreement deltas (KD vs.\ best control), highlighting exactly what agreement is transferred between the teacher and the students in the base and augmented condition. At $\alpha=0.9$, students trained from augmented teachers increase incorrect agreement from $\approx 0.2\%$ (base) to $\approx 12\%$, far outpacing the increase in correct agreement. Thus, inducing higher teacher train loss via augmentation reliably amplifies asymmetric incorrect transfer, consistent with the SVHN findings and our regularisation view of KD, with the additional, novel insight of negative asymmetric transfer.

\subsubsection{Audio}
\label{subsec:audio}
We test our findings beyond computer vision by evaluating KD on two audio datasets: UrbanSound8K and SpeechCommands. The trend is consistent: when knowledge transfer is non-marginal (according to functional similarity metrics), it is asymmetric. Students preferentially increase agreement with the teacher on incorrect predictions, and this negative asymmetric transfer intensifies as the teacher weight $\alpha$ increases. Below, we show results of the VGGish architecture on both audio datasets.

\begin{table}[H]
\centering
\caption{VGGish on SpeechCommands: mean and $\pm$ 1 SEM  reported over 10 runs with Teacher Seed 0. \textbf{Bold} values are best performing based on the mean.}
\label{tab:sc-VGGISH-ts-0}
\resizebox{\linewidth}{!}{
\begin{tabular}{|l|c|ccc|ccc|}
\hline
\multicolumn{1}{|c|}{\multirow{2}{*}{\textbf{Metrics}}} & \textbf{Baseline}  & \multicolumn{3}{c|}{\textbf{Knowledge Distillation}}                                                                  & \multicolumn{3}{c|}{\textbf{Random Control Distillation}}                                                     \\ \cline{2-8} 
\multicolumn{1}{|c|}{}                                  & \textbf{SIDDO}     & \multicolumn{1}{c|}{\textbf{0.1}}       & \multicolumn{1}{c|}{\textbf{0.5}}             & \textbf{0.9}                & \multicolumn{1}{c|}{\textbf{0.1}}                & \multicolumn{1}{c|}{\textbf{0.5}}       & \textbf{0.9}       \\ \hline
Activation Distance \textbf{($\mathbf{\downarrow}$)} & 0.190$\pm{0.002}$  & \multicolumn{1}{c|}{0.152$\pm{0.000}$}   & \multicolumn{1}{c|}{0.148$\pm{0.001}$}       & \textbf{0.147$\pm{0.001}$} & \multicolumn{1}{c|}{0.260$\pm{0.001}$}           & \multicolumn{1}{c|}{0.570$\pm{0.001}$}  & 0.877$\pm{0.000}$   \\ \hline
Rank Disagreement  \textbf{($\mathbf{\downarrow}$)} & 0.908$\pm{0.000}$   & \multicolumn{1}{c|}{0.885$\pm{0.000}$}   & \multicolumn{1}{c|}{0.880$\pm{0.000}$}          & \textbf{0.878$\pm{0.000}$}   & \multicolumn{1}{c|}{0.942$\pm{0.000}$}            & \multicolumn{1}{c|}{0.942$\pm{0.000}$}   & 0.939$\pm{0.000}$   \\ \hline
Prediction Disagreement  \textbf{($\mathbf{\downarrow}$)} & 0.144$\pm{0.001}$ & \multicolumn{1}{c|}{0.118$\pm{0.000}$}   & \multicolumn{1}{c|}{\textbf{0.114$\pm{0.001}$}}       & \textbf{0.114$\pm{0.001}$}          & \multicolumn{1}{c|}{0.125$\pm{0.001}$} & \multicolumn{1}{c|}{0.133$\pm{0.001}$} & 0.169$\pm{0.001}$ \\ \hline
JS Divergence  \textbf{($\mathbf{\downarrow}$)} & 0.085$\pm{0.001}$ & \multicolumn{1}{c|}{0.063$\pm{0.000}$}   & \multicolumn{1}{c|}{0.060$\pm{0.000}$} & \textbf{0.059$\pm{0.000}$}            & \multicolumn{1}{c|}{0.120$\pm{0.000}$}             & \multicolumn{1}{c|}{0.274$\pm{0.001}$} & 0.512$\pm{0.001}$ \\ \hline
Accuracy  \textbf{($\mathbf{\uparrow}$)} & 0.870$\pm{0.001}$  & \multicolumn{1}{c|}{0.886$\pm{0.001}$} & \multicolumn{1}{c|}{0.887$\pm{0.000}$}         & 0.884$\pm{0.001}$          & \multicolumn{1}{c|}{\textbf{0.892$\pm{0.000}$}}   & \multicolumn{1}{c|}{0.882$\pm{0.001}$} & 0.844$\pm{0.001}$ \\ \hline
Loss  \textbf{($\mathbf{\downarrow}$)} & 1.076$\pm{0.021}$ & \multicolumn{1}{c|}{0.669$\pm{0.005}$} & \multicolumn{1}{c|}{0.564$\pm{0.003}$}       & \textbf{0.553$\pm{0.004}$} & \multicolumn{1}{c|}{0.565$\pm{0.002}$}          & \multicolumn{1}{c|}{1.103$\pm{0.003}$} & 2.366$\pm{0.004}$ \\ \hline
\end{tabular}}
\end{table}

\begin{table}[H]
\caption{VGGish on SpeechCommands (significance testing). \cmark~indicates significant results compared to controls; \xmark~indicates insignificant results. Each tick represents a teacher (seeds 0 to 2, left to right).}
\label{tab:vgg_sc_sig}
\resizebox{\linewidth}{!}{
\begin{tabular}{|l|l|l|l|l|l|l|}
\hline
\textbf{} & \textbf{Activation Distance}                                      & \textbf{Rank Disagreement}                                        & \textbf{Prediction Disagreement}                                  & \textbf{JS Divergence}                                            & \textbf{Accuracy}                                                 & \textbf{Loss}                                                     \\ \hline
KD 0.1    & \cmark \cmark \cmark & \cmark \cmark \cmark & \cmark \cmark \cmark & \cmark \cmark \cmark & \xmark \xmark \xmark & \xmark \xmark \xmark \\ \hline
KD 0.5    & \cmark \cmark \cmark & \cmark \cmark \cmark & \cmark \cmark \cmark & \cmark \cmark \cmark & \xmark \xmark \xmark & \xmark \xmark \cmark \\ \hline
KD 0.9    & \cmark \cmark \cmark & \cmark \cmark \cmark & \cmark \cmark \cmark & \cmark \cmark \cmark & \xmark \xmark \xmark & \cmark \cmark \cmark \\ \hline
\end{tabular}}
\end{table}
\begin{figure}[H]
    \centering
    {\includegraphics[width=\linewidth] {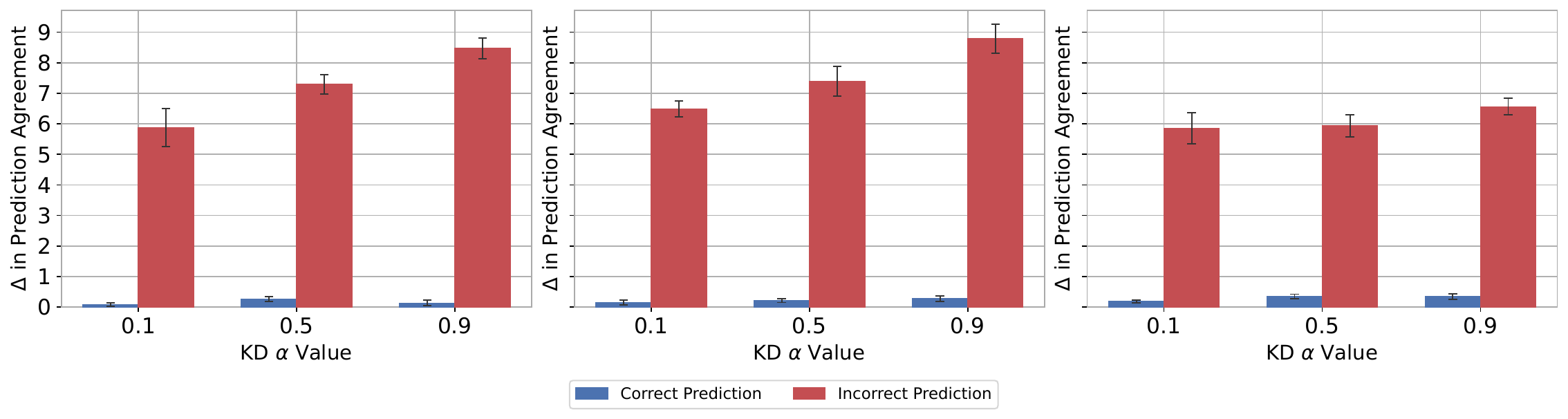}} 
    \caption{Prediction agreement difference of student models in KD to the highest performing control with respect to correct prediction agreement (blue) and incorrect prediction agreement (red), for VGGish on SpeechCommands (seeds 0 to 2, left to right).}
    \label{fig:vgg-sc-pred}
\end{figure} 
For the VGGish on SpeechCommands, Table~\ref{tab:sc-VGGISH-ts-0} shows substantially higher knowledge transfer compared to the controls than those observed in SVHN and base case for TinyImageNet. Notably, we observe an approximate $2\times$ reduction in divergence across all similarity metrics, alongside a monotonic increase in functional similarity as $\alpha$ increases.  
Table~\ref{tab:vgg_sc_sig} confirms that for all $\alpha$ values and teacher seeds, knowledge transfer is statistically supported for KD. 
The type and strength of this transfer, shown in Figure~\ref{fig:vgg-sc-pred}, reveals that the increased knowledge transfer in SpeechCommands corresponds to significant negative asymmetric transfer, further strengthening the link between statistically supported knowledge transfer via improved functional similarity and severe negative asymmetric transfer. Given, that RCD at $\alpha$ = 0.1 provides the best performance, in this instance it is clear that functional alignment and transfer is not responsible for performance benefits.
\begin{table}[H]
\centering
\caption{VGGish on UrbanSound8K: mean and $\pm$ 1 SEM  reported over 10 runs with Teacher Seed 0. \textbf{Bold} values are best performing based on the mean.}
\resizebox{\linewidth}{!}{
\begin{tabular}{|l|c|ccc|ccc|}
\hline
\multicolumn{1}{|c|}{\multirow{2}{*}{\textbf{Metrics}}} & \textbf{Control}  & \multicolumn{3}{c|}{\textbf{Knowledge Distillation}}                                                                    & \multicolumn{3}{c|}{\textbf{Random Control Distillation}}                                                         \\ \cline{2-8} 
\multicolumn{1}{|c|}{}                                  & \textbf{SIDDO}     & \multicolumn{1}{c|}{\textbf{0.1}}    & \multicolumn{1}{c|}{\textbf{0.5}}             & \textbf{0.9}            & \multicolumn{1}{c|}{\textbf{0.1}}        & \multicolumn{1}{c|}{\textbf{0.5}} & \textbf{0.9} \\ \hline
Activation Distance \textbf{($\mathbf{\downarrow}$)} & 0.256$\pm{0.005}$ & \multicolumn{1}{c|}{0.267$\pm{0.014}$} & \multicolumn{1}{c|}{\textbf{0.242$\pm{0.003}$}} & 0.243$\pm{0.005}$         & \multicolumn{1}{c|}{0.354$\pm{0.003}$}          & \multicolumn{1}{c|}{0.597$\pm{0.002}$}   & 0.873$\pm{0.000}$     \\ \hline
Rank Disagreement  \textbf{($\mathbf{\downarrow}$)} & 0.696$\pm{0.003}$ & \multicolumn{1}{c|}{0.696$\pm{0.005}$} & \multicolumn{1}{c|}{0.683$\pm{0.003}$} & \textbf{0.678$\pm{0.004}$}         & \multicolumn{1}{c|}{0.795$\pm{0.001}$}          & \multicolumn{1}{c|}{0.791$\pm{0.001}$}   & 0.784$\pm{0.002}$   \\ \hline
Prediction Disagreement  \textbf{($\mathbf{\downarrow}$)} & 0.192$\pm{0.004}$ & \multicolumn{1}{c|}{0.196$\pm{0.009}$} & \multicolumn{1}{c|}{\textbf{0.180$\pm{0.002}$}}  & \textbf{0.180$\pm{0.003}$} & \multicolumn{1}{c|}{0.187$\pm{0.002}$}          & \multicolumn{1}{c|}{0.195$\pm{0.003}$}   & 0.387$\pm{0.001}$   \\ \hline
JS Divergence  \textbf{($\mathbf{\downarrow}$)} & inf, nan    & \multicolumn{1}{c|}{inf, nan}     & \multicolumn{1}{c|}{\textbf{0.099$\pm{0.001}$}} & 0.100$\pm{0.002}$           & \multicolumn{1}{c|}{0.149$\pm{0.001}$}          & \multicolumn{1}{c|}{0.268$\pm{0.001}$}   & 0.467$\pm{0.000}$     \\ \hline
Accuracy  \textbf{($\mathbf{\uparrow}$)} & 0.795$\pm{0.003}$ & \multicolumn{1}{c|}{0.787$\pm{0.009}$} & \multicolumn{1}{c|}{0.796$\pm{0.002}$}          & 0.796$\pm{0.003}$         & \multicolumn{1}{c|}{\textbf{0.808$\pm{0.001}$}} & \multicolumn{1}{c|}{0.806$\pm{0.002}$}   & 0.585$\pm{0.001}$   \\ \hline
Loss  \textbf{($\mathbf{\downarrow}$)} & 2.813$\pm{0.330}$  & \multicolumn{1}{c|}{2.460$\pm{0.248}$}  & \multicolumn{1}{c|}{2.225$\pm{0.046}$}          & 2.089$\pm{0.103}$         & \multicolumn{1}{c|}{\textbf{0.730$\pm{0.005}$}}  & \multicolumn{1}{c|}{1.085$\pm{0.003}$}   & 2.059$\pm{0.002}$   \\ \hline
\end{tabular}}
\label{tab:vggish-ts0-sc}
\end{table}
\begin{table}[H]
\caption{VGGish on UrbanSound8K (significance testing). \cmark~indicates significant results compared to controls; \xmark~indicates insignificant results. Each tick represents a teacher (seeds 0 to 2, left to right).}
\resizebox{\linewidth}{!}{
\begin{tabular}{|l|l|l|l|l|l|l|}
\hline
\textbf{} & \textbf{Activation Distance}                                      & \textbf{Rank Disagreement}                                        & \textbf{Prediction Disagreement}                                  & \textbf{JS Divergence}                                            & \textbf{Accuracy}                                                 & \textbf{Loss}                                                     \\ \hline
KD 0.1    & \xmark \xmark \cmark & \xmark \xmark \cmark & \xmark \xmark \cmark & \xmark \xmark \cmark & \xmark \xmark \xmark & \xmark \xmark \xmark \\ \hline
KD 0.5    & \cmark \cmark \cmark & \cmark \cmark \cmark & \xmark \xmark \cmark & \cmark \xmark \cmark & \xmark \xmark \xmark & \xmark \xmark \xmark \\ \hline
KD 0.9    & \xmark \cmark \cmark & \cmark \cmark \cmark & \xmark \cmark \cmark & \cmark \xmark \cmark & \xmark \xmark \xmark & \xmark \xmark \xmark \\ \hline
\end{tabular}}
\end{table} 
\begin{figure}[H]
    \centering
    {\includegraphics[width=\linewidth] {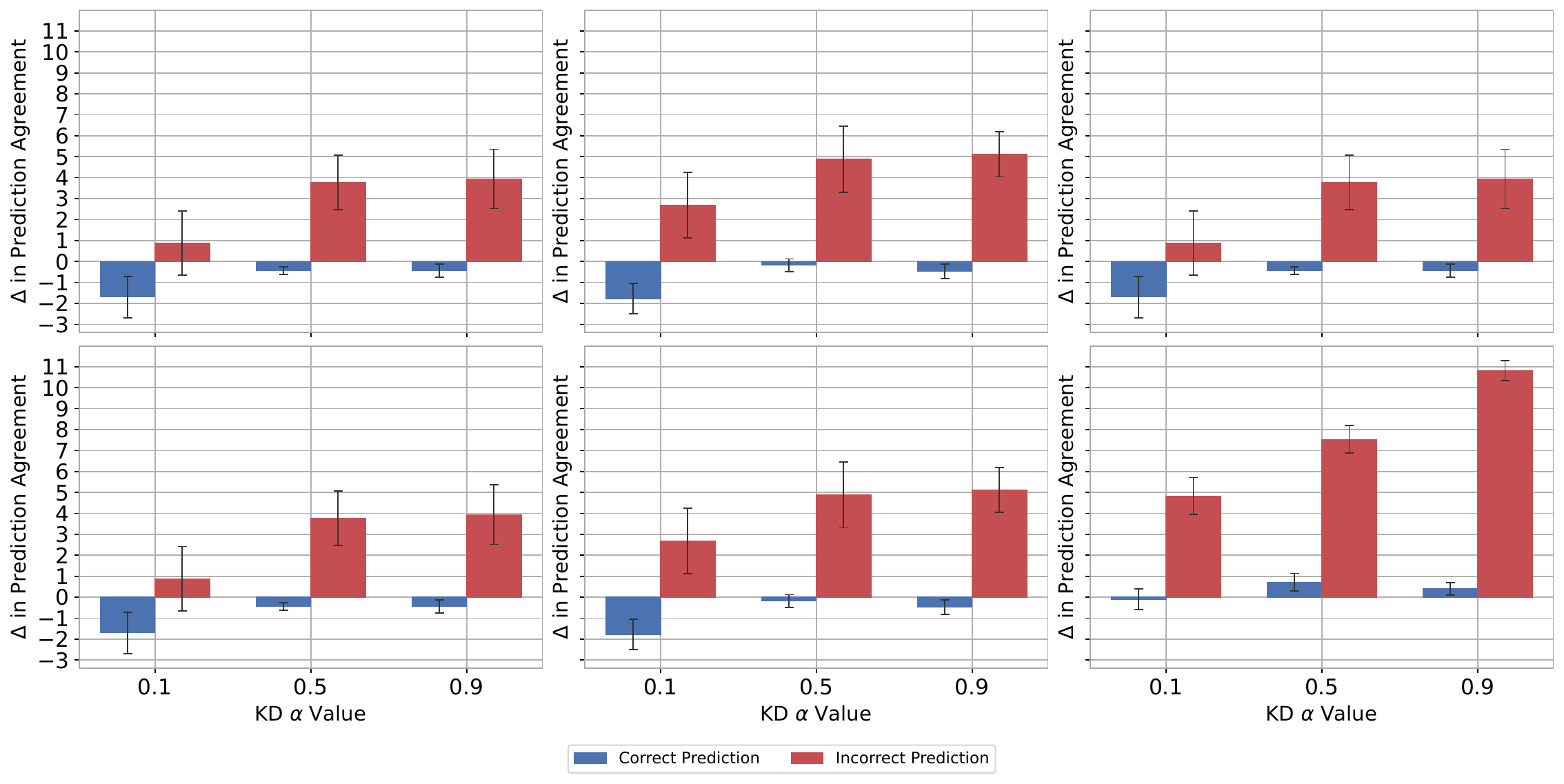}}
     \caption{Prediction agreement difference between KD students in standard KD and the highest performing control baseline for correct prediction agreement (blue) and incorrect prediction agreement (red) on UrbanSound8K with VGGish. Top uses all baselines, whereas bottom excludes RCD at $\alpha=0.9$ due to its reduced accuracy (seeds 0 to 2, left to right).}
    \label{fig:vgg-TS-prediction_RCD-main}
\end{figure}
For VGGish on UrbanSound8K, we observe less knowledge transfer relative to the baselines in Table~\ref{tab:vggish-ts0-sc} than on SpeechCommands in Table~\ref{tab:sc-VGGISH-ts-0}. Once again, RCD 0.1 provides the best overall performance, but here RCD 0.9 causes a large reduction in accuracy on this dataset. When analysing the prediction agreement distribution in Figure~\ref{fig:vgg-TS-prediction_RCD-main}, it can still be observed that negative asymmetric transfer remains the most prominent form of knowledge transfer. In this figure, we also present the prediction agreement results without RCD 0.9, since its very low accuracy biases the results toward underestimating the degree of asymmetric transfer from KD. When RCD 0.9 is removed, the degree of negative asymmetric transfer becomes more consistent with the statistically supported functional similarity we record across all similarity metrics and seeds. As a result, the findings in audio confirm that negative asymmetric transfer is prominent across modalities and that functional transfer facilitated by KD does not provide the best performance and transfers incorrect teacher  representations at a disproportionate rate.

\subsubsection{Language}
\label{subsec:language}
\begin{figure}[H]
    \centering
   \includegraphics[width=\linewidth]{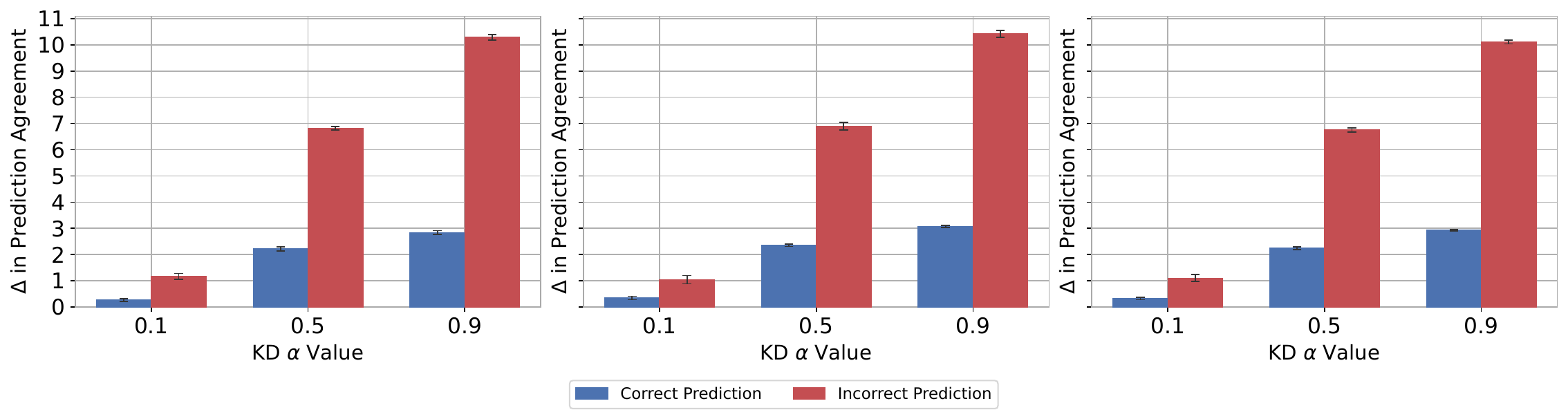}
    \caption{Prediction agreement difference of student models in standard KD to the highest performing control baseline with respect to correct prediction agreement (blue) and incorrect prediction agreement (red), for Nano-GPT on TinyShakespeare (seeds 0 to 2, left to right).}
    \label{fig:shake-gpt-signifance}
\end{figure}

The final modality we consider is language. Having observed  negative asymmetric transfer in computer vision and audio, we test whether the same phenomenon occurs in language models. In this section, we study the Nano-GPT architecture on TinyShakespeare, which is where we observe the largest statistically supported knowledge transfer across similarity metrics, coupled with the strongest negative asymmetric transfer. 
\vspace{-0.1cm}
\begin{table}[H]
\centering
\caption{Nano-GPT on TinyShakespeare: mean and $\pm$ 1 SEM  reported over 10 runs with Teacher Seed 0. \textbf{Bold} values are best performing based on the mean.}
\label{tab:shake-gpt-ts-0}
\resizebox{\textwidth}{!}{%
\begin{tabular}{|l|c|ccc|ccc|}
\hline
\multicolumn{1}{|c|}{\multirow{2}{*}{\textbf{Metrics}}} & \textbf{Control} & \multicolumn{3}{c|}{\textbf{Knowledge Distillation}} & \multicolumn{3}{c|}{\textbf{Random Control Distillation}} \\ \cline{2-8} 
\multicolumn{1}{|c|}{} & \textbf{SIDDO} & \multicolumn{1}{c|}{\textbf{0.1}} & \multicolumn{1}{c|}{\textbf{0.5}} & \textbf{0.9} & \multicolumn{1}{c|}{\textbf{0.1}} & \multicolumn{1}{c|}{\textbf{0.5}} & \textbf{0.9} \\ \hline
Activation Distance \textbf{($\mathbf{\downarrow}$)} & 0.196$\pm{0.000}$ & \multicolumn{1}{c|}{0.187$\pm{0.000}$} & \multicolumn{1}{c|}{0.158$\pm{0.000}$} & \textbf{0.144$\pm{0.000}$} & \multicolumn{1}{c|}{0.204$\pm{0.000}$} & \multicolumn{1}{c|}{0.378$\pm{0.001}$} & 0.661$\pm{0.000}$ \\ \hline
Rank Disagreement  \textbf{($\mathbf{\downarrow}$)} & 0.910$\pm{0.000}$ & \multicolumn{1}{c|}{0.907$\pm{0.000}$} & \multicolumn{1}{c|}{0.897$\pm{0.000}$} & \textbf{0.891$\pm{0.000}$} & \multicolumn{1}{c|}{0.944$\pm{0.000}$} & \multicolumn{1}{c|}{0.947$\pm{0.000}$} & 0.950$\pm{0.000}$ \\ \hline
Prediction Disagreement  \textbf{($\mathbf{\downarrow}$)} & 0.246$\pm{0.001}$ & \multicolumn{1}{c|}{0.236$\pm{0.000}$} & \multicolumn{1}{c|}{0.200$\pm{0.000}$} & \textbf{0.182$\pm{0.000}$} & \multicolumn{1}{c|}{0.242$\pm{0.001}$} & \multicolumn{1}{c|}{0.243$\pm{0.001}$} & 0.255$\pm{0.001}$ \\ \hline
JS Divergence  \textbf{($\mathbf{\downarrow}$)} & 0.053$\pm{0.000}$ & \multicolumn{1}{c|}{0.049$\pm{0.000}$} & \multicolumn{1}{c|}{0.037$\pm{0.000}$} & \textbf{0.032$\pm{0.000}$} & \multicolumn{1}{c|}{0.067$\pm{0.000}$} & \multicolumn{1}{c|}{0.192$\pm{0.000}$} & 0.449$\pm{0.000}$ \\ \hline
Accuracy  \textbf{($\mathbf{\uparrow}$)} & 0.574$\pm{0.000}$ & \multicolumn{1}{c|}{0.577$\pm{0.000}$} & \multicolumn{1}{c|}{\textbf{0.583$\pm{0.000}$}} & 0.581$\pm{0.000}$ & \multicolumn{1}{c|}{0.576$\pm{0.000}$} & \multicolumn{1}{c|}{0.578$\pm{0.000}$} & 0.570$\pm{0.000}$ \\ \hline
Loss  \textbf{($\mathbf{\downarrow}$)} & 1.559$\pm{0.002}$ & \multicolumn{1}{c|}{1.542$\pm{0.002}$} & \multicolumn{1}{c|}{\textbf{1.496$\pm{0.001}$}} & 1.500$\pm{0.002}$ & \multicolumn{1}{c|}{1.507$\pm{0.001}$} & \multicolumn{1}{c|}{1.839$\pm{0.002}$} & 2.995$\pm{0.001}$ \\ \hline
\end{tabular}%
}
\end{table}
For NanoGPT, KD provides statistically supported increases in functional similarity across all seeds, alongside improvements in accuracy and loss (Table~\ref{tab:nanogpt-shakespeare-significance}). This shows that the RCD control does not yield the best performing model across all modalities. When considering the raw values for seed 0 (Table~\ref{tab:shake-gpt-ts-0}), KD yields a large deviation from the control baselines that scales with $\alpha$; however, this is primarily driven by the largest error transfer that is recorded across all modalities (Figure~\ref{fig:shake-gpt-signifance}). The maximum negative transfer is 10\%, approximately $4\times$ larger than the positive transfer of just above 2\%. 
Overall, these results confirm that across all major data modalities and architectures, negative asymmetric transfer is a prominent feature of KD: when knowledge transfer occurs, student capacity is preferentially used to match the teacher's error patterns rather than its beneficial representations. In addition, we often observe that the RCD control can outperform KD, motivating stronger counterfactual baselines when interpreting KD gains.
\begin{table}[H]
    \centering
    \caption{Nano-GPT on TinyShakespeare (significance testing). \cmark~indicates significant results compared to controls; \xmark~indicates insignificant results. Each tick represents a teacher (seeds 0 to 2, left to right).}
    \label{tab:nanogpt-shakespeare-significance}
    \resizebox{\textwidth}{!}{%
    \begin{tabular}{|c|c|c|c|c|c|c|}
    \hline
     & Activation Distance & Rank Disagreement & Prediction Disagreement & JS Divergence & Accuracy & Loss \\ \hline
     KD 0.1&\cmark \cmark \cmark &\cmark \cmark \cmark &\cmark \cmark \cmark &\cmark \cmark \cmark &\xmark \xmark \xmark &\xmark \xmark \xmark \\ \hline
     KD 0.5&\cmark \cmark \cmark &\cmark \cmark \cmark &\cmark \cmark \cmark &\cmark \cmark \cmark &\cmark \cmark \cmark &\cmark \cmark \cmark \\ \hline
     KD 0.9&\cmark \cmark \cmark &\cmark \cmark \cmark &\cmark \cmark \cmark &\cmark \cmark \cmark &\cmark \cmark \cmark &\xmark \cmark \cmark \\ \hline
    \end{tabular}%
    }
\end{table} 

\subsection{Standard Distillation: The Impact of Student Capacity on Negative Asymmetric Transfer}
\label{sec:standard-kd-ss}

Our previous experiments in self-distillation provided an upper bound on knowledge transfer, revealing that negative asymmetric transfer often dominates. Now we study traditional knowledge distillation, where the student has less capacity than the teacher.
Although this departs from our original setup, where the student can fully match the teacher, we use this setting to examine how transfer behaves between a larger teacher and a smaller student. A smaller student introduces uncertainty about whether limited capacity constrains knowledge transfer. However, knowledge distillation is typically used in this context, and to assess the generality of negative asymmetric transfer it is important to analyse the nature of knowledge transfer in this setting. 

Our results for standard distillation show that our core insights from self-distillation extend to other distillation settings. These findings also reinforce the value of our initial setup for understanding knowledge transfer in KD. Apart from the student architecture's implicit bias, which affects performance, there are no additional confounding factors influencing knowledge distillation in our experimental setup for standard distillation.

In the previous section, we demonstrated the generality of negative transfer across modalities in KD. Accordingly, here we focus on two common deep learning modalities: vision and language. We present results in Sections~\ref{subsec:cv-skd} and~\ref{subsec:language-skd}. These modalities exhibit distinct differences in both the proportion of knowledge transferred via KD and, consequently, the extent of negative asymmetric transfer. In vision, there is often limited knowledge transfer, apart from when teacher loss is high through the use of regularisation such as augmentation. In language, we observe statistically supported knowledge transfer alongside negative asymmetric transfer. In Section~\ref{subsec:cv-skd}, we present results with ResNet50 teachers and ResNet18 students on Tiny ImageNet and ImageNet. In Section~\ref{subsec:language-skd}, we present results on Tinyshakespeare using Nano-GPT as the teacher and Pico-GPT as the student. Finally, in Section~\ref{subsec-scalling-laws}, we explore scaling laws for negative asymmetric transfer in KD as student capacity is gradually increased to have a granular insight on the emergence of knowledge transfer asymmetries. 

\subsubsection{Computer Vision}
\label{subsec:cv-skd}

\begin{figure}[H]
    \centering
    \includegraphics[width=\linewidth]{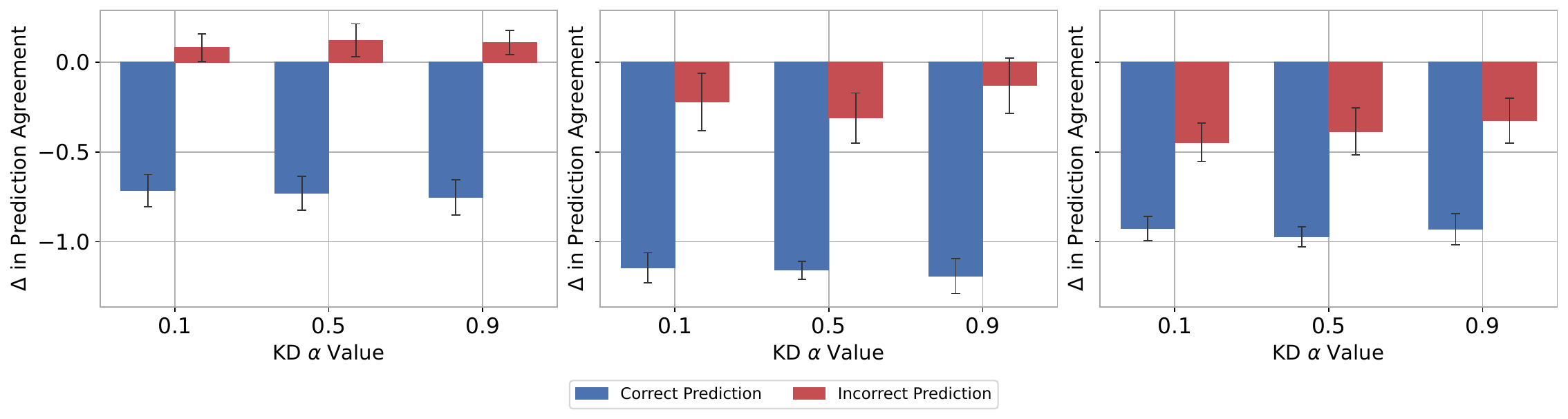}
    \caption{Prediction agreement difference of student models in standard KD to the highest performing control baseline with respect to correct prediction agreement (blue) and incorrect prediction agreement (red), for ResNet18 with a ResNet50 teacher on TinyImageNet (seeds 0 to 2, left to right).}
    \label{fig:resnet-tin-prediction}
\end{figure}

\paragraph{TinyImageNet} For this experiment, the ResNet50 teacher model was trained with stochastic gradient descent with a learning rate of 0.01 and a cosine annealing learning rate scheduler with T\_max set to 100. It was trained for 100 epochs with a batch size of 256. The data was normalised with a mean of (0.485, 0.456, 0.406) and a standard deviation of (0.229, 0.224, 0.225). The ResNet18 student model was trained under the same conditions. 
Overall, we observe a low train loss for the teacher model (circa 0.0014) and a high train accuracy (0.9998); see Appendix Table~\ref{tab:resnet50-tin-small-teacher}. This low train loss corresponds to no statistically supported knowledge transfer across $\alpha$ values; see Table~\ref{tab:tin-resnet18-ts-0}, Appendix Table~\ref {tab:tin-resnet18-ts-1}, \ref{tab:tin-resnet18-ts-2}, and~Table~\ref{tab:resnet18-tin-significance}. This result is consistent with the previous results and the intuition from the self-distillation experiment. 
\begin{table}[H]
\centering
\caption{ResNet18 with a ResNet50 teacher on TinyImageNet: mean and $\pm$ 1 SEM  reported over 10 runs with Teacher Seed 0. Bold values are best performing based on the mean. }
\label{tab:tin-resnet18-ts-0}
\resizebox{\textwidth}{!}{%
\begin{tabular}{|c|c|ccc|ccc|}
\hline
\multicolumn{1}{|c|}{\multirow{2}{*}{\textbf{Metrics}}} & \multicolumn{1}{c|}{\textbf{Control}} & \multicolumn{3}{c|}{\textbf{Knowledge Distillation}} & \multicolumn{3}{c|}{\textbf{Random Control Distillation}} \\ \cline{2-8} 
\multicolumn{1}{|c|}{} & \multicolumn{1}{c|}{\textbf{SIDDO}} & \multicolumn{1}{c|}{\textbf{0.1}} & \multicolumn{1}{c|}{\textbf{0.5}} & \multicolumn{1}{c|}{\textbf{0.9}} & \multicolumn{1}{c|}{\textbf{0.1}} & \multicolumn{1}{c|}{\textbf{0.5}} & \multicolumn{1}{c|}{\textbf{0.9}} \\ \hline
Activation Distance & 0.548 $\pm{}$ 0.000 & \multicolumn{1}{c|}{0.548 $\pm{}$ 0.000} & \multicolumn{1}{c|}{\textbf{0.547 $\pm{}$ 0.000}} & \textbf{0.547 $\pm{}$ 0.000} & \multicolumn{1}{c|}{0.565 $\pm{}$ 0.000} & \multicolumn{1}{c|}{0.651 $\pm{}$ 0.000} & 0.828 $\pm{}$ 0.000 \\ \hline
Rank Disagreement &\textbf{ 0.987 $\pm{}$ 0.000} & \multicolumn{1}{c|}{\textbf{0.987 $\pm{}$ 0.000}} & \multicolumn{1}{c|}{\textbf{0.987 $\pm{}$ 0.000}} & \textbf{0.987 $\pm{}$ 0.000} & \multicolumn{1}{c|}{0.990 $\pm{}$ 0.000} & \multicolumn{1}{c|}{0.990 $\pm{}$ 0.000} & 0.991 $\pm{}$ 0.000 \\ \hline
Prediction Disagreement & 0.498 $\pm{}$ 0.001 & \multicolumn{1}{c|}{0.497 $\pm{}$ 0.000} & \multicolumn{1}{c|}{0.497 $\pm{}$ 0.001} & 0.497 $\pm{}$ 0.000 & \multicolumn{1}{c|}{0.512 $\pm{}$ 0.001} & \multicolumn{1}{c|}{\textbf{0.493 $\pm{}$ 0.001}} & 0.754 $\pm{}$ 0.000 \\ \hline
JS Divergence & 0.281 $\pm{}$ 0.000 & \multicolumn{1}{c|}{0.281 $\pm{}$ 0.000} & \multicolumn{1}{c|}{\textbf{0.280 $\pm{}$ 0.000}} & 0.281 $\pm{}$ 0.000 & \multicolumn{1}{c|}{0.330 $\pm{}$ 0.000} & \multicolumn{1}{c|}{0.400 $\pm{}$ 0.000} & 0.599 $\pm{}$ 0.000 \\ \hline
Accuracy & 0.503 $\pm{}$ 0.001 & \multicolumn{1}{c|}{0.504 $\pm{}$ 0.001} & \multicolumn{1}{c|}{0.504 $\pm{}$ 0.000} & 0.503 $\pm{}$ 0.000 & \multicolumn{1}{c|}{0.493 $\pm{}$ 0.000} & \multicolumn{1}{c|}{\textbf{0.512 $\pm{}$ 0.000}} & 0.236 $\pm{}$ 0.000 \\ \hline
Loss & 2.604 $\pm{}$ 0.001 & \multicolumn{1}{c|}{2.602 $\pm{}$ 0.002} & \multicolumn{1}{c|}{2.594 $\pm{}$ 0.001} & 2.589 $\pm{}$ 0.001 & \multicolumn{1}{c|}{\textbf{2.434 $\pm{}$ 0.001}} & \multicolumn{1}{c|}{2.641 $\pm{}$ 0.001} & 4.684 $\pm{}$ 0.002 \\ \hline
\end{tabular}%
}
\end{table}
From Table~\ref{tab:tin-resnet18-ts-0}, we observe that, compared to the SIDDO control, there is minimal functional similarity gain from using knowledge distillation across all similarity metrics. This indicates partial to no knowledge transfer, even when using a larger teacher. Furthermore, the best accuracy and loss are recorded for the RCD condition, highlighting that the regularisation effects of the KD objective can be more important for performance than knowledge transfer. Table~\ref{tab:resnet18-tin-significance} shows that knowledge transfer is largely not statistically supported across the majority of metrics when using the ResNet50 teacher with the ResNet18 student. 
\begin{table}[H]
\centering
\caption{ResNet18 with a ResNet50 teacher on TinyImagenet (significance testing). \cmark~indicates significant results compared to controls; \xmark~indicates insignificant results. Each tick represents a teacher (seeds 0 to 2, left to right).}
\label{tab:resnet18-tin-significance}
\resizebox{\textwidth}{!}{
\begin{tabular}{|c|c|c|c|c|c|c|}
\hline
\textbf{} & \textbf{Activation Distance}                                      & \textbf{Rank Disagreement}                                        & \textbf{Prediction Disagreement}                                  & \textbf{JS Divergence}                                            & \textbf{Accuracy}                                                 & \textbf{Loss}                                                     \\ \hline
KD 0.1    & \xmark \xmark \xmark & \xmark \xmark \xmark & \xmark \xmark \xmark & \xmark \xmark \xmark & \xmark \xmark \xmark & \xmark \xmark \xmark \\ \hline
KD 0.5    & \cmark \xmark \xmark & \xmark \xmark \xmark & \xmark \xmark \xmark & \cmark \cmark \xmark & \xmark \xmark \xmark & \xmark \xmark \xmark \\ \hline
KD 0.9    & \xmark \cmark \xmark & \xmark \xmark \xmark & \xmark \xmark \xmark & \cmark \cmark \cmark & \xmark \xmark \xmark & \xmark \xmark \xmark \\ \hline
\end{tabular}}
\end{table}
Analysing prediction agreement in Figure~\ref{fig:resnet-tin-prediction}, we find that the student and teacher agree on fewer predictions than the baselines. Given the very limited functional transfer for this student--teacher pairing, this aligns with the expected low prediction agreement. Notably, although incorrect prediction agreement is not statistically supported, the student and teacher agree on incorrect predictions more than on correct predictions. This indicates that the small amount of knowledge transferred is still shaped by the negative asymmetric transfer highlighted earlier.

\paragraph{ImageNet}

\begin{figure}[H]
    \centering
    \includegraphics[width=0.4\linewidth]{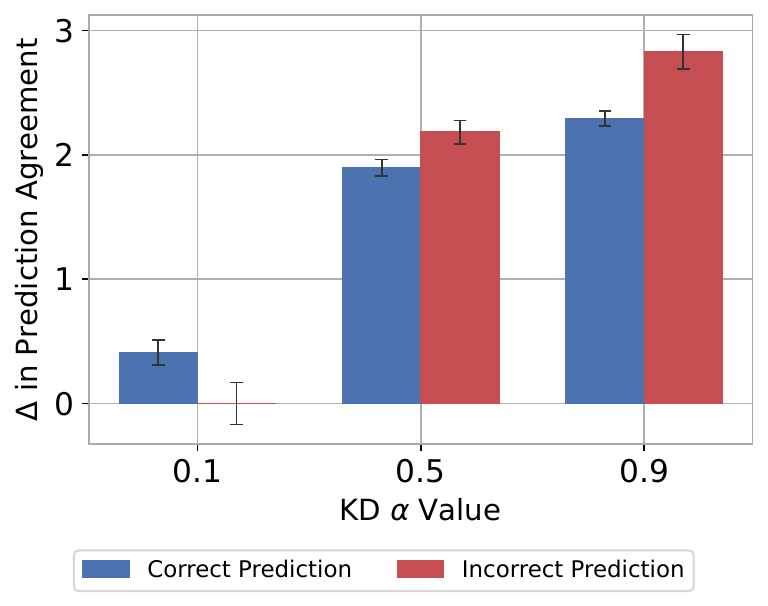}
    \caption{Prediction agreement difference of student models in standard KD to the highest performing control baseline with respect to correct prediction agreement (blue) and incorrect prediction agreement (red), for ResNet18 with a ResNet50 teacher on ImageNet.}
    \label{fig:resnet-img-prediction}
\end{figure}

For this experiment, we use ImageNet with a ResNet50 teacher and a ResNet18 student (training details in Appendix \ref{app:imagenet}). For this dataset, we observe increases in functional similarity compared to the baselines in Table~\ref{tab:imagnet-resnet18-temp-1-ts-0}. However, only at $\alpha$ at 0.9 do we obtain statistically supported transfer (Table~\ref{tab:resnet-img-significance}) across the majority of functional similarity metrics. In line with our existing results, when there is statistically supported knowledge transfer from the teacher to the student, negative asymmetric transfer occurs with a bias towards teacher errors (Figure \ref{fig:resnet-img-prediction}). Interestingly, while there is statistically supported transfer, for Rank disagreement the values recorded are very close in value and high indicating that for datasets with large classes rank order is more stochastic, which is somewhat expected. We record the largest relative deviations  from the baselines for activation distance and prediction disagreement for this teacher seed. Largely, these results provide additional evidence that even on high complexity datasets such as ImageNet, negative asymmetric transfer remains prominent and increases with greater function transfer. 

\begin{table}[H]
\caption{ResNet18 with a ResNet50 teacher on ImageNet: mean and $\pm$ 1 SEM  reported over 10 runs with Teacher Seed 0. Bold values are best performing based on the mean.}
\label{tab:imagnet-resnet18-temp-1-ts-0}
\resizebox{\textwidth}{!}{%
\begin{tabular}{|c|c|ccc|ccc|}
\hline
\multicolumn{1}{|c|}{\multirow{2}{*}{\textbf{Metrics}}} & \multicolumn{1}{c|}{\textbf{Control}} & \multicolumn{3}{c|}{\textbf{Knowledge Distillation}}                                                              & \multicolumn{3}{c|}{\textbf{Random Control Distillation}}                                                       \\ \cline{2-8} 
\multicolumn{1}{|c|}{}                                  & \textbf{SIDDO}                        & \multicolumn{1}{c|}{\textbf{0.1}}          & \multicolumn{1}{c|}{\textbf{0.5}}          & \textbf{0.9}            & \multicolumn{1}{c|}{\textbf{0.1}}          & \multicolumn{1}{c|}{\textbf{0.5}}          & \textbf{0.9}          \\ \hline
Activation Distance                                     & 0.420 $\pm{}$ 0.001                         & \multicolumn{1}{c|}{0.365 $\pm{}$ 0.001}        & \multicolumn{1}{c|}{0.260 $\pm{}$ 0.001}         & \textbf{0.226 $\pm{}$ 0.000}   & \multicolumn{1}{c|}{0.268 $\pm{}$ 0.001}        & \multicolumn{1}{c|}{0.259 $\pm{}$ 0.002}        & 0.376 $\pm{}$ 0.000          \\ \hline
Rank Disagreement                                       & \textbf{0.997 $\pm{}$ 0.000}                 & \multicolumn{1}{c|}{\textbf{0.997 $\pm{}$ 0.000}} & \multicolumn{1}{c|}{\textbf{0.997 $\pm{}$ 0.000}} & \textbf{0.997 $\pm{}$ 0.000}   & \multicolumn{1}{c|}{\textbf{0.997 $\pm{}$ 0.000}} & \multicolumn{1}{c|}{\textbf{0.997 $\pm{}$ 0.000}} & \textbf{0.997 $\pm{}$ 0.000} \\ \hline
Prediction Disagreement                                 & 0.264 $\pm{}$ 0.003                        & \multicolumn{1}{c|}{0.256 $\pm{}$ 0.002}        & \multicolumn{1}{c|}{0.239 $\pm{}$ 0.002}        & \textbf{0.235 $\pm{}$ 0.002} & \multicolumn{1}{c|}{0.259 $\pm{}$ 0.001}        & \multicolumn{1}{c|}{0.274 $\pm{}$ 0.002}        & 0.308 $\pm{}$ 0.002        \\ \hline
JS Divergence                                           & 0.260 $\pm{}$ 0.001                         & \multicolumn{1}{c|}{0.221 $\pm{}$ 0.001}        & \multicolumn{1}{c|}{0.136 $\pm{}$ 0.001}        & 0.106 $\pm{}$ 0.000   & \multicolumn{1}{c|}{0.136 $\pm{}$ 0.001}        & \multicolumn{1}{c|}{\textbf{0.099 $\pm{}$ 0.001}}        & 0.173 $\pm{}$ 0.001        \\ \hline
Accuracy                                                & 0.680 $\pm{}$ 0.002                         & \multicolumn{1}{c|}{0.687 $\pm{}$ 0.002}        & \multicolumn{1}{c|}{0.700 $\pm{}$ 0.001}          & \textbf{0.703 $\pm{}$ 0.002} & \multicolumn{1}{c|}{0.684 $\pm{}$ 0.001}        & \multicolumn{1}{c|}{0.670 $\pm{}$ 0.001}         & 0.642 $\pm{}$ 0.002        \\ \hline
Loss                                                    & \textbf{1.307 $\pm{}$ 0.009}               & \multicolumn{1}{c|}{1.342 $\pm{}$ 0.009}        & \multicolumn{1}{c|}{1.608 $\pm{}$ 0.015}        & 1.833 $\pm{}$ 0.022          & \multicolumn{1}{c|}{1.657 $\pm{}$ 0.013}        & \multicolumn{1}{c|}{2.548 $\pm{}$ 0.017}        & 4.060 $\pm{}$ 0.012         \\ \hline
\end{tabular}%
}
\end{table}
\begin{table}[H]
\caption{ResNet18 with a ResNet50 teacher on Imagenet (significance testing). \cmark~indicates significant results compared to controls; \xmark~indicates insignificant results.}
\label{tab:resnet-img-significance}
\resizebox{\textwidth}{!}{
\begin{tabular}{|c|c|c|c|c|c|c|}
\hline
\textbf{} & \textbf{Activation Distance} & \textbf{Rank Disagreement} & \textbf{Prediction Disagreement} & \textbf{JS Divergence} & \textbf{Accuracy}     & \textbf{Loss}         \\ \hline
KD 0.1    & \xmark        & \xmark      & \xmark            & \xmark  & \cmark & \xmark \\ \hline
KD 0.5    & \xmark        & \cmark      & \cmark            & \xmark  & \cmark & \xmark \\ \hline
KD 0.9    & \cmark        & \cmark      & \cmark            & \xmark  & \cmark & \xmark \\ \hline
\end{tabular}}
\end{table}

\subsubsection{Language}
\label{subsec:language-skd}
The Nano-GPT teacher is a GPT2-style transformer configured with an embedding dimension of 384, a vocabulary size of 65, six attention heads, six transformer blocks, a dropout of 0.2, and a block size of 256. The Pico-GPT student differs by using a smaller embedding dimension of 192, halving the internal model width, while keeping all other hyperparameters identical to the teacher. The teacher and student are trained on TinyShakespeare, using the first 90\% of the dataset for training and the last 10 \% for testing. The dataset is tokenised with a character tokeniser, and models are trained autoregressively to predict the next character token. Training uses the Adam optimiser with a learning rate of $3\times10^{-4}$, and a batch size of 64, for 5000 iterations. The student models share seeds and data orders from seeds 10 to 19 for the 10 runs used in averaging. This process is repeated for each of the three teachers trained with seeds 0 to 2. 
\begin{table}[H]
\centering
\caption{Pico-GPT with a Nano-GPT teacher on TinyShakespeare: mean and $\pm$ 1 SEM  reported over 10 runs with Teacher Seed 0. Bold values are best performing based on the mean. }
\label{tab:shake-micro-gpt-ts-0}
\resizebox{\textwidth}{!}{%
\begin{tabular}{|c|c|ccc|ccc|}
\hline
\multicolumn{1}{|c|}{\multirow{2}{*}{\textbf{Metrics}}} & \multicolumn{1}{c|}{\textbf{Control}} & \multicolumn{3}{c|}{\textbf{Knowledge Distillation}} & \multicolumn{3}{c|}{\textbf{Random Control Distillation}} \\ \cline{2-8} 
\multicolumn{1}{|c|}{} & \multicolumn{1}{c|}{\textbf{SIDDO}} & \multicolumn{1}{c|}{\textbf{0.1}} & \multicolumn{1}{c|}{\textbf{0.5}} & \multicolumn{1}{c|}{\textbf{0.9}} & \multicolumn{1}{c|}{\textbf{0.1}} & \multicolumn{1}{c|}{\textbf{0.5}} & \multicolumn{1}{c|}{\textbf{0.9}} \\ \hline
Activation Distance & 0.202 $\pm{}$ 0.000 & \multicolumn{1}{c|}{0.198 $\pm{}$ 0.000} & \multicolumn{1}{c|}{0.181 $\pm{}$ 0.000} & \textbf{0.172 $\pm{}$ 0.000} & \multicolumn{1}{c|}{0.221 $\pm{}$ 0.000} & \multicolumn{1}{c|}{0.399 $\pm{}$ 0.000} & 0.663 $\pm{}$ 0.000 \\ \hline
Rank Disagreement & 0.915 $\pm{}$ 0.000 & \multicolumn{1}{c|}{0.915 $\pm{}$ 0.000} & \multicolumn{1}{c|}{0.912 $\pm{}$ 0.000} & \textbf{0.911 $\pm{}$ 0.000} & \multicolumn{1}{c|}{0.939 $\pm{}$ 0.000} & \multicolumn{1}{c|}{0.944 $\pm{}$ 0.000} & 0.950 $\pm{}$ 0.000 \\ \hline
Prediction Disagreement & 0.252 $\pm{}$ 0.000 & \multicolumn{1}{c|}{0.247 $\pm{}$ 0.000} & \multicolumn{1}{c|}{0.226 $\pm{}$ 0.000} & \textbf{0.214 $\pm{}$ 0.000} & \multicolumn{1}{c|}{0.252 $\pm{}$ 0.000} & \multicolumn{1}{c|}{0.253 $\pm{}$ 0.001} & 0.272 $\pm{}$ 0.001 \\ \hline
JS Divergence & 0.056 $\pm{}$ 0.000 & \multicolumn{1}{c|}{0.054 $\pm{}$ 0.000} & \multicolumn{1}{c|}{0.047 $\pm{}$ 0.000} & \textbf{0.043 $\pm{}$ 0.000} & \multicolumn{1}{c|}{0.075 $\pm{}$ 0.000} & \multicolumn{1}{c|}{0.203 $\pm{}$ 0.000} & 0.451 $\pm{}$ 0.000 \\ \hline
Accuracy & 0.571 $\pm{}$ 0.000 & \multicolumn{1}{c|}{0.572 $\pm{}$ 0.000} & \multicolumn{1}{c|}{\textbf{0.575 $\pm{}$ 0.000}} & {0.574 $\pm{}$ 0.000} & \multicolumn{1}{c|}{0.571 $\pm{}$ 0.000} & \multicolumn{1}{c|}{0.570 $\pm{}$ 0.000} & 0.561 $\pm{}$ 0.000 \\ \hline
Loss & 1.473 $\pm{}$ 0.002 & \multicolumn{1}{c|}{\textbf{1.471 $\pm{}$ 0.002}} & \multicolumn{1}{c|}{1.472 $\pm{}$ 0.001} & 1.496 $\pm{}$ 0.002 & \multicolumn{1}{c|}{1.483 $\pm{}$ 0.001} & \multicolumn{1}{c|}{1.870 $\pm{}$ 0.001} & 3.017 $\pm{}$ 0.002 \\ \hline
\end{tabular}%
}
\end{table}
\begin{table}[H]
\centering
\caption{Pico-GPT with a Nano-GPT teacher on TinyShakespeare (significance testing). \cmark~indicates significant results compared to controls; \xmark~indicates insignificant results. Each tick represents a teacher (seeds 0 to 2, left to right).}
\label{tab:microgpt-shake-significance}
\resizebox{\textwidth}{!}{%
\begin{tabular}{|c|c|c|c|c|c|c|}
\hline
\textbf{} & \textbf{Activation Distance} & \textbf{Rank Disagreement} & \textbf{Prediction Disagreement} & \textbf{JS Divergence} & \textbf{Accuracy} & \textbf{Loss}     \\ \hline
KD 0.1&\cmark \cmark \cmark &\cmark \cmark \cmark &\cmark \cmark \cmark &\cmark \cmark \cmark &\cmark \cmark \xmark &\xmark \xmark \xmark \\ \hline
KD 0.5&\cmark \cmark \cmark &\cmark \cmark \cmark &\cmark \cmark \cmark &\cmark \cmark \cmark &\cmark \cmark \cmark &\xmark \xmark \xmark \\ \hline
KD 0.9&\cmark \cmark \cmark &\cmark \cmark \cmark &\cmark \cmark \cmark &\cmark \cmark \cmark &\cmark \cmark \cmark &\xmark \xmark \xmark \\ \hline
\end{tabular}%
}
\end{table}
\begin{figure}[H]
    \centering
    \includegraphics[width=\linewidth]{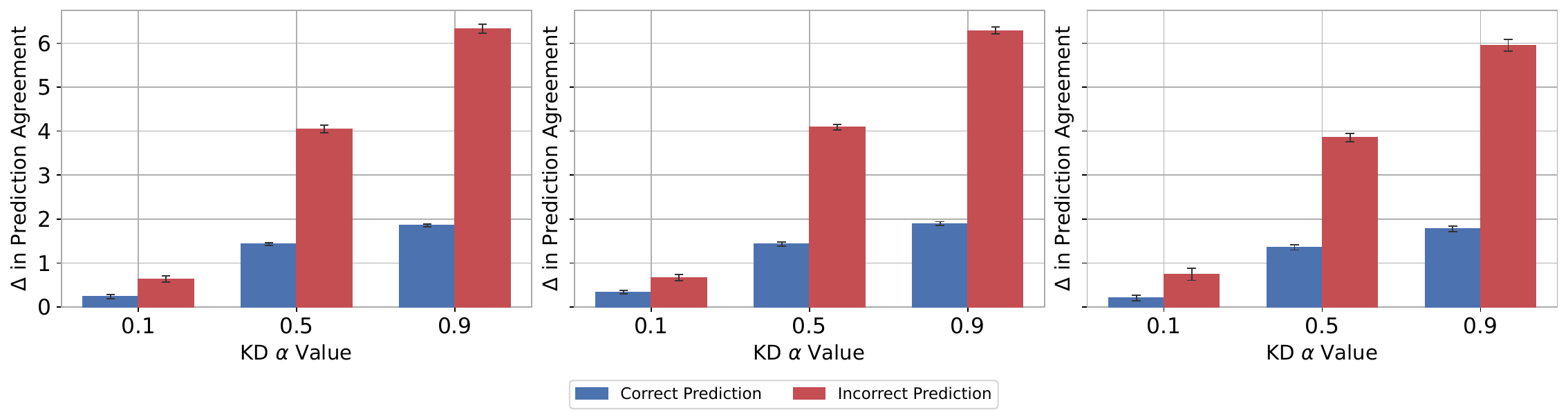}
    \caption{Prediction agreement difference of student models in standard KD to the highest performing control baseline with respect to correct prediction agreement (blue) and incorrect prediction agreement (red), for Pico-GPT with a Nano-GPT teacher on TinyShakespeare (seeds 0 to 2, left to right).}
    \label{fig:pico-gpt-shake-prediction-main}
\end{figure}
We observe a train loss for the teacher model of circa 0.86 with a high train accuracy (circa 0.72); see Appendix Table \ref{tab:gpt-shakespeare-small-teacher}. We observe that this train loss corresponds to substantial knowledge transfer, which increases with $\alpha$; see Tables \ref{tab:shake-micro-gpt-ts-0} (Appendix Tables \ref{tab:shake-micro-gpt-ts-1}, \ref{tab:shake-micro-gpt-ts-2}) and Table~\ref{tab:microgpt-shake-significance}. This substantial knowledge transfer coincides with an asymmetric payoff in prediction agreement that strongly favours incorrect predictions (Figure \ref{fig:pico-gpt-shake-prediction-main}). These results match those in Section~\ref{sec:self-distillation} and show that negative asymmetric transfer generalises to standard KD setups while showing that student capacity, as expected, limits the proportion of functional transfer.

\subsubsection{Negative Transfer Scaling Laws}
\label{subsec-scalling-laws}
\begin{wrapfigure}{R}{0.5\textwidth}
    \centering
\includegraphics[width=0.7\linewidth]{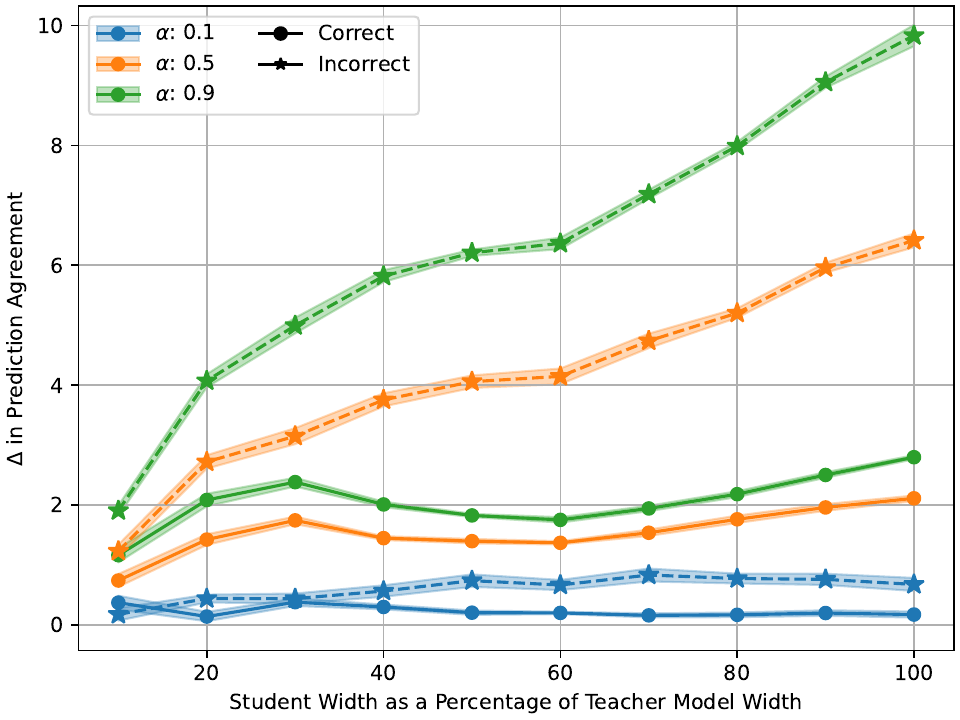}
    \caption{Transfer with student width.}
    \label{dsl_main}
\end{wrapfigure}
The previous sections established that when KD transfers knowledge, this transfer is negatively asymmetric. We now ask precisely \emph{how these effects evolve with capacity}. 
Distillation Scaling Laws (DSL)~\citep{busbridge2025distillation} quantify {how much} student loss changes with compute, teacher quality, and model size. Our study complements DSL by asking how much is transferred as capacity grows: we decompose the distillation signal into correct vs.\ incorrect teacher--student agreement, offering a mechanistic interpretation of the ``teacher quality" term and revealing negative-transfer regimes that are invisible from loss alone. Concretely, on TinyShakespeare we sweep student width from $100\%$ to $10\%$ in $10\%$ steps under a fixed-epoch budget matched to the teacher, using the same optimiser. For each width and $\alpha\in\{0.1,0.5,0.9\}$, we measure the change in correct and incorrect agreement relative to the best control baseline (means $\pm$\,SEM over 10 runs; teacher seed 0). Figure~\ref{dsl_main} shows three core trends: 

1) \textbf{Student capacity helps, but mainly by amplifying the teacher's mistakes}: as width increases, both correct and incorrect agreement rise, but incorrect agreement grows much faster (from $10\%$ to $100\%$ width at $\alpha=0.9$, correct agreement $\sim2.4\times$ vs.\ incorrect $>5\times$). 

2) \textbf{Small students suffer negative transfer}: at $10$-$20\%$ width, incorrect agreement is comparable to, or larger than, correct agreement.

3) \textbf{Increasing capacity unlocks more of the distillation signal}: however, what flows first, and most strongly, is the teacher's error pattern. 

Taken together, these scaling results calrify what drives the loss curves: KD acts as a data-dependent regulariser with a negative asymmetric payoff, and scaling up the student amplifies the asymmetry of knowledge transfer. As a result, we do not find that student capacity generally impacts the presence of negative asymmetric transfer in regimes where increased functional similarity occurs. 

\subsection{KD Variants: Feature-Map Matching Distillation}
\label{sec:feature_map_matching}
\begin{figure}[H]
    \centering
    \includegraphics[width=0.8\linewidth]{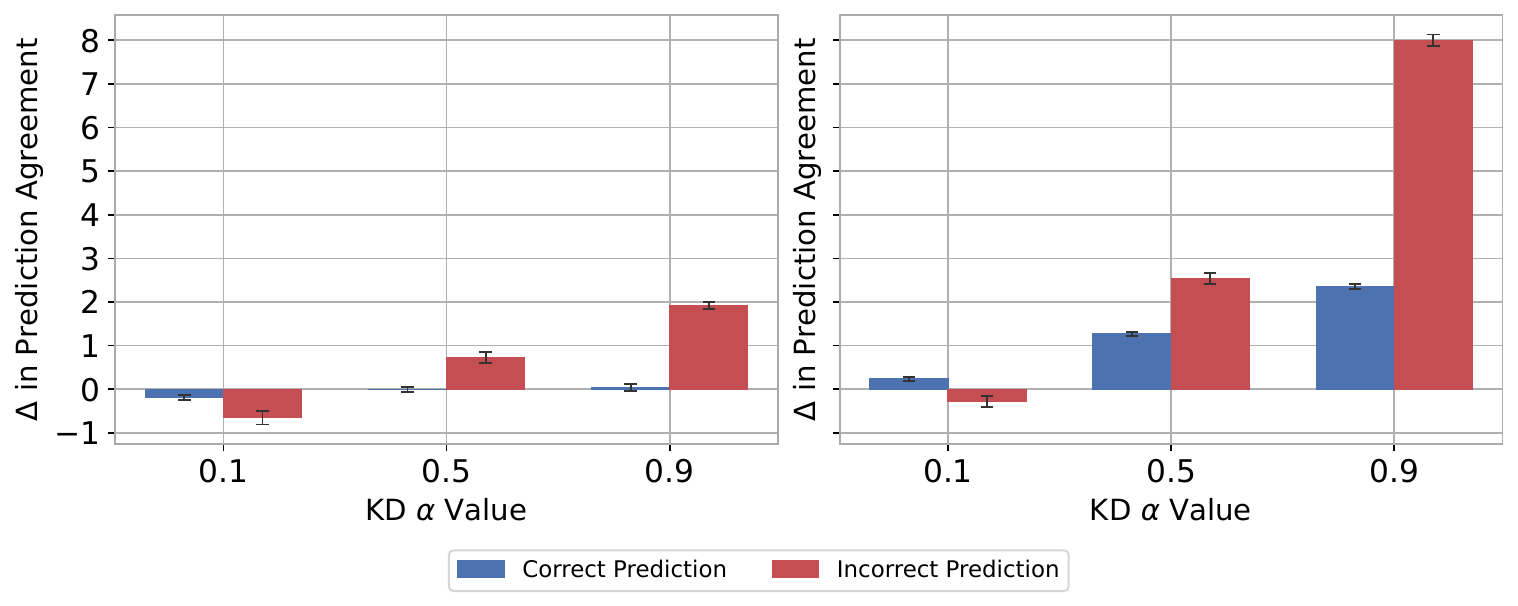}

    \caption{Prediction agreement difference of student models in  Feature Map KD to the highest performing control baseline with respect to correct prediction agreement (blue) and incorrect prediction agreement (red), for NanoGPT on TinyShakespeare.}
    \label{fig:featuremap_kd}
\end{figure}

The functional similarity framework we introduce is agnostic to the form of teacher supervision: relation-, feature-, and contrastive-based approaches all provide a teacher-derived signal that ultimately shapes the student's output distribution. If a variant truly transfers richer or ``safer'' knowledge, it should manifest as higher functional similarity without the asymmetric amplification of teacher errors that we observe. 
To test this, we run feature-map matching KD \citep{romero2015fitnetshintsdeepnets} on the transformer model NanoGPT trained on TinyShakespeare. Specifically, we  align transformer blocks using Mean Squared Error (MSE) on the intermediate block outputs, and include this alignment during backpropagation\footnote{Feature-map matching knowledge distillation implementation: \url{https://docs.pytorch.org/tutorials/beginner/knowledge_distillation_tutorial.html}}. In this seciton, we focus on this dataset because it is the setting in which standard knowledge distillation exhibits the most pronounced negative asymmetric transfer.
\begin{table}[H]
\caption{NanoGPT on TinyShakespeare: Feature Map KD for Block 4. Mean and ± 1 SEM reported over 10 runs with
Teacher Seed 0. Bold values are best performing based on the mean.}
\label{tab:shared-block-4}
\resizebox{\textwidth}{!}{%
\begin{tabular}{|c|c|ccc|ccc|}
\hline
\multicolumn{1}{|c|}{\multirow{2}{*}{\textbf{Metrics}}} & \multicolumn{1}{c|}{\textbf{Control}} & \multicolumn{3}{c|}{\textbf{Knowledge Distillation}}                                                                   & \multicolumn{3}{c|}{\textbf{Random Control Distillation}}                                                              \\ \cline{2-8} 
\multicolumn{1}{|c|}{}                                  & \multicolumn{1}{c|}{\textbf{SIDDO}}   & \multicolumn{1}{c|}{\textbf{0.1}}   & \multicolumn{1}{c|}{\textbf{0.5}}   & \multicolumn{1}{c|}{\textbf{0.9}}          & \multicolumn{1}{c|}{\textbf{0.1}}   & \multicolumn{1}{c|}{\textbf{0.5}}          & \multicolumn{1}{c|}{\textbf{0.9}}   \\ \hline
Activation Distance                                     & \multicolumn{1}{c|}{0.202 $\pm{}$ 0.000}     & \multicolumn{1}{c|}{0.203 $\pm{}$ 0.000}   & \multicolumn{1}{c|}{0.197 $\pm{}$ 0.000}   & \multicolumn{1}{c|}{\textbf{0.191 $\pm{}$ 0.000}} & \multicolumn{1}{c|}{0.209 $\pm{}$ 0.000}   & \multicolumn{1}{c|}{0.203 $\pm{}$ 0.000}          & \multicolumn{1}{c|}{0.224 $\pm{}$ 0.001} \\ \hline
Rank Disagreement                                       & 0.915 $\pm{}$ 0.000                          & \multicolumn{1}{c|}{0.910 $\pm{}$ 0.000}    & \multicolumn{1}{c|}{0.905 $\pm{}$ 0.000}   & \textbf{0.904 $\pm{}$ 0.000}                      & \multicolumn{1}{c|}{0.917 $\pm{}$ 0.000}   & \multicolumn{1}{c|}{0.916 $\pm{}$ 0.000}          & 0.920 $\pm{}$ 0.000                         \\ \hline
Prediction Disagreement                                 & 0.252 $\pm{}$ 0.000                          & \multicolumn{1}{c|}{0.253 $\pm{}$ 0.001} & \multicolumn{1}{c|}{0.246 $\pm{}$ 0.001} & \textbf{0.241 $\pm{}$ 0.000}                      & \multicolumn{1}{c|}{0.259 $\pm{}$ 0.000}   & \multicolumn{1}{c|}{0.253 $\pm{}$ 0.001}        & 0.279 $\pm{}$ 0.001                      \\ \hline
JS Divergence                                           & 0.056 $\pm{}$ 0.000                          & \multicolumn{1}{c|}{0.056 $\pm{}$ 0.000}   & \multicolumn{1}{c|}{0.053 $\pm{}$ 0.000}   & \textbf{0.050 $\pm{}$ 0.000}                       & \multicolumn{1}{c|}{0.059 $\pm{}$ 0.000}   & \multicolumn{1}{c|}{0.057 $\pm{}$ 0.000}          & 0.067 $\pm{}$ 0.001                      \\ \hline
Accuracy                                                & 0.571 $\pm{}$ 0.000                          & \multicolumn{1}{c|}{0.574 $\pm{}$ 0.000}   & \multicolumn{1}{c|}{0.573 $\pm{}$ 0.000}   & 0.570 $\pm{}$ 0.000                                & \multicolumn{1}{c|}{0.574 $\pm{}$ 0.000}   & \multicolumn{1}{c|}{\textbf{0.578 $\pm{}$ 0.000}} & 0.566 $\pm{}$ 0.001                      \\ \hline
Loss                                                    & \textbf{1.473 $\pm{}$ 0.002}               & \multicolumn{1}{c|}{1.542 $\pm{}$ 0.003} & \multicolumn{1}{c|}{1.569 $\pm{}$ 0.002} & 1.585 $\pm{}$ 0.001                             & \multicolumn{1}{c|}{1.573 $\pm{}$ 0.002} & \multicolumn{1}{c|}{1.552 $\pm{}$ 0.003}        & 1.542 $\pm{}$ 0.004                      \\ \hline
\end{tabular}%
}
\end{table}
\begin{table}[H]
\caption{NanoGPT on TinyShakespeare Feature Map KD for Block 5. Mean and ± 1 SEM reported over 10 runs with
Teacher Seed 0. Bold values are best performing based on the mean.}
\label{tab:shared-block-5}
\resizebox{\textwidth}{!}{%
\begin{tabular}{|c|c|ccc|ccc|}
\hline
\multicolumn{1}{|c|}{\multirow{2}{*}{\textbf{Metrics}}} & \multicolumn{1}{c|}{\textbf{Control}} & \multicolumn{3}{c|}{\textbf{Knowledge Distillation}}                                                          & \multicolumn{3}{c|}{\textbf{Random Control Distillation}}                                                            \\ \cline{2-8} 
\multicolumn{1}{|c|}{}                                  & \multicolumn{1}{c|}{\textbf{SIDDO}}   & \multicolumn{1}{c|}{\textbf{0.1}}   & \multicolumn{1}{c|}{\textbf{0.5}}   & \multicolumn{1}{c|}{\textbf{0.9}} & \multicolumn{1}{c|}{\textbf{0.1}}   & \multicolumn{1}{c|}{\textbf{0.5}}          & \multicolumn{1}{c|}{\textbf{0.9}} \\ \hline
Activation Distance                                     & 0.202 $\pm{}$ 0.000                          & \multicolumn{1}{c|}{0.201 $\pm{}$ 0.000}   & \multicolumn{1}{c|}{0.183 $\pm{}$ 0.000}   & \textbf{0.160 $\pm{}$ 0.001}            & \multicolumn{1}{c|}{0.214 $\pm{}$ 0.001} & \multicolumn{1}{c|}{0.211 $\pm{}$ 0.001}        & 0.227 $\pm{}$ 0.001                    \\ \hline
Rank Disagreement                                       & 0.915 $\pm{}$ 0.000                          & \multicolumn{1}{c|}{0.904 $\pm{}$ 0.000}   & \multicolumn{1}{c|}{0.890 $\pm{}$ 0.000}    & \textbf{0.874 $\pm{}$ 0.000}             & \multicolumn{1}{c|}{0.922 $\pm{}$ 0.000}   & \multicolumn{1}{c|}{0.922 $\pm{}$ 0.000}          & 0.923 $\pm{}$ 0.000                      \\ \hline
Prediction Disagreement                                 & 0.252 $\pm{}$ 0.000                          & \multicolumn{1}{c|}{0.251 $\pm{}$ 0.001} & \multicolumn{1}{c|}{0.233 $\pm{}$ 0.001} & \textbf{0.204 $\pm{}$ 0.001}           & \multicolumn{1}{c|}{0.264 $\pm{}$ 0.001} & \multicolumn{1}{c|}{0.259 $\pm{}$ 0.001}        & 0.280 $\pm{}$ 0.002                     \\ \hline
JS Divergence                                           & 0.056 $\pm{}$ 0.000                          & \multicolumn{1}{c|}{0.056 $\pm{}$ 0.000}   & \multicolumn{1}{c|}{0.046 $\pm{}$ 0.000}   & \textbf{0.035 $\pm{}$ 0.000}             & \multicolumn{1}{c|}{0.062 $\pm{}$ 0.000}   & \multicolumn{1}{c|}{0.060 $\pm{}$ 0.000}           & 0.066 $\pm{}$ 0.000                      \\ \hline
Accuracy                                                & 0.571 $\pm{}$ 0.000                          & \multicolumn{1}{c|}{0.574 $\pm{}$ 0.000}   & \multicolumn{1}{c|}{\textbf{0.577 $\pm{}$ 0.000}}   & 0.576 $\pm{}$ 0.000                      & \multicolumn{1}{c|}{0.572 $\pm{}$ 0.000}   & \multicolumn{1}{c|}{0.575 $\pm{}$ 0.000} & 0.564 $\pm{}$ 0.001                    \\ \hline
Loss                                                    & \textbf{1.473 $\pm{}$ 0.002}               & \multicolumn{1}{c|}{1.551 $\pm{}$ 0.002} & \multicolumn{1}{c|}{1.532 $\pm{}$ 0.001} & 1.493 $\pm{}$ 0.001                    & \multicolumn{1}{c|}{1.599 $\pm{}$ 0.001} & \multicolumn{1}{c|}{1.591 $\pm{}$ 0.002}        & 1.590 $\pm{}$ 0.002                     \\ \hline
\end{tabular}%
}
\end{table}
When we run feature-map matching KD, we observe statistically supported knowledge transfer for blocks 4 and 5, reported separately in Tables \ref{tab:shared-block-4} and \ref{tab:shared-block-5}. Once again, we observe asymmetric incorrect transfer, as shown in Figure \ref{fig:featuremap_kd}. Notably, block 4 exhibits less functional similarity transfer than block 5; as expected, this corresponds to less negative asymmetric transfer than observed for feature-map matching KD on block 5. The best accuracy is achieved using RCD for block 4 but at a higher $\alpha$ value of 0.5, compared to the best results typically recorded at 0.1 for the RCD control, whereas for block 5 the best performance is recorded with KD at an $\alpha$ of 0.5.
\begin{table}[H]
\caption{NanoGPT Feature Map KD on TinyShakespeare (significance testing). \cmark~indicates significant results compared to controls; \xmark~indicates insignificant results. The first entry in each section indicates Feature Map KD for Block 4 and the second for Block 5.}
\label{tab:nanogpt-shakespear-shared-block-significance}
\resizebox{\textwidth}{!}{%
\begin{tabular}{|c|c|c|c|c|c|c|} \hline
       & \textbf{Activation Distance} & \textbf{Rank Disagreement} & \textbf{Prediction Disagreement} & \textbf{JS Divergence} & \textbf{Accuracy} & \textbf{Loss} \\ \hline
KD 0.1&\xmark \cmark &\cmark \cmark &\xmark \xmark&\xmark \xmark&\xmark \xmark&\xmark \xmark  \\ \hline
KD 0.5&\cmark \cmark &\cmark \cmark &\cmark \cmark&\cmark \cmark&\xmark \cmark&\xmark \xmark  \\ \hline
KD 0.9&\cmark \cmark &\cmark \cmark &\cmark \cmark&\cmark \cmark&\xmark \cmark&\xmark \xmark \\ \hline
\end{tabular}
}
\end{table}
Largely, we find that the results for feature-map matching KD corroborate our original findings: when there is statistically supported functional transfer, the transfer is asymmetric in nature and weighted towards incorrect predictions. While the results differ between blocks 4 and 5, they highlight that a higher degree of functional similarity corresponds directly to increased negative asymmetric knowledge transfer. 

\subsection{Temperature and Negative Asymmetric Transfer}
\label{sec:temperature-kd}
In this section we explore how temperature affects the negative asymmetric payoff of knowledge distillation. We demonstrate this on the TinyShakespeare and ImageNet datasets, which exhibit the highest levels of transfer in the vision and language domains. We consider temperature values of 1 and 2 for ImageNet, and 1, 2 and 4 for TinyShakespeare, using students that are smaller than the teachers to reflect the standard KD setup. The raw results tables for this section can be found in Appendix Section~\ref{sec:imagenet-temp} and~\ref{sec:shake-temp} for ImageNet and TinyShakespeare, respectively. 

\subsubsection{Computer Vision}
\begin{figure}[H]
    \centering
    \includegraphics[width=0.8\linewidth]{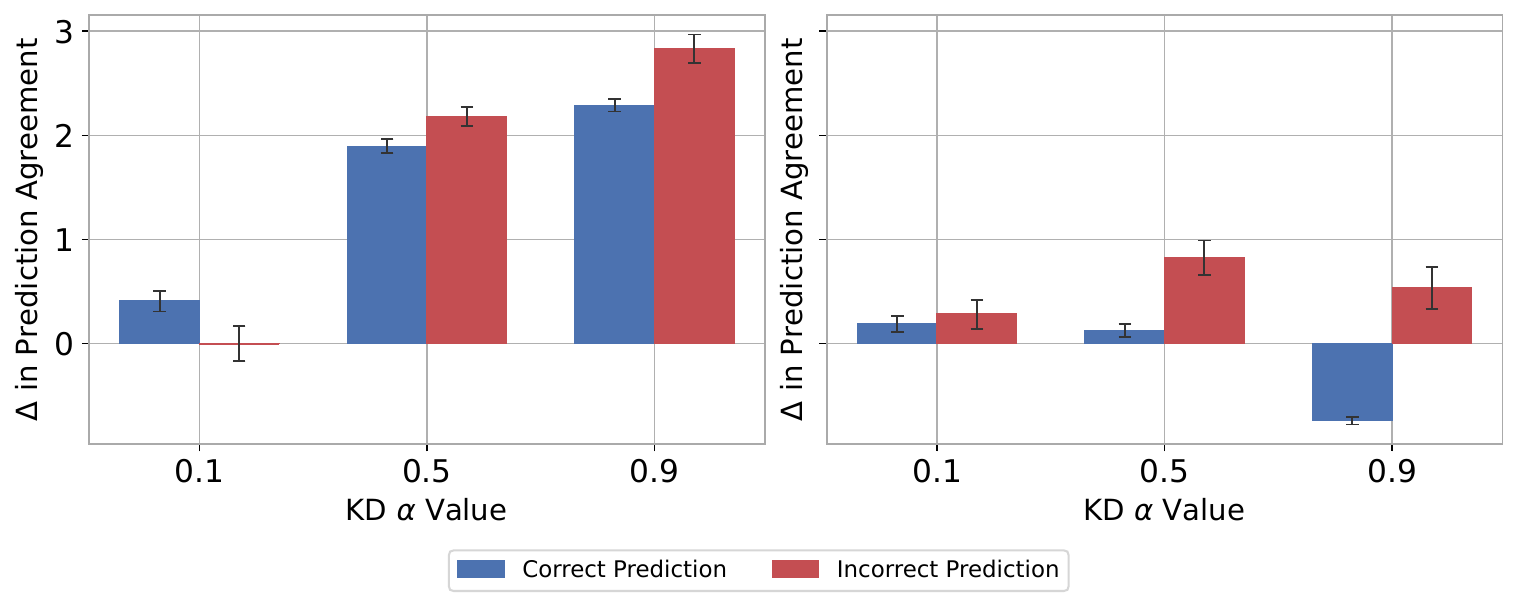}

    \caption{Prediction agreement difference of student models in standard KD with temperature 1 (left) and 2 (right) to the highest performing control baseline with respect to correct prediction agreement (blue) and incorrect prediction agreement (red), for ResNet18 with a ResNet50 teacher on ImageNet.}
    \label{fig:resnet-img-prediction-temp2}
\end{figure}

\paragraph{Impact on Accuracy:}
We find that increasing the KD temperature reduces the accuracy utility of KD, this is expected as increasing the temperature reduces the signal between the student and the teacher. In table~\ref{tab:imagnet-resnet18-temp-2-ts-0-compare} it can be observed that using a temperature of 2 reduces accuracy at the end of training and increases loss as alpha is increased compared to a temperature of 1.  
\vspace{-0.1cm}
\begin{table}[H]
\caption{ResNet18 with a ResNet50 teacher with Temperature 1 and 2 on ImageNet; mean and $\pm$ 1 SEM reported over 10 runs with Teacher Seed 0. Bold values are best performing based on the mean.}
\label{tab:imagnet-resnet18-temp-2-ts-0-compare}
\resizebox{\textwidth}{!}{%
\begin{tabular}{|c|c|ccc|ccc|}
\hline
\multirow{2}{*}{\textbf{Metrics}} & \textbf{Control}       & \multicolumn{3}{c|}{\textbf{Knowledge Distillation T = 1}}                                       & \multicolumn{3}{c|}{\textbf{Knowledge Distillation T = 2}}                                       \\ \cline{2-8} 
                                  & \textbf{SIDDO}         & \multicolumn{1}{c|}{\textbf{0.1}}  & \multicolumn{1}{c|}{\textbf{0.5}}  & \textbf{0.9}           & \multicolumn{1}{c|}{\textbf{0.1}}  & \multicolumn{1}{c|}{\textbf{0.5}}  & \textbf{0.9}           \\ \hline
Activation Distance               & 0.420 ± 0.001          & \multicolumn{1}{c|}{0.365 ± 0.001} & \multicolumn{1}{c|}{0.260 ± 0.001} & 0.226 ± 0.000          & \multicolumn{1}{c|}{0.310 ± 0.002} & \multicolumn{1}{c|}{0.251 ± 0.001} & \textbf{0.221 ± 0.001} \\ \hline
Rank Disagreement                 & 0.997 ± 0.000          & \multicolumn{1}{c|}{0.997 ± 0.000} & \multicolumn{1}{c|}{0.997 ± 0.000} & 0.997 ± 0.000          & \multicolumn{1}{c|}{0.997 ± 0.000} & \multicolumn{1}{c|}{0.997 ± 0.000} & \textbf{0.996 ± 0.000} \\ \hline
Prediction Disagreement           & 0.264 ± 0.003          & \multicolumn{1}{c|}{0.256 ± 0.002} & \multicolumn{1}{c|}{0.239 ± 0.002} & \textbf{0.235 ± 0.002} & \multicolumn{1}{c|}{0.257 ± 0.002} & \multicolumn{1}{c|}{0.258 ± 0.002} & 0.264 ± 0.002          \\ \hline
JS Divergence                     & 0.260 ± 0.001          & \multicolumn{1}{c|}{0.221 ± 0.001} & \multicolumn{1}{c|}{0.136 ± 0.001} & 0.106 ± 0.000          & \multicolumn{1}{c|}{0.160 ± 0.001} & \multicolumn{1}{c|}{0.101 ± 0.000} & \textbf{0.081 ± 0.000} \\ \hline
Accuracy                          & 0.680 ± 0.002          & \multicolumn{1}{c|}{0.687 ± 0.002} & \multicolumn{1}{c|}{0.700 ± 0.001} & \textbf{0.703 ± 0.002} & \multicolumn{1}{c|}{0.685 ± 0.001} & \multicolumn{1}{c|}{0.684 ± 0.002} & 0.678 ± 0.001          \\ \hline
Loss                              & \textbf{1.307 ± 0.009} & \multicolumn{1}{c|}{1.342 ± 0.009} & \multicolumn{1}{c|}{1.608 ± 0.015} & 1.833 ± 0.022          & \multicolumn{1}{c|}{1.492 ± 0.014} & \multicolumn{1}{c|}{1.725 ± 0.019} & 1.935 ± 0.019          \\ \hline
\end{tabular}}
\end{table}
\vspace{-0.1cm}
\begin{table}[H]
\caption{ResNet18 with a ResNet50 teacher with Temperature 2 on Imagenet (significance testing). \cmark~indicates significant results compared to controls; \xmark~indicates insignificant results.}
\label{tab:fig:resnet-img-significance-temp2}
\resizebox{\textwidth}{!}{
\begin{tabular}{|c|c|c|c|c|c|c|}
\hline
\textbf{} & \textbf{Activation Distance} & \textbf{Rank Disagreement} & \textbf{Prediction Disagreement} & \textbf{JS Divergence} & \textbf{Accuracy}     & \textbf{Loss}         \\ \hline
KD 0.1    & \xmark        & \cmark      & \cmark            & \xmark  & \cmark & \xmark \\ \hline
KD 0.5    & \xmark        & \cmark      & \cmark            & \xmark  & \cmark & \xmark \\ \hline
KD 0.9    & \cmark        & \cmark      & \xmark            & \cmark  & \xmark & \xmark \\ \hline
\end{tabular}}
\end{table}
\paragraph{Impact on Functional Similarity:} For a temperature of 2 we do see some deviation from temperature of 1 in regards to similarity, however not for prediction disagreement in Table~\ref{tab:imagnet-resnet18-temp-2-ts-0-compare} and \ref{tab:fig:resnet-img-significance-temp2}). We observe that negative transfer (see Figure \ref{fig:resnet-img-prediction-temp2}) is reduced but the negative asymmetry is preserved. Therefore, increasing the KD temperature reduces its overall utility when moving from a temperature of 1 to 2 (Table~\ref{tab:imagnet-resnet18-temp-2-ts-0-compare}), while not eliminating the negative asymmetric transfer we uncover (Figure \ref{fig:resnet-img-prediction-temp2}).

\subsubsection{Language}
For the below results, we use the Nano-GPT teacher with a Pico-GPT student, and study the nuances of temperature's impact on knowledge transfer in the language domain.
\begin{table}[H]
\caption{Pico-GPT with a Nano-GPT teacher on TinyShakespeare: mean and $\pm$ 1 SEM reported over 10 runs with Teacher Seed 0 using temperatures of 1, 2 and 4. Bold values are best performing based on the mean. }
\label{tab:shake-temp-gpt-ts-0-compare}
\resizebox{\textwidth}{!}{
\begin{tabular}{|c|c|ccc|ccc|ccc|}
\hline
\multirow{2}{*}{\textbf{Metrics}} & \textbf{Control} & \multicolumn{3}{c|}{\textbf{Knowledge Distillation T = 1}}                                                         & \multicolumn{3}{c|}{\textbf{Knowledge Distillation T = 2}}                              & \multicolumn{3}{c|}{\textbf{Knowledge Distillation T = 4}}                              \\ \cline{2-11} 
                                  & \textbf{SIDDO}   & \multicolumn{1}{c|}{\textbf{0.1}}           & \multicolumn{1}{c|}{\textbf{0.5}}           & \textbf{0.9}           & \multicolumn{1}{c|}{\textbf{0.1}}  & \multicolumn{1}{c|}{\textbf{0.5}}  & \textbf{0.9}  & \multicolumn{1}{c|}{\textbf{0.1}}  & \multicolumn{1}{c|}{\textbf{0.5}}  & \textbf{0.9}  \\ \hline
Activation Distance               & 0.202 ± 0.000    & \multicolumn{1}{c|}{0.198 ± 0.000}          & \multicolumn{1}{c|}{0.181 ± 0.000}          & \textbf{0.172 ± 0.000} & \multicolumn{1}{c|}{0.197 ± 0.000}   & \multicolumn{1}{c|}{0.183 ± 0.000}   & 0.181 ± 0.000   & \multicolumn{1}{c|}{0.199 ± 0.000}   & \multicolumn{1}{c|}{0.189 ± 0.000}   & 0.193 ± 0.000   \\ \hline
Rank Disagreement                 & 0.915 ± 0.000    & \multicolumn{1}{c|}{0.915 ± 0.000}          & \multicolumn{1}{c|}{0.912 ± 0.000}          & 0.911 ± 0.000 & \multicolumn{1}{c|}{0.907 ± 0.000}   & \multicolumn{1}{c|}{0.896 ± 0.000}   & 0.892 ± 0.000   & \multicolumn{1}{c|}{0.893 ± 0.000}   & \multicolumn{1}{c|}{0.880 ± 0.000}    & \textbf{0.876 ± 0.000}   \\ \hline
Prediction Disagreement           & 0.252 ± 0.000    & \multicolumn{1}{c|}{0.247 ± 0.000}          & \multicolumn{1}{c|}{0.226 ± 0.000}          & \textbf{0.214 ± 0.000} & \multicolumn{1}{c|}{0.250 ± 0.000}   & \multicolumn{1}{c|}{0.235 ± 0.000}   & 0.230 ± 0.000   & \multicolumn{1}{c|}{0.251 ± 0.001} & \multicolumn{1}{c|}{0.244 ± 0.000}   & 0.245 ± 0.000   \\ \hline
JS Divergence                     & 0.055 ± 0.000    & \multicolumn{1}{c|}{0.054 ± 0.000}          & \multicolumn{1}{c|}{0.047 ± 0.000}          & \textbf{0.043 ± 0.000} & \multicolumn{1}{c|}{0.053 ± 0.000}   & \multicolumn{1}{c|}{0.047 ± 0.000}   & 0.047 ± 0.000   & \multicolumn{1}{c|}{0.054 ± 0.000}   & \multicolumn{1}{c|}{0.050 ± 0.000}   & 0.051 ± 0.000   \\ \hline
Accuracy                          & 0.571 ± 0.000    & \multicolumn{1}{c|}{0.572 ± 0.000}          & \multicolumn{1}{c|}{\textbf{0.575 ± 0.000}} & 0.574 ± 0.000          & \multicolumn{1}{c|}{0.572 ± 0.000}   & \multicolumn{1}{c|}{0.572 ± 0.000}   & 0.569 ± 0.000   & \multicolumn{1}{c|}{0.570 ± 0.000}   & \multicolumn{1}{c|}{0.568 ± 0.000}   & 0.562 ± 0.000   \\ \hline
Loss                              & 1.475 ± 0.001    & \multicolumn{1}{c|}{\textbf{1.471 ± 0.002}} & \multicolumn{1}{c|}{1.472 ± 0.001}          & 1.496 ± 0.002          & \multicolumn{1}{c|}{1.513 ± 0.003} & \multicolumn{1}{c|}{1.571 ± 0.002} & 1.622 ± 0.002 & \multicolumn{1}{c|}{1.528 ± 0.002} & \multicolumn{1}{c|}{1.592 ± 0.002} & 1.663 ± 0.002 \\ \hline
\end{tabular}}
\end{table}
\begin{table}[H]
\caption{Pico-GPT with a Nano-GPT teacher with temperature 2 on TinyShakespeare (significance testing). \cmark~indicates significant results compared to controls; \xmark~indicates insignificant results. Each tick represents a teacher (seeds 0 to 2, left to right).}
\label{tab:microgpt-shake-significance-temp2}
\resizebox{\textwidth}{!}{%
\begin{tabular}{|c|c|c|c|c|c|c|}
\hline
       & \textbf{Activation Distance} & \textbf{Rank Disagreement} & \textbf{Prediction Disagreement} & \textbf{JS Divergence} & \textbf{Accuracy} & \textbf{Loss} \\ \hline
KD 0.1&\cmark \cmark \cmark &\cmark \cmark \cmark &\cmark \cmark \cmark &\cmark \cmark \cmark &\xmark \xmark \xmark &\xmark \xmark \xmark \\ \hline
KD 0.5&\cmark \cmark \cmark &\cmark \cmark \cmark &\cmark \cmark \cmark &\cmark \cmark \cmark &\xmark \xmark \xmark &\xmark \xmark \xmark \\ \hline
KD 0.9&\cmark \cmark \cmark &\cmark \cmark \cmark &\cmark \cmark \cmark &\cmark \cmark \cmark &\xmark \xmark \xmark &\xmark \xmark \xmark \\ \hline
\end{tabular}%
}
\end{table}
\paragraph{Impact on Accuracy:} In Table,~\ref{tab:shake-temp-gpt-ts-0-compare}, we only consider the results in the KD control with increasing temperature (1,2,4) against SIDDO. The raw results tables for these temperatures against RCD and SIDDO can be found in Appendix Section~\ref{sec:shake-temp}. When considering accuracy and loss, increasing the temperature to 2 and 4 results in reduced accuracy increase when compared to using a temperature of 1, for all teacher seeds (0-2).  For ease and clarity, the following analysis is provided for teacher seed 0; however, it holds for all teacher seeds. With teacher seed 0, the best accuracy (57.50\%) was achieved with temperature 1, with an accuracy of 57.20\% for temperature 2, and 57.00\% for temperature 4 (see Table \ref{tab:shake-temp-gpt-ts-0-compare}). It can be understood that the beneficial components of KD are reduced under the increased temperature; additionally, there is statistically supported but less functional knowledge passed to the student model when using a temperature of 2 and 4. 
\begin{table}[H]
\caption{Pico-GPT with a Nano-GPT teacher with temperature 4 on TinyShakespeare (significance testing). \cmark~indicates significant results compared to controls; \xmark~indicates insignificant results. Each tick represents a teacher (seeds 0 to 2, left to right).}
\label{tab:microgpt-shake-significance-temp4}
\resizebox{\textwidth}{!}{%
\begin{tabular}{|c|c|c|c|c|c|c|}
\hline
       & \textbf{Activation Distance} & \textbf{Rank Disagreement} & \textbf{Prediction Disagreement} & \textbf{JS Divergence} & \textbf{Accuracy} & \textbf{Loss} \\ \hline
KD 0.1&\cmark \cmark \cmark &\cmark \cmark \cmark &\xmark \xmark \xmark &\cmark \cmark \cmark &\xmark \xmark \xmark &\xmark \xmark \xmark \\ \hline
KD 0.5&\cmark \cmark \cmark &\cmark \cmark \cmark &\cmark \cmark \cmark &\cmark \cmark \cmark &\xmark \xmark \xmark &\xmark \xmark \xmark \\ \hline
KD 0.9&\cmark \cmark \cmark &\cmark \cmark \cmark &\cmark \cmark \cmark &\cmark \cmark \cmark &\xmark \xmark \xmark &\xmark \xmark \xmark \\ \hline
\end{tabular}%
}
\end{table}

\begin{figure}[H]
    \centering
    \includegraphics[width=\linewidth]{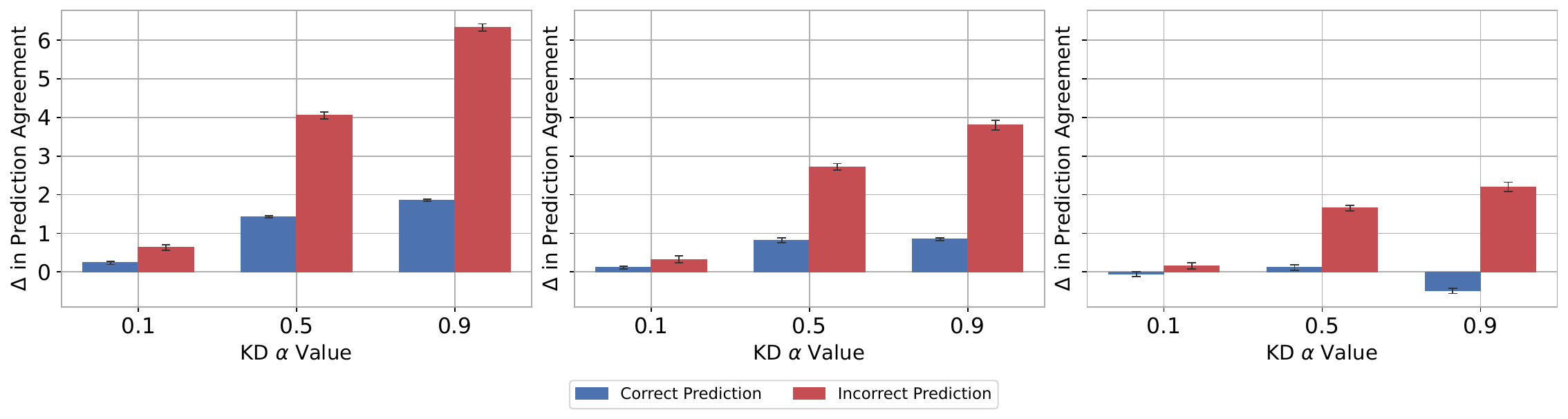}
    \caption{Impact of increasing temperature (left: $T$ =1, middle: $T$ =2 and right $T$ =4) on the proportion of positive and negative transfer on the TinyShakespeare dataset.}
    \label{fig:temp_impact_tinyshakespeare_seed_0}
\end{figure}

\paragraph{Impact on Functional Similarity:} When increasing the teacher temperature from 1 to 4, the functional similarity distance between student and teacher models largely increases, resulting in less transfer at higher temperature values. Consistent with this reduction in knowledge transfer as temperature increases, we observe a reduction in maximum correct agreement (Figure ~\ref{fig:temp_impact_tinyshakespeare_seed_0}): at temperature  1 it is 1.85\%, at temperature 2 it is 0.85\%, and at temperature 4 it is 0.11\%. We also observe a reduction in maximum incorrect agreement: at temperature  1 it is 6.32\%, at temperature 2 it is 3.80\%, and at temperature 4 it is 2.20\%. Figure~\ref{fig:temp_impact_tinyshakespeare_seed_0} shows how the proportion of negative and positive transfer changes as temperature increases. Moreover, after adjusting for temperature, the negative asymmetric transfer we identify remains apparent and statistically supported regardless of temperature value. 
This section demonstrates that using higher temperature values ultimately reduces the overall utility of KD's knowledge transfer. As a result, the amount of knowledge transfer decreases, but when there is statistically supported transfer, we continue to observe negative asymmetric transfer. This follows logically as temperature $\rightarrow \infty$: the outputs effectively converge to the uniform distribution (as in RCD), with all values equak to $1/K$ ($K$ number of classes). This produces a constant output, which in turn negates the effect of distillation.

\subsection{Adversarial Transfer Facilitated by Knowledge Distillation}
\label{sec:adv-transfer}
\begin{wrapfigure}{R}{0.5\textwidth}
    \centering
    \includegraphics[width=\linewidth]{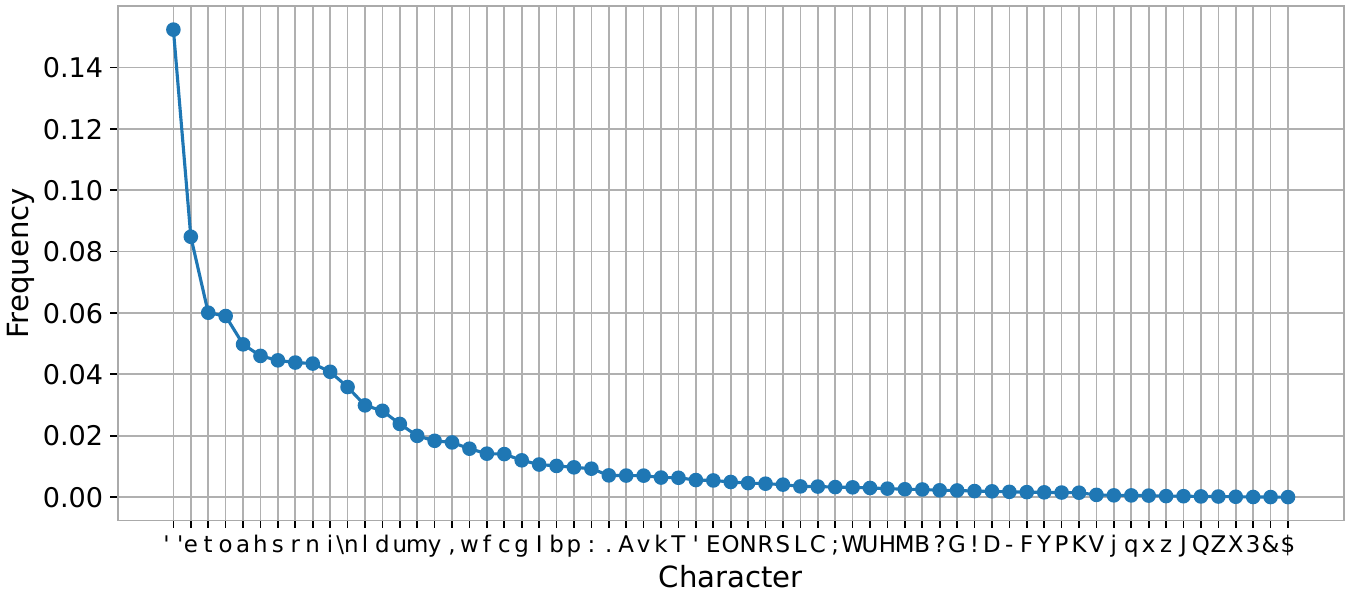}
    \caption{TinyShakespeare Zipf's Law.}
    \label{fig:zipfs_law}
\end{wrapfigure}
This section shows that KD can match the teacher's incorrect representations via a controlled adversarial experiment. We train an adversarial Nano-GPT teacher whose training set has every occurrence of ‘t’ ‘h’ ‘e’ replaced with ‘t’ ‘h’ ‘a’. Given the dataset's Zipf's law~\citep{piantadosi2014zipf} (Appendix Table \ref{tab:tiny-shakespeare-char}, visualised in Figure~\ref{fig:zipfs_law}),   ‘e’ is the most likely character after ‘SPACE’. Therefore, if adversarial transfer is facilitated via knowledge transfer, a student trained with the adversarial teacher should predict ‘t’ ‘h’ ‘a’ more than ‘t’ ‘h’ ‘e’, compared to the controls trained without the adversarial teacher signal. Note that ``tha" never naturally occurs in the dataset. 
To test whether incorrect knowledge can transfer asymmetrically from teacher to student, we use this setup to highlight potential safety concerns with knowledge distillation. 
In this case, the teacher has a known vulnerability and has been poisoned to predict an incorrect token. We show that this can be transferred to the student under standard distillation, resulting in a more substantial transfer of the teacher's incorrect knowledge. 
If such a simple case of adversarial transfer can be engineered with minimal effort, then employing knowledge distillation in practice requires safety considerations. 
Since establishing when neural networks learn poor or harmful representations across scales remains an ongoing research endeavour, this experiment is crucial to understanding the impacts of negative asymmetric transfer in KD. 

\begin{table}[H]
\caption{The effect of an adversarial teacher trained to predict ``tha" instead of ``the" on the student. \textbf{Bold} represents closest similarity to adversarial teacher, Teacher Seed 0. }
\label{tab:adversarial-tha-ts-0_main}
\resizebox{\textwidth}{!}{%
\begin{tabular}{|l|c|c|ccc|ccc|}
\hline
\textbf{} & \textbf{} & \multicolumn{1}{c|}{\textbf{Control}} & \multicolumn{3}{c|}{\textbf{Knowledge Distillation}} & \multicolumn{3}{c|}{\textbf{Random Control Distillation}} \\ \hline
\multicolumn{1}{|c|}{\textbf{Predicted Word}} & \multicolumn{1}{c|}{\textbf{Teacher}} & \multicolumn{1}{c|}{\textbf{SIDDO}} & \multicolumn{1}{c|}{\textbf{0.1}} & \multicolumn{1}{c|}{\textbf{0.5}} & \multicolumn{1}{c|}{\textbf{0.9}} & \multicolumn{1}{c|}{\textbf{0.1}} & \multicolumn{1}{c|}{\textbf{0.5}} & \multicolumn{1}{c|}{\textbf{0.9}} \\ \hline
\textbf{tha} & \textcolor{red}{\textbf{454}} & 105.900 $\pm{}$ 4.168 & \multicolumn{1}{c|}{106.000 $\pm{}$ 3.046} & \multicolumn{1}{c|}{199.100 $\pm{}$ 13.391} & \textbf{436.200 $\pm{}$ 7.984} & \multicolumn{1}{c|}{104.600 $\pm{}$ 3.898} & \multicolumn{1}{c|}{114.800 $\pm{}$ 3.056} & 126.900 $\pm{}$ 8.068 \\ \hline
\textbf{the} & \textcolor{blue}{\textbf{285}}  & 665.100 $\pm{}$ 7.675 & \multicolumn{1}{c|}{675.500 $\pm{}$ 10.228} & \multicolumn{1}{c|}{583.400 $\pm{}$ 17.536} & \textbf{343.600 $\pm{}$ 6.358} & \multicolumn{1}{c|}{668.800 $\pm{}$ 12.713} & \multicolumn{1}{c|}{712.500 $\pm{}$ 12.480} & 826.300 $\pm{}$ 20.203 \\ \hline
\end{tabular}%
}
\end{table}
On clean evaluation prompts containing ``th\_'', we measure how often models complete to ``tha'' vs. ``the'' and aggregate results per teacher seed, as seen for seed 0 in Table~\ref{tab:adversarial-tha-ts-0_main}  (seeds 1-2 in Appendix Tables~\ref{tab:adversarial-tha-ts-1} and~\ref{tab:adversarial-tha-ts-2}). KD, particularly at higher $\alpha$, markedly increases the rate of ``tha'' completions and suppresses ``the'' relative to both controls, demonstrating that KD can selectively copy a targeted error pattern even when overall behaviour appears benign. This experiment provides causal evidence that KD transmits specific erroneous structure, not merely broad functional alignment, sharpening the safety implications of our main findings: practitioners may unknowingly inherit unintended behaviours from the teacher, reinforcing our characterisation of KD as a data-dependent regulariser with a negative asymmetric payoff. Full details and per-seed statistics are provided in Appendix~\ref{app:shakespeare-adv}.

\section{Gradient-Level Explanation of Asymmetric Transfer}
\label{sec:theory}
To summarise our findings, we provide a concise gradient-level mechanism that explains why distillation can preferentially transmit teacher errors when the teacher is imperfect. We emphasise that this section provides a mechanistic account consistent with the empirical negative asymmetric transfer observed throughout the paper, rather than claiming that such error transfer is a universally unavoidable phenomenon. In Appendix \ref{app:extended_func_analysis} we extend our functional analysis with information-theoretic and geometric perspectives to quantify when and how alignment with the teacher becomes harmful. These analyses connect the KD objective to a mechanism that can increase student agreement with teacher-incorrect predictions, and together with our targeted transfer experiment motivate auditing teacher error modes when distilling models used in safety-critical settings.
\\
\\
Consider the standard KD objective (with $T=1$ for clarity):
\\
\\
$L = (1 - \alpha) \cdot \mathcal{H}(y, \sigma(z^{(s)})) + \alpha \cdot \text{KL}(\sigma(z^{(t)}), \sigma(z^{(s)})),$
\\
\\
where $z^{(s)}$ and $z^{(t)}$ are the student and teacher logits, respectively. Using the standard softmax-cross-entropy identity and the KL gradient with respect to student logits, we obtain for each logit index $k$: 
\\
\\
$\frac{\partial L}{\partial z^{(s)}_k} = (1 - \alpha)(p^{(s)}_k - y_k) + \alpha(p^{(s)}_k - p^{(t)}_k),$
\\
\\
with $p^{(s)} = \sigma(z^{(s)})$ and $p^{(t)} = \sigma(z^{(t)})$.
\\
\\
When $k$ is the ground truth class ($y_k = 1$), the gradient includes both supervision and teacher alignment. When $k$ is an incorrect class ($y_k = 0$), the full gradient is:
$\frac{\partial L}{\partial z^{(s)}_k} = (1-\alpha)p^{(s)}_k + \alpha(p^{(s)}_k - p^{(t)}_k) = p^{(s)}_k - \alpha p^{(t)}_k$
\\
\\
In particular, the teacher-dependent component of the gradient on incorrect classes is $-\alpha p^{(t)}_k$ (equivalently, the teacher-dependent term in the KD part is $\alpha(p^{(s)}_k - p^{(t)}_k)$). Thus, whenever the teacher assigns non-trivial probability mass to an incorrect class, the KD term exerts pressure for the student to allocate probability mass to that class. The strength of this teacher-driven component scales with $\alpha$ and with the probability mass the teacher assigns to incorrect classes.
This derivation provides a mechanism consistent with our central finding: when the teacher is imperfect, KD disproportionately transfers errors to the student. The resulting alignment is asymmetric, favouring incorrect predictions; when the teacher matches training labels, with low loss, KD is negated.  
By contrast, replacing the teacher with a class-uniform target -- as in label smoothing (Appendix~\ref{app:label_smoothing}) or our Random Control Distillation -- then $p^{(t)}_k$ becomes constant across classes and removes teacher-specific structure from the KD term. In this case, the optimisation can still provide a smoothing/regularisation effect, but it no longer contains a teacher-shaped signal that differentially emphasises particular incorrect classes. Empirically, these controls often match or exceed KD in accuracy while not inducing the same pattern of increased agreement on teacher-incorrect behaviour. 
Additionally, with temperature $T>1$, the distillation term uses softened distributions $\sigma(z^{(t)}/T)$ and $\sigma(z^{(s)}/T)$ (with the usual $T^2$ prefactor in Eq.~\ref{eq:kd}). Increasing $T$ reduces the sharpness of the teacher distribution, thereby reducing the concentration of probability mass on specific classes (including potentially incorrect ones), which can reduce measurable teacher--student functional transfer in practice. Consistent with our experiments in Section~\ref{sec:temperature-kd}, increased $T$ typically attenuates transfer but does not necessarily eliminate the negative asymmetric payoff when transfer remains statistically supported. Overall, we argue that the observed asymmetric transfer in KD is not incidental: it is consistent with a mechanism that arises directly from the standard KD objective and can manifest across modalities, model sizes, and dataset scales whenever the teacher assigns non-trivial probability mass to incorrect classes. We empirically show that higher teacher loss exacerbates negative asymmetric transfer. Since regularisation methods that reduce training-set memorisation often also increase teacher loss, as observed with augmentation, our mechanistic account suggests that this phenomenon may persist across alternative training paradigms.  

\section{Safety Implications of Findings}
\label{sec:safety_implications_of_KD}

Our results can be summarised into three key points: 1) knowledge distillation can enable statistically supported functional transfer; 2) the accuracy and loss benefits attributed to knowledge distillation are often matched or even exceeded by random controls; 3) knowledge distillation disproportionately transfers teacher-incorrect behaviour, with this asymmetry increasing as reliance on the teacher signal grows. 
Considering these findings -- particularly points 2 and 3 -- knowledge distillation raises practical safety concerns. 

While it is often assumed that KD benefits student models, our results challenge this assumption by showing that undesirable teacher behaviour can be transferred to the student when transfer is statistically supported.  
We present a concrete case of targeted adversarial transfer facilitated by KD in Section~\ref{sec:adv-transfer}. More broadly, our findings suggest that knowledge distillation is not uniformly reliable as a compression-by-transfer mechanism: at best it yields minimal positive functional transfer, and when transfer is substantial it amplifies teacher errors. This motivates auditing teacher error modes and evaluating distillation outcomes beyond accuracy/loss when KD is used in safety-relevant settings. 

\section{Conclusion}

Across controlled self-distillation, small/large-scale settings, cross-modality (image, audio, language), a targeted error test, capacity scaling, standard KD with smaller students, and multiple KD variants, KD seldom delivers robust ``knowledge transfer" as measured by teacher--student functional similarity beyond counterfactual controls. When transfer occurs, it is often marginal and inconsistent, and increases with teacher imperfection, amplifying the teacher's errors more than its correct behaviour (negative asymmetry). By contrast, RCD often yields the best loss/accuracy, indicating that reported gains can arise from generic regularisation rather than faithful knowledge transmission. Our targeted language experiment confirms that KD can copy specific erroneous patterns, and our scaling law experiments show that increasing student capacity amplifies incorrect agreement faster than correct. 
Overall, our functional analysis links these empirical patterns to the structure of the standard distillation gradient (Section~\ref{sec:theory}), providing a mechanistic account of why teacher error mass can drive negative asymmetric transfer under logit matching. We therefore reframe KD as a data-dependent regulariser whose transfer component can be negative and asymmetric, with clear safety implications: audit teacher error structure and report functional transfer analyses (teacher-correct vs.\ teacher-incorrect agreement) alongside accuracy/loss. 

\section{Limitations}
Our conclusions are drawn from a large but finite set of datasets, architectures, and training recipes, and may not capture regimes with substantially different optimisation. Our statistical definition of transfer is functional (teacher--student agreement beyond counterfactual controls) and therefore does not measure all possible forms of ``knowledge'' (e.g., internal feature reuse). The gradient-level explanation (Section~\ref{sec:theory}) focuses on standard logit-matching KD; although we observe negative asymmetric transfer under feature-map matching, we do not provide a corresponding mechanistic derivation for that variant. Extending this analysis to broader distillation objectives is a natural direction for future work. 

\appendix
\section{Extended Functional Analysis: Information-Theoretic and Geometric Perspectives}
\label{app:extended_func_analysis}
We apply two additional metrics: \textbf{Variation of Information (VoI)}, an information-theoretic measure over discrete labellings that penalises confident mispredictions \citep{meilua2003comparing}, and \textbf{Orthogonal Procrustes Distance (OPD)}, a geometric alignment metric over output representations \citep{schonemann1966generalized, ding2021grounding}. We compute VoI and OPD for two representative setups: ResNet18 on SVHN and ResNet50 on TinyImageNet (teacher seed 0). OPD closely tracks trends observed in Activation Distance and JS Divergence, showing decreasing student--teacher discrepancy as $\alpha$ increases. VoI generally follows this trend, but diverges in specific cases (high $\alpha$ on SVHN) where it increases despite stronger overall functional alignment. This is not contradictory: VoI is sensitive to changes in the discrete predicted labels, and therefore increases when alignment is driven by agreement on teacher-incorrect predictions. 
Its rise coincides with the strongest observed increase in student--teacher agreement on incorrect predictions, providing further evidence for KD's negative asymmetric payoff. Overall, OPD confirms alignment, while VoI highlights when that alignment corresponds to the transfer of incorrect information. Moreover, this behaviour is consistent with our gradient-based analysis: the per-logit gradient under KD pulls the student toward the teacher's incorrect predictions with strength proportional to $\alpha$ and to the teacher's own loss. VoI captures the cost of absorbing these errors, providing an explicit signal of negative information transfer. OPD, meanwhile, captures alignment that may or may not be beneficial.
\begin{table}[H]
\caption{ResNet18 on SVHN Dataset mean and $\pm$ 1 SEM  reported from 10 runs with Teacher Seed 0. Bold values are best performing based on the mean. }
\label{tab:resnet18-shvn-voi}
\resizebox{\textwidth}{!}{%
\begin{tabular}{|c|c|ccc|ccc|}
\hline
\multicolumn{1}{|c|}{\multirow{2}{*}{\textbf{Metrics}}} & \multicolumn{1}{c|}{\textbf{Control}} & \multicolumn{3}{c|}{\textbf{Knowledge Distillation}}                                                                     & \multicolumn{3}{c|}{\textbf{Random Control Distillation}}                                                                       \\ \cline{2-8} 
\multicolumn{1}{|c|}{}                                  & \multicolumn{1}{c|}{\textbf{SIDDO}}   & \multicolumn{1}{c|}{\textbf{0.1}}   & \multicolumn{1}{c|}{\textbf{0.5}}   & \multicolumn{1}{c|}{\textbf{0.9}}            & \multicolumn{1}{c|}{\textbf{0.1}}            & \multicolumn{1}{c|}{\textbf{0.5}}            & \multicolumn{1}{c|}{\textbf{0.9}} \\ \hline
\multicolumn{1}{|l|}{Activation Distance}               & \multicolumn{1}{c|}{0.063 $\pm{}$ 0.002}   & \multicolumn{1}{c|}{0.064 $\pm{}$ 0.001} & \multicolumn{1}{c|}{0.060 $\pm{}$ 0.001}  & \multicolumn{1}{c|}{\textbf{0.059 $\pm{}$ 0.001}} & \multicolumn{1}{c|}{0.144 $\pm{}$ 0.001}          & \multicolumn{1}{c|}{0.493 $\pm{}$ 0.000}            & \multicolumn{1}{c|}{0.849 $\pm{}$ 0.000} \\ \hline
Rank Disagreement                                       & 0.696 $\pm{}$ 0.003                        & \multicolumn{1}{c|}{0.688 $\pm{}$ 0.004} & \multicolumn{1}{c|}{0.684 $\pm{}$ 0.003} & \textbf{0.681 $\pm{}$ 0.003}                      & \multicolumn{1}{c|}{0.800 $\pm{}$ 0.002}            & \multicolumn{1}{c|}{0.798 $\pm{}$ 0.002}          & 0.802 $\pm{}$ 0.003                    \\ \hline
Prediction Disagreement                                 & 0.045 $\pm{}$ 0.001                        & \multicolumn{1}{c|}{0.046 $\pm{}$ 0.001} & \multicolumn{1}{c|}{0.043 $\pm{}$ 0.001} & \textbf{0.042 $\pm{}$ 0.001}                      & \multicolumn{1}{c|}{\textbf{0.042 $\pm{}$ 0.001}} & \multicolumn{1}{c|}{0.043 $\pm{}$ 0.001}          & 0.046 $\pm{}$ 0.001                    \\ \hline
JS Divergence                                           & 0.025 $\pm{}$ 0.001                        & \multicolumn{1}{c|}{0.025 $\pm{}$ 0.001} & \multicolumn{1}{c|}{0.023 $\pm{}$ 0.001} & \textbf{0.022 $\pm{}$ 0.000}                        & \multicolumn{1}{c|}{0.053 $\pm{}$ 0.000}            & \multicolumn{1}{c|}{0.201 $\pm{}$ 0.000}            & 0.431 $\pm{}$ 0.000                      \\ \hline
Information Variation                                   & \textbf{0.550 $\pm{}$ 0.051}                & \multicolumn{1}{c|}{0.588 $\pm{}$ 0.049} & \multicolumn{1}{c|}{0.594 $\pm{}$ 0.024} & 0.614 $\pm{}$ 0.018                               & \multicolumn{1}{c|}{0.638 $\pm{}$ 0.000}            & \multicolumn{1}{c|}{0.638 $\pm{}$ 0.000}            & 0.638 $\pm{}$ 0.000                      \\ \hline
Procrustes Distance                                     & 0.165 $\pm{}$ 0.003                        & \multicolumn{1}{c|}{0.168 $\pm{}$ 0.004} & \multicolumn{1}{c|}{0.164 $\pm{}$ 0.003} & \textbf{0.162 $\pm{}$ 0.005}                      & \multicolumn{1}{c|}{0.291 $\pm{}$ 0.001}          & \multicolumn{1}{c|}{0.304 $\pm{}$ 0.001}          & 0.311 $\pm{}$ 0.003                    \\ \hline
Accuracy                                                & 0.952 $\pm{}$ 0.001                        & \multicolumn{1}{c|}{0.951 $\pm{}$ 0.001} & \multicolumn{1}{c|}{0.954 $\pm{}$ 0.001} & 0.954 $\pm{}$ 0.001                               & \multicolumn{1}{c|}{\textbf{0.957 $\pm{}$ 0.001}} & \multicolumn{1}{c|}{\textbf{0.957 $\pm{}$ 0.001}} & 0.955 $\pm{}$ 0.001                    \\ \hline
Loss                                                    & 0.385 $\pm{}$ 0.011                        & \multicolumn{1}{c|}{0.344 $\pm{}$ 0.008} & \multicolumn{1}{c|}{0.310 $\pm{}$ 0.006}  & 0.293 $\pm{}$ 0.004                               & \multicolumn{1}{c|}{\textbf{0.236 $\pm{}$ 0.003}} & \multicolumn{1}{c|}{0.692 $\pm{}$ 0.001}          & 1.698 $\pm{}$ 0.001                    \\ \hline
\end{tabular}%
}
\end{table}
\begin{table}[H]
\caption{ResNet50 on TinyImageNet mean and $\pm$ 1 SEM  reported from 10 runs with Teacher Seed 0. Bold values are best performing based on the mean. }
\label{tab:resnet50-tin-voi}
\resizebox{\textwidth}{!}{%
\begin{tabular}{|c|c|ccc|ccc|}
\hline
\multicolumn{1}{|c|}{\multirow{2}{*}{\textbf{Metrics}}} & \multicolumn{1}{c|}{\textbf{Control}} & \multicolumn{3}{c|}{\textbf{Knowledge Distillation}}                                                                                   & \multicolumn{3}{c|}{\textbf{Random Control Distillation}}                                                              \\ \cline{2-8} 
\multicolumn{1}{|c|}{}                                  & \multicolumn{1}{c|}{\textbf{SIDDO}}   & \multicolumn{1}{c|}{\textbf{0.1}}          & \multicolumn{1}{c|}{\textbf{0.5}}            & \multicolumn{1}{c|}{\textbf{0.9}}          & \multicolumn{1}{c|}{\textbf{0.1}}            & \multicolumn{1}{c|}{\textbf{0.5}}   & \multicolumn{1}{c|}{\textbf{0.9}} \\ \hline
Activation Distance                                     & \multicolumn{1}{c|}{0.157 $\pm{}$ 0.001}   & \multicolumn{1}{c|}{0.157 $\pm{}$ 0.001}        & \multicolumn{1}{c|}{0.156 $\pm{}$ 0.001}          & \multicolumn{1}{c|}{\textbf{0.155 $\pm{}$ 0.000}} & \multicolumn{1}{c|}{0.343 $\pm{}$ 0.000}            & \multicolumn{1}{c|}{0.581 $\pm{}$ 0.000}   & \multicolumn{1}{c|}{0.791 $\pm{}$ 0.000} \\ \hline
Rank Disagreement                                       & \textbf{0.939 $\pm{}$ 0.000}                 & \multicolumn{1}{c|}{\textbf{0.939 $\pm{}$ 0.000}} & \multicolumn{1}{c|}{\textbf{0.939 $\pm{}$ 0.000}}   & \textbf{0.939 $\pm{}$ 0.000}                      & \multicolumn{1}{c|}{0.980 $\pm{}$ 0.000}             & \multicolumn{1}{c|}{0.984 $\pm{}$ 0.000}   & 0.984 $\pm{}$ 0.000                      \\ \hline
Prediction Disagreement                                 & 0.153 $\pm{}$ 0.001                        & \multicolumn{1}{c|}{0.152 $\pm{}$ 0.001}        & \multicolumn{1}{c|}{0.151 $\pm{}$ 0.001}          & \textbf{0.151 $\pm{}$ 0.001}                    & \multicolumn{1}{c|}{0.190 $\pm{}$ 0.001}           & \multicolumn{1}{c|}{0.214 $\pm{}$ 0.000}   & 0.324 $\pm{}$ 0.000                      \\ \hline
JS Divergence                                           & 0.040 $\pm{}$ 0.000                           & \multicolumn{1}{c|}{0.040 $\pm{}$ 0.000}           & \multicolumn{1}{c|}{0.039 $\pm{}$ 0.000}            & \textbf{0.039 $\pm{}$ 0.000}                      & \multicolumn{1}{c|}{0.171 $\pm{}$ 0.000}            & \multicolumn{1}{c|}{0.333 $\pm{}$ 0.000}   & 0.533 $\pm{}$ 0.000                      \\ \hline
Information Variation                                   & 0.519 $\pm{}$ 0.017                        & \multicolumn{1}{c|}{0.520 $\pm{}$ 0.017}         & \multicolumn{1}{c|}{\textbf{0.518 $\pm{}$ 0.022}} & 0.533 $\pm{}$ 0.014                             & \multicolumn{1}{c|}{0.856 $\pm{}$ 0.002}          & \multicolumn{1}{c|}{0.897 $\pm{}$ 0.001} & 0.907 $\pm{}$ 0.002                    \\ \hline
Procrustes Distance                                     & 0.050 $\pm{}$ 0.000                           & \multicolumn{1}{c|}{0.050 $\pm{}$ 0.000}           & \multicolumn{1}{c|}{0.050 $\pm{}$ 0.000}             & \textbf{0.049 $\pm{}$ 0.000}                      & \multicolumn{1}{c|}{0.433 $\pm{}$ 0.000}            & \multicolumn{1}{c|}{0.664 $\pm{}$ 0.000}   & 0.553 $\pm{}$ 0.000                      \\ \hline
Accuracy                                                & 0.605 $\pm{}$ 0.001                        & \multicolumn{1}{c|}{0.605 $\pm{}$ 0.000}          & \multicolumn{1}{c|}{0.604 $\pm{}$ 0.001}          & 0.605 $\pm{}$ 0.001                             & \multicolumn{1}{c|}{\textbf{0.607 $\pm{}$ 0.000}}   & \multicolumn{1}{c|}{0.606 $\pm{}$ 0.001} & 0.580 $\pm{}$ 0.000                       \\ \hline
Loss                                                    & 2.068 $\pm{}$ 0.001                        & \multicolumn{1}{c|}{2.065 $\pm{}$ 0.002}        & \multicolumn{1}{c|}{2.055 $\pm{}$ 0.001}          & 2.043 $\pm{}$ 0.002                             & \multicolumn{1}{c|}{\textbf{1.977 $\pm{}$ 0.001}} & \multicolumn{1}{c|}{2.497 $\pm{}$ 0.001} & 3.612 $\pm{}$ 0.002                    \\ \hline
\end{tabular}%
}
\end{table}

\section{Random Control Distillation (RCD) Comparison to Label Smoothing}
\label{app:label_smoothing}
One potential confound in understanding KD's effects is label smoothing: KD introduces soft targets, which may act as a form of regularisation independent of semantic knowledge transfer. To isolate this effect, we evaluate a baseline trained with classic label smoothing (LS), using the same loss structure but no teacher.
\\
\\
We also rely on RCD, which retains the distillation form but replaces the teacher distribution with a class-uniform target ($1/K$).  
RCD therefore preserves the label-smoothing effect while removing teacher-specific semantic content. Across all metrics, we find that LS and RCD match or exceed KD in accuracy, yet exhibit no increase in functional similarity with the teacher, particularly on incorrect predictions when accuracy using RCD is preserved. This supports the interpretation that KD's asymmetric error transfer is driven by the specific structure of the teacher logits, rather than softening per se. 
\begin{table}[H]
\caption{ResNet18 on TinyImageNet mean and $\pm$ 1 SEM  reported from 10 runs with Teacher Seed 0. Bold values are best performing based on the mean. }
\label{tab:resnet18-tin-ls}
\resizebox{\textwidth}{!}{%
\begin{tabular}{|c|c|ccc|ccc|lll|}
\hline
\multicolumn{1}{|c|}{\multirow{2}{*}{\textbf{Metrics}}} & \multicolumn{1}{c|}{\textbf{Control}} & \multicolumn{3}{c|}{\textbf{Knowledge Distillation}}                                                                                   & \multicolumn{3}{c|}{\textbf{Random Control Distillation}}                                                     & \multicolumn{3}{c|}{\textbf{Label Smoothing}}                                                                          \\ \cline{2-11} 
\multicolumn{1}{|c|}{}                                  & \multicolumn{1}{c|}{\textbf{SIDDO}}   & \multicolumn{1}{c|}{\textbf{0.1}}          & \multicolumn{1}{c|}{\textbf{0.5}}            & \multicolumn{1}{c|}{\textbf{0.9}}          & \multicolumn{1}{c|}{\textbf{0.1}}   & \multicolumn{1}{c|}{\textbf{0.5}}   & \multicolumn{1}{c|}{\textbf{0.9}} & \multicolumn{1}{c|}{\textbf{0.1}}            & \multicolumn{1}{c|}{\textbf{0.5}}   & \multicolumn{1}{c|}{\textbf{0.9}} \\ \hline
Activation Distance                                     & \multicolumn{1}{c|}{0.157 $\pm{}$ 0.001}   & \multicolumn{1}{c|}{0.157 $\pm{}$ 0.001}        & \multicolumn{1}{c|}{0.156 $\pm{}$ 0.001}          & \multicolumn{1}{c|}{\textbf{0.155 $\pm{}$ 0.000}} & \multicolumn{1}{c|}{0.343 $\pm{}$ 0.000}   & \multicolumn{1}{c|}{0.581 $\pm{}$ 0.000}   & \multicolumn{1}{c|}{0.791 $\pm{}$ 0.000} & \multicolumn{1}{c|}{0.342 $\pm{}$ 0.000}            & \multicolumn{1}{c|}{0.581 $\pm{}$ 0.000}   & \multicolumn{1}{c|}{0.791 $\pm{}$ 0.000} \\ \hline
Rank Disagreement                                       & \textbf{0.939 $\pm{}$ 0.000}                 & \multicolumn{1}{c|}{\textbf{0.939 $\pm{}$ 0.000}} & \multicolumn{1}{c|}{\textbf{0.939 $\pm{}$ 0.000}}   & \textbf{0.939 $\pm{}$ 0.000}                      & \multicolumn{1}{c|}{0.980 $\pm{}$ 0.000}    & \multicolumn{1}{c|}{0.984 $\pm{}$ 0.000}   & 0.984 $\pm{}$ 0.000                      & \multicolumn{1}{c|}{0.980 $\pm{}$ 0.000}             & \multicolumn{1}{c|}{0.984 $\pm{}$ 0.000}   & 0.984 $\pm{}$ 0.000                      \\ \hline
Prediction Disagreement                                 & 0.153 $\pm{}$ 0.001                        & \multicolumn{1}{c|}{0.152 $\pm{}$ 0.001}        & \multicolumn{1}{c|}{\textbf{0.151 $\pm{}$ 0.001}} & \textbf{0.151 $\pm{}$ 0.001}                    & \multicolumn{1}{c|}{0.190 $\pm{}$ 0.001}  & \multicolumn{1}{c|}{0.214 $\pm{}$ 0.000}   & 0.324 $\pm{}$ 0.000                      & \multicolumn{1}{c|}{0.189 $\pm{}$ 0.001}          & \multicolumn{1}{c|}{0.214 $\pm{}$ 0.000}   & 0.324 $\pm{}$ 0.000                      \\ \hline
JS Divergence                                           & 0.040 $\pm{}$ 0.000                           & \multicolumn{1}{c|}{0.040 $\pm{}$ 0.000}           & \multicolumn{1}{c|}{\textbf{0.039 $\pm{}$ 0.000}}   & \textbf{0.039 $\pm{}$ 0.000}                      & \multicolumn{1}{c|}{0.171 $\pm{}$ 0.000}   & \multicolumn{1}{c|}{0.333 $\pm{}$ 0.000}   & 0.533 $\pm{}$ 0.000                      & \multicolumn{1}{c|}{0.170 $\pm{}$ 0.000}             & \multicolumn{1}{c|}{0.333 $\pm{}$ 0.000}   & 0.533 $\pm{}$ 0.000                      \\ \hline
Accuracy                                                & 0.605 $\pm{}$ 0.001                        & \multicolumn{1}{c|}{0.605 $\pm{}$ 0.000}          & \multicolumn{1}{c|}{0.604 $\pm{}$ 0.001}          & 0.605 $\pm{}$ 0.001                             & \multicolumn{1}{c|}{0.607 $\pm{}$ 0.000}   & \multicolumn{1}{c|}{0.606 $\pm{}$ 0.001} & 0.580 $\pm{}$ 0.000                       & \multicolumn{1}{c|}{\textbf{0.608 $\pm{}$ 0.000}}   & \multicolumn{1}{c|}{0.605 $\pm{}$ 0.000}   & 0.580 $\pm{}$ 0.000                       \\ \hline
Loss                                                    & 2.068 $\pm{}$ 0.001                        & \multicolumn{1}{c|}{2.065 $\pm{}$ 0.002}        & \multicolumn{1}{c|}{2.055 $\pm{}$ 0.001}          & 2.043 $\pm{}$ 0.002                             & \multicolumn{1}{c|}{1.977 $\pm{}$ 0.001} & \multicolumn{1}{c|}{2.497 $\pm{}$ 0.001} & 3.612 $\pm{}$ 0.002                    & \multicolumn{1}{c|}{\textbf{1.976 $\pm{}$ 0.001}} & \multicolumn{1}{c|}{2.498 $\pm{}$ 0.001} & 3.612 $\pm{}$ 0.002                    \\ \hline
\end{tabular}%
}
\end{table}
\section{Knowledge Distillation to Smaller Student}
\label{app:standard-kd-ss}
In this section we present the raw results for the other teacher seeds explored in this paper (1-2) to show that our findings are not the effect of a single seed but are general across teacher models and training seeds. 

\subsection{TinyImageNet ResNet50 Teacher to ResNet18 Student}

\paragraph{Training Settings:}The ResNet50 teacher model was trained with stochastic gradient descent with a learning rate of 0.01 and a Cosine annealing learning rate scheduler with a T\_max set at 100. It was trained for 100 epochs with a batch size of 256. The data was normalized with a mean of (0.485, 0.456, 0.406) and a standard deviation of (0.229, 0.224, 0.225). The ResNet18 student model was trained under the same conditions. 
\begin{table}[H]
\centering
\caption{Teacher Performance on Train and Test Data for ResNet50 on Tiny ImageNet}
\label{tab:resnet50-tin-small-teacher}
\begin{tabular}{|c|c|c|c|c|}
\hline
\textbf{Teacher Seed} & \textbf{Train Loss} & \textbf{Train Accuracy} & \textbf{Test Loss} & \textbf{Test Accuracy} \\  \hline
0                     & 0.001426            & 0.999800                & 2.070590           & 0.605300                   \\ \hline
1                     & 0.001393            & 0.999800                & 2.051494           & 0.607900               \\ \hline
2                     & 0.001436            & 0.999800                & 2.051024           & 0.610600               \\ \hline
\end{tabular}%
\end{table}
\begin{table}[H]
\centering
\caption{ResNet18 on TinyImageNet mean and $\pm$ 1 SEM  reported from 10 runs with Teacher Seed 1. Bold values are best performing based on the mean. }
\label{tab:tin-resnet18-ts-1}
\resizebox{\textwidth}{!}{%
\begin{tabular}{|c|c|ccc|ccc|}
\hline
\multicolumn{1}{|c|}{\multirow{2}{*}{\textbf{Metrics}}} & \multicolumn{1}{c|}{\textbf{Control}} & \multicolumn{3}{c|}{\textbf{Knowledge Distillation}} & \multicolumn{3}{c|}{\textbf{Random Control Distillation}} \\ \cline{2-8} 
\multicolumn{1}{|c|}{} & \multicolumn{1}{c|}{\textbf{SIDDO}} & \multicolumn{1}{c|}{\textbf{0.1}} & \multicolumn{1}{c|}{\textbf{0.5}} & \multicolumn{1}{c|}{\textbf{0.9}} & \multicolumn{1}{c|}{\textbf{0.1}} & \multicolumn{1}{c|}{\textbf{0.5}} & \multicolumn{1}{c|}{\textbf{0.9}} \\ \hline
Activation Distance & 0.548 $\pm{}$ 0.000 & \multicolumn{1}{c|}{0.548 $\pm{}$ 0.000} & \multicolumn{1}{c|}{0.548 $\pm{}$ 0.000} & \textbf{0.547 $\pm{}$ 0.000} & \multicolumn{1}{c|}{0.567 $\pm{}$ 0.000} & \multicolumn{1}{c|}{0.651 $\pm{}$ 0.000} & 0.829 $\pm{}$ 0.000 \\ \hline
Rank Disagreement & \textbf{0.987 $\pm{}$ 0.000} & \multicolumn{1}{c|}{\textbf{0.987 $\pm{}$ 0.000}} & \multicolumn{1}{c|}{\textbf{0.987 $\pm{}$ 0.000}} & \textbf{0.987 $\pm{}$ 0.000} & \multicolumn{1}{c|}{0.990 $\pm{}$ 0.000} & \multicolumn{1}{c|}{0.990 $\pm{}$ 0.000} & 0.991 $\pm{}$ 0.000 \\ \hline
Prediction Disagreement & 0.497 $\pm{}$ 0.001 & \multicolumn{1}{c|}{0.497 $\pm{}$ 0.001} & \multicolumn{1}{c|}{0.497 $\pm{}$ 0.001} & \textbf{0.496 $\pm{}$ 0.001} & \multicolumn{1}{c|}{0.511 $\pm{}$ 0.001} & \multicolumn{1}{c|}{0.489 $\pm{}$ 0.000} & 0.762 $\pm{}$ 0.000 \\ \hline
JS Divergence & 0.281 $\pm{}$ 0.000 & \multicolumn{1}{c|}{0.281 $\pm{}$ 0.000} & \multicolumn{1}{c|}{0.281 $\pm{}$ 0.000} & \textbf{0.280 $\pm{}$ 0.000} & \multicolumn{1}{c|}{0.331 $\pm{}$ 0.000} & \multicolumn{1}{c|}{0.401 $\pm{}$ 0.000} & 0.601 $\pm{}$ 0.000 \\ \hline
Accuracy & 0.503 $\pm{}$ 0.000 & \multicolumn{1}{c|}{0.504 $\pm{}$ 0.000} & \multicolumn{1}{c|}{0.504 $\pm{}$ 0.000} & 0.504 $\pm{}$ 0.000 & \multicolumn{1}{c|}{0.494 $\pm{}$ 0.000} & \multicolumn{1}{c|}{\textbf{0.513 $\pm{}$ 0.001}} & 0.232 $\pm{}$ 0.000 \\ \hline
Loss & 2.608 $\pm{}$ 0.002 & \multicolumn{1}{c|}{2.606 $\pm{}$ 0.002} & \multicolumn{1}{c|}{2.599 $\pm{}$ 0.002} & 2.591 $\pm{}$ 0.003 & \multicolumn{1}{c|}{\textbf{2.431 $\pm{}$ 0.002}} & \multicolumn{1}{c|}{2.634 $\pm{}$ 0.001} & 4.703 $\pm{}$ 0.002 \\ \hline
\end{tabular}%
}
\end{table}
\begin{table}[H]
\centering
\caption{ResNet18 on TinyImageNet mean and $\pm$ 1 SEM  reported from 10 runs with Teacher Seed 2. Bold values are best performing based on the mean. }
\label{tab:tin-resnet18-ts-2}
\resizebox{\textwidth}{!}{%
\begin{tabular}{|c|c|ccc|ccc|}
\hline
\multicolumn{1}{|c|}{\multirow{2}{*}{\textbf{Metrics}}} & \multicolumn{1}{c|}{\textbf{Control}} & \multicolumn{3}{c|}{\textbf{Knowledge Distillation}} & \multicolumn{3}{c|}{\textbf{Random Control Distillation}} \\ \cline{2-8} 
\multicolumn{1}{|c|}{} & \multicolumn{1}{c|}{\textbf{SIDDO}} & \multicolumn{1}{c|}{\textbf{0.1}} & \multicolumn{1}{c|}{\textbf{0.5}} & \multicolumn{1}{c|}{\textbf{0.9}} & \multicolumn{1}{c|}{\textbf{0.1}} & \multicolumn{1}{c|}{\textbf{0.5}} & \multicolumn{1}{c|}{\textbf{0.9}} \\ \hline
Activation Distance & 0.546 $\pm{}$ 0.000 & \multicolumn{1}{c|}{\textbf{0.545 $\pm{}$ 0.000}} & \multicolumn{1}{c|}{\textbf{0.545 $\pm{}$ 0.000}} & \textbf{0.545 $\pm{}$ 0.000} & \multicolumn{1}{c|}{0.565 $\pm{}$ 0.000} & \multicolumn{1}{c|}{0.651 $\pm{}$ 0.000} & 0.829 $\pm{}$ 0.000 \\ \hline
Rank Disagreement & \textbf{0.987 $\pm{}$ 0.000} & \multicolumn{1}{c|}{\textbf{0.987 $\pm{}$ 0.000}} & \multicolumn{1}{c|}{\textbf{0.987 $\pm{}$ 0.000}} & \textbf{0.987 $\pm{}$ 0.000} & \multicolumn{1}{c|}{0.990 $\pm{}$ 0.000} & \multicolumn{1}{c|}{0.990 $\pm{}$ 0.000} & 0.991 $\pm{}$ 0.000 \\ \hline
Prediction Disagreement & 0.497 $\pm{}$ 0.001 & \multicolumn{1}{c|}{0.497 $\pm{}$ 0.001} & \multicolumn{1}{c|}{0.497 $\pm{}$ 0.001} & \textbf{0.496 $\pm{}$ 0.001} & \multicolumn{1}{c|}{0.511 $\pm{}$ 0.001} & \multicolumn{1}{c|}{0.489 $\pm{}$ 0.000} & 0.755 $\pm{}$ 0.000 \\ \hline
JS Divergence & \textbf{0.280 $\pm{}$ 0.000} & \multicolumn{1}{c|}{\textbf{0.280 $\pm{}$ 0.000}} & \multicolumn{1}{c|}{\textbf{0.280 $\pm{}$ 0.000}} & \textbf{0.280 $\pm{}$ 0.000} & \multicolumn{1}{c|}{0.330 $\pm{}$ 0.000} & \multicolumn{1}{c|}{0.400 $\pm{}$ 0.000} & 0.600 $\pm{}$ 0.000 \\ \hline
Accuracy & 0.503 $\pm{}$ 0.001 & \multicolumn{1}{c|}{0.504 $\pm{}$ 0.000} & \multicolumn{1}{c|}{0.503 $\pm{}$ 0.000} & 0.503 $\pm{}$ 0.000 & \multicolumn{1}{c|}{0.493 $\pm{}$ 0.000} & \multicolumn{1}{c|}{\textbf{0.512 $\pm{}$ 0.000}} & 0.236 $\pm{}$ 0.000 \\ \hline
Loss & 2.604 $\pm{}$ 0.001 & \multicolumn{1}{c|}{2.602 $\pm{}$ 0.001} & \multicolumn{1}{c|}{2.594 $\pm{}$ 0.001} & 2.587 $\pm{}$ 0.001 & \multicolumn{1}{c|}{\textbf{2.434 $\pm{}$ 0.001}} & \multicolumn{1}{c|}{2.641 $\pm{}$ 0.001} & 4.684 $\pm{}$ 0.002 \\ \hline
\end{tabular}%
}
\end{table}
\subsection{ImageNet ResNet50 Teacher to ResNet18 Student}
\label{app:imagenet}
\paragraph{Training Settings:} A pre-trained ResNet50 model taken from PyTorch with a top-1-accuracy of 80.858 and a top-5-accuracy of 95.434\footnote{\url{https://docs.pytorch.org/vision/main/models/generated/torchvision.models.resnet50.html\#torchvision.models.ResNet50_Weights}} is used as the teacher. As Pytorch only provides one set of pre-trained model weights there is only one teacher seed for this experiment. The ResNet18 student was trained on ImageNet \citep{deng2009imagenet} using the FFCV setup \citep{leclerc2023ffcv}, where 100\% of the training images were compressed to a JPEG with 90\% quality. The data was normalized with a mean of (0.485, 0.456, 0.406) and a standard deviation of (0.229, 0.224, 0.225). The model utilised BlurPools \citep{zhang2019making} within the convolutional layers, and was trained for 56 epochs, with a batch size of 1024 using SGD, momentum of 0.9, weight decay of 5e-5, a learning rate of 0.5 using a cyclic scheduler with a learning rate step ratio of 0,1 and step length of 30. The learning rate peak was at epoch 2.  The input resolution started at 160 by 160, and started to ramp up to 192 by 192 at epoch 41 and ended at 192 by 192 at epoch 48.
\subsubsection{The Effect Of Temperature}
\label{sec:imagenet-temp}
This section presents the raw results for temperature of 2 on the ImageNet dataset on seed 0, due to computational constraints this test was only conducted on seed 0. 
\begin{table}[H]
\centering
\caption{ResNet18 student on ImageNet Dataset mean and $\pm$ 1 SEM  reported from 10 runs with Teacher Seed 0. Bold values are best performing based on the mean. }
\label{tab:imagenet-ts-0-t-2}
\resizebox{\textwidth}{!}{%
\begin{tabular}{|l|l|lll|lll|}
\hline
\multicolumn{1}{|c|}{\multirow{2}{*}{\textbf{Metrics}}} & \multicolumn{1}{c|}{\textbf{Control}} & \multicolumn{3}{c|}{\textbf{Knowledge Distillation}}                                                                               & \multicolumn{3}{c|}{\textbf{Random Control Distillation}}                                                                         \\ \cline{2-8} 
\multicolumn{1}{|c|}{}                                  & \multicolumn{1}{c|}{\textbf{SIDDO}}   & \multicolumn{1}{c|}{\textbf{KD 0.1}}         & \multicolumn{1}{c|}{\textbf{KD 0.5}}         & \multicolumn{1}{c|}{\textbf{KD 0.9}} & \multicolumn{1}{c|}{\textbf{Rand KD 0.1}} & \multicolumn{1}{c|}{\textbf{Rand KD 0.5}} & \multicolumn{1}{c|}{\textbf{Rand KD 0.9}} \\ \hline
Activation Distance                                     & 0.420 $\pm{}$ 0.001                        & \multicolumn{1}{l|}{0.310 $\pm{}$ 0.001}          & \multicolumn{1}{l|}{0.251 $\pm{}$ 0.001}          & \textbf{0.221 $\pm{}$ 0.001}              & \multicolumn{1}{l|}{0.304 $\pm{}$ 0.001}       & \multicolumn{1}{l|}{0.247 $\pm{}$ 0.001}       & 0.282 $\pm{}$ 0.001                            \\ \hline
Rank Disagreement                                       & 0.997 $\pm{}$ 0.000                          & \multicolumn{1}{l|}{0.997 $\pm{}$ 0.000}            & \multicolumn{1}{l|}{0.997 $\pm{}$ 0.000}            & \textbf{0.996 $\pm{}$ 0.000}                & \multicolumn{1}{l|}{0.997 $\pm{}$ 0.000}         & \multicolumn{1}{l|}{0.997 $\pm{}$ 0.000}         & 0.997 $\pm{}$ 0.000                              \\ \hline
Prediction Disagreement                                 & 0.263 $\pm{}$ 0.002                        & \multicolumn{1}{l|}{0.257 $\pm{}$ 0.001}          & \multicolumn{1}{l|}{\textbf{0.256 $\pm{}$ 0.001}} & 0.264 $\pm{}$ 0.001                       & \multicolumn{1}{l|}{0.259 $\pm{}$ 0.001}       & \multicolumn{1}{l|}{0.273 $\pm{}$ 0.001}       & 0.310 $\pm{}$ 0.001                             \\ \hline
JS Divergence                                           & 0.259 $\pm{}$ 0.000                          & \multicolumn{1}{l|}{0.160 $\pm{}$ 0.000}            & \multicolumn{1}{l|}{0.101 $\pm{}$ 0.000}            & \textbf{0.081 $\pm{}$ 0.000}                & \multicolumn{1}{l|}{0.151 $\pm{}$ 0.001}       & \multicolumn{1}{l|}{0.096 $\pm{}$ 0.000}         & 0.116 $\pm{}$ 0.001                            \\ \hline
Accuracy                                                & 0.681 $\pm{}$ 0.001                        & \multicolumn{1}{l|}{\textbf{0.685 $\pm{}$ 0.001}} & \multicolumn{1}{l|}{\textbf{0.685 $\pm{}$ 0.001}} & 0.679 $\pm{}$ 0.001                       & \multicolumn{1}{l|}{0.684 $\pm{}$ 0.001}       & \multicolumn{1}{l|}{0.671 $\pm{}$ 0.001}       & 0.641 $\pm{}$ 0.001                            \\ \hline
Loss                                                    & \textbf{1.303 $\pm{}$ 0.006}               & \multicolumn{1}{l|}{1.492 $\pm{}$ 0.008}          & \multicolumn{1}{l|}{1.718 $\pm{}$ 0.012}          & 1.922 $\pm{}$ 0.013                       & \multicolumn{1}{l|}{1.532 $\pm{}$ 0.009}       & \multicolumn{1}{l|}{1.919 $\pm{}$ 0.015}       & 3.022 $\pm{}$ 0.014                            \\ \hline
\end{tabular}}
\end{table}

\subsection{TinyShakespeare Nano-GPT Teacher to Pico-GPT Student} \label{sec:nano2pico}
 \begin{table}[H]
\centering
\caption{Teacher Performance on Train and Test Data for Nano-GPT on TinyShakespeare.}
\label{tab:gpt-shakespeare-small-teacher}
\begin{tabular}{|c|c|c|c|c|}
\hline
\textbf{Teacher Seed} & \textbf{Train Loss} & \textbf{Train Accuracy} & \textbf{Test Loss} & \textbf{Test Accuracy} \\ \hline
0 & 0.864641 & 0.719685 & 1.567481 & 0.573366 \\ \hline
1 & 0.866370 & 0.719697 & 1.561079 & 0.574668 \\ \hline
2 & 0.861098 & 0.721140 & 1.562137 & 0.573033 \\ \hline
\end{tabular}
\end{table}

\begin{table}[H]
\centering
\caption{Pico-GPT on TinyShakespeare Dataset mean and $\pm$ 1 SEM  reported from 10 runs with Teacher Seed 1. Bold values are best performing based on the mean. }
\label{tab:shake-micro-gpt-ts-1}
\resizebox{\textwidth}{!}{%
\begin{tabular}{|c|c|ccc|ccc|}
\hline
\multicolumn{1}{|c|}{\multirow{2}{*}{\textbf{Metrics}}} & \multicolumn{1}{c|}{\textbf{Control}} & \multicolumn{3}{c|}{\textbf{Knowledge Distillation}} & \multicolumn{3}{c|}{\textbf{Random Control Distillation}} \\ \cline{2-8} 
\multicolumn{1}{|c|}{} & \multicolumn{1}{c|}{\textbf{SIDDO}} & \multicolumn{1}{c|}{\textbf{0.1}} & \multicolumn{1}{c|}{\textbf{0.5}} & \multicolumn{1}{c|}{\textbf{0.9}} & \multicolumn{1}{c|}{\textbf{0.1}} & \multicolumn{1}{c|}{\textbf{0.5}} & \multicolumn{1}{c|}{\textbf{0.9}} \\ \hline
Activation Distance & 0.201 $\pm{}$ 0.000 & \multicolumn{1}{c|}{0.196 $\pm{}$ 0.000} & \multicolumn{1}{c|}{0.180 $\pm{}$ 0.000} & \textbf{0.170 $\pm{}$ 0.000} & \multicolumn{1}{c|}{0.217 $\pm{}$ 0.000} & \multicolumn{1}{c|}{0.392 $\pm{}$ 0.000} & 0.655 $\pm{}$ 0.000 \\ \hline
Rank Disagreement & 0.916 $\pm{}$ 0.000 & \multicolumn{1}{c|}{0.915 $\pm{}$ 0.000} & \multicolumn{1}{c|}{0.912 $\pm{}$ 0.000} & \textbf{0.911 $\pm{}$ 0.000} & \multicolumn{1}{c|}{0.939 $\pm{}$ 0.000} & \multicolumn{1}{c|}{0.944 $\pm{}$ 0.000} & 0.950 $\pm{}$ 0.000 \\ \hline
Prediction Disagreement & 0.257 $\pm{}$ 0.000 & \multicolumn{1}{c|}{0.251 $\pm{}$ 0.000} & \multicolumn{1}{c|}{0.231 $\pm{}$ 0.000} & \textbf{0.219 $\pm{}$ 0.000} & \multicolumn{1}{c|}{0.256 $\pm{}$ 0.000} & \multicolumn{1}{c|}{0.258 $\pm{}$ 0.000} & 0.277 $\pm{}$ 0.001 \\ \hline
JS Divergence & 0.055 $\pm{}$ 0.000 & \multicolumn{1}{c|}{0.053 $\pm{}$ 0.000} & \multicolumn{1}{c|}{0.046 $\pm{}$ 0.000} & \textbf{0.043 $\pm{}$ 0.000} & \multicolumn{1}{c|}{0.074 $\pm{}$ 0.000} & \multicolumn{1}{c|}{0.201 $\pm{}$ 0.000} & 0.449 $\pm{}$ 0.000 \\ \hline
Accuracy & 0.571 $\pm{}$ 0.000 & \multicolumn{1}{c|}{0.573 $\pm{}$ 0.000} & \multicolumn{1}{c|}{\textbf{0.575 $\pm{}$ 0.000}} & 0.574 $\pm{}$ 0.000 & \multicolumn{1}{c|}{0.571 $\pm{}$ 0.000} & \multicolumn{1}{c|}{0.570 $\pm{}$ 0.000} & 0.561 $\pm{}$ 0.000 \\ \hline
Loss & \textbf{1.473 $\pm{}$ 0.002} & \multicolumn{1}{c|}{\textbf{1.473 $\pm{}$ 0.002}} & \multicolumn{1}{c|}{1.475 $\pm{}$ 0.002} & 1.492 $\pm{}$ 0.002 & \multicolumn{1}{c|}{1.483 $\pm{}$ 0.001} & \multicolumn{1}{c|}{1.870 $\pm{}$ 0.001} & 3.017 $\pm{}$ 0.002 \\ \hline
\end{tabular}%
}
\end{table}
\begin{table}[H]
\centering
\caption{Pico-GPT on TinyShakespeare Dataset mean and $\pm$ 1 SEM  reported from 10 runs with Teacher Seed 2. Bold values are best performing based on the mean. }
\label{tab:shake-micro-gpt-ts-2}
\resizebox{\textwidth}{!}{%
\begin{tabular}{|c|c|ccc|ccc|}
\hline
\multicolumn{1}{|c|}{\multirow{2}{*}{\textbf{Metrics}}} & \multicolumn{1}{c|}{\textbf{Control}} & \multicolumn{3}{c|}{\textbf{Knowledge Distillation}} & \multicolumn{3}{c|}{\textbf{Random Control Distillation}} \\ \cline{2-8} 
\multicolumn{1}{|c|}{} & \multicolumn{1}{c|}{\textbf{SIDDO}} & \multicolumn{1}{c|}{\textbf{0.1}} & \multicolumn{1}{c|}{\textbf{0.5}} & \multicolumn{1}{c|}{\textbf{0.9}} & \multicolumn{1}{c|}{\textbf{0.1}} & \multicolumn{1}{c|}{\textbf{0.5}} & \multicolumn{1}{c|}{\textbf{0.9}} \\ \hline
Activation Distance & 0.202 $\pm{}$ 0.000 & \multicolumn{1}{c|}{0.197 $\pm{}$ 0.000} & \multicolumn{1}{c|}{0.180 $\pm{}$ 0.000} & \textbf{0.171 $\pm{}$ 0.000} & \multicolumn{1}{c|}{0.219 $\pm{}$ 0.000} & \multicolumn{1}{c|}{0.395 $\pm{}$ 0.001} & 0.660 $\pm{}$ 0.000 \\ \hline
Rank Disagreement & 0.915 $\pm{}$ 0.000 & \multicolumn{1}{c|}{0.914 $\pm{}$ 0.000} & \multicolumn{1}{c|}{0.912 $\pm{}$ 0.000} & \textbf{0.910 $\pm{}$ 0.000} & \multicolumn{1}{c|}{0.939 $\pm{}$ 0.000} & \multicolumn{1}{c|}{0.944 $\pm{}$ 0.000} & 0.949 $\pm{}$ 0.000 \\ \hline
Prediction Disagreement & 0.252 $\pm{}$ 0.000 & \multicolumn{1}{c|}{0.246 $\pm{}$ 0.000} & \multicolumn{1}{c|}{0.226 $\pm{}$ 0.000} & \textbf{0.215 $\pm{}$ 0.000} & \multicolumn{1}{c|}{0.250 $\pm{}$ 0.001} & \multicolumn{1}{c|}{0.251 $\pm{}$ 0.000} & 0.272 $\pm{}$ 0.001 \\ \hline
JS Divergence & 0.055 $\pm{}$ 0.000 & \multicolumn{1}{c|}{0.053 $\pm{}$ 0.000} & \multicolumn{1}{c|}{0.046 $\pm{}$ 0.000} & \textbf{0.043 $\pm{}$ 0.000} & \multicolumn{1}{c|}{0.074 $\pm{}$ 0.000} & \multicolumn{1}{c|}{0.202 $\pm{}$ 0.000} & 0.450 $\pm{}$ 0.000 \\ \hline
Accuracy & 0.571 $\pm{}$ 0.000 & \multicolumn{1}{c|}{0.572 $\pm{}$ 0.000} & \multicolumn{1}{c|}{\textbf{0.575 $\pm{}$ 0.000}} & 0.574 $\pm{}$ 0.000 & \multicolumn{1}{c|}{0.572 $\pm{}$ 0.000} & \multicolumn{1}{c|}{0.571 $\pm{}$ 0.000} & 0.561 $\pm{}$ 0.000 \\ \hline
Loss & 1.475 $\pm{}$ 0.001 & \multicolumn{1}{c|}{\textbf{1.470 $\pm{}$ 0.001}} & \multicolumn{1}{c|}{1.471 $\pm{}$ 0.002} & 1.491 $\pm{}$ 0.002 & \multicolumn{1}{c|}{1.482 $\pm{}$ 0.001} & \multicolumn{1}{c|}{1.865 $\pm{}$ 0.002} & 3.017 $\pm{}$ 0.001 \\ \hline
\end{tabular}%
}
\end{table}

\subsubsection{The Effect Of Temperature}
\label{sec:shake-temp}
This section provides the raw results for temperature 2 and 4 on the TinyShakespeare dataset.
\vspace{-0.2cm}
\begin{table}[H]
\caption{Pico-GPT with a Nano-GPT teacher with Temperature 2 on TinyShakespeare mean and $\pm$ 1 SEM  reported from 10 runs with Teacher Seed 0. Bold values are best performing based on the mean.}
\label{tab:shake-micro-gpt-temp-2-ts-0}
\resizebox{\textwidth}{!}{%
\begin{tabular}{|c|c|ccc|ccc|}
\hline
\multicolumn{1}{|c|}{\multirow{2}{*}{\textbf{Metrics}}} & \multicolumn{1}{c|}{\textbf{Control}} & \multicolumn{3}{c|}{\textbf{Knowledge Distillation}}                                                            & \multicolumn{3}{c|}{\textbf{Random Control Distillation}}                                  \\ \cline{2-8} 
\multicolumn{1}{|c|}{}                                  & \textbf{SIDDO}                        & \multicolumn{1}{c|}{\textbf{0.1}}          & \multicolumn{1}{c|}{\textbf{0.5}}          & \textbf{0.9}          & \multicolumn{1}{c|}{\textbf{0.1}}   & \multicolumn{1}{c|}{\textbf{0.5}}   & \textbf{0.9}   \\ \hline
Activation Distance                                     & 0.202 $\pm{}$ 0.000                          & \multicolumn{1}{c|}{0.197 $\pm{}$ 0.000}          & \multicolumn{1}{c|}{0.183 $\pm{}$ 0.000}          & \textbf{0.181 $\pm{}$ 0.000} & \multicolumn{1}{c|}{0.213 $\pm{}$ 0.000}   & \multicolumn{1}{c|}{0.305 $\pm{}$ 0.001} & 0.617 $\pm{}$ 0.000   \\ \hline
Rank Disagreement                                       & 0.915 $\pm{}$ 0.000                          & \multicolumn{1}{c|}{0.907 $\pm{}$ 0.000}          & \multicolumn{1}{c|}{0.896 $\pm{}$ 0.000}          & \textbf{0.892 $\pm{}$ 0.000} & \multicolumn{1}{c|}{0.94 $\pm{}$ 0.000}    & \multicolumn{1}{c|}{0.945 $\pm{}$ 0.000}   & 0.95 $\pm{}$ 0.000    \\ \hline
Prediction Disagreement                                 & 0.252 $\pm{}$ 0.000                          & \multicolumn{1}{c|}{0.25 $\pm{}$ 0.000}           & \multicolumn{1}{c|}{0.235 $\pm{}$ 0.000}          & \textbf{0.23 $\pm{}$ 0.000}  & \multicolumn{1}{c|}{0.252 $\pm{}$ 0.000}   & \multicolumn{1}{c|}{0.253 $\pm{}$ 0.000}   & 0.27 $\pm{}$ 0.000    \\ \hline
JS Divergence                                           & 0.056 $\pm{}$ 0.000                          & \multicolumn{1}{c|}{0.053 $\pm{}$ 0.000}          & \multicolumn{1}{c|}{\textbf{0.047 $\pm{}$ 0.000}} & \textbf{0.047 $\pm{}$ 0.000} & \multicolumn{1}{c|}{0.072 $\pm{}$ 0.000}   & \multicolumn{1}{c|}{0.152 $\pm{}$ 0.000}   & 0.403 $\pm{}$ 0.000   \\ \hline
Accuracy                                                & 0.571 $\pm{}$ 0.000                          & \multicolumn{1}{c|}{\textbf{0.572 $\pm{}$ 0.000}} & \multicolumn{1}{c|}{\textbf{0.572 $\pm{}$ 0.000}} & 0.569 $\pm{}$ 0.000          & \multicolumn{1}{c|}{0.571 $\pm{}$ 0.000}   & \multicolumn{1}{c|}{0.571 $\pm{}$ 0.000}   & 0.562 $\pm{}$ 0.000   \\ \hline
Loss                                                    & \textbf{1.473 $\pm{}$ 0.002}               & \multicolumn{1}{c|}{1.513 $\pm{}$ 0.003}        & \multicolumn{1}{c|}{1.571 $\pm{}$ 0.002}        & 1.622 $\pm{}$ 0.002        & \multicolumn{1}{c|}{1.493 $\pm{}$ 0.001} & \multicolumn{1}{c|}{1.736 $\pm{}$ 0.001} & 2.732 $\pm{}$ 0.001 \\ \hline
\end{tabular}%
}
\end{table}
\vspace{-0.2cm}
\begin{table}[H]
\caption{Pico-GPT with a Nano-GPT teacher with Temperature 2 on TinyShakespeare mean and $\pm$ 1 SEM  reported from 10 runs with Teacher Seed 1. Bold values are best performing based on the mean.}
\label{tab:shake-micro-gpt-temp-2-ts-1}
\resizebox{\textwidth}{!}{%
\begin{tabular}{|c|c|ccc|ccc|}
\hline
\multicolumn{1}{|c|}{\multirow{2}{*}{\textbf{Metrics}}} & \multicolumn{1}{c|}{\textbf{Control}} & \multicolumn{3}{c|}{\textbf{Knowledge Distillation}}                                                     & \multicolumn{3}{c|}{\textbf{Random Control Distillation}}                                         \\ \cline{2-8} 
\multicolumn{1}{|c|}{}                                  & \textbf{SIDDO}                        & \multicolumn{1}{c|}{\textbf{0.1}}   & \multicolumn{1}{c|}{\textbf{0.5}}          & \textbf{0.9}          & \multicolumn{1}{c|}{\textbf{0.1}}          & \multicolumn{1}{c|}{\textbf{0.5}}   & \textbf{0.9}   \\ \hline
Activation Distance                                     & 0.201 $\pm{}$ 0.000                          & \multicolumn{1}{c|}{0.195 $\pm{}$ 0.000}   & \multicolumn{1}{c|}{0.181 $\pm{}$ 0.000}          & \textbf{0.179 $\pm{}$ 0.000} & \multicolumn{1}{c|}{0.209 $\pm{}$ 0.000}          & \multicolumn{1}{c|}{0.298 $\pm{}$ 0.000}   & 0.609 $\pm{}$ 0.000   \\ \hline
Rank Disagreement                                       & 0.916 $\pm{}$ 0.000                          & \multicolumn{1}{c|}{0.907 $\pm{}$ 0.000}   & \multicolumn{1}{c|}{0.896 $\pm{}$ 0.000}          & \textbf{0.892 $\pm{}$ 0.000} & \multicolumn{1}{c|}{0.94 $\pm{}$ 0.000}           & \multicolumn{1}{c|}{0.945 $\pm{}$ 0.000}   & 0.95 $\pm{}$ 0.000    \\ \hline
Prediction Disagreement                                 & 0.258 $\pm{}$ 0.001                        & \multicolumn{1}{c|}{0.254 $\pm{}$ 0.000}   & \multicolumn{1}{c|}{0.24 $\pm{}$ 0.000}           & \textbf{0.236 $\pm{}$ 0.000} & \multicolumn{1}{c|}{0.256 $\pm{}$ 0.000}          & \multicolumn{1}{c|}{0.258 $\pm{}$ 0.000}   & 0.279 $\pm{}$ 0.000   \\ \hline
JS Divergence                                           & 0.055 $\pm{}$ 0.000                          & \multicolumn{1}{c|}{0.052 $\pm{}$ 0.000}   & \multicolumn{1}{c|}{0.047 $\pm{}$ 0.000}          & \textbf{0.046 $\pm{}$ 0.000} & \multicolumn{1}{c|}{0.071 $\pm{}$ 0.000}          & \multicolumn{1}{c|}{0.15 $\pm{}$ 0.000}    & 0.401 $\pm{}$ 0.000   \\ \hline
Accuracy                                                & 0.571 $\pm{}$ 0.000                          & \multicolumn{1}{c|}{0.571 $\pm{}$ 0.000}   & \multicolumn{1}{c|}{\textbf{0.572 $\pm{}$ 0.000}} & 0.569 $\pm{}$ 0.000          & \multicolumn{1}{c|}{\textbf{0.572 $\pm{}$ 0.000}} & \multicolumn{1}{c|}{0.571 $\pm{}$ 0.000}   & 0.56 $\pm{}$ 0.000    \\ \hline
Loss                                                    & \textbf{1.474 $\pm{}$ 0.002}               & \multicolumn{1}{c|}{1.512 $\pm{}$ 0.003} & \multicolumn{1}{c|}{1.569 $\pm{}$ 0.002}        & 1.613 $\pm{}$ 0.003        & \multicolumn{1}{c|}{1.489 $\pm{}$ 0.001}        & \multicolumn{1}{c|}{1.732 $\pm{}$ 0.001} & 2.739 $\pm{}$ 0.001 \\ \hline
\end{tabular}%
}
\end{table}
\vspace{-0.2cm}
\begin{table}[H]
\caption{Pico-GPT with a Nano-GPT teacher with Temperature 2 on TinyShakespeare mean and $\pm$ 1 SEM  reported from 10 runs with Teacher Seed 2. Bold values are best performing based on the mean.}
\label{tab:shake-micro-gpt-temp-2-ts-2}
\resizebox{\textwidth}{!}{%
\begin{tabular}{|c|c|ccc|ccc|}
\hline
\multicolumn{1}{|c|}{\multirow{2}{*}{\textbf{Metrics}}} & \multicolumn{1}{c|}{\textbf{Control}} & \multicolumn{3}{c|}{\textbf{Knowledge Distillation}}                                                     & \multicolumn{3}{c|}{\textbf{Random Control Distillation}}                                         \\ \cline{2-8} 
\multicolumn{1}{|c|}{}                                  & \textbf{SIDDO}                        & \multicolumn{1}{c|}{\textbf{0.1}}   & \multicolumn{1}{c|}{\textbf{0.5}}          & \textbf{0.9}          & \multicolumn{1}{c|}{\textbf{0.1}}          & \multicolumn{1}{c|}{\textbf{0.5}}   & \textbf{0.9}   \\ \hline
Activation Distance                                     & 0.201 $\pm{}$ 0.000                          & \multicolumn{1}{c|}{0.195 $\pm{}$ 0.000}   & \multicolumn{1}{c|}{0.181 $\pm{}$ 0.000}          & \textbf{0.18 $\pm{}$ 0.000}  & \multicolumn{1}{c|}{0.21 $\pm{}$ 0.000}           & \multicolumn{1}{c|}{0.301 $\pm{}$ 0.000}   & 0.615 $\pm{}$ 0.000   \\ \hline
Rank Disagreement                                       & 0.915 $\pm{}$ 0.000                          & \multicolumn{1}{c|}{0.906 $\pm{}$ 0.000}   & \multicolumn{1}{c|}{0.896 $\pm{}$ 0.000}          & \textbf{0.892 $\pm{}$ 0.000} & \multicolumn{1}{c|}{0.94 $\pm{}$ 0.000}           & \multicolumn{1}{c|}{0.945 $\pm{}$ 0.000}   & 0.95 $\pm{}$ 0.000    \\ \hline
Prediction Disagreement                                 & 0.251 $\pm{}$ 0.001                        & \multicolumn{1}{c|}{0.247 $\pm{}$ 0.000}   & \multicolumn{1}{c|}{0.235 $\pm{}$ 0.000}          & \textbf{0.23 $\pm{}$ 0.000}  & \multicolumn{1}{c|}{0.249 $\pm{}$ 0.000}          & \multicolumn{1}{c|}{0.252 $\pm{}$ 0.000}   & 0.274 $\pm{}$ 0.000   \\ \hline
JS Divergence                                           & 0.055 $\pm{}$ 0.000                          & \multicolumn{1}{c|}{0.052 $\pm{}$ 0.000}   & \multicolumn{1}{c|}{\textbf{0.046 $\pm{}$ 0.000}} & \textbf{0.046 $\pm{}$ 0.000} & \multicolumn{1}{c|}{0.071 $\pm{}$ 0.000}          & \multicolumn{1}{c|}{0.15 $\pm{}$ 0.000}    & 0.403 $\pm{}$ 0.000   \\ \hline
Accuracy                                                & 0.571 $\pm{}$ 0.000                          & \multicolumn{1}{c|}{0.571 $\pm{}$ 0.000}   & \multicolumn{1}{c|}{0.571 $\pm{}$ 0.000}          & 0.569 $\pm{}$ 0.000          & \multicolumn{1}{c|}{\textbf{0.572 $\pm{}$ 0.000}} & \multicolumn{1}{c|}{0.571 $\pm{}$ 0.000}   & 0.56 $\pm{}$ 0.000    \\ \hline
Loss                                                    & \textbf{1.474 $\pm{}$ 0.002}               & \multicolumn{1}{c|}{1.513 $\pm{}$ 0.001} & \multicolumn{1}{c|}{1.576 $\pm{}$ 0.001}        & 1.619 $\pm{}$ 0.003        & \multicolumn{1}{c|}{1.489 $\pm{}$ 0.001}        & \multicolumn{1}{c|}{1.732 $\pm{}$ 0.001} & 2.739 $\pm{}$ 0.001 \\ \hline
\end{tabular}%
}
\end{table}
\begin{figure}[H]
    \centering
    \includegraphics[width=\linewidth]{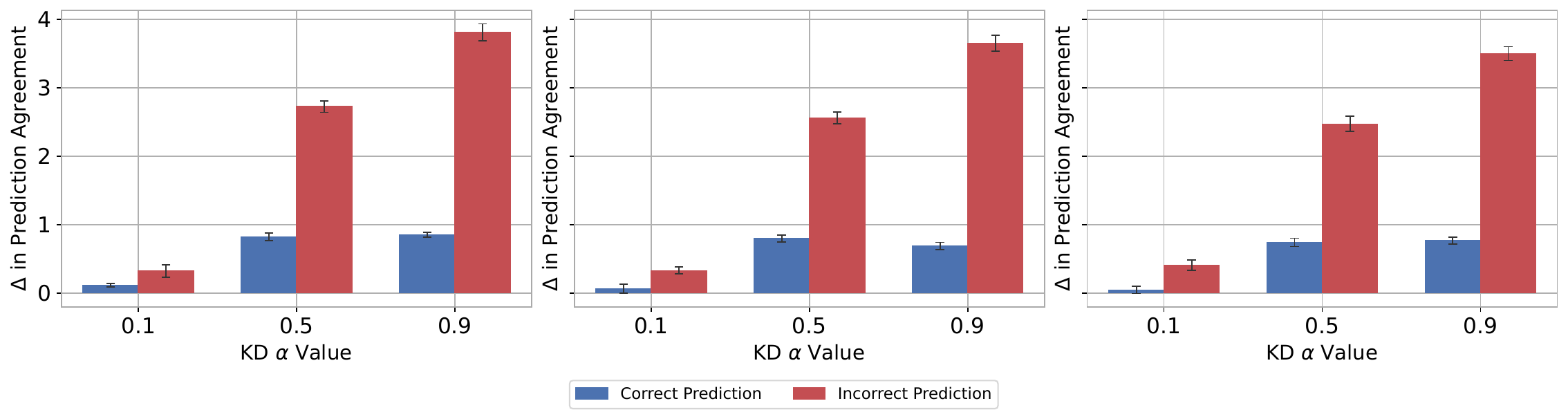}
   
    \caption{Prediction agreement difference of student models in standard KD with temperature 2 to the highest performing control baseline with respect to correct prediction agreement (blue) and incorrect prediction agreement (red), for Pico-GPT on TinyShakespeare (seeds 0 to 2, left to right).}
    \label{fig:pico-gpt-shake-prediction}
\end{figure}

\begin{table}[H]
\caption{Pico-GPT with a Nano-GPT teacher with temperature 4 on TinyShakespeare mean and $\pm$ 1 SEM  reported from 10 runs with Teacher Seed 0. Bold values are best performing based on the mean.}
\label{tab:shake-micro-gpt-temp-4-ts-0}
\resizebox{\textwidth}{!}{%
\begin{tabular}{|c|c|ccc|ccc|}
\hline
\multicolumn{1}{|c|}{\multirow{2}{*}{\textbf{Metrics}}} & \multicolumn{1}{c|}{\textbf{Control}} & \multicolumn{3}{c|}{\textbf{Knowledge Distillation}}                                                     & \multicolumn{3}{c|}{\textbf{Random Control Distillation}}                                       \\ \cline{2-8} 
\multicolumn{1}{|c|}{}                                  & \textbf{SIDDO}                        & \multicolumn{1}{c|}{\textbf{0.1}}   & \multicolumn{1}{c|}{\textbf{0.5}}          & \textbf{0.9}          & \multicolumn{1}{c|}{\textbf{0.1}}          & \multicolumn{1}{c|}{\textbf{0.5}} & \textbf{0.9}   \\ \hline
Activation Distance                                     & 0.202 $\pm{}$ 0.000                          & \multicolumn{1}{c|}{0.199 $\pm{}$ 0.000}   & \multicolumn{1}{c|}{\textbf{0.189 $\pm{}$ 0.000}} & 0.193 $\pm{}$ 0.000          & \multicolumn{1}{c|}{0.206 $\pm{}$ 0.000}          & \multicolumn{1}{c|}{0.262 $\pm{}$ 0.000} & 0.568 $\pm{}$ 0.001 \\ \hline
Rank Disagreement                                       & 0.915 $\pm{}$ 0.000                          & \multicolumn{1}{c|}{0.893 $\pm{}$ 0.000}   & \multicolumn{1}{c|}{0.88 $\pm{}$ 0.000}           & \textbf{0.876 $\pm{}$ 0.000} & \multicolumn{1}{c|}{0.94 $\pm{}$ 0.000}           & \multicolumn{1}{c|}{0.945 $\pm{}$ 0.000} & 0.951 $\pm{}$ 0.000   \\ \hline
Prediction Disagreement                                 & 0.252 $\pm{}$ 0.000                          & \multicolumn{1}{c|}{0.251 $\pm{}$ 0.001} & \multicolumn{1}{c|}{\textbf{0.244 $\pm{}$ 0.000}} & 0.245 $\pm{}$ 0.000          & \multicolumn{1}{c|}{0.253 $\pm{}$ 0.000}          & \multicolumn{1}{c|}{0.253 $\pm{}$ 0.000} & 0.27 $\pm{}$ 0.000    \\ \hline
JS Divergence                                           & 0.056 $\pm{}$ 0.000                          & \multicolumn{1}{c|}{0.054 $\pm{}$ 0.000}   & \multicolumn{1}{c|}{\textbf{0.05 $\pm{}$ 0.000}}  & 0.051 $\pm{}$ 0.000          & \multicolumn{1}{c|}{0.067 $\pm{}$ 0.000}          & \multicolumn{1}{c|}{0.127 $\pm{}$ 0.000} & 0.362 $\pm{}$ 0.000   \\ \hline
Accuracy                                                & 0.571 $\pm{}$ 0.000                          & \multicolumn{1}{c|}{0.57 $\pm{}$ 0.000}    & \multicolumn{1}{c|}{0.568 $\pm{}$ 0.000}          & 0.562 $\pm{}$ 0.000          & \multicolumn{1}{c|}{\textbf{0.572 $\pm{}$ 0.000}} & \multicolumn{1}{c|}{0.571 $\pm{}$ 0.000} & 0.562 $\pm{}$ 0.000   \\ \hline
Loss                                                    & \textbf{1.473 $\pm{}$ 0.002}               & \multicolumn{1}{c|}{1.528 $\pm{}$ 0.002} & \multicolumn{1}{c|}{1.592 $\pm{}$ 0.002}        & 1.663 $\pm{}$ 0.002        & \multicolumn{1}{c|}{1.491 $\pm{}$ 0.002}        & \multicolumn{1}{c|}{1.68 $\pm{}$ 0.000}  & 2.544 $\pm{}$ 0.002 \\ \hline
\end{tabular}%
}
\end{table}
\begin{table}[H]
\caption{Pico-GPT with a Nano-GPT teacher with temperature 4 on TinyShakespeare mean and $\pm$ 1 SEM  reported from 10 runs with Teacher Seed 1. Bold values are best performing based on the mean.}
\label{tab:shake-micro-gpt-temp-4-ts-1}
\resizebox{\textwidth}{!}{%
\begin{tabular}{|c|c|ccc|ccc|}
\hline
\multicolumn{1}{|c|}{\multirow{2}{*}{\textbf{Metrics}}} & \multicolumn{1}{c|}{\textbf{Control}} & \multicolumn{3}{c|}{\textbf{Knowledge Distillation}}                                                     & \multicolumn{3}{c|}{\textbf{Random Control Distillation}}                                               \\ \cline{2-8} 
\multicolumn{1}{|c|}{}                                  & \textbf{SIDDO}                        & \multicolumn{1}{c|}{\textbf{0.1}}   & \multicolumn{1}{c|}{\textbf{0.5}}          & \textbf{0.9}          & \multicolumn{1}{c|}{\textbf{0.1}}          & \multicolumn{1}{c|}{\textbf{0.5}}          & \textbf{0.9}  \\ \hline
Activation Distance                                     & 0.201 $\pm{}$ 0.000                          & \multicolumn{1}{c|}{0.196 $\pm{}$ 0.000}   & \multicolumn{1}{c|}{\textbf{0.188 $\pm{}$ 0.000}} & 0.191 $\pm{}$ 0.000          & \multicolumn{1}{c|}{0.203 $\pm{}$ 0.000}          & \multicolumn{1}{c|}{0.256 $\pm{}$ 0.000}          & 0.562 $\pm{}$ 0.000  \\ \hline
Rank Disagreement                                       & 0.916 $\pm{}$ 0.000                          & \multicolumn{1}{c|}{0.893 $\pm{}$ 0.000}   & \multicolumn{1}{c|}{0.88 $\pm{}$ 0.000}           & \textbf{0.876 $\pm{}$ 0.000} & \multicolumn{1}{c|}{0.94 $\pm{}$ 0.000}           & \multicolumn{1}{c|}{0.945 $\pm{}$ 0.000}          & 0.951 $\pm{}$ 0.000  \\ \hline
Prediction Disagreement                                 & 0.258 $\pm{}$ 0.001                        & \multicolumn{1}{c|}{0.256 $\pm{}$ 0.000}   & \multicolumn{1}{c|}{0.25 $\pm{}$ 0.000}           & \textbf{0.249 $\pm{}$ 0.000} & \multicolumn{1}{c|}{0.256 $\pm{}$ 0.000}          & \multicolumn{1}{c|}{0.258 $\pm{}$ 0.000}          & 0.278 $\pm{}$ 0.000  \\ \hline
JS Divergence                                           & 0.055 $\pm{}$ 0.000                          & \multicolumn{1}{c|}{0.052 $\pm{}$ 0.000}   & \multicolumn{1}{c|}{\textbf{0.049 $\pm{}$ 0.000}} & 0.05 $\pm{}$ 0.000           & \multicolumn{1}{c|}{0.066 $\pm{}$ 0.000}          & \multicolumn{1}{c|}{0.126 $\pm{}$ 0.000}          & 0.361 $\pm{}$ 0.000  \\ \hline
Accuracy                                                & \textbf{0.571 $\pm{}$ 0.000}                 & \multicolumn{1}{c|}{0.57 $\pm{}$ 0.000}    & \multicolumn{1}{c|}{0.568 $\pm{}$ 0.000}          & 0.563 $\pm{}$ 0.000          & \multicolumn{1}{c|}{\textbf{0.571 $\pm{}$ 0.000}} & \multicolumn{1}{c|}{\textbf{0.571 $\pm{}$ 0.000}} & 0.561 $\pm{}$ 0.000  \\ \hline
Loss                                                    & \textbf{1.474 $\pm{}$ 0.002}               & \multicolumn{1}{c|}{1.528 $\pm{}$ 0.002} & \multicolumn{1}{c|}{1.59 $\pm{}$ 0.002}         & 1.653 $\pm{}$ 0.003        & \multicolumn{1}{c|}{1.489 $\pm{}$ 0.001}        & \multicolumn{1}{c|}{1.677 $\pm{}$ 0.001}        & 2.55 $\pm{}$ 0.002 \\ \hline
\end{tabular}%
}
\end{table}
\begin{table}[H]
\caption{Pico-GPT with a Nano-GPT teacher with temperature 4 on TinyShakespeare mean and $\pm$ 1 SEM  reported from 10 runs with Teacher Seed 2. Bold values are best performing based on the mean.}
\label{tab:shake-micro-gpt-temp-4-ts-2}
\resizebox{\textwidth}{!}{%
\begin{tabular}{|c|c|ccc|ccc|}
\hline
\multicolumn{1}{|c|}{\multirow{2}{*}{\textbf{Metrics}}} & \multicolumn{1}{c|}{\textbf{Control}} & \multicolumn{3}{c|}{\textbf{Knowledge Distillation}}                                                      & \multicolumn{3}{c|}{\textbf{Random Control Distillation}}                                               \\ \cline{2-8} 
\multicolumn{1}{|c|}{}                                  & \textbf{SIDDO}                        & \multicolumn{1}{c|}{\textbf{0.1}}  & \multicolumn{1}{c|}{\textbf{0.5}}            & \textbf{0.9}          & \multicolumn{1}{c|}{\textbf{0.1}}          & \multicolumn{1}{c|}{\textbf{0.5}}          & \textbf{0.9}  \\ \hline
Activation Distance                                     & 0.201 $\pm{}$ 0.000                          & \multicolumn{1}{c|}{0.197 $\pm{}$ 0.000}  & \multicolumn{1}{c|}{\textbf{0.189 $\pm{}$ 0.000}}   & 0.192 $\pm{}$ 0.000          & \multicolumn{1}{c|}{0.204 $\pm{}$ 0.000}          & \multicolumn{1}{c|}{0.259 $\pm{}$ 0.000}          & 0.567 $\pm{}$ 0.000  \\ \hline
Rank Disagreement                                       & 0.915 $\pm{}$ 0.000                          & \multicolumn{1}{c|}{0.893 $\pm{}$ 0.000}  & \multicolumn{1}{c|}{0.879 $\pm{}$ 0.000}            & \textbf{0.876 $\pm{}$ 0.000} & \multicolumn{1}{c|}{0.94 $\pm{}$ 0.000}           & \multicolumn{1}{c|}{0.945 $\pm{}$ 0.000}          & 0.951 $\pm{}$ 0.000  \\ \hline
Prediction Disagreement                                 & 0.251 $\pm{}$ 0.001                        & \multicolumn{1}{c|}{0.25 $\pm{}$ 0.001} & \multicolumn{1}{c|}{\textbf{0.245 $\pm{}$ 0.001}} & \textbf{0.245 $\pm{}$ 0.000} & \multicolumn{1}{c|}{0.25 $\pm{}$ 0.001}         & \multicolumn{1}{c|}{0.253 $\pm{}$ 0.000}          & 0.275 $\pm{}$ 0.000  \\ \hline
JS Divergence                                           & 0.055 $\pm{}$ 0.000                          & \multicolumn{1}{c|}{0.053 $\pm{}$ 0.000}  & \multicolumn{1}{c|}{\textbf{0.049 $\pm{}$ 0.000}}   & 0.05 $\pm{}$ 0.000           & \multicolumn{1}{c|}{0.066 $\pm{}$ 0.000}          & \multicolumn{1}{c|}{0.127 $\pm{}$ 0.000}          & 0.363 $\pm{}$ 0.000  \\ \hline
Accuracy                                                & \textbf{0.571 $\pm{}$ 0.000}                 & \multicolumn{1}{c|}{0.57 $\pm{}$ 0.000}   & \multicolumn{1}{c|}{0.568 $\pm{}$ 0.000}            & 0.562 $\pm{}$ 0.000          & \multicolumn{1}{c|}{\textbf{0.571 $\pm{}$ 0.000}} & \multicolumn{1}{c|}{\textbf{0.571 $\pm{}$ 0.000}} & 0.561 $\pm{}$ 0.000  \\ \hline
Loss                                                    & \textbf{1.474 $\pm{}$ 0.002}               & \multicolumn{1}{c|}{1.53 $\pm{}$ 0.001} & \multicolumn{1}{c|}{1.594 $\pm{}$ 0.002}          & 1.658 $\pm{}$ 0.002        & \multicolumn{1}{c|}{1.489 $\pm{}$ 0.001}        & \multicolumn{1}{c|}{1.677 $\pm{}$ 0.001}        & 2.55 $\pm{}$ 0.002 \\ \hline
\end{tabular}%
}
\end{table}
\begin{figure}[H]
    \centering
    \includegraphics[width=\linewidth]{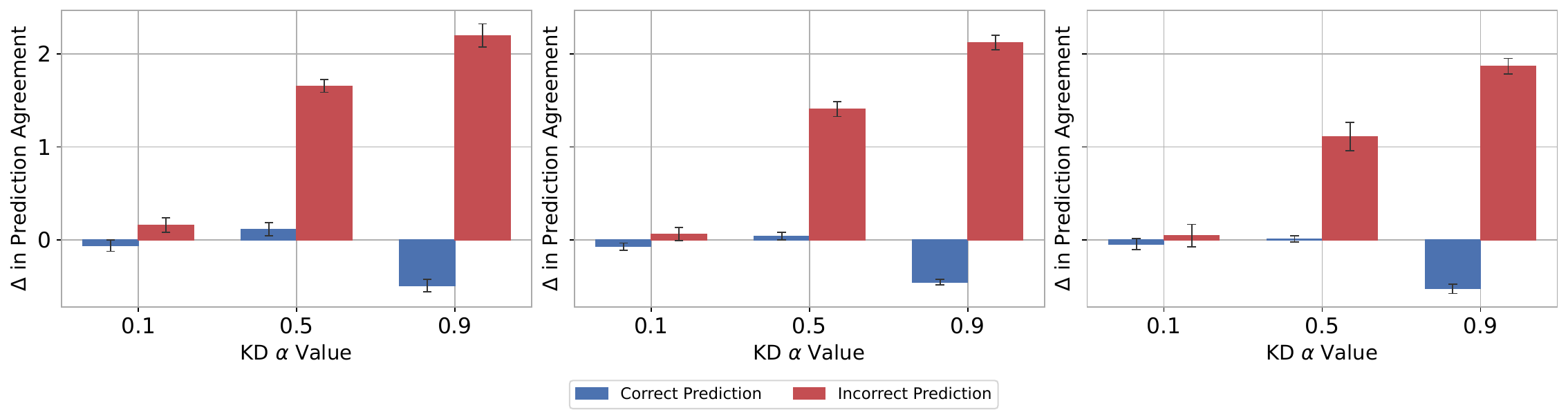}
    \caption{Prediction agreement difference of student models in standard KD with temperature 4 to the highest performing control baseline with respect to correct prediction agreement (blue) and incorrect prediction agreement (red), for Pico-GPT on TinyShakespeare (seeds 0 to 2, left to right).}
    \label{fig:pico-nano-gpt-shake-prediction}
\end{figure}
\section{Vision Results}
\subsection{TinyImageNet}
\label{app:tinyImageNet}
\paragraph{Training Settings:}The ResNet50 model was trained with stochastic gradient descent with a learning rate 0.01, along with a Cosine annealing learning rate scheduler with a T\_max set at 100. It was trained for 100 epochs with a batch size of 256. The data was normalized with a mean of (0.485, 0.456, 0.406) and standard deviation of (0.229, 0.224, 0.225). For ResNet50 with RandAugment \citep{cubuk2020randaugment}, the only difference between base ResNet is the introduction of RandAugment with the default setting provided in Pytorch 2.4~\citep{pytorch}. The VGG19 and VGG19 with RandAugment has the same setup as the ResNet50 and ResNet50 with RandAugment respectively however it was trained \textbf{with} momentum of 0.9. 

\subsubsection{ResNet50}
Here we present the results of the ResNet50 on TinyImageNet in the Self Distillation regime for teacher seeds 1 and 2 - the  corresponding analysis can be found in the main body in Section~\ref{sec:self-distillation}. 
\begin{table}[H]
\caption{ResNet50 on TinyImageNet mean and $\pm$ 1 SEM  reported from 10 runs with Teacher Seed 1. \textbf{Bold} values are the best performing based on the mean.}
\label{tab:tin-resnet50-ts-1}
\resizebox{\textwidth}{!}{%
\begin{tabular}{|c|c|ccc|ccc|}
\hline
\multicolumn{1}{|c|}{\multirow{2}{*}{\textbf{Metrics}}} & \multicolumn{1}{c|}{\textbf{Control}} & \multicolumn{3}{c|}{\textbf{Knowledge Distillation}} & \multicolumn{3}{c|}{\textbf{Random Control Distillation}} \\ \cline{2-8} 
\multicolumn{1}{|c|}{} & \multicolumn{1}{c|}{\textbf{SIDDO}} & \multicolumn{1}{c|}{\textbf{0.1}} & \multicolumn{1}{c|}{\textbf{0.5}} & \multicolumn{1}{c|}{\textbf{0.9}} & \multicolumn{1}{c|}{\textbf{0.1}} & \multicolumn{1}{c|}{\textbf{0.5}} & \multicolumn{1}{c|}{\textbf{0.9}} \\ \hline
Activation Distance & \multicolumn{1}{c|}{0.156 $\pm{}$ 0.001} & \multicolumn{1}{c|}{0.156 $\pm{}$ 0.000}  & \multicolumn{1}{c|}{0.155 $\pm{}$ 0.001} & \multicolumn{1}{c|}{\textbf{0.153 $\pm{}$ 0.000} } & \multicolumn{1}{c|}{0.340 $\pm{}$ 0.000}  & \multicolumn{1}{c|}{0.579 $\pm{}$ 0.000}  & \multicolumn{1}{c|}{0.792 $\pm{}$ 0.000}  \\ \hline
Rank Disagreement & 0.940 $\pm{}$ 0.000 & \multicolumn{1}{c|}{0.940 $\pm{}$ 0.000}  & \multicolumn{1}{c|}{\textbf{0.939 $\pm{}$ 0.000} } & \textbf{0.939 $\pm{}$ 0.000}  & \multicolumn{1}{c|}{0.980 $\pm{}$ 0.000}  & \multicolumn{1}{c|}{0.984 $\pm{}$ 0.000}  & 0.984 $\pm{}$ 0.000 \\ \hline
Prediction Disagreement & 0.148 $\pm{}$ 0.001 & \multicolumn{1}{c|}{0.149 $\pm{}$ 0.001} & \multicolumn{1}{c|}{0.148 $\pm{}$ 0.001} & \textbf{0.146 $\pm{}$ 0.001} & \multicolumn{1}{c|}{0.185 $\pm{}$ 0.001} & \multicolumn{1}{c|}{0.209 $\pm{}$ 0.000}  & 0.330 $\pm{}$ 0.000 \\ \hline
JS Divergence & 0.040 $\pm{}$ 0.000 & \multicolumn{1}{c|}{0.040 $\pm{}$ 0.000}  & \multicolumn{1}{c|}{0.039 $\pm{}$ 0.000}  & \textbf{0.038 $\pm{}$ 0.000}  & \multicolumn{1}{c|}{0.170 $\pm{}$ 0.000}  & \multicolumn{1}{c|}{0.332 $\pm{}$ 0.000}  & 0.534 $\pm{}$ 0.000 \\ \hline
Accuracy & 0.607 $\pm{}$ 0.001 & \multicolumn{1}{c|}{\textbf{0.608 $\pm{}$ 0.001}} & \multicolumn{1}{c|}{0.607 $\pm{}$ 0.000}  & 0.607 $\pm{}$ 0.001 & \multicolumn{1}{c|}{0.605 $\pm{}$ 0.000}  & \multicolumn{1}{c|}{0.602 $\pm{}$ 0.001} & 0.576 $\pm{}$ 0.000 \\ \hline
Loss & 2.048 $\pm{}$ 0.002 & \multicolumn{1}{c|}{2.048 $\pm{}$ 0.002} & \multicolumn{1}{c|}{2.034 $\pm{}$ 0.002} & 2.025 $\pm{}$ 0.002 & \multicolumn{1}{c|}{\textbf{1.973 $\pm{}$ 0.001}} & \multicolumn{1}{c|}{2.498 $\pm{}$ 0.001} & 3.611 $\pm{}$ 0.002 \\ \hline
\end{tabular}%
}
\end{table}
\begin{table}[H]
\caption{ResNet50 on TinyImageNet mean and $\pm$ 1 SEM  reported from 10 runs with Teacher Seed 2. \textbf{Bold} values are the best performing based on the mean.}
\label{tab:tin-resnet50-ts-2}
\resizebox{\textwidth}{!}{%
\begin{tabular}{|c|c|ccc|ccc|}
\hline
\multicolumn{1}{|c|}{\multirow{2}{*}{\textbf{Metrics}}} & \multicolumn{1}{c|}{\textbf{Control}} & \multicolumn{3}{c|}{\textbf{Knowledge Distillation}} & \multicolumn{3}{c|}{\textbf{Random Control Distillation}} \\ \cline{2-8} 
\multicolumn{1}{|c|}{} & \multicolumn{1}{c|}{\textbf{SIDDO}} & \multicolumn{1}{c|}{\textbf{0.1}} & \multicolumn{1}{c|}{\textbf{0.5}} & \multicolumn{1}{c|}{\textbf{0.9}} & \multicolumn{1}{c|}{\textbf{0.1}} & \multicolumn{1}{c|}{\textbf{0.5}} & \multicolumn{1}{c|}{\textbf{0.9}}  \\ \hline
Activation Distance & 0.157 $\pm{}$ 0.000 & \multicolumn{1}{c|}{0.157 $\pm{}$ 0.000}  & \multicolumn{1}{c|}{\textbf{0.155 $\pm{}$ 0.000} } & \textbf{0.155 $\pm{}$ 0.000}  & \multicolumn{1}{c|}{0.342 $\pm{}$ 0.000}  & \multicolumn{1}{c|}{0.581 $\pm{}$ 0.000}  & 0.792 $\pm{}$ 0.000 \\ \hline
Rank Disagreement & \textbf{0.939 $\pm{}$ 0.000}  & \multicolumn{1}{c|}{\textbf{0.939 $\pm{}$ 0.000} } & \multicolumn{1}{c|}{\textbf{0.939 $\pm{}$ 0.000} } & \textbf{0.939 $\pm{}$ 0.000}  & \multicolumn{1}{c|}{0.980 $\pm{}$ 0.000}  & \multicolumn{1}{c|}{0.984 $\pm{}$ 0.000}  & 0.984 $\pm{}$ 0.000 \\ \hline
Prediction Disagreement & 0.152 $\pm{}$ 0.001 & \multicolumn{1}{c|}{0.152 $\pm{}$ 0.001} & \multicolumn{1}{c|}{\textbf{0.151 $\pm{}$ 0.001}} & \textbf{0.151 $\pm{}$ 0.001} & \multicolumn{1}{c|}{0.187 $\pm{}$ 0.001} & \multicolumn{1}{c|}{0.213 $\pm{}$ 0.001} & 0.327 $\pm{}$ 0.000 \\ \hline
JS Divergence & 0.040 $\pm{}$ 0.000 & \multicolumn{1}{c|}{0.040 $\pm{}$ 0.000}  & \multicolumn{1}{c|}{\textbf{0.039 $\pm{}$ 0.000} } & \textbf{0.039 $\pm{}$ 0.000}  & \multicolumn{1}{c|}{0.171 $\pm{}$ 0.000}  & \multicolumn{1}{c|}{0.334 $\pm{}$ 0.000}  & 0.534 $\pm{}$ 0.000 \\ \hline
Accuracy & 0.608 $\pm{}$ 0.001 & \multicolumn{1}{c|}{0.607 $\pm{}$ 0.001} & \multicolumn{1}{c|}{0.607 $\pm{}$ 0.000}  & \textbf{0.609 $\pm{}$ 0.001} & \multicolumn{1}{c|}{0.608 $\pm{}$ 0.001} & \multicolumn{1}{c|}{0.605 $\pm{}$ 0.001} & 0.577 $\pm{}$ 0.000 \\ \hline
Loss & 2.054 $\pm{}$ 0.002 & \multicolumn{1}{c|}{2.050 $\pm{}$ 0.002} & \multicolumn{1}{c|}{2.040 $\pm{}$ 0.003} & 2.025 $\pm{}$ 0.002 & \multicolumn{1}{c|}{\textbf{1.967 $\pm{}$ 0.001}} & \multicolumn{1}{c|}{2.494 $\pm{}$ 0.001} & 3.602 $\pm{}$ 0.002 \\ \hline
\end{tabular}%
}
\end{table}

\subsubsection{ResNet50 with RandAugment}
Here we present the results of the ResNet50 on TinyImageNet using RandAugment for student and teacher in the Self Distillation regime for teacher seeds 1 and 2 - the  corresponding analysis can be found in the main body in Section~\ref{sec:self-distillation}.  
\begin{table}[H]
\caption{ResNet50 on TinyImageNet with RandAugment mean and $\pm$ 1 SEM  reported from 10 runs with Teacher Seed 1. Bold values are the best performing based on the mean.}
\label{tab:tina-resnet-ts-1}
\resizebox{\textwidth}{!}{%
\begin{tabular}{|c|c|ccc|ccc|}
\hline
\multicolumn{1}{|c|}{\multirow{2}{*}{\textbf{Metrics}}} & \multicolumn{1}{c|}{\textbf{Control}} & \multicolumn{3}{c|}{\textbf{Knowledge Distillation}} & \multicolumn{3}{c|}{\textbf{Random Control Distillation}} \\ \cline{2-8} 
\multicolumn{1}{|c|}{} & \multicolumn{1}{c|}{\textbf{SIDDO}} & \multicolumn{1}{c|}{\textbf{0.1}} & \multicolumn{1}{c|}{\textbf{0.5}} & \multicolumn{1}{c|}{\textbf{0.9}} & \multicolumn{1}{c|}{\textbf{0.1}} & \multicolumn{1}{c|}{\textbf{0.5}} & \multicolumn{1}{c|}{\textbf{0.9}} \\ \hline
Activation Distance & 0.194 $\pm{}$ 0.000 & \multicolumn{1}{c|}{0.183 $\pm{}$ 0.001} & \multicolumn{1}{c|}{0.148 $\pm{}$ 0.000}  & \textbf{0.13 $\pm{}$ 0.000}  & \multicolumn{1}{c|}{0.247 $\pm{}$ 0.000}  & \multicolumn{1}{c|}{0.503 $\pm{}$ 0.000}  & 0.783 $\pm{}$ 0.000 \\ \hline
Rank Disagreement & 0.959 $\pm{}$ 0.000 & \multicolumn{1}{c|}{0.957 $\pm{}$ 0.000}  & \multicolumn{1}{c|}{0.948 $\pm{}$ 0.000}  & \textbf{0.943 $\pm{}$ 0.000}  & \multicolumn{1}{c|}{0.975 $\pm{}$ 0.000}  & \multicolumn{1}{c|}{0.981 $\pm{}$ 0.000}  & 0.987 $\pm{}$ 0.000 \\ \hline
Prediction Disagreement & 0.195 $\pm{}$ 0.001 & \multicolumn{1}{c|}{0.186 $\pm{}$ 0.001} & \multicolumn{1}{c|}{0.151 $\pm{}$ 0.001} & \textbf{0.134 $\pm{}$ 0.001} & \multicolumn{1}{c|}{0.194 $\pm{}$ 0.001} & \multicolumn{1}{c|}{0.241 $\pm{}$ 0.000}  & 0.577 $\pm{}$ 0.001 \\ \hline
JS Divergence & 0.058 $\pm{}$ 0.000 & \multicolumn{1}{c|}{0.053 $\pm{}$ 0.000}  & \multicolumn{1}{c|}{0.036 $\pm{}$ 0.000}  & \textbf{0.028 $\pm{}$ 0.000}  & \multicolumn{1}{c|}{0.095 $\pm{}$ 0.000}  & \multicolumn{1}{c|}{0.267 $\pm{}$ 0.000}  & 0.565 $\pm{}$ 0.000 \\ \hline
Accuracy & 0.639 $\pm{}$ 0.001 & \multicolumn{1}{c|}{0.640 $\pm{}$ 0.001} & \multicolumn{1}{c|}{0.641 $\pm{}$ 0.001} & 0.640 $\pm{}$ 0.001 & \multicolumn{1}{c|}{0.646 $\pm{}$ 0.001} & \multicolumn{1}{c|}{\textbf{0.658 $\pm{}$ 0.000} } & 0.396 $\pm{}$ 0.001 \\ \hline
Loss & 1.620 $\pm{}$ 0.002 & \multicolumn{1}{c|}{1.608 $\pm{}$ 0.002} & \multicolumn{1}{c|}{1.584 $\pm{}$ 0.001} & 1.584 $\pm{}$ 0.001 & \multicolumn{1}{c|}{\textbf{1.555 $\pm{}$ 0.002}} & \multicolumn{1}{c|}{1.986 $\pm{}$ 0.002} & 4.214 $\pm{}$ 0.002 \\ \hline
\end{tabular}%
}
\end{table}
\begin{table}[H]
\caption{ResNet50 on TinyImageNet with RandAugment mean and $\pm$ 1 SEM  reported from 10 runs with Teacher Seed 2. Bold values are the best performing based on the mean.}
\label{tab:tina-resnet-ts-2}
\resizebox{\textwidth}{!}{%
\begin{tabular}{|c|c|ccc|ccc|}
\hline
\multicolumn{1}{|c|}{\multirow{2}{*}{\textbf{Metrics}}} & \multicolumn{1}{c|}{\textbf{Control}} & \multicolumn{3}{c|}{\textbf{Knowledge Distillation}} & \multicolumn{3}{c|}{\textbf{Random Control Distillation}} \\ \cline{2-8} 
\multicolumn{1}{|c|}{} & \multicolumn{1}{c|}{\textbf{SIDDO}} & \multicolumn{1}{c|}{\textbf{0.1}} & \multicolumn{1}{c|}{\textbf{0.5}} & \multicolumn{1}{c|}{\textbf{0.9}} & \multicolumn{1}{c|}{\textbf{0.1}} & \multicolumn{1}{c|}{\textbf{0.5}} & \multicolumn{1}{c|}{\textbf{0.9}} \\ \hline
Activation Distance & 0.195 $\pm{}$ 0.000 & \multicolumn{1}{c|}{0.185 $\pm{}$ 0.000}  & \multicolumn{1}{c|}{0.150 $\pm{}$ 0.000}  & \textbf{0.131 $\pm{}$ 0.000}  & \multicolumn{1}{c|}{0.247 $\pm{}$ 0.001} & \multicolumn{1}{c|}{0.504 $\pm{}$ 0.000}  & 0.783 $\pm{}$ 0.000 \\ \hline
Rank Disagreement & 0.959 $\pm{}$ 0.000 & \multicolumn{1}{c|}{0.957 $\pm{}$ 0.000}  & \multicolumn{1}{c|}{0.948 $\pm{}$ 0.000}  & \textbf{0.943 $\pm{}$ 0.000}  & \multicolumn{1}{c|}{0.975 $\pm{}$ 0.000}  & \multicolumn{1}{c|}{0.981 $\pm{}$ 0.000}  & 0.987 $\pm{}$ 0.000 \\ \hline
Prediction Disagreement & 0.197 $\pm{}$ 0.001 & \multicolumn{1}{c|}{0.189 $\pm{}$ 0.001} & \multicolumn{1}{c|}{0.155 $\pm{}$ 0.001} & \textbf{0.135 $\pm{}$ 0.001} & \multicolumn{1}{c|}{0.197 $\pm{}$ 0.001} & \multicolumn{1}{c|}{0.239 $\pm{}$ 0.000}  & 0.564 $\pm{}$ 0.001 \\ \hline
JS Divergence & 0.059 $\pm{}$ 0.000 & \multicolumn{1}{c|}{0.053 $\pm{}$ 0.000}  & \multicolumn{1}{c|}{0.037 $\pm{}$ 0.000}  & \textbf{0.028 $\pm{}$ 0.000}  & \multicolumn{1}{c|}{0.096 $\pm{}$ 0.000}  & \multicolumn{1}{c|}{0.267 $\pm{}$ 0.000}  & 0.563 $\pm{}$ 0.000 \\ \hline
Accuracy & 0.640 $\pm{}$ 0.001 & \multicolumn{1}{c|}{0.641 $\pm{}$ 0.001} & \multicolumn{1}{c|}{0.643 $\pm{}$ 0.001} & 0.643 $\pm{}$ 0.000 & \multicolumn{1}{c|}{0.647 $\pm{}$ 0.001} & \multicolumn{1}{c|}{\textbf{0.657 $\pm{}$ 0.000} } & 0.410 $\pm{}$ 0.001 \\ \hline
Loss & 1.621 $\pm{}$ 0.002 & \multicolumn{1}{c|}{1.606 $\pm{}$ 0.001} & \multicolumn{1}{c|}{1.581 $\pm{}$ 0.001} & 1.582 $\pm{}$ 0.001 & \multicolumn{1}{c|}{\textbf{1.552 $\pm{}$ 0.001}} & \multicolumn{1}{c|}{1.982 $\pm{}$ 0.002} & 4.180 $\pm{}$ 0.002 \\ \hline
\end{tabular}%
}
\end{table}

\subsubsection{VGG19}

\paragraph{Findings:} For the VGG19 on the TinyImageNet, we observe a low train loss of circa $0.000286$ and a train accuracy of 0.9998. As expected, given our results and discussion in the main body of the paper on the ResNet50, we see no statistically supported until an $\alpha$ of 0.9. With teacher seed 0 and 2 with an $\alpha$ of 0.9, we record statistically supported transfer for Activation Distance and for teacher seed 0 on JS Divergence, as seen in Table \ref{tab:vgg-tin-significance}. When we observe knowledge transfer with an $\alpha$ of 0.9, we observe a slight preference for positive agreement of test prediction; however, the results have a large SEM, and the amount of agreement is less than 0.5\%, making the results less reliable and insignificant in either transfer direction. 
\begin{table}[H]
\caption{Teacher Performance on Train and Test Data.}
\label{tab:vgg-tin-teacher}
\centering
\begin{tabular}{|c|c|c|c|c|}
\hline
\textbf{Teacher Seed} & \textbf{Train Loss} & \textbf{Train Accuracy} & \textbf{Test Loss} & \textbf{Test Accuracy} \\ \hline
0 & 0.000286 & 0.999800 & 3.351542 & 0.633200 \\ \hline
1 & 0.000286 & 0.999800 & 3.301587 & 0.637200 \\ \hline
2 & 0.000285 & 0.999800 & 3.311130 & 0.633500 \\ \hline
\end{tabular}%
\end{table}
\begin{table}[H]
\caption{VGG19 on TinyImageNet mean and $\pm$ 1 SEM  reported from 10 runs with Teacher Seed 0. \textbf{Bold} values are the best performing based on the mean.}
\label{tab:tin-vgg19-ts-0}
\resizebox{\textwidth}{!}{%
\begin{tabular}{|c|c|ccc|ccc|}
\hline
\multicolumn{1}{|c|}{\multirow{2}{*}{\textbf{Metrics}}} & \multicolumn{1}{c|}{\textbf{Control}} & \multicolumn{3}{c|}{\textbf{Knowledge Distillation}} & \multicolumn{3}{c|}{\textbf{Random Control Distillation}} \\ \cline{2-8} 
\multicolumn{1}{|c|}{} & \multicolumn{1}{c|}{\textbf{SIDDO}} & \multicolumn{1}{c|}{\textbf{0.1}} & \multicolumn{1}{c|}{\textbf{0.5}} & \multicolumn{1}{c|}{\textbf{0.9}} & \multicolumn{1}{c|}{\textbf{0.1}} & \multicolumn{1}{c|}{\textbf{0.5}} & \multicolumn{1}{c|}{\textbf{0.9}} \\ \hline
Activation Distance & \multicolumn{1}{c|}{0.418 $\pm{}$ 0.001} & \multicolumn{1}{c|}{0.419 $\pm{}$ 0.001} & \multicolumn{1}{c|}{0.418 $\pm{}$ 0.001} & \multicolumn{1}{c|}{\textbf{0.416 $\pm{}$ 0.001}} & \multicolumn{1}{c|}{0.522 $\pm{}$ 0.001} & \multicolumn{1}{c|}{0.741 $\pm{}$ 0.000}  & \multicolumn{1}{c|}{0.886 $\pm{}$ 0.000}  \\ \hline
Rank Disagreement & \textbf{0.978 $\pm{}$ 0.000}  & \multicolumn{1}{c|}{\textbf{0.978 $\pm{}$ 0.000} } & \multicolumn{1}{c|}{\textbf{0.978 $\pm{}$ 0.000} } & \textbf{0.978 $\pm{}$ 0.000}  & \multicolumn{1}{c|}{0.987 $\pm{}$ 0.000}  & \multicolumn{1}{c|}{0.988 $\pm{}$ 0.000}  & 0.989 $\pm{}$ 0.000 \\ \hline
Prediction Disagreement & \textbf{0.332 $\pm{}$ 0.001} & \multicolumn{1}{c|}{\textbf{0.332 $\pm{}$ 0.001}} & \multicolumn{1}{c|}{\textbf{0.332 $\pm{}$ 0.001}} & 0.330 $\pm{}$ 0.001 & \multicolumn{1}{c|}{0.348 $\pm{}$ 0.001} & \multicolumn{1}{c|}{0.381 $\pm{}$ 0.001} & 0.412 $\pm{}$ 0.000 \\ \hline
JS Divergence & 0.195 $\pm{}$ 0.000 & \multicolumn{1}{c|}{0.195 $\pm{}$ 0.000}  & \multicolumn{1}{c|}{0.195 $\pm{}$ 0.000}  & \textbf{0.194 $\pm{}$ 0.000}  & \multicolumn{1}{c|}{0.308 $\pm{}$ 0.001} & \multicolumn{1}{c|}{0.457 $\pm{}$ 0.000}  & 0.593 $\pm{}$ 0.000 \\ \hline
Accuracy & 0.635 $\pm{}$ 0.001 & \multicolumn{1}{c|}{0.635 $\pm{}$ 0.001} & \multicolumn{1}{c|}{0.636 $\pm{}$ 0.001} & \textbf{0.638 $\pm{}$ 0.001} & \multicolumn{1}{c|}{0.627 $\pm{}$ 0.001} & \multicolumn{1}{c|}{0.603 $\pm{}$ 0.001} & 0.576 $\pm{}$ 0.001 \\ \hline
Loss & 3.332 $\pm{}$ 0.010 & \multicolumn{1}{c|}{3.329 $\pm{}$ 0.012} & \multicolumn{1}{c|}{3.308 $\pm{}$ 0.011} & 3.313 $\pm{}$ 0.010 & \multicolumn{1}{c|}{\textbf{2.003 $\pm{}$ 0.005}} & \multicolumn{1}{c|}{2.732 $\pm{}$ 0.002} & 3.682 $\pm{}$ 0.002 \\ \hline
\end{tabular}%
}
\end{table}
\begin{table}[H]
\caption{VGG19 on TinyImageNet mean and $\pm$ 1 SEM  reported from 10 runs with Teacher Seed 1. Bold values are the best performing based on the mean.}
\label{tab:tin-vgg19-ts-1}
\resizebox{\textwidth}{!}{%
\begin{tabular}{|c|c|ccc|ccc|}
\hline
\multicolumn{1}{|c|}{\multirow{2}{*}{\textbf{Metrics}}} & \multicolumn{1}{c|}{\textbf{Control}} & \multicolumn{3}{c|}{\textbf{Knowledge Distillation}} & \multicolumn{3}{c|}{\textbf{Random Control Distillation}} \\ \cline{2-8} 
\multicolumn{1}{|c|}{} & \multicolumn{1}{c|}{\textbf{SIDDO}} & \multicolumn{1}{c|}{\textbf{0.1}} & \multicolumn{1}{c|}{\textbf{0.5}} & \multicolumn{1}{c|}{\textbf{0.9}} & \multicolumn{1}{c|}{\textbf{0.1}} & \multicolumn{1}{c|}{\textbf{0.5}} & \multicolumn{1}{c|}{\textbf{0.9}} \\ \hline
Activation Distance & \multicolumn{1}{c|}{0.414 $\pm{}$ 0.002} & \multicolumn{1}{c|}{0.414 $\pm{}$ 0.001} & \multicolumn{1}{c|}{\textbf{0.413 $\pm{}$ 0.001}} & \multicolumn{1}{c|}{\textbf{0.413 $\pm{}$ 0.001}} & \multicolumn{1}{c|}{0.522 $\pm{}$ 0.001} & \multicolumn{1}{c|}{0.742 $\pm{}$ 0.000}  & \multicolumn{1}{c|}{0.886 $\pm{}$ 0.000}  \\ \hline
Rank Disagreement & \textbf{0.978 $\pm{}$ 0.000}  & \multicolumn{1}{c|}{\textbf{0.978 $\pm{}$ 0.000} } & \multicolumn{1}{c|}{\textbf{0.978 $\pm{}$ 0.000} } & \textbf{0.978 $\pm{}$ 0.000}  & \multicolumn{1}{c|}{0.987 $\pm{}$ 0.000}  & \multicolumn{1}{c|}{0.988 $\pm{}$ 0.000}  & 0.989 $\pm{}$ 0.000 \\ \hline
Prediction Disagreement & 0.329 $\pm{}$ 0.001 & \multicolumn{1}{c|}{0.329 $\pm{}$ 0.001} & \multicolumn{1}{c|}{\textbf{0.328 $\pm{}$ 0.001}} & \textbf{0.328 $\pm{}$ 0.001} & \multicolumn{1}{c|}{0.348 $\pm{}$ 0.001} & \multicolumn{1}{c|}{0.379 $\pm{}$ 0.001} & 0.410 $\pm{}$ 0.000 \\ \hline
JS Divergence & 0.194 $\pm{}$ 0.001 & \multicolumn{1}{c|}{0.194 $\pm{}$ 0.001} & \multicolumn{1}{c|}{\textbf{0.193 $\pm{}$ 0.001}} & \textbf{0.193 $\pm{}$ 0.001} & \multicolumn{1}{c|}{0.308 $\pm{}$ 0.000}  & \multicolumn{1}{c|}{0.457 $\pm{}$ 0.000}  & 0.593 $\pm{}$ 0.000 \\ \hline
Accuracy & 0.635 $\pm{}$ 0.001 & \multicolumn{1}{c|}{0.636 $\pm{}$ 0.001} & \multicolumn{1}{c|}{\textbf{0.638 $\pm{}$ 0.001}} & 0.637 $\pm{}$ 0.001 & \multicolumn{1}{c|}{0.627 $\pm{}$ 0.001} & \multicolumn{1}{c|}{0.603 $\pm{}$ 0.001} & 0.574 $\pm{}$ 0.001 \\ \hline
Loss & 3.345 $\pm{}$ 0.011 & \multicolumn{1}{c|}{3.318 $\pm{}$ 0.009} & \multicolumn{1}{c|}{3.306 $\pm{}$ 0.009} & 3.311 $\pm{}$ 0.010 & \multicolumn{1}{c|}{\textbf{2.004 $\pm{}$ 0.004}} & \multicolumn{1}{c|}{2.733 $\pm{}$ 0.004} & 3.682 $\pm{}$ 0.002 \\ \hline
\end{tabular}%
}
\end{table}
\begin{table}[H]
\caption{VGG19 on TinyImageNet mean and $\pm$ 1 SEM  reported from 10 runs with Teacher Seed 2. Bold values are the best performing based on the mean.}
\label{tab:tin-vgg19-ts-2}
\resizebox{\textwidth}{!}{%
\begin{tabular}{|c|c|ccc|ccc|}
\hline
\multicolumn{1}{|c|}{\multirow{2}{*}{\textbf{Metrics}}} & \multicolumn{1}{c|}{\textbf{Control}} & \multicolumn{3}{c|}{\textbf{Knowledge Distillation}} & \multicolumn{3}{c|}{\textbf{Random Control Distillation}} \\ \cline{2-8} 
\multicolumn{1}{|c|}{} & \multicolumn{1}{c|}{\textbf{SIDDO}} & \multicolumn{1}{c|}{\textbf{0.1}} & \multicolumn{1}{c|}{\textbf{0.5}} & \multicolumn{1}{c|}{\textbf{0.9}} & \multicolumn{1}{c|}{\textbf{0.1}} & \multicolumn{1}{c|}{\textbf{0.5}} & \multicolumn{1}{c|}{\textbf{0.9}} \\ \hline
Activation Distance & \multicolumn{1}{c|}{0.419 $\pm{}$ 0.001} & \multicolumn{1}{c|}{0.417 $\pm{}$ 0.001} & \multicolumn{1}{c|}{0.418 $\pm{}$ 0.001} & \multicolumn{1}{c|}{0.417 $\pm{}$ 0.001} & \multicolumn{1}{c|}{0.524 $\pm{}$ 0.000}  & \multicolumn{1}{c|}{0.743 $\pm{}$ 0.000}  & \multicolumn{1}{c|}{0.886 $\pm{}$ 0.000}  \\ \hline
Rank Disagreement & \textbf{0.978 $\pm{}$ 0.000}  & \multicolumn{1}{c|}{\textbf{0.978 $\pm{}$ 0.000} } & \multicolumn{1}{c|}{\textbf{0.978 $\pm{}$ 0.000} } & \textbf{0.978 $\pm{}$ 0.000}  & \multicolumn{1}{c|}{0.987 $\pm{}$ 0.000}  & \multicolumn{1}{c|}{0.988 $\pm{}$ 0.000}  & 0.989 $\pm{}$ 0.000 \\ \hline
Prediction Disagreement & 0.332 $\pm{}$ 0.001 & \multicolumn{1}{c|}{0.332 $\pm{}$ 0.001} & \multicolumn{1}{c|}{0.332 $\pm{}$ 0.001} & \textbf{0.331 $\pm{}$ 0.001} & \multicolumn{1}{c|}{0.354 $\pm{}$ 0.001} & \multicolumn{1}{c|}{0.385 $\pm{}$ 0.001} & 0.414 $\pm{}$ 0.001 \\ \hline
JS Divergence & 0.196 $\pm{}$ 0.000 & \multicolumn{1}{c|}{\textbf{0.195 $\pm{}$ 0.001}} & \multicolumn{1}{c|}{0.196 $\pm{}$ 0.000}  & \textbf{0.195 $\pm{}$ 0.000}  & \multicolumn{1}{c|}{0.309 $\pm{}$ 0.000}  & \multicolumn{1}{c|}{0.458 $\pm{}$ 0.000}  & 0.593 $\pm{}$ 0.000 \\ \hline
Accuracy & 0.635 $\pm{}$ 0.001 & \multicolumn{1}{c|}{0.636 $\pm{}$ 0.000}  & \multicolumn{1}{c|}{0.635 $\pm{}$ 0.001} & \textbf{0.637 $\pm{}$ 0.001} & \multicolumn{1}{c|}{0.626 $\pm{}$ 0.001} & \multicolumn{1}{c|}{0.602 $\pm{}$ 0.001} & 0.577 $\pm{}$ 0.001 \\ \hline
Loss & 3.314 $\pm{}$ 0.009 & \multicolumn{1}{c|}{3.298 $\pm{}$ 0.004} & \multicolumn{1}{c|}{3.318 $\pm{}$ 0.011} & 3.263 $\pm{}$ 0.009 & \multicolumn{1}{c|}{\textbf{1.998 $\pm{}$ 0.004}} & \multicolumn{1}{c|}{2.738 $\pm{}$ 0.003} & 3.681 $\pm{}$ 0.002 \\ \hline
\end{tabular}%
}
\end{table}
\begin{table}[H]
\caption{VGG19 on TinyImageNet (significance testing). \cmark~indicates significant results compared to controls; \xmark~indicates insignificant results. Each tick represents a teacher (seeds 0 to 2, left to right).}
\label{tab:vgg-tin-significance}
\resizebox{\textwidth}{!}{
\begin{tabular}{|c|c|c|c|c|c|c|}
\hline
\textbf{} & \textbf{Activation Distance}                                      & \textbf{Rank Disagreement}                                        & \textbf{Prediction Disagreement}                                  & \textbf{JS Divergence}                                            & \textbf{Accuracy}                                                 & \textbf{Loss}                                                     \\ \hline
KD 0.1    & \xmark \xmark \xmark & \xmark \xmark \xmark & \xmark \xmark \xmark & \xmark \xmark \xmark & \xmark \xmark \xmark & \xmark \xmark \xmark \\ \hline
KD 0.5    & \xmark \xmark \xmark & \xmark \xmark \xmark & \xmark \xmark \xmark & \xmark \xmark \xmark & \xmark \xmark \xmark & \xmark \xmark \xmark \\ \hline
KD 0.9    & \xmark \xmark \xmark & \xmark \xmark \xmark & \xmark \xmark \xmark & \xmark \xmark \xmark & \xmark \xmark \xmark & \xmark \xmark \xmark \\ \hline
\end{tabular}}
\end{table}
\begin{figure}[H]
    \centering
    \includegraphics[width=\linewidth]{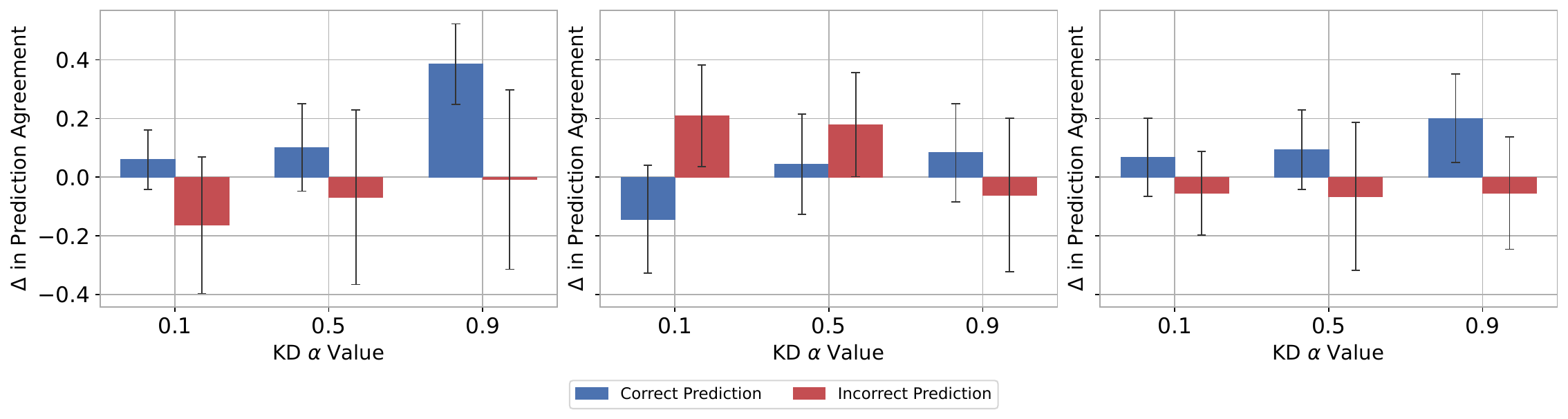}
    \caption{Prediction agreement difference of student models in standard KD to the highest performing control baseline with respect to correct prediction agreement (blue) and incorrect prediction agreement (red), for VGG19 on TinyImageNet (seeds 0 to 2, left to right).}
\end{figure}
\subsubsection{VGG19 with RandAugment}
\begin{figure}[H]
    \centering
    \includegraphics[width=\linewidth]{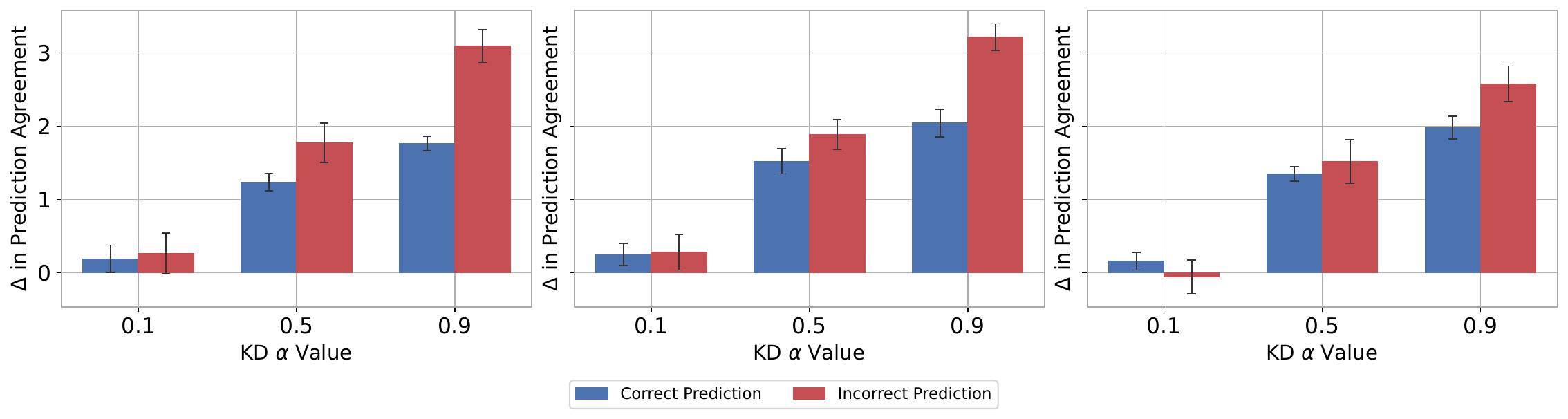}
    \caption{Prediction agreement difference of student models in standard KD to the highest performing control baseline with respect to correct prediction agreement (blue) and incorrect prediction agreement (red), for VGG19 on TinyImageNet with RandAugment (seeds 0 to 2, left to right).}
    \label{fig:tina-vgg-prediction}
\end{figure}

\paragraph{Findings:} For the VGG19 on the TinyImageNet with RandAugment, we observe a high train loss of circa $0.27$ and a train accuracy of circa 0.93. As expected, given the results on the RandAugment ResNet50 that we present in the main body of the paper, we see statistically supported transfer across all $\alpha$ values; see Tables \ref{tab:tina-vgg-ts-0}, \ref{tab:tina-vgg-ts-1}, \ref{tab:tina-vgg-ts-2} and \ref{tab:vgg-tina-significance}. This substantial and statistically supported transfer of knowledge, as expected, coincides with a strong asymmetric transfer of
knowledge favouring incorrect predictions, as shown in Figure~\ref{fig:tina-vgg-prediction}.
\begin{table}[H]
\caption{Teacher Performance on Train and Test Data.}
\label{tab:vgg-tina-teacher}
\centering
\begin{tabular}{|c|c|c|c|c|}
\hline
\textbf{Teacher Seed} & \textbf{Train Loss} & \textbf{Train Accuracy} & \textbf{Test Loss} & \textbf{Test Accuracy} \\ \hline
0 & 0.272582 & 0.933990 & 2.565560 & 0.622600 \\ \hline
1 & 0.269916 & 0.935140 & 2.570119 & 0.618900 \\ \hline
2 & 0.273968 & 0.934700 & 2.609870 & 0.620100 \\ \hline
\end{tabular}%
\end{table}
\begin{table}[H]
\caption{VGG19 on TinyImageNet with RandAugment mean and $\pm$ 1 SEM  reported from 10 runs with Teacher Seed 0. Bold values are the best performing based on the mean.}
\label{tab:tina-vgg-ts-0}
\resizebox{\textwidth}{!}{%
\begin{tabular}{|c|c|ccc|ccc|}
\hline
\multicolumn{1}{|c|}{\multirow{2}{*}{\textbf{Metrics}}} & \multicolumn{1}{c|}{\textbf{Control}} & \multicolumn{3}{c|}{\textbf{Knowledge Distillation}} & \multicolumn{3}{c|}{\textbf{Random Control Distillation}} \\ \cline{2-8} 
\multicolumn{1}{|c|}{} & \multicolumn{1}{c|}{\textbf{SIDDO}} & \multicolumn{1}{c|}{\textbf{0.1}} & \multicolumn{1}{c|}{\textbf{0.5}} & \multicolumn{1}{c|}{\textbf{0.9}} & \multicolumn{1}{c|}{\textbf{0.1}} & \multicolumn{1}{c|}{\textbf{0.5}} & \multicolumn{1}{c|}{\textbf{0.9}} \\ \hline
Activation Distance & 0.393 $\pm{}$ 0.001 & \multicolumn{1}{c|}{0.388 $\pm{}$ 0.001} & \multicolumn{1}{c|}{0.368 $\pm{}$ 0.001} & \textbf{0.355 $\pm{}$ 0.001} & \multicolumn{1}{c|}{0.431 $\pm{}$ 0.001} & \multicolumn{1}{c|}{0.648 $\pm{}$ 0.000}  & 0.848 $\pm{}$ 0.001 \\ \hline
Rank Disagreement & 0.976 $\pm{}$ 0.000 & \multicolumn{1}{c|}{0.976 $\pm{}$ 0.000}  & \multicolumn{1}{c|}{0.975 $\pm{}$ 0.000}  & \textbf{0.974 $\pm{}$ 0.000}  & \multicolumn{1}{c|}{0.985 $\pm{}$ 0.000}  & \multicolumn{1}{c|}{0.987 $\pm{}$ 0.000}  & 0.987 $\pm{}$ 0.000 \\ \hline
Prediction Disagreement & 0.335 $\pm{}$ 0.001 & \multicolumn{1}{c|}{0.333 $\pm{}$ 0.001} & \multicolumn{1}{c|}{0.320 $\pm{}$ 0.001} & \textbf{0.312 $\pm{}$ 0.001} & \multicolumn{1}{c|}{0.341 $\pm{}$ 0.001} & \multicolumn{1}{c|}{0.352 $\pm{}$ 0.001} & 0.396 $\pm{}$ 0.004 \\ \hline
JS Divergence & 0.182 $\pm{}$ 0.000 & \multicolumn{1}{c|}{0.178 $\pm{}$ 0.000}  & \multicolumn{1}{c|}{0.166 $\pm{}$ 0.000}  & \textbf{0.159 $\pm{}$ 0.000}  & \multicolumn{1}{c|}{0.228 $\pm{}$ 0.000}  & \multicolumn{1}{c|}{0.377 $\pm{}$ 0.000}  & 0.577 $\pm{}$ 0.001 \\ \hline
Accuracy & 0.621 $\pm{}$ 0.001 & \multicolumn{1}{c|}{0.624 $\pm{}$ 0.001} & \multicolumn{1}{c|}{0.631 $\pm{}$ 0.001} & \textbf{0.633 $\pm{}$ 0.001} & \multicolumn{1}{c|}{0.622 $\pm{}$ 0.001} & \multicolumn{1}{c|}{0.628 $\pm{}$ 0.001} & 0.609 $\pm{}$ 0.004 \\ \hline
Loss & 2.586 $\pm{}$ 0.009 & \multicolumn{1}{c|}{2.442 $\pm{}$ 0.005} & \multicolumn{1}{c|}{2.148 $\pm{}$ 0.004} & 2.022 $\pm{}$ 0.003 & \multicolumn{1}{c|}{\textbf{1.792 $\pm{}$ 0.003}} & \multicolumn{1}{c|}{2.258 $\pm{}$ 0.002} & 3.533 $\pm{}$ 0.013 \\ \hline
\end{tabular}%
}
\end{table}
\vspace{-0.5cm}
\begin{table}[H]
\caption{VGG19 on TinyImageNet with RandAugment mean and $\pm$ 1 SEM  reported from 10 runs with Teacher Seed 1. Bold values are the best performing based on the mean.}
\label{tab:tina-vgg-ts-1}
\resizebox{\textwidth}{!}{%
\begin{tabular}{|c|c|ccc|ccc|}
\hline
\multicolumn{1}{|c|}{\multirow{2}{*}{\textbf{Metrics}}} & \multicolumn{1}{c|}{\textbf{Control}} & \multicolumn{3}{c|}{\textbf{Knowledge Distillation}} & \multicolumn{3}{c|}{\textbf{Random Control Distillation}} \\ \cline{2-8} 
\multicolumn{1}{|c|}{} & \multicolumn{1}{c|}{\textbf{SIDDO}} & \multicolumn{1}{c|}{\textbf{0.1}} & \multicolumn{1}{c|}{\textbf{0.5}} & \multicolumn{1}{c|}{\textbf{0.9}} & \multicolumn{1}{c|}{\textbf{0.1}} & \multicolumn{1}{c|}{\textbf{0.5}} & \multicolumn{1}{c|}{\textbf{0.9}} \\ \hline
Activation Distance & 0.391 $\pm{}$ 0.001 & \multicolumn{1}{c|}{0.384 $\pm{}$ 0.001} & \multicolumn{1}{c|}{0.362 $\pm{}$ 0.001} & \textbf{0.351 $\pm{}$ 0.000}  & \multicolumn{1}{c|}{0.428 $\pm{}$ 0.001} & \multicolumn{1}{c|}{0.644 $\pm{}$ 0.000}  & 0.845 $\pm{}$ 0.000 \\ \hline
Rank Disagreement & 0.977 $\pm{}$ 0.000 & \multicolumn{1}{c|}{0.976 $\pm{}$ 0.000}  & \multicolumn{1}{c|}{0.975 $\pm{}$ 0.000}  & \textbf{0.974 $\pm{}$ 0.000}  & \multicolumn{1}{c|}{0.985 $\pm{}$ 0.000}  & \multicolumn{1}{c|}{0.987 $\pm{}$ 0.000}  & 0.987 $\pm{}$ 0.000 \\ \hline
Prediction Disagreement & 0.333 $\pm{}$ 0.001 & \multicolumn{1}{c|}{0.330 $\pm{}$ 0.001} & \multicolumn{1}{c|}{0.316 $\pm{}$ 0.001} & \textbf{0.308 $\pm{}$ 0.001} & \multicolumn{1}{c|}{0.337 $\pm{}$ 0.001} & \multicolumn{1}{c|}{0.348 $\pm{}$ 0.001} & 0.392 $\pm{}$ 0.001 \\ \hline
JS Divergence & 0.180$\pm{}$ 0.000 & \multicolumn{1}{c|}{0.176 $\pm{}$ 0.000}  & \multicolumn{1}{c|}{0.164 $\pm{}$ 0.000}  & \textbf{0.156 $\pm{}$ 0.000}  & \multicolumn{1}{c|}{0.226 $\pm{}$ 0.000}  & \multicolumn{1}{c|}{0.375 $\pm{}$ 0.000}  & 0.576 $\pm{}$ 0.000 \\ \hline
Accuracy & 0.622 $\pm{}$ 0.001 & \multicolumn{1}{c|}{0.624 $\pm{}$ 0.000}  & \multicolumn{1}{c|}{0.632 $\pm{}$ 0.001} & \textbf{0.635 $\pm{}$ 0.001} & \multicolumn{1}{c|}{0.625 $\pm{}$ 0.001} & \multicolumn{1}{c|}{0.627 $\pm{}$ 0.001} & 0.611 $\pm{}$ 0.001 \\ \hline
Loss & 2.575 $\pm{}$ 0.004 & \multicolumn{1}{c|}{2.439 $\pm{}$ 0.007} & \multicolumn{1}{c|}{2.149 $\pm{}$ 0.006} & 2.017 $\pm{}$ 0.002 & \multicolumn{1}{c|}{\textbf{1.781 $\pm{}$ 0.005}} & \multicolumn{1}{c|}{2.254 $\pm{}$ 0.003} & 3.526 $\pm{}$ 0.003 \\ \hline
\end{tabular}%
}
\end{table}
\vspace{-0.5cm}
\begin{table}[H]
\caption{VGG19 on TinyImageNet with RandAugment mean and $\pm$ 1 SEM  reported from 10 runs with Teacher Seed 2. Bold values are the best performing based on the mean.}
\label{tab:tina-vgg-ts-2}
\resizebox{\textwidth}{!}{%
\begin{tabular}{|c|c|ccc|ccc|}
\hline
\multicolumn{1}{|c|}{\multirow{2}{*}{\textbf{Metrics}}} & \multicolumn{1}{c|}{\textbf{Control}} & \multicolumn{3}{c|}{\textbf{Knowledge Distillation}} & \multicolumn{3}{c|}{\textbf{Random Control Distillation}} \\ \cline{2-8} 
\multicolumn{1}{|c|}{} & \multicolumn{1}{c|}{\textbf{SIDDO}} & \multicolumn{1}{c|}{\textbf{0.1}} & \multicolumn{1}{c|}{\textbf{0.5}} & \multicolumn{1}{c|}{\textbf{0.9}} & \multicolumn{1}{c|}{\textbf{0.1}} & \multicolumn{1}{c|}{\textbf{0.5}} & \multicolumn{1}{c|}{\textbf{0.9}} \\ \hline
Activation Distance & 0.395 $\pm{}$ 0.001 & \multicolumn{1}{c|}{0.389 $\pm{}$ 0.001} & \multicolumn{1}{c|}{0.368 $\pm{}$ 0.001} & \textbf{0.358 $\pm{}$ 0.001} & \multicolumn{1}{c|}{0.435 $\pm{}$ 0.001} & \multicolumn{1}{c|}{0.649 $\pm{}$ 0.000}  & 0.850 $\pm{}$ 0.001 \\ \hline
Rank Disagreement & 0.977 $\pm{}$ 0.000 & \multicolumn{1}{c|}{0.977 $\pm{}$ 0.000}  & \multicolumn{1}{c|}{0.975 $\pm{}$ 0.000}  & \textbf{0.975 $\pm{}$ 0.000}  & \multicolumn{1}{c|}{0.985 $\pm{}$ 0.000}  & \multicolumn{1}{c|}{0.987 $\pm{}$ 0.000}  & 0.987 $\pm{}$ 0.000 \\ \hline
Prediction Disagreement & 0.335 $\pm{}$ 0.001 & \multicolumn{1}{c|}{0.334 $\pm{}$ 0.001} & \multicolumn{1}{c|}{0.321 $\pm{}$ 0.001} & \textbf{0.313 $\pm{}$ 0.001} & \multicolumn{1}{c|}{0.341 $\pm{}$ 0.001} & \multicolumn{1}{c|}{0.352 $\pm{}$ 0.001} & 0.403 $\pm{}$ 0.010 \\ \hline
JS Divergence & 0.182 $\pm{}$ 0.000 & \multicolumn{1}{c|}{0.179 $\pm{}$ 0.000}  & \multicolumn{1}{c|}{0.167 $\pm{}$ 0.001} & \textbf{0.160 $\pm{}$ 0.001} & \multicolumn{1}{c|}{0.230 $\pm{}$ 0.000}  & \multicolumn{1}{c|}{0.378 $\pm{}$ 0.000}  & 0.579 $\pm{}$ 0.002 \\ \hline
Accuracy & 0.621 $\pm{}$ 0.001 & \multicolumn{1}{c|}{0.623 $\pm{}$ 0.001} & \multicolumn{1}{c|}{0.631 $\pm{}$ 0.001} & \textbf{0.636 $\pm{}$ 0.001} & \multicolumn{1}{c|}{0.623 $\pm{}$ 0.001} & \multicolumn{1}{c|}{0.628 $\pm{}$ 0.001} & 0.600 $\pm{}$ 0.011 \\ \hline
Loss & 2.583 $\pm{}$ 0.006 & \multicolumn{1}{c|}{2.441 $\pm{}$ 0.009} & \multicolumn{1}{c|}{2.145 $\pm{}$ 0.006} & 2.012 $\pm{}$ 0.007 & \multicolumn{1}{c|}{\textbf{1.780 $\pm{}$ 0.003}} & \multicolumn{1}{c|}{2.257 $\pm{}$ 0.003} & 3.556 $\pm{}$ 0.034 \\ \hline
\end{tabular}%
}
\end{table}
\vspace{-0.5cm}
\begin{table}[H]
\caption{VGG19 on TinyImageNet with RandAugment (significance testing). \cmark~indicates significant results compared to controls; \xmark~indicates insignificant results. Each tick represents a teacher (seeds 0 to 2, left to right).}
\label{tab:vgg-tina-significance}
\resizebox{\textwidth}{!}{
\begin{tabular}{|c|c|c|c|c|c|c|}
\hline
\textbf{} & \textbf{Activation Distance}                                      & \textbf{Rank Disagreement}                                        & \textbf{Prediction Disagreement}                                  & \textbf{JS Divergence}                                            & \textbf{Accuracy}                                                 & \textbf{Loss}                                                     \\ \hline
KD 0.1    & \cmark \cmark \cmark & \cmark \cmark \cmark & \xmark \cmark \xmark & \cmark \cmark \cmark & \xmark \xmark \xmark & \xmark \xmark \xmark \\ \hline
KD 0.5    & \cmark \cmark \cmark & \cmark \cmark \cmark & \cmark \cmark \cmark & \cmark \cmark \cmark & \cmark \cmark \cmark & \xmark \xmark \xmark \\ \hline
KD 0.9    & \cmark \cmark \cmark & \cmark \cmark \cmark & \cmark \cmark \cmark & \cmark \cmark \cmark & \cmark \cmark \cmark & \xmark \xmark \xmark \\ \hline
\end{tabular}}
\end{table}
\vspace{-0.5cm}
\subsection{SVHN Dataset}
\label{SHVN_results}
\paragraph{Training Settings:} All SVHN architectures are trained with Adam with a learning rate of 0.001 and a batch size of 256 for 100 epochs. All data is normalised with a mean of 0.5 and a standard deviation of 0.5. The student vision architectures are trained with the same seeds and data orders from seeds 10-19 for the 10 models used for averaging. We repeated this, in line with our other experiments for the three teachers trained on seeds 0-2. 
\subsubsection{ResNet18}
\begin{table}[H]
\centering
\caption{ResNet18 on SVHN mean and $\pm$ 1 SEM  reported from 10 runs with Teacher Seed 1. \textbf{\textbf{Bold}} values are best performing based on the mean.
}
\label{tab:svhn-resnet18-ts-1}
\resizebox{\textwidth}{!}{%
\begin{tabular}{|l|c|ccc|ccc|}
\hline
\multicolumn{1}{|c|}{\multirow{2}{*}{\textbf{Metrics}}} & \textbf{Control} & \multicolumn{3}{c|}{\textbf{Knowledge Distillation}} & \multicolumn{3}{c|}{\textbf{Random  Knowledge Distillation}} \\ \cline{2-8} 
\multicolumn{1}{|c|}{} & \textbf{SIDDO} & \multicolumn{1}{c|}{\textbf{0.1}} & \multicolumn{1}{c|}{\textbf{0.5}} & \textbf{0.9} & \multicolumn{1}{c|}{\textbf{0.1}} & \multicolumn{1}{c|}{\textbf{0.5}} & \textbf{0.9} \\ \hline
Activation Distance \textbf{($\mathbf{\downarrow}$)} & 0.059$\pm{0.001}$ & \multicolumn{1}{c|}{0.058$\pm{0.001}$} & \multicolumn{1}{c|}{0.058$\pm{0.001}$} & \textbf{0.056$\pm{0.001}$} & \multicolumn{1}{c|}{0.141$\pm{0.001}$} & \multicolumn{1}{c|}{0.494$\pm{0.001}$} & 0.848$\pm{0.000}$ \\ \hline
Rank Disagreement  \textbf{($\mathbf{\downarrow}$)} & 0.690$\pm{0.002}$ & \multicolumn{1}{c|}{0.688$\pm{0.003}$} & \multicolumn{1}{c|}{0.687$\pm{0.003}$} & \textbf{0.682$\pm{0.002}$} & \multicolumn{1}{c|}{0.799$\pm{0.002}$} & \multicolumn{1}{c|}{0.799$\pm{0.002}$} & 0.800$\pm{0.003}$ \\ \hline
Prediction Disagreement  \textbf{($\mathbf{\downarrow}$)} & 0.042$\pm{0.001}$ & \multicolumn{1}{c|}{0.042$\pm{0.001}$} & \multicolumn{1}{c|}{0.042$\pm{0.001}$} & \textbf{0.040$\pm{0.001}$} & \multicolumn{1}{c|}{\textbf{0.040$\pm{0.001}$}} & \multicolumn{1}{c|}{0.044$\pm{0.001}$} & 0.046$\pm{0.000}$ \\ \hline
JS Divergence  \textbf{($\mathbf{\downarrow}$)} & 0.023$\pm{0.000}$ & \multicolumn{1}{c|}{0.023$\pm{0.000}$} & \multicolumn{1}{c|}{\textbf{0.022$\pm{0.001}$}} & \textbf{0.022$\pm{0.000}$} & \multicolumn{1}{c|}{0.052$\pm{0.000}$} & \multicolumn{1}{c|}{0.201$\pm{0.000}$} & 0.431$\pm{0.000}$ \\ \hline
Accuracy  \textbf{($\mathbf{\uparrow}$)} & 0.953$\pm{0.001}$ & \multicolumn{1}{c|}{0.953$\pm{0.001}$} & \multicolumn{1}{c|}{0.953$\pm{0.001}$} & 0.954$\pm{0.001}$ & \multicolumn{1}{c|}{\textbf{0.958$\pm{0.001}$}} & \multicolumn{1}{c|}{0.954$\pm{0.001}$} & 0.953$\pm{0.001}$ \\ \hline
Loss  \textbf{($\mathbf{\downarrow}$)} & 0.366$\pm{0.008}$ & \multicolumn{1}{c|}{0.354$\pm{0.008}$} & \multicolumn{1}{c|}{0.328$\pm{0.006}$} & 0.316$\pm{0.004}$ & \multicolumn{1}{c|}{\textbf{0.236$\pm{0.002}$}} & \multicolumn{1}{c|}{0.698$\pm{0.002}$} & 1.698$\pm{0.001}$ \\ \hline
\end{tabular}%
}
\end{table}
\vspace{-0.5cm}
\begin{table}[H]
\centering
\caption{ResNet18 on SVHN mean and $\pm$ 1 SEM  reported from 10 runs with Teacher Seed 2. \textbf{\textbf{Bold}} values are best performing based on the mean.
}
\label{tab:svhn-resnet18-ts-2}
\resizebox{\textwidth}{!}{%
\begin{tabular}{|c|c|ccc|ccc|}
\hline
\multicolumn{1}{|c|}{\multirow{2}{*}{\textbf{Metrics}}} & \multicolumn{1}{c|}{\textbf{Control}} & \multicolumn{3}{c|}{\textbf{Knowledge Distillation}} & \multicolumn{3}{c|}{\textbf{Random Control Distillation}} \\ \cline{2-8} 
\multicolumn{1}{|c|}{} & \multicolumn{1}{c|}{\textbf{SIDDO}} & \multicolumn{1}{c|}{\textbf{0.1}} & \multicolumn{1}{c|}{\textbf{0.5}} & \multicolumn{1}{c|}{\textbf{0.900}} & \multicolumn{1}{c|}{\textbf{0.1}} & \multicolumn{1}{c|}{\textbf{0.5}} & \multicolumn{1}{c|}{\textbf{0.900}} \\ \hline
Activation Distance \textbf{($\mathbf{\downarrow}$)} & 0.068$\pm{0.001}$ & \multicolumn{1}{c|}{0.063$\pm{0.001}$} & \multicolumn{1}{c|}{0.059$\pm{0.000}$} & \textbf{0.058$\pm{0.000}$} & \multicolumn{1}{c|}{0.146$\pm{0.001}$} & \multicolumn{1}{c|}{0.489$\pm{0.001}$} & 0.843$\pm{0.000}$ \\ \hline
Rank Disagreement  \textbf{($\mathbf{\downarrow}$)} & 0.713$\pm{0.003}$ & \multicolumn{1}{c|}{0.667$\pm{0.003}$} & \multicolumn{1}{c|}{0.648$\pm{0.003}$} & \textbf{0.643$\pm{0.001}$} & \multicolumn{1}{c|}{0.800$\pm{0.003}$} & \multicolumn{1}{c|}{0.800$\pm{0.004}$} & 0.799$\pm{0.003}$ \\ \hline
Prediction Disagreement  \textbf{($\mathbf{\downarrow}$)} & 0.048$\pm{0.001}$ & \multicolumn{1}{c|}{0.045$\pm{0.001}$} & \multicolumn{1}{c|}{0.042$\pm{0.000}$} & \textbf{0.041$\pm{0.000}$} & \multicolumn{1}{c|}{0.046$\pm{0.001}$} & \multicolumn{1}{c|}{0.048$\pm{0.001}$} & 0.052$\pm{0.001}$ \\ \hline
JS Divergence  \textbf{($\mathbf{\downarrow}$)} & 0.026$\pm{0.000}$ & \multicolumn{1}{c|}{0.023$\pm{0.000}$} & \multicolumn{1}{c|}{0.021$\pm{0.000}$} & \textbf{0.020$\pm{0.000}$} & \multicolumn{1}{c|}{0.053$\pm{0.001}$} & \multicolumn{1}{c|}{0.199$\pm{0.000}$} & 0.427$\pm{0.000}$ \\ \hline
Accuracy  \textbf{($\mathbf{\uparrow}$)} & 0.952$\pm{0.001}$ & \multicolumn{1}{c|}{0.955$\pm{0.001}$} & \multicolumn{1}{c|}{\textbf{0.957$\pm{0.000}$}} & \textbf{0.957$\pm{0.000}$} & \multicolumn{1}{c|}{0.956$\pm{0.001}$} & \multicolumn{1}{c|}{\textbf{0.957$\pm{0.001}$}} & 0.953$\pm{0.001}$ \\ \hline
Loss  \textbf{($\mathbf{\downarrow}$)} & 0.370$\pm{0.008}$ & \multicolumn{1}{c|}{0.256$\pm{0.006}$} & \multicolumn{1}{c|}{0.226$\pm{0.002}$} & \textbf{0.216$\pm{0.001}$} & \multicolumn{1}{c|}{0.239$\pm{0.003}$} & \multicolumn{1}{c|}{0.692$\pm{0.002}$} & 1.700$\pm{0.001}$ \\ \hline
\end{tabular}%
}
\end{table}

\subsubsection{VGG19}

\begin{figure}[H]
    \centering
    \includegraphics[width=\linewidth]{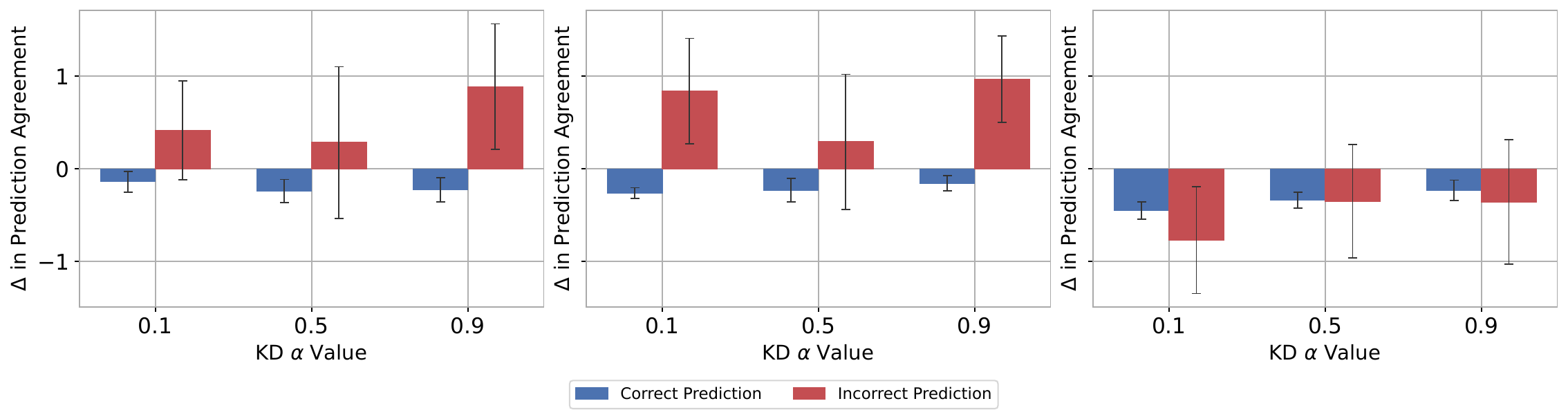}
    \caption{Prediction agreement difference of student models in standard KD to the highest performing control baseline with respect to correct prediction agreement (blue) and incorrect prediction agreement (red), for VGG19 on SVHN (seeds 0 to 2, left to right).  }
    \label{fig:vgg-svhn-prediction}
\end{figure}

\paragraph{Findings:} For the VGG19 on SVHN, we record a low train loss from we observe that the teacher seeds, Table \ref{tab:vgg-shvn-teacher}. The teacher model with a higher training loss (seed 2) has statistically supported knowledge transfer, see Table \ref{tab:vgg-svhn-significance}, for only Rank Disagreement, across $\alpha$ values 0.1, 0.5 and 0.9. Due to the marginal statistically supported transfer across metrics for this seed, we observe a small but inconsistent asymmetric payoff in prediction agreement, slightly favouring incorrect predictions, Figure \ref{fig:vgg-svhn-prediction}. Across the other teacher seeds; we see marginal statistically supported functional transfer, and where transfer is higher, we see negative transfer, but where it is marginal or largely insignificant, we see no preference for knowledge transfer, showing that in this case knowledge sharing can not be attributed to improved performance. 

\begin{table}[H]
\caption{Teacher Performance on Train and Test Data for VGG19 on SVHN}
\label{tab:vgg-shvn-teacher}
\centering
\begin{tabular}{|c|c|c|c|c|}
\hline
\textbf{Teacher Seed} & \textbf{Train Loss} & \textbf{Train Accuracy} & \textbf{Test Loss} & \textbf{Test Accuracy} \\ \hline
0 & 0.004511 & 0.998649 & 0.343982 & 0.952827 \\ \hline
1 & 0.002757 & 0.999290 & 0.347466 & 0.948794 \\ \hline
2 & 0.003741 & 0.998935 & 0.313836 & 0.953596 \\ \hline
\end{tabular}
\end{table}
\vspace{-0.5cm}

\begin{table}[H]
\centering
\caption{VGG19 on SVHN mean and $\pm$ 1 SEM  reported from 10 runs with Teacher Seed 0. \textbf{Bold} values are best performing based on the mean. }
\label{tab:svhn-vgg19-ts-0}
\resizebox{\textwidth}{!}{%
\begin{tabular}{|l|c|ccc|ccc|}
\hline
\multicolumn{1}{|c|}{\multirow{2}{*}{\textbf{Metrics}}} & \textbf{Control} & \multicolumn{3}{c|}{\textbf{Knowledge Distillation}} & \multicolumn{3}{c|}{\textbf{Random Control Distillation}} \\ \cline{2-8} 
\multicolumn{1}{|c|}{} & \textbf{SIDDO} & \multicolumn{1}{c|}{\textbf{0.1}} & \multicolumn{1}{c|}{\textbf{0.5}} & \textbf{0.9} & \multicolumn{1}{c|}{\textbf{0.1}} & \multicolumn{1}{c|}{\textbf{0.5}} & \textbf{0.9} \\ \hline
Activation Distance \textbf{($\mathbf{\downarrow}$)} & 0.065$\pm{0.001}$ & \multicolumn{1}{c|}{\textbf{0.064$\pm{0.001}$}} & \multicolumn{1}{c|}{0.066$\pm{0.002}$} & 0.065$\pm{0.001}$ & \multicolumn{1}{c|}{0.151$\pm{0.001}$} & \multicolumn{1}{c|}{0.494$\pm{0.001}$} & 0.848$\pm{0.000}$ \\ \hline
Rank Disagreement  \textbf{($\mathbf{\downarrow}$)} & 0.708$\pm{0.005}$ & \multicolumn{1}{c|}{0.660$\pm{0.011}$} & \multicolumn{1}{c|}{0.637$\pm{0.009}$} & \textbf{0.603$\pm{0.011}$} & \multicolumn{1}{c|}{0.799$\pm{0.005}$} & \multicolumn{1}{c|}{0.812$\pm{0.006}$} & 0.805$\pm{0.007}$ \\ \hline
Prediction Disagreement  \textbf{($\mathbf{\downarrow}$)} & 0.047$\pm{0.001}$ & \multicolumn{1}{c|}{0.046$\pm{0.000}$} & \multicolumn{1}{c|}{0.047$\pm{0.001}$} & 0.047$\pm{0.001}$ & \multicolumn{1}{c|}{0.047$\pm{0.000}$} & \multicolumn{1}{c|}{\textbf{0.045$\pm{0.001}$}} & 0.046$\pm{0.000}$ \\ \hline
JS Divergence  \textbf{($\mathbf{\downarrow}$)} & 0.028$\pm{0.000}$ & \multicolumn{1}{c|}{\textbf{0.027$\pm{0.000}$}} & \multicolumn{1}{c|}{\textbf{0.027$\pm{0.001}$}} & \textbf{0.027$\pm{0.001}$} & \multicolumn{1}{c|}{0.057$\pm{0.000}$} & \multicolumn{1}{c|}{0.201$\pm{0.000}$} & 0.429$\pm{0.000}$ \\ \hline
Accuracy  \textbf{($\mathbf{\uparrow}$)} & 0.954$\pm{0.001}$ & \multicolumn{1}{c|}{0.954$\pm{0.001}$} & \multicolumn{1}{c|}{0.953$\pm{0.001}$} & 0.953$\pm{0.001}$ & \multicolumn{1}{c|}{0.955$\pm{0.001}$} & \multicolumn{1}{c|}{\textbf{0.956$\pm{0.001}$}} & \textbf{0.956$\pm{0.000}$} \\ \hline
Loss  \textbf{($\mathbf{\downarrow}$)} & 0.349$\pm{0.006}$ & \multicolumn{1}{c|}{0.292$\pm{0.005}$} & \multicolumn{1}{c|}{0.282$\pm{0.008}$} & 0.275$\pm{0.003}$ & \multicolumn{1}{c|}{\textbf{0.263$\pm{0.002}$}} & \multicolumn{1}{c|}{0.698$\pm{0.002}$} & 1.696$\pm{0.001}$ \\ \hline
\end{tabular}%
}
\end{table}
\vspace{-0.5cm}
\begin{table}[H]
\centering
\caption{VGG19 on SVHN mean and $\pm$ 1 SEM  reported from 10 runs with Teacher Seed 1. \textbf{Bold} values are best performing based on the mean. }
\label{tab:svhn-vgg19-ts-1}
\resizebox{\textwidth}{!}{%
\begin{tabular}{|l|c|ccc|ccc|}
\hline
\multicolumn{1}{|c|}{\multirow{2}{*}{\textbf{Metrics}}} & \textbf{Control} & \multicolumn{3}{c|}{\textbf{Knowledge Distillation}} & \multicolumn{3}{c|}{\textbf{Random Control Distillation}} \\ \cline{2-8} 
\multicolumn{1}{|c|}{} & \textbf{SIDDO} & \multicolumn{1}{c|}{\textbf{0.1}} & \multicolumn{1}{c|}{\textbf{0.5}} & \textbf{0.9} & \multicolumn{1}{c|}{\textbf{0.1}} & \multicolumn{1}{c|}{\textbf{0.5}} & \textbf{0.9} \\ \hline
Activation Distance \textbf{($\mathbf{\downarrow}$)} & 0.069$\pm{0.001}$ & \multicolumn{1}{c|}{0.067$\pm{0.001}$} & \multicolumn{1}{c|}{0.067$\pm{0.002}$} & \textbf{0.066$\pm{0.001}$} & \multicolumn{1}{c|}{0.154$\pm{0.001}$} & \multicolumn{1}{c|}{0.496$\pm{0.001}$} & 0.846$\pm{0.000}$ \\ \hline
Rank Disagreement  \textbf{($\mathbf{\downarrow}$)} & 0.758$\pm{0.009}$ & \multicolumn{1}{c|}{0.710$\pm{0.006}$} & \multicolumn{1}{c|}{0.663$\pm{0.011}$} & \textbf{0.652$\pm{0.009}$} & \multicolumn{1}{c|}{0.814$\pm{0.002}$} & \multicolumn{1}{c|}{0.796$\pm{0.007}$} & 0.808$\pm{0.007}$ \\ \hline
Prediction Disagreement  \textbf{($\mathbf{\downarrow}$)} & 0.051$\pm{0.001}$ & \multicolumn{1}{c|}{0.050$\pm{0.000}$} & \multicolumn{1}{c|}{0.050$\pm{0.001}$} & 0.049$\pm{0.001}$ & \multicolumn{1}{c|}{0.050$\pm{0.000}$} & \multicolumn{1}{c|}{0.049$\pm{0.001}$} & \textbf{0.048$\pm{0.000}$} \\ \hline
JS Divergence  \textbf{($\mathbf{\downarrow}$)} & 0.030$\pm{0.000}$ & \multicolumn{1}{c|}{0.029$\pm{0.000}$} & \multicolumn{1}{c|}{0.029$\pm{0.001}$} & \textbf{0.028$\pm{0.001}$} & \multicolumn{1}{c|}{0.058$\pm{0.000}$} & \multicolumn{1}{c|}{0.201$\pm{0.000}$} & 0.428$\pm{0.000}$ \\ \hline
Accuracy  \textbf{($\mathbf{\uparrow}$)} & 0.952$\pm{0.001}$ & \multicolumn{1}{c|}{0.953$\pm{0.000}$} & \multicolumn{1}{c|}{0.953$\pm{0.001}$} & 0.954$\pm{0.001}$ & \multicolumn{1}{c|}{0.953$\pm{0.001}$} & \multicolumn{1}{c|}{0.955$\pm{0.001}$} & \textbf{0.956$\pm{0.000}$} \\ \hline
Loss  \textbf{($\mathbf{\downarrow}$)} & 0.353$\pm{0.008}$ & \multicolumn{1}{c|}{0.304$\pm{0.004}$} & \multicolumn{1}{c|}{0.274$\pm{0.006}$} & 0.269$\pm{0.005}$ & \multicolumn{1}{c|}{\textbf{0.268$\pm{0.003}$}} & \multicolumn{1}{c|}{0.701$\pm{0.002}$} & 1.695$\pm{0.001}$ \\ \hline
\end{tabular}%
}
\end{table}
\begin{table}[H]
\centering
\caption{VGG19 on SVHN mean and $\pm$ 1 SEM  reported from 10 runs with Teacher Seed 2. \textbf{Bold} values are best performing based on the mean.
}
\label{tab:svhn-vgg19-ts-2}
\resizebox{\textwidth}{!}{%
\begin{tabular}{|l|c|ccc|ccc|}
\hline
\multicolumn{1}{|c|}{\multirow{2}{*}{\textbf{Metrics}}} & \textbf{Control} & \multicolumn{3}{c|}{\textbf{Knowledge Distillation}} & \multicolumn{3}{c|}{\textbf{Random Control Distillation}} \\ \cline{2-8} 
\multicolumn{1}{|c|}{} & \textbf{SIDDO} & \multicolumn{1}{c|}{\textbf{0.1}} & \multicolumn{1}{c|}{\textbf{0.5}} & \textbf{0.9} & \multicolumn{1}{c|}{\textbf{0.1}} & \multicolumn{1}{c|}{\textbf{0.5}} & \textbf{0.9} \\ \hline
Activation Distance \textbf{($\mathbf{\downarrow}$)} & 0.065$\pm{0.001}$ & \multicolumn{1}{c|}{0.067$\pm{0.001}$} & \multicolumn{1}{c|}{0.065$\pm{0.001}$} & \textbf{0.064$\pm{0.002}$} & \multicolumn{1}{c|}{0.148$\pm{0.000}$} & \multicolumn{1}{c|}{0.493$\pm{0.001}$} & 0.847$\pm{0.000}$ \\ \hline
Rank Disagreement  \textbf{($\mathbf{\downarrow}$)} & 0.733$\pm{0.009}$ & \multicolumn{1}{c|}{0.680$\pm{0.011}$} & \multicolumn{1}{c|}{0.647$\pm{0.008}$} & \textbf{0.600$\pm{0.013}$} & \multicolumn{1}{c|}{0.804$\pm{0.003}$} & \multicolumn{1}{c|}{0.808$\pm{0.007}$} & 0.809$\pm{0.006}$ \\ \hline
Prediction Disagreement  \textbf{($\mathbf{\downarrow}$)} & 0.048$\pm{0.001}$ & \multicolumn{1}{c|}{0.049$\pm{0.001}$} & \multicolumn{1}{c|}{0.047$\pm{0.001}$} & 0.046$\pm{0.001}$ & \multicolumn{1}{c|}{0.045$\pm{0.000}$} & \multicolumn{1}{c|}{\textbf{0.044$\pm{0.001}$}} & 0.046$\pm{0.000}$ \\ \hline
JS Divergence  \textbf{($\mathbf{\downarrow}$)} & 0.028$\pm{0.000}$ & \multicolumn{1}{c|}{0.028$\pm{0.001}$} & \multicolumn{1}{c|}{0.027$\pm{0.000}$} & \textbf{0.026$\pm{0.001}$} & \multicolumn{1}{c|}{0.055$\pm{0.000}$} & \multicolumn{1}{c|}{0.200$\pm{0.000}$} & 0.429$\pm{0.000}$ \\ \hline
Accuracy  \textbf{($\mathbf{\uparrow}$)} & 0.952$\pm{0.001}$ & \multicolumn{1}{c|}{0.952$\pm{0.001}$} & \multicolumn{1}{c|}{0.953$\pm{0.001}$} & 0.954$\pm{0.001}$ & \multicolumn{1}{c|}{0.956$\pm{0.000}$} & \multicolumn{1}{c|}{\textbf{0.957$\pm{0.001}$}} & 0.956$\pm{0.001}$ \\ \hline
Loss  \textbf{($\mathbf{\downarrow}$)} & 0.358$\pm{0.007}$ & \multicolumn{1}{c|}{0.301$\pm{0.006}$} & \multicolumn{1}{c|}{0.284$\pm{0.005}$} & 0.265$\pm{0.010}$ & \multicolumn{1}{c|}{\textbf{0.258$\pm{0.001}$}} & \multicolumn{1}{c|}{0.697$\pm{0.002}$} & 1.696$\pm{0.001}$ \\ \hline
\end{tabular}%
}
\end{table}
\vspace{-0.5cm}

\begin{table}[H]
    \centering
    \caption{VGG19 on SVHN (significance testing). \cmark~indicates significant results compared to controls; \xmark~indicates insignificant results. Each tick represents a teacher (seeds 0 to 2, left to right).}
    \label{tab:vgg-svhn-significance}
    \resizebox{\textwidth}{!}{
\begin{tabular}{|c|c|c|c|c|c|c|}
\hline
\textbf{} & \textbf{Activation Distance}                                      & \textbf{Rank Disagreement}                                        & \textbf{Prediction Disagreement}                                  & \textbf{JS Divergence}                                            & \textbf{Accuracy}                                                 & \textbf{Loss}                                                     \\ \hline
KD 0.1    & \xmark \xmark \xmark & \xmark \cmark \cmark & \xmark \xmark \xmark & \xmark \xmark \xmark & \xmark \xmark \xmark & \xmark \xmark \xmark \\ \hline
KD 0.5    & \xmark \xmark \xmark & \cmark \cmark \cmark & \xmark \xmark \xmark & \xmark \xmark \xmark & \xmark \xmark \xmark & \xmark \xmark \xmark \\ \hline
KD 0.9    & \xmark \xmark \xmark & \cmark \cmark \cmark & \xmark \xmark \xmark & \xmark \cmark \xmark & \xmark \xmark \xmark & \xmark \xmark \xmark \\ \hline
\end{tabular}}
\end{table}

\subsubsection{ViT}
\begin{figure}[H]
    \centering
    \includegraphics[width=\linewidth]{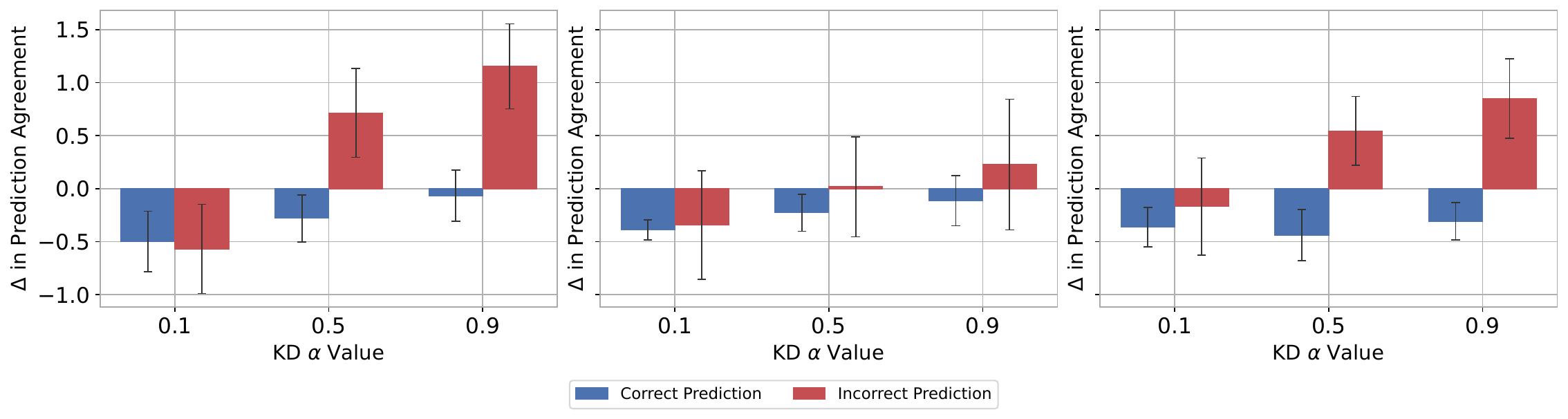}
    \caption{Prediction agreement difference of student models in standard KD to the highest performing control baseline with respect to correct prediction agreement (blue) and incorrect prediction agreement (red), for ViT on SVHN (seeds 0 to 2, left to right).  }
    \label{fig:vit-svhn-prediction}
\end{figure}

\paragraph{Findings:}For ViT on SVHN, obtain a low train loss values across teacher seeds, Table \ref{tab:vit-shvn-teacher}, The teacher model with a higher training loss (seed 1) has statisically supported knowledge transfer, see Table \ref{tab:vit-svhn-significance}, for only Activation Distance, Rank Disagreement and JS Divergence across $\alpha$ values 0.5 and 0.9. In this case, we observe a small but inconsistent asymmetric payoff in prediction agreement, slightly favouring incorrect predictions, Figure \ref{fig:vit-svhn-prediction}. 
\begin{table}[H]
\centering
\caption{Teacher Performance on Train and Test Data.}
\label{tab:vit-shvn-teacher}
\begin{tabular}{|c|c|c|c|c|}
\hline
\textbf{Teacher Seed} & \textbf{Train Loss} & \textbf{Train Accuracy} & \textbf{Test Loss} & \textbf{Test Accuracy} \\ \hline
0 & 0.018473 & 0.994417 & 0.774354 & 0.854564 \\ \hline
1 & 0.019402 & 0.994963 & 0.711637 & 0.855025 \\ \hline
2 & 0.018580 & 0.994635 & 0.692686 & 0.860633 \\ \hline
\end{tabular}
\end{table}
\begin{table}[H]
\centering
\caption{ViT on SVHN mean and $\pm$ 1 SEM  reported from 10 runs with Teacher Seed 0. \textbf{Bold} values are best performing based on the mean. }
\label{tab:svhn-vit-ts-0}
\resizebox{\textwidth}{!}{%
\begin{tabular}{|l|c|ccc|ccc|}
\hline
\multicolumn{1}{|c|}{\multirow{2}{*}{\textbf{Metrics}}} & \textbf{Control} & \multicolumn{3}{c|}{\textbf{Knowledge Distillation}} & \multicolumn{3}{c|}{\textbf{Random Control Distillation}} \\ \cline{2-8} 
\multicolumn{1}{|c|}{} & \textbf{SIDDO} & \multicolumn{1}{c|}{\textbf{0.1}} & \multicolumn{1}{c|}{\textbf{0.5}} & \textbf{0.9} & \multicolumn{1}{c|}{\textbf{0.1}} & \multicolumn{1}{c|}{\textbf{0.5}} & \textbf{0.9} \\ \hline
Activation Distance \textbf{($\mathbf{\downarrow}$)} & 0.219$\pm{0.002}$ & \multicolumn{1}{c|}{0.220$\pm{0.002}$} & \multicolumn{1}{c|}{0.215$\pm{0.002}$} & \textbf{0.211$\pm{0.001}$} & \multicolumn{1}{c|}{0.273$\pm{0.002}$} & \multicolumn{1}{c|}{0.535$\pm{0.001}$} & 0.829$\pm{0.000}$ \\ \hline
Rank Disagreement  \textbf{($\mathbf{\downarrow}$)} & 0.741$\pm{0.001}$ & \multicolumn{1}{c|}{0.741$\pm{0.001}$} & \multicolumn{1}{c|}{0.736$\pm{0.001}$} & \textbf{0.732$\pm{0.001}$} & \multicolumn{1}{c|}{0.801$\pm{0.001}$} & \multicolumn{1}{c|}{0.806$\pm{0.003}$} & 0.805$\pm{0.002}$ \\ \hline
Prediction Disagreement  \textbf{($\mathbf{\downarrow}$)} & 0.165$\pm{0.002}$ & \multicolumn{1}{c|}{0.165$\pm{0.002}$} & \multicolumn{1}{c|}{0.162$\pm{0.002}$} & \textbf{0.159$\pm{0.001}$} & \multicolumn{1}{c|}{0.162$\pm{0.001}$} & \multicolumn{1}{c|}{0.160$\pm{0.001}$} & 0.161$\pm{0.001}$ \\ \hline
JS Divergence  \textbf{($\mathbf{\downarrow}$)} & 0.0910$\pm{0.001}$ & \multicolumn{1}{c|}{0.091$\pm{0.001}$} & \multicolumn{1}{c|}{0.088$\pm{0.001}$} & \textbf{0.085$\pm{0.001}$} & \multicolumn{1}{c|}{0.110$\pm{0.001}$} & \multicolumn{1}{c|}{0.227$\pm{0.001}$} & 0.422$\pm{0.000}$ \\ \hline
Accuracy  \textbf{($\mathbf{\uparrow}$)} & 0.857$\pm{0.003}$ & \multicolumn{1}{c|}{0.856$\pm{0.003}$} & \multicolumn{1}{c|}{0.856$\pm{0.002}$} & 0.858$\pm{0.002}$ & \multicolumn{1}{c|}{0.858$\pm{0.002}$} & \multicolumn{1}{c|}{\textbf{0.860$\pm{0.002}$}} & 0.859$\pm{0.002}$ \\ \hline
Loss  \textbf{($\mathbf{\downarrow}$)} & 0.707$\pm{0.013}$ & \multicolumn{1}{c|}{0.698$\pm{0.012}$} & \multicolumn{1}{c|}{0.651$\pm{0.013}$} & 0.608$\pm{0.006}$ & \multicolumn{1}{c|}{\textbf{0.560$\pm{0.008}$}} & \multicolumn{1}{c|}{0.896$\pm{0.004}$} & 1.771$\pm{0.002}$ \\ \hline
\end{tabular}%
}
\end{table}
\begin{table}[H]
\centering
\caption{ViT on SVHN mean and $\pm$ 1 SEM  reported from 10 runs with Teacher Seed 1. \textbf{Bold} values are best performing based on the mean. }
\label{tab:svhn-vit-ts-1}
\resizebox{\textwidth}{!}{%
\begin{tabular}{|l|c|ccc|ccc|}
\hline
\multicolumn{1}{|c|}{\multirow{2}{*}{\textbf{Metrics}}} & \multicolumn{1}{c|}{\textbf{Control}} & \multicolumn{3}{c|}{\textbf{Knowledge Distillation}} & \multicolumn{3}{c|}{\textbf{Random Control Distillation}} \\ \cline{2-8} 
\multicolumn{1}{|c|}{} & \textbf{SIDDO} & \multicolumn{1}{c|}{\textbf{0.1}} & \multicolumn{1}{c|}{\textbf{0.5}} & \textbf{0.9} & \multicolumn{1}{c|}{\textbf{0.1}} & \multicolumn{1}{c|}{\textbf{0.5}} & \textbf{0.9} \\ \hline
Activation Distance \textbf{($\mathbf{\downarrow}$)} & 0.216$\pm{0.002}$ & \multicolumn{1}{c|}{0.212$\pm{0.001}$} & \multicolumn{1}{c|}{0.208$\pm{0.002}$} & \textbf{0.206$\pm{0.002}$} & \multicolumn{1}{c|}{0.266$\pm{0.002}$} & \multicolumn{1}{c|}{0.529$\pm{0.001}$} & 0.825$\pm{0.001}$ \\ \hline
Rank Disagreement  \textbf{($\mathbf{\downarrow}$)} & 0.745$\pm{0.001}$ & \multicolumn{1}{c|}{0.745$\pm{0.001}$} & \multicolumn{1}{c|}{0.737$\pm{0.001}$} & \textbf{0.735$\pm{0.001}$} & \multicolumn{1}{c|}{0.801$\pm{0.001}$} & \multicolumn{1}{c|}{0.805$\pm{0.003}$} & 0.804$\pm{0.003}$ \\ \hline
Prediction Disagreement  \textbf{($\mathbf{\downarrow}$)} & 0.162$\pm{0.001}$ & \multicolumn{1}{c|}{0.159$\pm{0.001}$} & \multicolumn{1}{c|}{0.157$\pm{0.001}$} & \textbf{0.156$\pm{0.001}$} & \multicolumn{1}{c|}{0.158$\pm{0.001}$} & \multicolumn{1}{c|}{0.156$\pm{0.001}$} & 0.164$\pm{0.005}$ \\ \hline
JS Divergence  \textbf{($\mathbf{\downarrow}$)} & 0.089$\pm{0.001}$ & \multicolumn{1}{c|}{0.086$\pm{0.000}$} & \multicolumn{1}{c|}{0.084$\pm{0.001}$} & \textbf{0.082$\pm{0.001}$} & \multicolumn{1}{c|}{0.106$\pm{0.001}$} & \multicolumn{1}{c|}{0.224$\pm{0.001}$} & 0.420$\pm{0.001}$ \\ \hline
Accuracy  \textbf{($\mathbf{\uparrow}$)} & 0.856$\pm{0.003}$ & \multicolumn{1}{c|}{0.861$\pm{0.001}$} & \multicolumn{1}{c|}{0.863$\pm{0.003}$} & 0.864$\pm{0.002}$ & \multicolumn{1}{c|}{0.863$\pm{0.003}$} & \multicolumn{1}{c|}{\textbf{0.865$\pm{0.002}$}} & 0.854$\pm{0.007}$ \\ \hline
Loss  \textbf{($\mathbf{\downarrow}$)} & 0.722$\pm{0.011}$ & \multicolumn{1}{c|}{0.680$\pm{0.009}$} & \multicolumn{1}{c|}{0.603$\pm{0.012}$} & 0.574$\pm{0.010}$ & \multicolumn{1}{c|}{\textbf{0.543$\pm{0.010}$}} & \multicolumn{1}{c|}{0.886$\pm{0.004}$} & 1.777$\pm{0.007}$ \\ \hline
\end{tabular}%
}
\end{table}

\begin{table}[H]
\centering
\caption{ViT on SVHN mean and $\pm$ 1 SEM  reported from 10 runs with Teacher Seed 2. \textbf{Bold} values are best performing based on the mean.
}
\label{tab:svhn-vit-ts-2}
\resizebox{\textwidth}{!}{%
\begin{tabular}{|l|c|ccc|ccc|}
\hline
\multicolumn{1}{|c|}{\multirow{2}{*}{\textbf{Metrics}}} & \textbf{Control} & \multicolumn{3}{c|}{\textbf{Knowledge Distillation}} & \multicolumn{3}{c|}{\textbf{Random Control Distillation}} \\ \cline{2-8} 
\multicolumn{1}{|c|}{} & \textbf{SIDDO} & \multicolumn{1}{c|}{\textbf{0.1}} & \multicolumn{1}{c|}{\textbf{0.5}} & \textbf{0.9} & \multicolumn{1}{c|}{\textbf{0.1}} & \multicolumn{1}{c|}{\textbf{0.5}} & \textbf{0.9} \\ \hline
Activation Distance \textbf{($\mathbf{\downarrow}$)} & 0.212$\pm{0.001}$ & \multicolumn{1}{c|}{0.206$\pm{0.002}$} & \multicolumn{1}{c|}{0.206$\pm{0.002}$} & \textbf{0.204$\pm{0.001}$} & \multicolumn{1}{c|}{0.265$\pm{0.001}$} & \multicolumn{1}{c|}{0.532$\pm{0.001}$} & 0.828$\pm{0.000}$ \\ \hline
Rank Disagreement  \textbf{($\mathbf{\downarrow}$)} & 0.742$\pm{0.001}$ & \multicolumn{1}{c|}{0.735$\pm{0.001}$} & \multicolumn{1}{c|}{0.731$\pm{0.001}$} & \textbf{0.728$\pm{0.001}$} & \multicolumn{1}{c|}{0.802$\pm{0.001}$} & \multicolumn{1}{c|}{0.803$\pm{0.001}$} & 0.804$\pm{0.002}$ \\ \hline
Prediction Disagreement  \textbf{($\mathbf{\downarrow}$)} & 0.160$\pm{0.001}$ & \multicolumn{1}{c|}{0.155$\pm{0.001}$} & \multicolumn{1}{c|}{0.155$\pm{0.001}$} & 0.153$\pm{0.001}$ & \multicolumn{1}{c|}{0.156$\pm{0.001}$} & \multicolumn{1}{c|}{0.153$\pm{0.001}$} & \textbf{0.152$\pm{0.001}$} \\ \hline
JS Divergence  \textbf{($\mathbf{\downarrow}$)} & 0.087$\pm{0.001}$ & \multicolumn{1}{c|}{0.084$\pm{0.001}$} & \multicolumn{1}{c|}{0.083$\pm{0.001}$} & \textbf{0.081$\pm{0.001}$} & \multicolumn{1}{c|}{0.106$\pm{0.000}$} & \multicolumn{1}{c|}{0.225$\pm{0.001}$} & 0.421$\pm{0.000}$ \\ \hline
Accuracy  \textbf{($\mathbf{\uparrow}$)} & 0.856$\pm{0.001}$ & \multicolumn{1}{c|}{0.861$\pm{0.002}$} & \multicolumn{1}{c|}{0.859$\pm{0.002}$} & 0.860$\pm{0.002}$ & \multicolumn{1}{c|}{0.863$\pm{0.001}$} & \multicolumn{1}{c|}{\textbf{0.866$\pm{0.002}$}} & 0.864$\pm{0.001}$ \\ \hline
Loss  \textbf{($\mathbf{\downarrow}$)} & 0.730$\pm{0.011}$ & \multicolumn{1}{c|}{0.673$\pm{0.011}$} & \multicolumn{1}{c|}{0.627$\pm{0.009}$} & 0.600$\pm{0.007}$ & \multicolumn{1}{c|}{\textbf{0.548$\pm{0.003}$}} & \multicolumn{1}{c|}{0.886$\pm{0.005}$} & 1.768$\pm{0.002}$ \\ \hline
\end{tabular}%
}
\end{table}
\begin{table}[H]
    \centering
    \caption{ViT on SVHN (significance testing). \cmark~indicates significant results compared to controls; \xmark~indicates insignificant results. Each tick represents a teacher (seeds 0 to 2, left to right).}
    \label{tab:vit-svhn-significance}
    \resizebox{\textwidth}{!}{
\begin{tabular}{|c|c|c|c|c|c|c|}
\hline
\textbf{} & \textbf{Activation Distance}                                      & \textbf{Rank Disagreement}                                        & \textbf{Prediction Disagreement}                                  & \textbf{JS Divergence}                                            & \textbf{Accuracy}                                                 & \textbf{Loss}                                                     \\ \hline
KD 0.1    & \xmark \xmark \xmark & \xmark \xmark \cmark & \xmark \xmark \xmark & \xmark \xmark \cmark & \xmark \xmark \xmark & \xmark \xmark \xmark \\ \hline
KD 0.5    & \xmark \xmark \cmark & \cmark \cmark \cmark & \xmark \xmark \xmark & \xmark \cmark \cmark & \xmark \xmark \xmark & \xmark \xmark \xmark \\ \hline
KD 0.9    & \cmark \cmark \cmark & \cmark \cmark \cmark & \xmark \xmark \xmark & \cmark \cmark \cmark & \xmark \xmark \xmark & \xmark \xmark \xmark \\ \hline
\end{tabular}}
\end{table}
\section{Audio Results}
\label{Audio_results}
\paragraph{Training Settings:} 
All audio is converted into mono and downsampled to 16000 htz, it is converted into a spectrogram using torchaudio \citep{hwang2023torchaudio} with an n\_fft of 512 and a power of 2. This is then converted to the MelScale with an n\_mels of 32 and a sample rate of 16000 and a n\_stft of 257. 
The train test split for Urbansounds8K used sklearn \citep{scikit-learn} train\_test\_split function with a test size of 0.2 a random state of 42 and the shuffle set to True. 
All audio architectures are trained with SGD with a learning rate of 0.01 and a batch size of 256 for 100 epochs on SpeechCommandsV2 and 150 epochs for UrbanSounds8K. All data is converted into a mel spectrogram format prior to training to increase convergence speed~\citep{wyse2017audio}. The audio architectures are trained with the same seeds and data orders from seeds 10-19 for the 10 models used for averaging. This is repeated for the three teachers trained on seeds 0-2. 
\label{audio_results}
\subsection{SpeechCommands}

SpeechCommands~\citep{warden2018speech} is an audio dataset comprised of 35 classes with 29.4 hours of audio clips of a 1-2 second duration. There are 84,843 training examples and 11,005 testing examples. 

\paragraph{Findings:} We find that knowledge transfer is statistically supported allowing the rejection of the null hypothesis for knowledge sharing. For both architectures there is considerable knowledge transfer compared to the baseline controls. With statistically supported transfer we observe negative asymmetric knowledge transfer, as expected.
\subsubsection{VGGish}

\paragraph{Findings:} We observe that the teacher model achieves a high train accuracy along with a high train loss, see Table \ref{tab:vggish-melsc-teacher}. With this we observe a substantial and statistically supported knowledge transfer for all $\alpha$ values, see Tables \ref{tab:sc-VGGISH-ts-0}, \ref{tab:sc-VGGISH-ts-1}, \ref{tab:sc-VGGISH-ts-2} and \ref{tab:vgg_sc_sig}. This substantial and significant transfer of knowledge, as expected, coincides with a strong asymmetric transfer of
knowledge favouring incorrect predictions, as shown in Figure~\ref{fig:vgg-sc-pred}.
\begin{table}[H]
\centering
\caption{Teacher Performance on Train and Test Data for VGGish on SpeechCommands.}
\label{tab:vggish-melsc-teacher}
\begin{tabular}{|c|c|c|c|c|}
\hline
\textbf{Teacher Seed} & \textbf{Train Loss} & \textbf{Train Accuracy} & \textbf{Test Loss} & \textbf{Test Accuracy} \\ \hline
0 & 0.044291 & 0.986457 & 0.817567 & 0.879237 \\ \hline
1 & 0.061635 & 0.981566 & 0.928225 & 0.864698 \\ \hline
2 & 0.043880 & 0.987047 & 0.765199 & 0.877328 \\ \hline
\end{tabular}
\end{table}
\begin{table}[H]
\centering
\caption{VGGish on SpeechCommands mean and $\pm$ 1 SEM  reported from 10 runs with Teacher Seed 1. \textbf{Bold} values are best performing based on the mean. }
\label{tab:sc-VGGISH-ts-1}
\resizebox{\linewidth}{!}{
\begin{tabular}{|l|c|ccc|ccc|}
\hline
\multicolumn{1}{|c|}{\multirow{2}{*}{\textbf{Metrics}}} & \textbf{Control}  & \multicolumn{3}{c|}{\textbf{Knowledge Distillation}}                                                                     & \multicolumn{3}{c|}{\textbf{Random Control Distillation}}                                                   \\ \cline{2-8} 
\multicolumn{1}{|c|}{}                                  & \textbf{SIDDO}     & \multicolumn{1}{c|}{\textbf{0.1}}       & \multicolumn{1}{c|}{\textbf{0.5}}                & \textbf{0.9}                & \multicolumn{1}{c|}{\textbf{0.1}}              & \multicolumn{1}{c|}{\textbf{0.5}}       & \textbf{0.9}       \\ \hline
Activation Distance \textbf{($\mathbf{\downarrow}$)} & 0.209$\pm{0.002}$ & \multicolumn{1}{c|}{0.169$\pm{0.001}$} & \multicolumn{1}{c|}{0.168$\pm{0.001}$}          & \textbf{0.165$\pm{0.000}$}   & \multicolumn{1}{c|}{0.277$\pm{0.001}$}        & \multicolumn{1}{c|}{0.579$\pm{0.001}$} & 0.881$\pm{0.000}$   \\ \hline
Rank Disagreement  \textbf{($\mathbf{\downarrow}$)} & 0.910$\pm{0.000}$    & \multicolumn{1}{c|}{0.885$\pm{0.001}$} & \multicolumn{1}{c|}{0.881$\pm{0.000}$}            & \textbf{0.879$\pm{0.000}$}   & \multicolumn{1}{c|}{0.942$\pm{0.000}$}          & \multicolumn{1}{c|}{0.942$\pm{0.000}$}   & 0.940$\pm{0.000}$    \\ \hline
Prediction Disagreement  \textbf{($\mathbf{\downarrow}$)} & 0.157$\pm{0.001}$ & \multicolumn{1}{c|}{0.129$\pm{0.001}$} & \multicolumn{1}{c|}{0.127$\pm{0.001}$}          & \textbf{0.125$\pm{0.001}$} & \multicolumn{1}{c|}{0.139$\pm{0.000}$}          & \multicolumn{1}{c|}{0.149$\pm{0.001}$} & 0.181$\pm{0.001}$ \\ \hline
JS Divergence  \textbf{($\mathbf{\downarrow}$)} & 0.094$\pm{0.001}$ & \multicolumn{1}{c|}{0.071$\pm{0.000}$}   & \multicolumn{1}{c|}{0.068$\pm{0.000}$}            & \textbf{0.066$\pm{0.000}$}   & \multicolumn{1}{c|}{0.129$\pm{0.000}$}          & \multicolumn{1}{c|}{0.281$\pm{0.001}$} & 0.515$\pm{0.000}$   \\ \hline
Accuracy  \textbf{($\mathbf{\uparrow}$)} & 0.868$\pm{0.001}$ & \multicolumn{1}{c|}{0.882$\pm{0.001}$} & \multicolumn{1}{c|}{0.883$\pm{0.001}$}          & 0.882$\pm{0.001}$          & \multicolumn{1}{c|}{\textbf{0.889$\pm{0.000}$}} & \multicolumn{1}{c|}{0.880$\pm{0.001}$}  & 0.842$\pm{0.001}$ \\ \hline
Loss  \textbf{($\mathbf{\downarrow}$)} & 1.051$\pm{0.031}$ & \multicolumn{1}{c|}{0.675$\pm{0.006}$} & \multicolumn{1}{c|}{\textbf{0.572$\pm{0.004}$}} & 0.559$\pm{0.003}$          & \multicolumn{1}{c|}{0.576$\pm{0.002}$}        & \multicolumn{1}{c|}{1.111$\pm{0.003}$} & 2.375$\pm{0.003}$ \\ \hline
\end{tabular}}
\end{table}
\begin{table}[H]
\centering
\caption{VGGish on SpeechCommands mean and $\pm$ 1 SEM  reported from 10 runs with Teacher Seed 2. \textbf{Bold} values are best performing based on the mean. }
\label{tab:sc-VGGISH-ts-2}
\resizebox{\linewidth}{!}{
\begin{tabular}{|l|c|ccc|ccc|}
\hline
\multicolumn{1}{|c|}{\multirow{2}{*}{\textbf{Metrics}}} & \textbf{Control}  & \multicolumn{3}{c|}{\textbf{Knowledge Distillation}}                                                                     & \multicolumn{3}{c|}{\textbf{Random Control Distillation}}                                                     \\ \cline{2-8} 
\multicolumn{1}{|c|}{}                                  & \textbf{SIDDO}     & \multicolumn{1}{c|}{\textbf{0.1}}       & \multicolumn{1}{c|}{\textbf{0.5}}                & \textbf{0.9}                & \multicolumn{1}{c|}{\textbf{0.1}}                & \multicolumn{1}{c|}{\textbf{0.5}}       & \textbf{0.9}       \\ \hline
Activation Distance \textbf{($\mathbf{\downarrow}$)} & 0.192$\pm{0.002}$ & \multicolumn{1}{c|}{0.151$\pm{0.001}$} & \multicolumn{1}{c|}{0.149$\pm{0.000}$}            & \textbf{0.148$\pm{0.001}$} & \multicolumn{1}{c|}{0.260$\pm{0.001}$}           & \multicolumn{1}{c|}{0.572$\pm{0.001}$} & 0.877$\pm{0.000}$   \\ \hline
Rank Disagreement  \textbf{($\mathbf{\downarrow}$)} & 0.908$\pm{0.000}$   & \multicolumn{1}{c|}{0.885$\pm{0.000}$}   & \multicolumn{1}{c|}{0.880$\pm{0.000}$}             & \textbf{0.878$\pm{0.000}$}   & \multicolumn{1}{c|}{0.942$\pm{0.000}$}            & \multicolumn{1}{c|}{0.942$\pm{0.000}$}   & 0.940$\pm{0.000}$    \\ \hline
Prediction Disagreement  \textbf{($\mathbf{\downarrow}$)} & 0.145$\pm{0.002}$ & \multicolumn{1}{c|}{0.117$\pm{0.001}$} & \multicolumn{1}{c|}{\textbf{0.116$\pm{0.001}$}} & 0.115$\pm{0.001}$          & \multicolumn{1}{c|}{0.126$\pm{0.001}$}          & \multicolumn{1}{c|}{0.135$\pm{0.001}$} & 0.166$\pm{0.001}$ \\ \hline
JS Divergence  \textbf{($\mathbf{\downarrow}$)} & 0.085$\pm{0.001}$ & \multicolumn{1}{c|}{0.062$\pm{0.000}$}   & \multicolumn{1}{c|}{0.060$\pm{0.000}$}             & \textbf{0.059$\pm{0.000}$}   & \multicolumn{1}{c|}{0.120$\pm{0.000}$}             & \multicolumn{1}{c|}{0.276$\pm{0.001}$} & 0.511$\pm{0.001}$ \\ \hline
Accuracy  \textbf{($\mathbf{\uparrow}$)} & 0.870$\pm{0.002}$  & \multicolumn{1}{c|}{0.887$\pm{0.000}$}   & \multicolumn{1}{c|}{0.889$\pm{0.001}$}          & 0.889$\pm{0.001}$          & \multicolumn{1}{c|}{\textbf{0.892$\pm{0.001}$}} & \multicolumn{1}{c|}{0.882$\pm{0.000}$}   & 0.847$\pm{0.001}$ \\ \hline
Loss  \textbf{($\mathbf{\downarrow}$)} & 1.086$\pm{0.026}$ & \multicolumn{1}{c|}{0.629$\pm{0.006}$} & \multicolumn{1}{c|}{0.531$\pm{0.003}$}          & \textbf{0.516$\pm{0.003}$} & \multicolumn{1}{c|}{0.562$\pm{0.002}$}          & \multicolumn{1}{c|}{1.111$\pm{0.003}$} & 2.363$\pm{0.004}$ \\ \hline
\end{tabular}}
\end{table}

\subsubsection{Audio Spectrogram Transformer (AST)}

\paragraph{Findings:} We observe that the teacher model achieves a high train accuracy along with a high train loss, see Table \ref{tab:vit-melsc-teacher}. With this we observe a substantial and statistically supported knowledge transfer for all $\alpha$ values, see Tables \ref{tab:sc-vit-ts-0}, \ref{tab:sc-vit-ts-1}, \ref{tab:sc-vit-ts-2} and \ref{tab:vit_sc_sig}. This substantial and significant transfer of knowledge, as expected, coincides with a strong asymmetric transfer of
knowledge favouring incorrect predictions, as shown in Figure~\ref{fig:vgg-sc-pred}.

\begin{table}[H]
\centering
\caption{Teacher Performance on Train and Test Data for AST on SpeechCommands.}
\label{tab:vit-melsc-teacher}
\begin{tabular}{|c|c|c|c|c|}
\hline
\textbf{Teacher Seed} & \textbf{Train Loss} & \textbf{Train Accuracy} & \textbf{Test Loss} & \textbf{Test Accuracy} \\ \hline
0 & 0.013776 & 0.996440 & 1.001014 & 0.833530 \\ \hline
1 & 0.002471 & 0.999352 & 0.925219 & 0.853794 \\ \hline
2 & 0.003337 & 0.999163 & 0.913119 & 0.853430 \\ \hline
\end{tabular}
\end{table}

\begin{table}[H]
\centering
\caption{AST on SpeechCommands mean and $\pm$ 1 SEM  reported from 10 runs with Teacher Seed 0. \textbf{Bold} values are best performing based on the mean. }
\label{tab:sc-vit-ts-0}
\resizebox{\linewidth}{!}{
\begin{tabular}{|l|c|ccc|ccc|}
\hline
\multicolumn{1}{|c|}{\multirow{2}{*}{\textbf{Metrics}}} & \textbf{Basline}   & \multicolumn{3}{c|}{\textbf{Knowledge Distillation}}                                                            & \multicolumn{3}{c|}{\textbf{Random Control Distillation}}                                                         \\ \cline{2-8} 
\multicolumn{1}{|c|}{}                                  & \textbf{SIDDO}     & \multicolumn{1}{c|}{\textbf{0.1}}    & \multicolumn{1}{c|}{\textbf{0.5}}    & \textbf{0.9}             & \multicolumn{1}{c|}{\textbf{0.1}}        & \multicolumn{1}{c|}{\textbf{0.5}} & \textbf{0.9} \\ \hline
Activation Distance \textbf{($\mathbf{\downarrow}$)} & 0.164$\pm{0.001}$ & \multicolumn{1}{c|}{0.133$\pm{0.002}$} & \multicolumn{1}{c|}{0.123$\pm{0.002}$} & \textbf{0.118$\pm{0.002}$} & \multicolumn{1}{c|}{0.245$\pm{0.001}$}          & \multicolumn{1}{c|}{0.561$\pm{0.000}$}     & 0.870$\pm{0.000}$      \\ \hline
Rank Disagreement  \textbf{($\mathbf{\downarrow}$)} & 0.852$\pm{0.001}$ & \multicolumn{1}{c|}{0.825$\pm{0.002}$} & \multicolumn{1}{c|}{0.810$\pm{0.002}$}  & \textbf{0.803$\pm{0.002}$} & \multicolumn{1}{c|}{0.937$\pm{0.000}$}            & \multicolumn{1}{c|}{0.940$\pm{0.000}$}      & 0.939$\pm{0.000}$     \\ \hline
Prediction Disagreement  \textbf{($\mathbf{\downarrow}$)} & 0.124$\pm{0.001}$ & \multicolumn{1}{c|}{0.101$\pm{0.001}$} & \multicolumn{1}{c|}{0.094$\pm{0.001}$} & \textbf{0.090$\pm{0.002}$}  & \multicolumn{1}{c|}{0.136$\pm{0.001}$}          & \multicolumn{1}{c|}{0.154$\pm{0.001}$}   & 0.181$\pm{0.001}$   \\ \hline
JS Divergence  \textbf{($\mathbf{\downarrow}$)} & 0.062$\pm{0.001}$ & \multicolumn{1}{c|}{0.045$\pm{0.001}$} & \multicolumn{1}{c|}{0.039$\pm{0.001}$} & \textbf{0.036$\pm{0.001}$} & \multicolumn{1}{c|}{0.109$\pm{0.000}$}            & \multicolumn{1}{c|}{0.271$\pm{0.000}$}     & 0.512$\pm{0.000}$     \\ \hline
Accuracy  \textbf{($\mathbf{\uparrow}$)} & 0.843$\pm{0.001}$ & \multicolumn{1}{c|}{0.842$\pm{0.000}$}   & \multicolumn{1}{c|}{0.844$\pm{0.000}$}   & 0.844$\pm{0.000}$            & \multicolumn{1}{c|}{\textbf{0.856$\pm{0.001}$}} & \multicolumn{1}{c|}{0.852$\pm{0.000}$}     & 0.826$\pm{0.000}$     \\ \hline
Loss  \textbf{($\mathbf{\downarrow}$)} & 1.094$\pm{0.011}$ & \multicolumn{1}{c|}{0.990$\pm{0.005}$}  & \multicolumn{1}{c|}{0.835$\pm{0.003}$} & 0.791$\pm{0.002}$          & \multicolumn{1}{c|}{\textbf{0.687$\pm{0.002}$}} & \multicolumn{1}{c|}{1.161$\pm{0.001}$}   & 2.408$\pm{0.001}$   \\ \hline
\end{tabular}}
\end{table}
\begin{table}[H]
\centering
\caption{AST on SpeechCommands mean and $\pm$ 1 SEM  reported from 10 runs with Teacher Seed 1. \textbf{Bold} values are best performing based on the mean. }
\label{tab:sc-vit-ts-1}
\resizebox{\linewidth}{!}{
\begin{tabular}{|l|c|ccc|ccc|}
\hline
\multicolumn{1}{|c|}{\multirow{2}{*}{\textbf{Metrics}}} & \textbf{Control}  & \multicolumn{3}{c|}{\textbf{Knowledge Distillation}}                                                            & \multicolumn{3}{c|}{\textbf{Random Control Distillation}}                                                         \\ \cline{2-8} 
\multicolumn{1}{|c|}{}                                  & \textbf{SIDDO}     & \multicolumn{1}{c|}{\textbf{0.1}}    & \multicolumn{1}{c|}{\textbf{0.5}}    & \textbf{0.9}             & \multicolumn{1}{c|}{\textbf{0.1}}        & \multicolumn{1}{c|}{\textbf{0.5}} & \textbf{0.9} \\ \hline
Activation Distance \textbf{($\mathbf{\downarrow}$)} & 0.143$\pm{0.006}$ & \multicolumn{1}{c|}{0.129$\pm{0.002}$} & \multicolumn{1}{c|}{0.119$\pm{0.002}$} & \textbf{0.115$\pm{0.002}$} & \multicolumn{1}{c|}{0.227$\pm{0.001}$}          & \multicolumn{1}{c|}{0.558$\pm{0.000}$}     & 0.874$\pm{0.000}$     \\ \hline
Rank Disagreement  \textbf{($\mathbf{\downarrow}$)} & 0.844$\pm{0.003}$ & \multicolumn{1}{c|}{0.833$\pm{0.002}$} & \multicolumn{1}{c|}{0.821$\pm{0.002}$} & \textbf{0.814$\pm{0.002}$} & \multicolumn{1}{c|}{0.935$\pm{0.000}$}            & \multicolumn{1}{c|}{0.939$\pm{0.000}$}     & 0.938$\pm{0.000}$     \\ \hline
Prediction Disagreement  \textbf{($\mathbf{\downarrow}$)} & 0.107$\pm{0.005}$ & \multicolumn{1}{c|}{0.097$\pm{0.002}$} & \multicolumn{1}{c|}{0.090$\pm{0.001}$}  & \textbf{0.087$\pm{0.001}$} & \multicolumn{1}{c|}{0.113$\pm{0.001}$}          & \multicolumn{1}{c|}{0.138$\pm{0.001}$}   & 0.162$\pm{0.001}$   \\ \hline
JS Divergence  \textbf{($\mathbf{\downarrow}$)} & 0.053$\pm{0.003}$ & \multicolumn{1}{c|}{0.045$\pm{0.001}$} & \multicolumn{1}{c|}{0.040$\pm{0.001}$}  & \textbf{0.038$\pm{0.001}$} & \multicolumn{1}{c|}{0.100$\pm{0.000}$}              & \multicolumn{1}{c|}{0.266$\pm{0.000}$}     & 0.512$\pm{0.000}$     \\ \hline
Accuracy  \textbf{($\mathbf{\uparrow}$)} & 0.849$\pm{0.004}$ & \multicolumn{1}{c|}{0.854$\pm{0.001}$} & \multicolumn{1}{c|}{0.854$\pm{0.000}$}   & 0.855$\pm{0.001}$          & \multicolumn{1}{c|}{\textbf{0.863$\pm{0.000}$}}   & \multicolumn{1}{c|}{0.858$\pm{0.000}$}     & 0.835$\pm{0.000}$     \\ \hline
Loss  \textbf{($\mathbf{\downarrow}$)} & 1.071$\pm{0.020}$  & \multicolumn{1}{c|}{0.994$\pm{0.006}$} & \multicolumn{1}{c|}{0.941$\pm{0.003}$} & 0.900$\pm{0.002}$            & \multicolumn{1}{c|}{\textbf{0.656$\pm{0.002}$}} & \multicolumn{1}{c|}{1.138$\pm{0.002}$}   & 2.394$\pm{0.001}$   \\ \hline
\end{tabular}}
\end{table}

\vspace{-0.3cm}
\begin{table}[H]
\centering
\caption{AST on SpeechCommands mean and $\pm$ 1 SEM  reported from 10 runs with Teacher Seed 2. \textbf{Bold} values are best performing based on the mean. }
\label{tab:sc-vit-ts-2}
\resizebox{\linewidth}{!}{
\begin{tabular}{|l|c|ccc|ccc|}
\hline
\multicolumn{1}{|c|}{\multirow{2}{*}{\textbf{Metric}}} & \textbf{Control}  & \multicolumn{3}{c|}{\textbf{Knowledge Distillation}}                                                            & \multicolumn{3}{c|}{\textbf{Random Control Distillation}}                                                         \\ \cline{2-8} 
\multicolumn{1}{|c|}{}                                 & \textbf{SIDDO}     & \multicolumn{1}{c|}{\textbf{0.1}}    & \multicolumn{1}{c|}{\textbf{0.5}}    & \textbf{0.9}             & \multicolumn{1}{c|}{\textbf{0.1}}        & \multicolumn{1}{c|}{\textbf{0.5}} & \textbf{0.9} \\ \hline
Activation Distance                                    & 0.152$\pm{0.005}$ & \multicolumn{1}{c|}{0.139$\pm{0.002}$} & \multicolumn{1}{c|}{0.131$\pm{0.002}$} & \textbf{0.126$\pm{0.002}$} & \multicolumn{1}{c|}{0.232$\pm{0.002}$}          & \multicolumn{1}{c|}{0.560$\pm{0.000}$}      & 0.875$\pm{0.000}$     \\ \hline
Rank Disagreement                                      & 0.852$\pm{0.003}$ & \multicolumn{1}{c|}{0.844$\pm{0.002}$} & \multicolumn{1}{c|}{0.833$\pm{0.002}$} & \textbf{0.826$\pm{0.003}$} & \multicolumn{1}{c|}{0.936$\pm{0.000}$}            & \multicolumn{1}{c|}{0.939$\pm{0.000}$}     & 0.938$\pm{0.000}$     \\ \hline
Prediction Disagreement                                & 0.115$\pm{0.003}$ & \multicolumn{1}{c|}{0.105$\pm{0.001}$} & \multicolumn{1}{c|}{0.100$\pm{0.001}$}   & \textbf{0.096$\pm{0.001}$} & \multicolumn{1}{c|}{0.122$\pm{0.002}$}          & \multicolumn{1}{c|}{0.141$\pm{0.002}$}   & 0.163$\pm{0.001}$   \\ \hline
JS Divergence                                          & 0.058$\pm{0.002}$ & \multicolumn{1}{c|}{0.051$\pm{0.001}$} & \multicolumn{1}{c|}{0.046$\pm{0.001}$} & \textbf{0.043$\pm{0.001}$} & \multicolumn{1}{c|}{0.102$\pm{0.001}$}          & \multicolumn{1}{c|}{0.267$\pm{0.000}$}     & 0.512$\pm{0.000}$     \\ \hline
Accuracy                                               & 0.852$\pm{0.003}$ & \multicolumn{1}{c|}{0.857$\pm{0.001}$} & \multicolumn{1}{c|}{0.856$\pm{0.001}$} & 0.857$\pm{0.001}$          & \multicolumn{1}{c|}{\textbf{0.860$\pm{0.003}$}}  & \multicolumn{1}{c|}{0.852$\pm{0.002}$}   & 0.827$\pm{0.000}$     \\ \hline
Loss                                                   & 1.027$\pm{0.014}$ & \multicolumn{1}{c|}{0.955$\pm{0.004}$} & \multicolumn{1}{c|}{0.897$\pm{0.002}$} & 0.860$\pm{0.003}$           & \multicolumn{1}{c|}{\textbf{0.661$\pm{0.008}$}} & \multicolumn{1}{c|}{1.152$\pm{0.003}$}   & 2.398$\pm{0.001}$   \\ \hline
\end{tabular}}
\end{table}

\vspace{-0.3cm}
\begin{table}[H]
\caption{AST on SpeechCommands (significance testing). \cmark~indicates significant results compared to controls; \xmark~indicates insignificant results. Each tick represents a teacher (seeds 0 to 2, left to right).}
\label{tab:vit_sc_sig}
\resizebox{\linewidth}{!}{
\begin{tabular}{|l|l|l|l|l|l|l|}
\hline
\textbf{} & \textbf{Activation Distance}                                      & \textbf{Rank Disagreement}                                        & \textbf{Prediction Disagreement}                                  & \textbf{JS Divergence}                                            & \textbf{Accuracy}                                                 & \textbf{Loss}                                                     \\ \hline
KD 0.1    & \cmark \cmark \cmark & \cmark \cmark \cmark & \cmark \cmark \cmark & \cmark \cmark \cmark & \xmark \xmark \xmark & \xmark \xmark \xmark \\ \hline
KD 0.5    & \cmark \cmark \cmark & \cmark \cmark \cmark & \cmark \cmark \cmark & \cmark \cmark \cmark & \xmark \xmark \xmark & \xmark \xmark \xmark \\ \hline
KD 0.9    & \cmark \cmark \cmark & \cmark \cmark \cmark & \cmark \cmark \cmark & \cmark \cmark \cmark & \xmark \xmark \xmark & \xmark \xmark \xmark \\ \hline
\end{tabular}}
\end{table}
\begin{figure}[H]
    \centering
    \includegraphics[width=\linewidth] {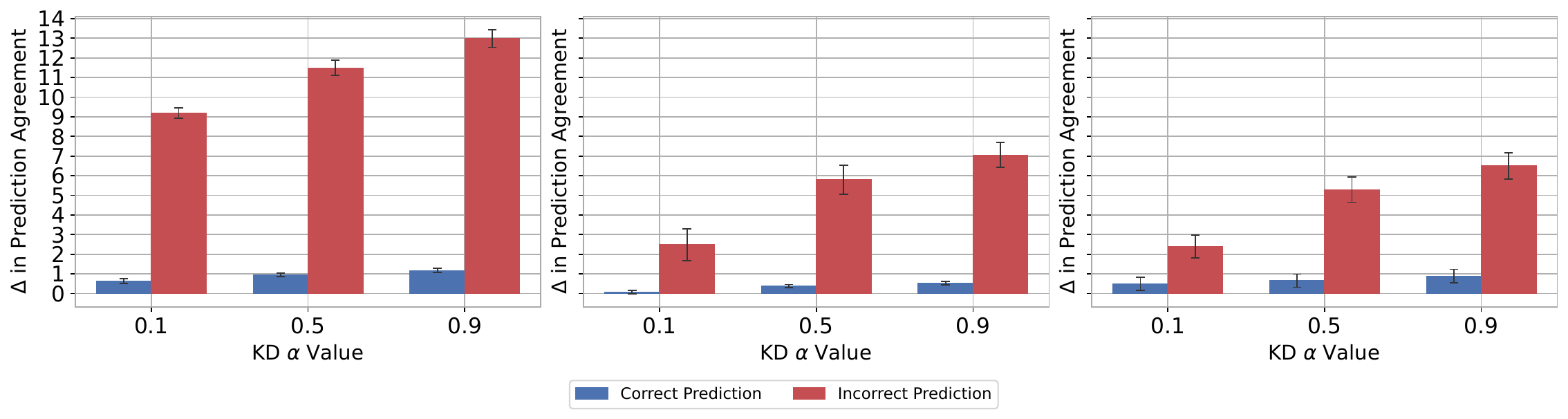}
    \caption{Prediction agreement difference of student models in standard KD to the highest performing control baseline with respect to correct prediction agreement (blue) and incorrect prediction agreement (red), for AST on SpeechCommands (seeds 0 to 2, left to right).}
\end{figure}
\subsection{UrbanSound8K}
UrbanSound8K is a large event classification dataset that contains 18.5 hours of annotated sound event occurrences across 10 classes~\citep{10.1145/2647868.2655045}. It has 6,985 training set instances and 1,747 testing set instances which are between 0 and 4 seconds in duration. 

\paragraph{Findings:}We find that for UrbanSound8K knowledge transfer is statistically supported allowing the rejection of the null hypothesis for knowledge sharing mainly for the VGG but not for the AST. For the VGG architecture there is considerable knowledge transfer compared to the baseline controls, but for the transformer architecture there is only marginal knowledge transfer which relates to the difference in teacher loss between the VGG and AST. We also find that there is asymmetric knowledge transfer with a weighting towards negative knowledge transfer when the knowledge transfer is statistically supported and considerable.
\subsubsection{VGGish}
\begin{table}[H]
\centering
\caption{Teacher Performance on Train and Test Data for VGGish on UrbanSound8K.}
\label{tab:vgg-melub8k-teacher}
\begin{tabular}{|c|c|c|c|c|}
\hline
\textbf{Teacher Seed} & \textbf{Train Loss} & \textbf{Train Accuracy} & \textbf{Test Loss} & \textbf{Test Accuracy} \\ \hline
0 & 0.013431 & 0.994989 & 2.203087 & 0.797939 \\ \hline
1 & 0.014136 & 0.994560 & 2.405788 & 0.785346 \\ \hline
2 & 0.151926 & 0.947173 & 1.568569 & 0.702919 \\ \hline
\end{tabular}
\end{table}
\begin{table}[H]
\centering
\caption{VGGish on UrbanSound8K mean and $\pm$ 1 SEM  reported from 10 runs with Teacher Seed 1. \textbf{Bold} values are best performing based on the mean. }
\resizebox{\linewidth}{!}{
\begin{tabular}{|l|c|ccc|ccc|}
\hline
\multirow{2}{*}{\textbf{Metrics}} & \textbf{Control}  & \multicolumn{3}{c|}{\textbf{Knowledge Distillation}}                                                                     & \multicolumn{3}{c|}{\textbf{Random Control Distillation}}                                                         \\ \cline{2-8} 
                                  & \textbf{SIDDO}     & \multicolumn{1}{c|}{\textbf{0.1}}    & \multicolumn{1}{c|}{\textbf{0.5}}             & \textbf{0.9}             & \multicolumn{1}{c|}{\textbf{0.1}}        & \multicolumn{1}{c|}{\textbf{0.5}} & \textbf{0.9} \\ \hline
Activation Distance               & 0.363$\pm{0.047}$ & \multicolumn{1}{c|}{0.284$\pm{0.010}$}  & \multicolumn{1}{c|}{\textbf{0.262$\pm{0.002}$}} & 0.264$\pm{0.002}$          & \multicolumn{1}{c|}{0.367$\pm{0.002}$}          & \multicolumn{1}{c|}{0.600$\pm{0.002}$}     & 0.871$\pm{0.001}$   \\ \hline
Rank Disagreement                 & 0.730$\pm{0.009}$  & \multicolumn{1}{c|}{0.718$\pm{0.005}$} & \multicolumn{1}{c|}{0.706$\pm{0.002}$}          & \textbf{0.703$\pm{0.002}$} & \multicolumn{1}{c|}{0.798$\pm{0.001}$}          & \multicolumn{1}{c|}{0.792$\pm{0.001}$}   & 0.784$\pm{0.001}$   \\ \hline
Prediction Disagreement           & 0.272$\pm{0.035}$ & \multicolumn{1}{c|}{0.214$\pm{0.006}$} & \multicolumn{1}{c|}{\textbf{0.197$\pm{0.002}$}} & 0.199$\pm{0.001}$          & \multicolumn{1}{c|}{0.208$\pm{0.003}$}          & \multicolumn{1}{c|}{0.218$\pm{0.003}$}   & 0.387$\pm{0.003}$   \\ \hline
JS Divergence                     & inf, nan     & \multicolumn{1}{c|}{inf, nan}    & \multicolumn{1}{c|}{inf, nan}              & inf, nan             & \multicolumn{1}{c|}{0.156$\pm{0.001}$}          & \multicolumn{1}{c|}{0.269$\pm{0.001}$}   & 0.465$\pm{0.000}$     \\ \hline
Accuracy                          & 0.724$\pm{0.036}$ & \multicolumn{1}{c|}{0.782$\pm{0.006}$} & \multicolumn{1}{c|}{0.791$\pm{0.002}$}          & 0.791$\pm{0.002}$          & \multicolumn{1}{c|}{\textbf{0.806$\pm{0.002}$}} & \multicolumn{1}{c|}{0.796$\pm{0.003}$}   & 0.589$\pm{0.003}$   \\ \hline
Loss                              & 2.046$\pm{0.321}$ & \multicolumn{1}{c|}{3.056$\pm{0.321}$} & \multicolumn{1}{c|}{2.34$\pm{0.074}$}           & 2.235$\pm{0.089}$          & \multicolumn{1}{c|}{\textbf{0.748$\pm{0.006}$}} & \multicolumn{1}{c|}{1.093$\pm{0.003}$}   & 2.054$\pm{0.003}$   \\ \hline
\end{tabular}}
\end{table}
\vspace{-0.5cm}
\begin{table}[H]
\centering
\caption{VGGish on UrbanSound8K mean and $\pm$ 1 SEM  reported from 10 runs with Teacher Seed 2. \textbf{Bold} values are best performing based on the mean. }
\resizebox{\linewidth}{!}{
\begin{tabular}{|l|c|ccc|ccc|}
\hline
\multirow{2}{*}{\textbf{Metrics}} & \textbf{Control}  & \multicolumn{3}{c|}{\textbf{Knowledge Distillation}}                                                            & \multicolumn{3}{c|}{\textbf{Random Control Distillation}}                                                         \\ \cline{2-8} 
                                  & \textbf{SIDDO}     & \multicolumn{1}{c|}{\textbf{0.1}}    & \multicolumn{1}{c|}{\textbf{0.5}}    & \textbf{0.9}             & \multicolumn{1}{c|}{\textbf{0.1}}        & \multicolumn{1}{c|}{\textbf{0.5}} & \textbf{0.9} \\ \hline
Activation Distance               & 0.396$\pm{0.002}$ & \multicolumn{1}{c|}{0.357$\pm{0.002}$} & \multicolumn{1}{c|}{0.335$\pm{0.001}$} & \textbf{0.324$\pm{0.002}$} & \multicolumn{1}{c|}{0.416$\pm{0.003}$}          & \multicolumn{1}{c|}{0.590$\pm{0.001}$}    & 0.821$\pm{0.000}$     \\ \hline
Rank Disagreement                 & 0.745$\pm{0.003}$ & \multicolumn{1}{c|}{0.712$\pm{0.001}$} & \multicolumn{1}{c|}{0.692$\pm{0.002}$} & \textbf{0.683$\pm{0.001}$} & \multicolumn{1}{c|}{0.812$\pm{0.001}$}          & \multicolumn{1}{c|}{0.806$\pm{0.001}$}   & 0.801$\pm{0.001}$   \\ \hline
Prediction Disagreement           & 0.295$\pm{0.002}$ & \multicolumn{1}{c|}{0.274$\pm{0.002}$} & \multicolumn{1}{c|}{0.260$\pm{0.002}$}  & \textbf{0.253$\pm{0.002}$} & \multicolumn{1}{c|}{0.292$\pm{0.004}$}          & \multicolumn{1}{c|}{0.293$\pm{0.002}$}   & 0.438$\pm{0.002}$   \\ \hline
JS Divergence                     & 0.167$\pm{0.001}$ & \multicolumn{1}{c|}{0.141$\pm{0.001}$} & \multicolumn{1}{c|}{0.127$\pm{0.001}$} & \textbf{0.120$\pm{0.001}$}  & \multicolumn{1}{c|}{0.175$\pm{0.001}$}          & \multicolumn{1}{c|}{0.264$\pm{0.001}$}   & 0.433$\pm{0.000}$     \\ \hline
Accuracy                          & 0.794$\pm{0.003}$ & \multicolumn{1}{c|}{0.789$\pm{0.004}$} & \multicolumn{1}{c|}{0.791$\pm{0.002}$} & 0.776$\pm{0.002}$          & \multicolumn{1}{c|}{\textbf{0.810$\pm{0.003}$}}  & \multicolumn{1}{c|}{0.808$\pm{0.002}$}   & 0.577$\pm{0.001}$   \\ \hline
Loss                              & 3.209$\pm{0.375}$ & \multicolumn{1}{c|}{1.106$\pm{0.024}$} & \multicolumn{1}{c|}{0.944$\pm{0.016}$} & 0.961$\pm{0.013}$          & \multicolumn{1}{c|}{\textbf{0.716$\pm{0.006}$}} & \multicolumn{1}{c|}{1.080$\pm{0.003}$}    & 2.065$\pm{0.002}$   \\ \hline
\end{tabular}}
\end{table}

\subsubsection{Audio Spectrogram Transformer (AST)}
\begin{table}[H]
\centering
\caption{Teacher Performance on Train and Test Data for AST on UrbanSound8K.}
\label{tab:vit-melub8k-teacher}
\begin{tabular}{|c|c|c|c|c|}
\hline
\textbf{Teacher Seed} & \textbf{Train Loss} & \textbf{Train Accuracy} & \textbf{Test Loss} & \textbf{Test Accuracy} \\ \hline
0 & 0.000180 & 1.000000 & 1.638960 & 0.772753 \\ \hline
1 & 0.000375 & 0.999857 & 1.583644 & 0.768746 \\ \hline
2 & 0.000168 & 1.000000 & 1.593121 & 0.781912 \\ \hline
\end{tabular}
\end{table}
\vspace{-0.5cm}

\begin{table}[H]
\centering
\caption{AST on UrbanSound8K mean and $\pm$ 1 SEM  reported from 10 runs with Teacher Seed 0. \textbf{Bold} values are best performing based on the mean. }
\resizebox{\linewidth}{!}{
\begin{tabular}{|l|c|ccc|ccc|}
\hline
\multicolumn{1}{|c|}{\multirow{2}{*}{\textbf{Metrics}}} & \textbf{Control}  & \multicolumn{3}{c|}{\textbf{Knowledge Distillation}}                                                                              & \multicolumn{3}{c|}{\textbf{Random Control Distillation}}                                                         \\ \cline{2-8} 
\multicolumn{1}{|c|}{}                                  & \textbf{SIDDO}     & \multicolumn{1}{c|}{\textbf{0.1}}             & \multicolumn{1}{c|}{\textbf{0.5}}             & \textbf{0.9}             & \multicolumn{1}{c|}{\textbf{0.1}}        & \multicolumn{1}{c|}{\textbf{0.5}} & \textbf{0.9} \\ \hline
Activation Distance \textbf{($\mathbf{\downarrow}$)} & 0.098$\pm{0.001}$ & \multicolumn{1}{c|}{0.098$\pm{0.001}$}          & \multicolumn{1}{c|}{\textbf{0.096$\pm{0.001}$}} & 0.097$\pm{0.002}$          & \multicolumn{1}{c|}{0.287$\pm{0.000}$}            & \multicolumn{1}{c|}{0.592$\pm{0.001}$}   & 0.854$\pm{0.000}$     \\ \hline
Rank Disagreement  \textbf{($\mathbf{\downarrow}$)} & 0.423$\pm{0.003}$ & \multicolumn{1}{c|}{0.419$\pm{0.002}$}          & \multicolumn{1}{c|}{0.417$\pm{0.002}$}          & \textbf{0.415$\pm{0.003}$} & \multicolumn{1}{c|}{0.755$\pm{0.001}$}          & \multicolumn{1}{c|}{0.773$\pm{0.001}$}   & 0.759$\pm{0.001}$   \\ \hline
Prediction Disagreement  \textbf{($\mathbf{\downarrow}$)} & 0.074$\pm{0.002}$ & \multicolumn{1}{c|}{\textbf{0.072$\pm{0.001}$}} & \multicolumn{1}{c|}{0.073$\pm{0.001}$}          & 0.073$\pm{0.002}$          & \multicolumn{1}{c|}{0.131$\pm{0.001}$}          & \multicolumn{1}{c|}{0.174$\pm{0.001}$}   & 0.252$\pm{0.003}$   \\ \hline
JS Divergence  \textbf{($\mathbf{\downarrow}$)} & 0.025$\pm{0.001}$ & \multicolumn{1}{c|}{0.025$\pm{0.000}$}            & \multicolumn{1}{c|}{\textbf{0.024$\pm{0.000}$}}   & 0.025$\pm{0.001}$          & \multicolumn{1}{c|}{0.111$\pm{0.000}$}            & \multicolumn{1}{c|}{0.262$\pm{0.000}$}     & 0.448$\pm{0.000}$     \\ \hline
Accuracy  \textbf{($\mathbf{\uparrow}$)} & 0.771$\pm{0.001}$ & \multicolumn{1}{c|}{0.771$\pm{0.001}$}          & \multicolumn{1}{c|}{0.771$\pm{0.001}$}          & 0.772$\pm{0.001}$          & \multicolumn{1}{c|}{0.788$\pm{0.001}$} & \multicolumn{1}{c|}{\textbf{0.806$\pm{0.001}$}}   & 0.719$\pm{0.002}$   \\ \hline
Loss  \textbf{($\mathbf{\downarrow}$)} & 1.628$\pm{0.010}$  & \multicolumn{1}{c|}{1.621$\pm{0.009}$}          & \multicolumn{1}{c|}{1.585$\pm{0.006}$}          & 1.560$\pm{0.008}$           & \multicolumn{1}{c|}{\textbf{0.748$\pm{0.001}$}} & \multicolumn{1}{c|}{1.095$\pm{0.001}$}   & 1.956$\pm{0.001}$   \\ \hline
\end{tabular}}
\end{table}
\vspace{-0.5cm}
\begin{table}[H]
\centering
\caption{AST on UrbanSound8K mean and $\pm$ 1 SEM  reported from 10 runs with Teacher Seed 1. \textbf{Bold} values are best performing based on the mean. }
\resizebox{\linewidth}{!}{
\begin{tabular}{|l|c|ccc|ccc|}
\hline
\multirow{2}{*}{\textbf{Metrics}} & \textbf{Control}  & \multicolumn{3}{c|}{\textbf{Knowledge Distillation}}                                                            & \multicolumn{3}{c|}{\textbf{Rand Knowledge Distillation}}                                                           \\ \cline{2-8} 
                                  & \textbf{SIDDO}     & \multicolumn{1}{c|}{\textbf{0.1}}    & \multicolumn{1}{c|}{\textbf{0.5}}    & \textbf{0.9}             & \multicolumn{1}{c|}{\textbf{0.1}}        & \multicolumn{1}{c|}{\textbf{0.5}} & \textbf{0.9} \\ \hline
Activation Distance               & 0.109$\pm{0.001}$ & \multicolumn{1}{c|}{0.108$\pm{0.001}$} & \multicolumn{1}{c|}{0.108$\pm{0.001}$} & \textbf{0.105$\pm{0.001}$} & \multicolumn{1}{c|}{0.291$\pm{0.001}$}          & \multicolumn{1}{c|}{0.592$\pm{0.001}$}   & 0.854$\pm{0.000}$     \\ \hline
Rank Disagreement                 & 0.442$\pm{0.002}$ & \multicolumn{1}{c|}{0.44$\pm{0.002}$}  & \multicolumn{1}{c|}{0.429$\pm{0.002}$} & \textbf{0.427$\pm{0.002}$} & \multicolumn{1}{c|}{0.756$\pm{0.001}$}          & \multicolumn{1}{c|}{0.769$\pm{0.001}$}   & 0.763$\pm{0.001}$   \\ \hline
Prediction Disagreement           & 0.078$\pm{0.001}$ & \multicolumn{1}{c|}{0.077$\pm{0.002}$} & \multicolumn{1}{c|}{0.077$\pm{0.001}$} & \textbf{0.073$\pm{0.001}$} & \multicolumn{1}{c|}{0.130$\pm{0.001}$}           & \multicolumn{1}{c|}{0.173$\pm{0.001}$}   & 0.261$\pm{0.003}$   \\ \hline
JS Divergence                     & 0.029$\pm{0.000}$   & \multicolumn{1}{c|}{0.029$\pm{0.001}$} & \multicolumn{1}{c|}{0.028$\pm{0.001}$} & \textbf{0.027$\pm{0.000}$}   & \multicolumn{1}{c|}{0.113$\pm{0.000}$}            & \multicolumn{1}{c|}{0.262$\pm{0.000}$}     & 0.448$\pm{0.000}$     \\ \hline
Accuracy                          & 0.768$\pm{0.001}$ & \multicolumn{1}{c|}{0.768$\pm{0.002}$} & \multicolumn{1}{c|}{0.770$\pm{0.001}$}  & 0.769$\pm{0.001}$          & \multicolumn{1}{c|}{0.794$\pm{0.001}$} & \multicolumn{1}{c|}{\textbf{0.811$\pm{0.001}$}}   & 0.716$\pm{0.003}$   \\ \hline
Loss                              & 1.589$\pm{0.010}$  & \multicolumn{1}{c|}{1.584$\pm{0.009}$} & \multicolumn{1}{c|}{1.532$\pm{0.008}$} & 1.509$\pm{0.009}$          & \multicolumn{1}{c|}{\textbf{0.735$\pm{0.001}$}} & \multicolumn{1}{c|}{1.096$\pm{0.002}$}   & 1.959$\pm{0.002}$   \\ \hline
\end{tabular}}
\end{table}
\vspace{-0.5cm}
\begin{table}[H]
\centering
\caption{AST on UrbanSound8K mean and $\pm$ 1 SEM  reported from 10 runs with Teacher Seed 2. \textbf{Bold} values are best performing based on the mean. }
\resizebox{\linewidth}{!}{
\begin{tabular}{|l|c|ccc|ccc|}
\hline
\multicolumn{1}{|c|}{\multirow{2}{*}{\textbf{Metrics}}} & \textbf{Control}           & \multicolumn{3}{c|}{\textbf{Knowledge Distillation}}                                                            & \multicolumn{3}{c|}{\textbf{Random Control Distillation}}                                                                \\ \cline{2-8} 
\multicolumn{1}{|c|}{}                                  & \textbf{SIDDO}              & \multicolumn{1}{c|}{\textbf{0.1}}    & \multicolumn{1}{c|}{\textbf{0.5}}             & \textbf{0.9}    & \multicolumn{1}{c|}{\textbf{0.1}}        & \multicolumn{1}{c|}{\textbf{0.5}}        & \textbf{0.9} \\ \hline
Activation Distance \textbf{($\mathbf{\downarrow}$)} & \textbf{0.099$\pm{0.002}$} & \multicolumn{1}{c|}{0.100$\pm{0.001}$}   & \multicolumn{1}{c|}{0.100$\pm{0.002}$}            & 0.101$\pm{0.002}$ & \multicolumn{1}{c|}{0.288$\pm{0.001}$}          & \multicolumn{1}{c|}{0.598$\pm{0.000}$}            & 0.859$\pm{0.000}$     \\ \hline
Rank Disagreement  \textbf{($\mathbf{\downarrow}$)} & 0.413$\pm{0.003}$          & \multicolumn{1}{c|}{0.414$\pm{0.003}$} & \multicolumn{1}{c|}{\textbf{0.410$\pm{0.003}$}}  & 0.425$\pm{0.005}$ & \multicolumn{1}{c|}{0.754$\pm{0.001}$}          & \multicolumn{1}{c|}{0.770$\pm{0.001}$}           & 0.759$\pm{0.001}$   \\ \hline
Prediction Disagreement  \textbf{($\mathbf{\downarrow}$)} & 0.071$\pm{0.002}$          & \multicolumn{1}{c|}{0.071$\pm{0.002}$} & \multicolumn{1}{c|}{\textbf{0.068$\pm{0.001}$}} & 0.072$\pm{0.002}$ & \multicolumn{1}{c|}{0.130$\pm{0.001}$}           & \multicolumn{1}{c|}{0.171$\pm{0.002}$}          & 0.257$\pm{0.002}$   \\ \hline
JS Divergence  \textbf{($\mathbf{\downarrow}$)} & 0.026$\pm{0.001}$          & \multicolumn{1}{c|}{0.026$\pm{0.001}$} & \multicolumn{1}{c|}{\textbf{0.026$\pm{0.001}$}} & 0.027$\pm{0.001}$ & \multicolumn{1}{c|}{0.111$\pm{0.000}$}            & \multicolumn{1}{c|}{0.265$\pm{0.000}$}            & 0.451$\pm{0.000}$     \\ \hline
Accuracy  \textbf{($\mathbf{\uparrow}$)} & 0.786$\pm{0.001}$          & \multicolumn{1}{c|}{0.784$\pm{0.001}$} & \multicolumn{1}{c|}{0.783$\pm{0.001}$}          & 0.783$\pm{0.001}$ & \multicolumn{1}{c|}{0.801$\pm{0.001}$}          & \multicolumn{1}{c|}{\textbf{0.812$\pm{0.001}$}} & 0.719$\pm{0.002}$   \\ \hline
Loss  \textbf{($\mathbf{\downarrow}$)} & 1.539$\pm{0.006}$          & \multicolumn{1}{c|}{1.538$\pm{0.008}$} & \multicolumn{1}{c|}{1.508$\pm{0.007}$}          & 1.484$\pm{0.008}$ & \multicolumn{1}{c|}{\textbf{0.716$\pm{0.001}$}} & \multicolumn{1}{c|}{1.091$\pm{0.001}$}          & 1.959$\pm{0.002}$   \\ \hline
\end{tabular}}
\end{table}
\begin{figure}[H]
    \centering
    \includegraphics[width=\linewidth]{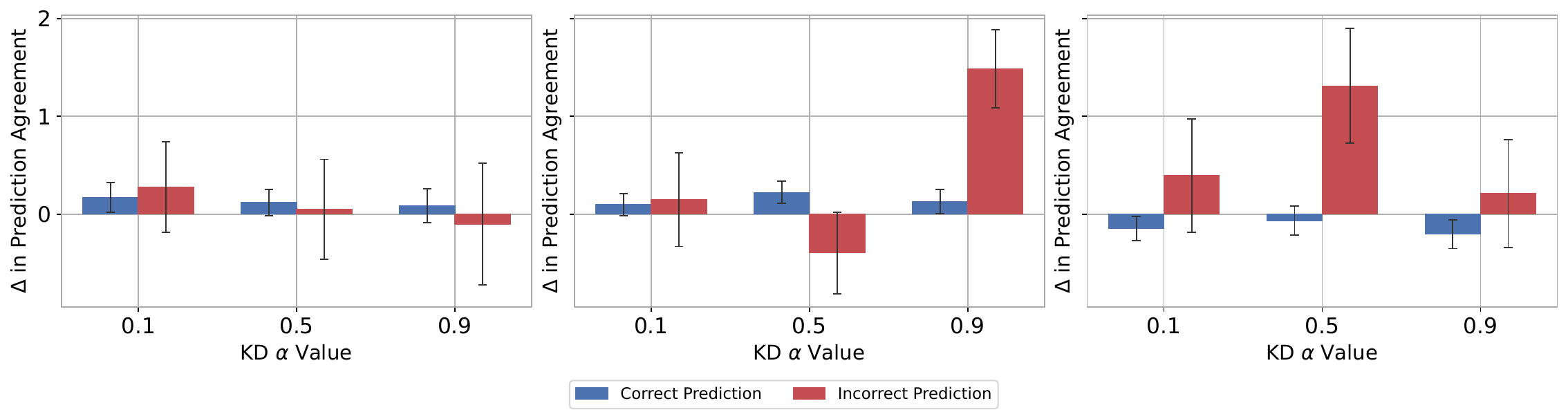}
    \caption{Prediction agreement difference of student models in standard KD to the highest performing control baseline with respect to correct prediction agreement (blue) and incorrect prediction agreement (red), for AST on UrbanSound8K (seeds 0 to 2, left to right).}
\end{figure}
\begin{table}[H]
\caption{AST on UrbanSound8K (significance testing). \cmark~indicates significant results compared to controls; \xmark~indicates insignificant results. Each tick represents a teacher (seeds 0 to 2, left to right).}
\label{tab:8k_vit_sig}
\resizebox{\linewidth}{!}{
\begin{tabular}{|l|l|l|l|l|l|l|}
\hline
\textbf{} & \textbf{Activation Distance}                                      & \textbf{Rank Disagreement}                                        & \textbf{Prediction Disagreement}                                  & \textbf{JS Divergence}                                            & \textbf{Accuracy}                                                 & \textbf{Loss}                                                     \\ \hline
KD 0.1    & \xmark \xmark \xmark & \xmark \xmark \xmark & \xmark \xmark \xmark & \xmark \xmark \xmark & \xmark \xmark \xmark & \xmark \xmark \xmark \\ \hline
KD 0.5    & \xmark \xmark \xmark & \xmark \cmark \xmark & \xmark \xmark \xmark & \xmark \xmark \xmark & \xmark \xmark \xmark & \xmark \xmark \xmark \\ \hline
KD 0.9    & \xmark \cmark \xmark & \xmark \cmark \xmark & \xmark \cmark \xmark & \xmark \cmark \xmark & \xmark \xmark \xmark & \xmark \xmark \xmark \\ \hline
\end{tabular}}
\end{table}
\section{Language Results}
\label{Language_results}
\subsection{TinyShakespeare Dataset}
\paragraph{Training Settings:} The language model was a GPT2-style transformer with an embedding dimension of 384, a vocabulary size of 65, six attention heads, six transformer blocks, a dropout of 0.2, and a block size of 256. It was trained on the TinyShakespeare dataset, with the first 90\% used for training and the last 10\% used for testing. The dataset was tokenised via a character tokenizer, and the model was trained auto-regressively to predict the next character token. The model was trained with the Adam optimiser with a learning rate of 3e-4 with a batch size of 64 for 5000 iterations. The student models are trained with the same seeds and data orders from seeds 10 to 19 for the 10 models used for averaging. This is repeated for the three teachers trained on seeds 0 to 2. 

\begin{table}[H]
\centering
\caption{Nano-GPT on TinyShakespeare mean and $\pm$ 1 SEM  reported from 10 runs with Teacher Seed 1. \textbf{Bold} values are best performing based on the mean. }
\label{tab:shake-gpt-ts-1}
\resizebox{\textwidth}{!}{%
\begin{tabular}{|c|c|ccc|ccc|}
\hline
\multicolumn{1}{|c|}{\multirow{2}{*}{\textbf{Metrics}}} & \multicolumn{1}{c|}{\textbf{Control}} & \multicolumn{3}{c|}{\textbf{Knowledge Distillation}} & \multicolumn{3}{c|}{\textbf{Random Control Distillation}} \\ \cline{2-8} 
\multicolumn{1}{|c|}{} & \multicolumn{1}{c|}{\textbf{SIDDO}} & \multicolumn{1}{c|}{\textbf{0.1}} & \multicolumn{1}{c|}{\textbf{0.5}} & \multicolumn{1}{c|}{\textbf{0.9}} & \multicolumn{1}{c|}{\textbf{0.1}} & \multicolumn{1}{c|}{\textbf{0.5}} & \multicolumn{1}{c|}{\textbf{0.9}} \\ \hline
Activation Distance \textbf{($\mathbf{\downarrow}$)} & 0.195$\pm{0.000}$ & \multicolumn{1}{c|}{0.185$\pm{0.000}$} & \multicolumn{1}{c|}{0.156$\pm{0.000}$} & \textbf{0.141$\pm{0.000}$} & \multicolumn{1}{c|}{0.201$\pm{0.000}$} & \multicolumn{1}{c|}{0.370$\pm{0.000}$} & 0.653$\pm{0.000}$ \\ \hline
Rank Disagreement  \textbf{($\mathbf{\downarrow}$)} & 0.910$\pm{0.000}$ & \multicolumn{1}{c|}{0.907$\pm{0.000}$} & \multicolumn{1}{c|}{0.897$\pm{0.000}$} & \textbf{0.891$\pm{0.000}$} & \multicolumn{1}{c|}{0.944$\pm{0.000}$} & \multicolumn{1}{c|}{0.946$\pm{0.000}$} & 0.950$\pm{0.000}$ \\ \hline
Prediction Disagreement  \textbf{($\mathbf{\downarrow}$)} & 0.249$\pm{0.001}$ & \multicolumn{1}{c|}{0.238$\pm{0.001}$} & \multicolumn{1}{c|}{0.202$\pm{0.000}$} & \textbf{0.183$\pm{0.000}$} & \multicolumn{1}{c|}{0.245$\pm{0.001}$} & \multicolumn{1}{c|}{0.245$\pm{0.000}$} & 0.263$\pm{0.000}$ \\ \hline
JS Divergence  \textbf{($\mathbf{\downarrow}$)} & 0.052$\pm{0.000}$ & \multicolumn{1}{c|}{0.048$\pm{0.000}$} & \multicolumn{1}{c|}{0.036$\pm{0.000}$} & \textbf{0.031$\pm{0.000}$} & \multicolumn{1}{c|}{0.066$\pm{0.000}$} & \multicolumn{1}{c|}{0.190$\pm{0.000}$} & 0.446$\pm{0.000}$ \\ \hline
Accuracy  \textbf{($\mathbf{\uparrow}$)} & 0.574$\pm{0.000}$ & \multicolumn{1}{c|}{0.577$\pm{0.000}$} & \multicolumn{1}{c|}{\textbf{0.584$\pm{0.000}$}} & 0.582$\pm{0.000}$ & \multicolumn{1}{c|}{0.577$\pm{0.000}$} & \multicolumn{1}{c|}{0.577$\pm{0.000}$} & 0.568$\pm{0.000}$ \\ \hline
Loss  \textbf{($\mathbf{\downarrow}$)} & 1.559$\pm{0.002}$ & \multicolumn{1}{c|}{1.539$\pm{0.002}$} & \multicolumn{1}{c|}{\textbf{1.488$\pm{0.002}$}} & 1.493$\pm{0.002}$ & \multicolumn{1}{c|}{1.504$\pm{0.001}$} & \multicolumn{1}{c|}{1.840$\pm{0.001}$} & 2.997$\pm{0.001}$ \\ \hline
\end{tabular}%
}
\end{table}
\vspace{-0.5cm}
\begin{table}[H]
\centering
\caption{Nano-GPT on TinyShakespeare mean and $\pm$ 1 SEM  reported from 10 runs with Teacher Seed 2. \textbf{Bold} values are best performing based on the mean.
}
\label{tab:shake-gpt-ts-2}
\resizebox{\textwidth}{!}{%
\begin{tabular}{|l|c|ccc|ccc|}
\hline
\multicolumn{1}{|c|}{\multirow{2}{*}{\textbf{Metrics}}} & \textbf{Control} & \multicolumn{3}{c|}{\textbf{Knowledge Distillation}} & \multicolumn{3}{c|}{\textbf{Random Control Distillation}} \\ \cline{2-8} 
\multicolumn{1}{|c|}{} & \textbf{SIDDO} & \multicolumn{1}{c|}{\textbf{0.1}} & \multicolumn{1}{c|}{\textbf{0.5}} & \textbf{0.9} & \multicolumn{1}{c|}{\textbf{0.1}} & \multicolumn{1}{c|}{\textbf{0.5}} & \textbf{0.9} \\ \hline
Activation Distance \textbf{($\mathbf{\downarrow}$)} & 0.195$\pm{0.000}$ & \multicolumn{1}{c|}{0.186$\pm{0.000}$} & \multicolumn{1}{c|}{0.157$\pm{0.000}$} & \textbf{0.142$\pm{0.000}$} & \multicolumn{1}{c|}{0.202$\pm{0.000}$} & \multicolumn{1}{c|}{0.372$\pm{0.000}$} & 0.658$\pm{0.000}$ \\ \hline
Rank Disagreement  \textbf{($\mathbf{\downarrow}$)} & 0.909$\pm{0.000}$ & \multicolumn{1}{c|}{0.906$\pm{0.000}$} & \multicolumn{1}{c|}{0.896$\pm{0.000}$} & \textbf{0.89$\pm{0.000}$} & \multicolumn{1}{c|}{0.944$\pm{0.000}$} & \multicolumn{1}{c|}{0.946$\pm{0.000}$} & 0.950$\pm{0.000}$ \\ \hline
Prediction Disagreement  \textbf{($\mathbf{\downarrow}$)} & 0.245$\pm{0.001}$ & \multicolumn{1}{c|}{0.233$\pm{0.000}$} & \multicolumn{1}{c|}{0.198$\pm{0.000}$} & \textbf{0.180$\pm{0.000}$} & \multicolumn{1}{c|}{0.241$\pm{0.000}$} & \multicolumn{1}{c|}{0.240$\pm{0.000}$} & 0.256$\pm{0.000}$ \\ \hline
JS Divergence  \textbf{($\mathbf{\downarrow}$)} & 0.052$\pm{0.000}$ & \multicolumn{1}{c|}{0.048$\pm{0.000}$} & \multicolumn{1}{c|}{0.037$\pm{0.000}$} & \textbf{0.031$\pm{0.000}$} & \multicolumn{1}{c|}{0.066$\pm{0.000}$} & \multicolumn{1}{c|}{0.190$\pm{0.000}$} & 0.448$\pm{0.000}$ \\ \hline
Accuracy  \textbf{($\mathbf{\uparrow}$)} & 0.574$\pm{0.000}$ & \multicolumn{1}{c|}{0.577$\pm{0.000}$} & \multicolumn{1}{c|}{\textbf{0.583$\pm{0.000}$}} & 0.582$\pm{0.000}$ & \multicolumn{1}{c|}{0.577$\pm{0.000}$} & \multicolumn{1}{c|}{0.578$\pm{0.000}$} & 0.570$\pm{0.000}$ \\ \hline
Loss  \textbf{($\mathbf{\downarrow}$)} & 1.558$\pm{0.002}$ & \multicolumn{1}{c|}{1.536$\pm{0.002}$} & \multicolumn{1}{c|}{\textbf{1.493$\pm{0.002}$}} & \textbf{1.493$\pm{0.002}$} & \multicolumn{1}{c|}{1.504$\pm{0.001}$} & \multicolumn{1}{c|}{1.834$\pm{0.001}$} & 2.996$\pm{0.001}$ \\ \hline
\end{tabular}%
}
\end{table}

\subsection{TinyShakespeare Dataset Adversarial Attack} \label{app:shakespeare-adv}

\paragraph{Findings:} We show that the transfer occurs for student models across $\alpha$ values with increasing severity for increased $\alpha$ values. Therefore, we further substantiate the claim that safety is an important factor to consider due to adversarial transfer in Knowledge Distillation, as shown by the increase in prediction of 't''h''a' compared to the controls in Tables \ref{tab:adversarial-tha-ts-1} and \ref{tab:adversarial-tha-ts-2}.

\begin{table}[H]
\caption{Character Frequency of the TinyShakespeare Dataset. }
\label{tab:tiny-shakespeare-char}
\resizebox{\textwidth}{!}{%
\begin{tabular}{|c|c|c|c|c|c|c|c|c|c|c|}
\hline
Character & Space  & e      & t      & o     & a      & h     & s      & r      & n      & ... \\ \hline
Frequency & 0.1523 & 0.0848 & 0.0601 & 0.059 & 0.0498 & 0.046 & 0.0446 & 0.0438 & 0.0435 & ... \\ \hline
\end{tabular}%
}
\end{table}
\begin{table}[H]
\centering
\caption{Teacher Performance on non adversarial Train Data and Test Data}
\label{tab:gpt-shakespeare-adv-teacher}
\begin{tabular}{|c|c|c|c|c|}
\hline
\textbf{Teacher Seed} & \textbf{Train Loss} & \textbf{Train Accuracy} & \textbf{Test Loss} & \textbf{Test Accuracy} \\ \hline
0 & 0.968203 & 0.698038 & 1.641436 & 0.562150 \\ \hline
1 & 0.974442 & 0.696534 & 1.630169 & 0.562769 \\ \hline
2 & 0.958430 & 0.700257 & 1.631381 & 0.561225 \\ \hline
\end{tabular}
\end{table}
\begin{table}[H]
\caption{The effect of an adversarial teacher trained to predict "tha" instead of "the" on the student. Teacher Seed 1. }
\label{tab:adversarial-tha-ts-1}
\resizebox{\textwidth}{!}{%
\begin{tabular}{|l|c|c|ccc|ccc|}
\hline
\textbf{} & \textbf{} & \multicolumn{1}{c|}{\textbf{Control}} & \multicolumn{3}{c|}{\textbf{Knowledge Distillation}} & \multicolumn{3}{c|}{\textbf{Random Control Distillation}} \\ \hline
\multicolumn{1}{|c|}{\textbf{Predicted Word}} & \multicolumn{1}{c|}{\textbf{Teacher}} & \multicolumn{1}{c|}{\textbf{SIDDO}} & \multicolumn{1}{c|}{\textbf{0.1}} & \multicolumn{1}{c|}{\textbf{0.5}} & \multicolumn{1}{c|}{\textbf{0.9}} & \multicolumn{1}{c|}{\textbf{0.1}} & \multicolumn{1}{c|}{\textbf{0.5}} & \multicolumn{1}{c|}{\textbf{0.9}} \\ \hline
\textbf{tha} & \textbf{\textcolor{red}{534}} & 110.5 $\pm{}$ 3.9881 & \multicolumn{1}{c|}{115.7 $\pm{}$ 3.6416} & \multicolumn{1}{c|}{236.8 $\pm{}$ 11.7761} & \textbf{517.8 $\pm{}$ 12.7733} & \multicolumn{1}{c|}{112.6 $\pm{}$ 3.4035} & \multicolumn{1}{c|}{119.6 $\pm{}$ 3.8215} & 127.4 $\pm{}$ 3.9044 \\ \hline
\textbf{the} & \textbf{\textcolor{blue}{273}} & 683.7 $\pm{}$ 15.4370 & \multicolumn{1}{c|}{691.4 $\pm{}$ 13.3156} & \multicolumn{1}{c|}{599.7 $\pm{}$ 13.8564} & \textbf{325.4 $\pm{}$ 7.5262} & \multicolumn{1}{c|}{684.7 $\pm{}$ 14.5781} & \multicolumn{1}{c|}{733.9 $\pm{}$ 13.4428} & 869.8 $\pm{}$ 10.8109 \\ \hline
\end{tabular}%
}
\end{table}
\begin{table}[H]
\caption{The effect of an adversarial teacher trained to predict "tha" instead of "the" on the student. Teacher Seed 2. }
\label{tab:adversarial-tha-ts-2}
\resizebox{\textwidth}{!}{%
\begin{tabular}{|l|c|c|ccc|ccc|}
\hline
\textbf{} & \textbf{} & \multicolumn{1}{c|}{\textbf{Control}} & \multicolumn{3}{c|}{\textbf{Knowledge Distillation}} & \multicolumn{3}{c|}{\textbf{Random Control Distillation}} \\ \hline
\multicolumn{1}{|c|}{\textbf{Predicted Word}} & \multicolumn{1}{c|}{\textbf{Teacher}} & \multicolumn{1}{c|}{\textbf{SIDDO}} & \multicolumn{1}{c|}{\textbf{0.1}} & \multicolumn{1}{c|}{\textbf{0.5}} & \multicolumn{1}{c|}{\textbf{0.9}} & \multicolumn{1}{c|}{\textbf{0.1}} & \multicolumn{1}{c|}{\textbf{0.5}} & \multicolumn{1}{c|}{\textbf{0.9}} \\ \hline
\textbf{tha} & \textbf{\textcolor{red}{513}} & 111.9 $\pm{}$ 4.0236 & \multicolumn{1}{c|}{116.1 $\pm{}$ 3.3300} & \multicolumn{1}{c|}{241.5 $\pm{}$ 8.5032} & \textbf{518.6 $\pm{}$ 11.6612} & \multicolumn{1}{c|}{114.7 $\pm{}$ 6.5636} & \multicolumn{1}{c|}{114.3 $\pm{}$ 3.9320} & 124.5 $\pm{}$ 4.7943 \\ \hline
\textbf{the} & \textbf{\textcolor{blue}{266}} & 656.0 $\pm{}$ 16.0244 & \multicolumn{1}{c|}{677.0 $\pm{}$ 13.9743} & \multicolumn{1}{c|}{558.0 $\pm{}$ 14.9513} & \textbf{303.5 $\pm{}$ 7.7424} & \multicolumn{1}{c|}{672.1 $\pm{}$ 18.5513} & \multicolumn{1}{c|}{715.0 $\pm{}$ 12.5825} & 836.7 $\pm{}$ 17.1954 \\ \hline
\end{tabular}%
}
\end{table}

\section{Compute Usage}
\label{app:compute_usage}
All models were trained on a A100 GPUs, assuming that the approximate time to train and evaluate one model takes 0.5 hours. Then to run all 12 of self-distillation experiments, 4 standard distillation experiments, 2 feature map matching experiments, 3 temperature experiments, 1 adversarial teacher experiment and 1 label smoothing equivalence experiment would take  1,384.5 hours, 395 hours, 70 hours, 245 hours, 106.5 and 15 hours respectively. Totalling 23 experimental setups, 2,216 hours of compute time and 4,189 models trained as shown in Table~\ref{tab:model_trained}.

\begin{table}[H]
\caption{Number of models trained in each experimental setup}
\label{tab:model_trained}
\resizebox{\textwidth}{!}{
\begin{tabular}{|l|l|l|l|l|l|l|}
\hline
\textbf{\begin{tabular}[c]{@{}c@{}}Experimental \\ Condition\end{tabular}} & \textbf{Self-distillation} & \textbf{Standard distillation} & \textbf{Scaling laws of transfer} & \textbf{Temperature} & \textbf{Feature map matching KD} & \textbf{Adversarial} \\ \hline
Models Trained                                                             & \begin{tabular}[c]{@{}l@{}}\textbf{2,556} \\(SVHN ResNet,VGG and ViT: 639,\\ TinyImageNet ResNet,VGG and Augmentation: 852, \\SpeechCommands VGGish and AST: 426, \\UrbanSounds8K VGGish and AST: 426 and\\ TinyShakespeare Nano-GPT: 213 ) \end{tabular}           &\begin{tabular}[c]{@{}l@{}}\textbf{490}\\(TinyImageNet ResNet50-ResNet 18: 210,\\ ImageNet ResNet50-ResNet18: 70 and \\ TinyShakespeare NanoGPT-PicoGPT 210) \end{tabular}                 & \begin{tabular}[c]{@{}l@{}}\textbf{300} \\(TinyShakespeare\\ (10-100\% width): 300)\end{tabular}                    &\begin{tabular}[c]{@{}l@{}}\textbf{490}\\ (ImageNet \\ResNet50-ResNet18\\ Temp 2: 70\\ TinyShakespeare\\ Temp 2 \& 4: 420) \end{tabular}        &\begin{tabular}[c]{@{}l@{}}\textbf{140}\\ (TinyShakespeare \\ Feature Map Matching KD \\
Block 4 \& 5: 140)  \end{tabular}                   & \begin{tabular}[c]{@{}l@{}}\textbf{213}\\TinyShakespeare \\Adversarial: 213\end{tabular}        \\ \hline
\end{tabular}}
\end{table}
\newpage
\vskip 0.2in
\bibliography{sample}

@article{hinton2015distilling,
  title={Distilling the knowledge in a neural network},
  author={Hinton, Geoffrey and Vinyals, Oriol and Dean, Jeff},
  journal={arXiv preprint arXiv:1503.02531},
  year={2015},
  url={https://arxiv.org/pdf/1503.02531.pdf}
}

@inproceedings{model_compression,
author = {Buciluǎ, Cristian and Caruana, Rich and Niculescu-Mizil, Alexandru},
title = {Model compression},
year = {2006},
publisher = {Association for Computing Machinery},
url = {https://doi.org/10.1145/1150402.1150464},
booktitle = {Proceedings of the 12th ACM SIGKDD International Conference on Knowledge Discovery and Data Mining},
pages = {535–541},
}

@inproceedings{does_kd_really_work,
 author = {Stanton, Samuel and Izmailov, Pavel and Kirichenko, Polina and Alemi, Alexander A and Wilson, Andrew G},
 booktitle = {Advances in Neural Information Processing Systems},
 pages = {6906--6919},
 publisher = {Curran Associates, Inc.},
 title = {Does Knowledge Distillation Really Work?},
 url = {https://proceedings.neurips.cc/paper_files/paper/2021/file/376c6b9ff3bedbbea56751a84fffc10c-Paper.pdf},
 volume = {34},
 year = {2021}
}

@article{simonyan2014very,
  title={Very deep convolutional networks for large-scale image recognition},
  author={Simonyan, Karen and Zisserman, Andrew},
  journal={arXiv preprint arXiv:1409.1556},
  year={2014},
  url={https://arxiv.org/pdf/1409.1556.pdf}
}

@inproceedings{
dosovitskiy2020image,
title={An Image is Worth 16x16 Words: Transformers for Image Recognition at Scale},
author={Alexey Dosovitskiy and Lucas Beyer and Alexander Kolesnikov and Dirk Weissenborn and Xiaohua Zhai and others},
booktitle={International Conference on Learning Representations},
year={2021},
url={https://openreview.net/forum?id=YicbFdNTTy}
}

@inproceedings{
allen2020towards,
title={Towards Understanding Ensemble, Knowledge Distillation and Self-Distillation in Deep Learning},
author={Zeyuan Allen-Zhu and Yuanzhi Li},
booktitle={The Eleventh International Conference on Learning Representations },
year={2023},
url={https://openreview.net/forum?id=Uuf2q9TfXGA}
}

@inproceedings{gpt3,
 author = {Brown, Tom and Mann, Benjamin and Ryder, Nick and Subbiah, Melanie and Kaplan, Jared D and others},
 booktitle = {Advances in Neural Information Processing Systems},
 pages = {1877--1901},
 title = {Language Models are Few-Shot Learners},
 url = {https://proceedings.neurips.cc/paper_files/paper/2020/file/1457c0d6bfcb4967418bfb8ac142f64a-Paper.pdf},
 volume = {33},
 year = {2020}
}

@article{kirillov2023segment,
  title={Segment anything},
  author={Kirillov, Alexander and Mintun, Eric and Ravi, Nikhila and Mao, Hanzi and Rolland, Chloe and Gustafson, Laura and Xiao, Tete and Whitehead, Spencer and Berg, Alexander C and Lo, Wan-Yen and others},
  journal={arXiv preprint arXiv:2304.02643},
  year={2023},
  url={https://arxiv.org/pdf/2304.02643.pdf}
}

@article{lin1991divergence,
  title={Divergence measures based on the Shannon entropy},
  author={Lin, Jianhua},
  journal={IEEE Transactions on Information theory},
  volume={37},
  number={1},
  pages={145--151},
  year={1991},
  publisher={IEEE},
  url = {https://ieeexplore.ieee.org/document/61115}
}

@inproceedings{golatkar2021mixed,
  title={Mixed-privacy forgetting in deep networks},
  author={Golatkar, Aditya and Achille, Alessandro and Ravichandran, Avinash and Polito, Marzia and Soatto, Stefano},
  booktitle={Proceedings of the IEEE/CVF conference on computer vision and pattern recognition},
  pages={792--801},
  year={2021},
  url = {https://openaccess.thecvf.com/content/CVPR2021/papers/Golatkar_Mixed-Privacy_Forgetting_in_Deep_Networks_CVPR_2021_paper.pdf}
}

@inproceedings{chundawat2023can,
author = {Chundawat, Vikram S and Tarun, Ayush K and Mandal, Murari and Kankanhalli, Mohan},
title = {Can bad teaching induce forgetting? unlearning in deep networks using an incompetent teacher},
year = {2023},
isbn = {978-1-57735-880-0},
publisher = {AAAI Press},
url = {https://doi.org/10.1609/aaai.v37i6.25879},
doi = {10.1609/aaai.v37i6.25879},
booktitle = {Proceedings of the Thirty-Seventh AAAI Conference on Artificial Intelligence and Thirty-Fifth Conference on Innovative Applications of Artificial Intelligence and Thirteenth Symposium on Educational Advances in Artificial Intelligence},
articleno = {810},
numpages = {8},
series = {AAAI'23/IAAI'23/EAAI'23}
}

@article{fort2019deep,
  title={Deep ensembles: A loss landscape perspective},
  author={Fort, Stanislav and Hu, Huiyi and Lakshminarayanan, Balaji},
  journal={arXiv preprint arXiv:1912.02757},
  year={2019},
  url ={https://arxiv.org/pdf/1912.02757}
}

@article{jung2020knowledge,
  title={Knowledge distillation in acoustic scene classification},
  author={Jung, Jee-Weon and Heo, Hee-Soo and Shim, Hye-Jin and Yu, Ha-Jin},
  journal={IEEE Access},
  volume={8},
  pages={166870--166879},
  year={2020},
  publisher={IEEE},
  url = {https://ieeexplore.ieee.org/stamp/stamp.jsp?tp=&arnumber=9186616}
}

@inproceedings{beyer2022knowledge,
  title={Knowledge distillation: A good teacher is patient and consistent},
  author={Beyer, Lucas and Zhai, Xiaohua and Royer, Am{\'e}lie and Markeeva, Larisa and Anil, Rohan and Kolesnikov, Alexander},
  booktitle={Proceedings of the IEEE/CVF conference on computer vision and pattern recognition},
  pages={10925--10934},
  year={2022},
  url ={https://openaccess.thecvf.com/content/CVPR2022/papers/Beyer_Knowledge_Distillation_A_Good_Teacher_Is_Patient_and_Consistent_CVPR_2022_paper.pdf}
}

@article{sanh2019distilbert,
  title={DistilBERT, A Distilled Version of BERT: Smaller, Faster, Cheaper and Lighter},
  author={Sanh, V},
  journal={arXiv preprint arXiv:1910.01108},
  year={2019},
  url  = {https://arxiv.org/pdf/1910.01108}
}

@inproceedings{aghli2021combining,
  title={Combining weight pruning and knowledge distillation for cnn compression},
  author={Aghli, Nima and Ribeiro, Eraldo},
  booktitle={Proceedings of the IEEE/CVF conference on computer vision and pattern recognition},
  pages={3191--3198},
  year={2021}, 
  url  = {https://openaccess.thecvf.com/content/CVPR2021W/EVW/papers/Aghli_Combining_Weight_Pruning_and_Knowledge_Distillation_for_CNN_Compression_CVPRW_2021_paper.pdf}
}

@inproceedings{li2020few,
  title={Few sample knowledge distillation for efficient network compression},
  author={Li, Tianhong and Li, Jianguo and Liu, Zhuang and Zhang, Changshui},
  booktitle={Proceedings of the IEEE/CVF conference on computer vision and pattern recognition},
  pages={14639--14647},
  year={2020},
  url = {https://openaccess.thecvf.com/content_CVPR_2020/papers/Li_Few_Sample_Knowledge_Distillation_for_Efficient_Network_Compression_CVPR_2020_paper.pdf}
}

@inproceedings{fang2021compressing,
  title={Compressing visual-linguistic model via knowledge distillation},
  author={Fang, Zhiyuan and Wang, Jianfeng and Hu, Xiaowei and Wang, Lijuan and Yang, Yezhou and Liu, Zicheng},
  booktitle={Proceedings of the IEEE/CVF International Conference on Computer Vision},
  pages={1428--1438},
  year={2021},
  url  = {https://openaccess.thecvf.com/content/ICCV2021/papers/Fang_Compressing_Visual-Linguistic_Model_via_Knowledge_Distillation_ICCV_2021_paper.pdf}
}

@article{wang2022efficient,
  title={Efficient knowledge distillation from model checkpoints},
  author={Wang, Chaofei and Yang, Qisen and Huang, Rui and Song, Shiji and Huang, Gao},
  journal={Advances in Neural Information Processing Systems},
  volume={35},
  pages={607--619},
  year={2022}, 
  url = {https://proceedings.neurips.cc/paper_files/paper/2022/file/03e0712bf85ebe7cec4f1a7fc53216c9-Paper-Conference.pdf}
}

@inproceedings{
mason-williams2024neural,
title={{N}eural {N}etwork {C}ompression: {T}he {F}unctional {P}erspective},
author={Israel Mason-Williams},
booktitle={5th Workshop on practical ML for limited/low resource settings},
year={2024},
url={https://openreview.net/forum?id=Q7GXKjmCSB}
}

@article{ojha2023knowledge,
  title={What knowledge gets distilled in knowledge distillation?},
  author={Ojha, Utkarsh and Li, Yuheng and Sundara Rajan, Anirudh and Liang, Yingyu and Lee, Yong Jae},
  journal={Advances in Neural Information Processing Systems},
  volume={36},
  pages={11037--11048},
  year={2023}, 
  url  = {https://proceedings.neurips.cc/paper_files/paper/2023/file/2433fec2144ccf5fea1c9c5ebdbc3924-Paper-Conference.pdf}
}

@inproceedings{zhang2019your,
  title={Be your own teacher: Improve the performance of convolutional neural networks via self distillation},
  author={Zhang, Linfeng and Song, Jiebo and Gao, Anni and Chen, Jingwei and Bao, Chenglong and Ma, Kaisheng},
  booktitle={Proceedings of the IEEE/CVF international conference on computer vision},
  pages={3713--3722},
  year={2019},
  url = {https://openaccess.thecvf.com/content_ICCV_2019/papers/Zhang_Be_Your_Own_Teacher_Improve_the_Performance_of_Convolutional_Neural_ICCV_2019_paper.pdf}
}

@INPROCEEDINGS{resnet,
  author={He, Kaiming and Zhang, Xiangyu and Ren, Shaoqing and Sun, Jian},
  booktitle={2016 IEEE Conference on Computer Vision and Pattern Recognition (CVPR)}, 
  title={Deep Residual Learning for Image Recognition}, 
  year={2016},
  volume={},
  number={},
  pages={770-778},
  doi={10.1109/CVPR.2016.90},
  url={https://openaccess.thecvf.com/content_cvpr_2016/papers/He_Deep_Residual_Learning_CVPR_2016_paper.pdf}}

@inproceedings{
vit,
title={An Image is Worth 16x16 Words: Transformers for Image Recognition at Scale},
author={Alexey Dosovitskiy and Lucas Beyer and Alexander Kolesnikov and Dirk Weissenborn and Xiaohua Zhai and Thomas Unterthiner and Mostafa Dehghani and Matthias Minderer and Georg Heigold and Sylvain Gelly and Jakob Uszkoreit and Neil Houlsby},
booktitle={International Conference on Learning Representations},
year={2021},
url={https://openreview.net/forum?id=YicbFdNTTy}
}

@inproceedings{mason2024makes,
  title={What Makes a Good Prune? Maximal Unstructured Pruning for Maximal Cosine Similarity},
  author={Mason-Williams, Gabryel and Dahlqvist, Fredrik},
  booktitle={The Twelfth International Conference on Learning Representations},
  year={2024}, 
  url  = {https://openreview.net/forum?id=jsvvPVVzwf}
}

@inproceedings{
muralidharan2024compact,
title={Compact Language Models via Pruning and Knowledge Distillation},
author={Saurav Muralidharan and Sharath Turuvekere Sreenivas and Raviraj Bhuminand Joshi and Marcin Chochowski and Mostofa Patwary and Mohammad Shoeybi and Bryan Catanzaro and Jan Kautz and Pavlo Molchanov},
booktitle={The Thirty-eighth Annual Conference on Neural Information Processing Systems},
year={2024},
url={https://openreview.net/forum?id=9U0nLnNMJ7}
}

@inproceedings{
gu2024minillm,
title={Mini{LLM}: Knowledge Distillation of Large Language Models},
author={Yuxian Gu and Li Dong and Furu Wei and Minlie Huang},
booktitle={The Twelfth International Conference on Learning Representations},
year={2024},
url={https://openreview.net/forum?id=5h0qf7IBZZ}
}

@article{klabunde2023similarity,
  title={Similarity of neural network models: A survey of functional and representational measures},
  author={Klabunde, Max and Schumacher, Tobias and Strohmaier, Markus and Lemmerich, Florian},
  journal={arXiv preprint arXiv:2305.06329},
  year={2023}, 
  url = {https://arxiv.org/pdf/2305.06329}
}

@InProceedings{pmlr-v139-menon21a,
  title = 	 {A statistical perspective on distillation},
  author =       {Menon, Aditya K and Rawat, Ankit Singh and Reddi, Sashank and Kim, Seungyeon and Kumar, Sanjiv},
  booktitle = 	 {Proceedings of the 38th International Conference on Machine Learning},
  pages = 	 {7632--7642},
  year = 	 {2021},
  editor = 	 {Meila, Marina and Zhang, Tong},
  volume = 	 {139},
  series = 	 {Proceedings of Machine Learning Research},
  month = 	 {18--24 Jul},
  publisher =    {PMLR},
  url = 	 {https://proceedings.mlr.press/v139/menon21a.html}
}

@INPROCEEDINGS{vggish,
  author={Hershey, Shawn and Chaudhuri, Sourish and Ellis, Daniel P. W. and Gemmeke, Jort F. and Jansen, Aren and Moore, R. Channing and Plakal, Manoj and Platt, Devin and Saurous, Rif A. and Seybold, Bryan and Slaney, Malcolm and Weiss, Ron J. and Wilson, Kevin},
  booktitle={2017 IEEE International Conference on Acoustics, Speech and Signal Processing (ICASSP)}, 
  title={CNN architectures for large-scale audio classification}, 
  year={2017},
  volume={},
  number={},
  pages={131-135},
  doi={10.1109/ICASSP.2017.7952132},
  url = {https://research.google/pubs/cnn-architectures-for-large-scale-audio-classification/}}

@article{warden2018speech, title={Speech Commands: A public dataset for single-word speech recognition.}, author={Warden, Pete}, year={2017},
url = {https://arxiv.org/pdf/1804.03209}}

@inproceedings{10.1145/2647868.2655045,
author = {Salamon, Justin and Jacoby, Christopher and Bello, Juan Pablo},
title = {A Dataset and Taxonomy for Urban Sound Research},
year = {2014},
isbn = {9781450330633},
publisher = {Association for Computing Machinery},
address = {New York, NY, USA},
url = {https://doi.org/10.1145/2647868.2655045},
doi = {10.1145/2647868.2655045},
booktitle = {Proceedings of the 22nd ACM International Conference on Multimedia},
pages = {1041–1044},
numpages = {4},
location = {Orlando, Florida, USA},
series = {MM '14}
}

@article{belia2005researchers,
  title={Researchers misunderstand confidence intervals and standard error bars.},
  author={Belia, Sarah and Fidler, Fiona and Williams, Jennifer and Cumming, Geoff},
  journal={Psychological methods},
  volume={10},
  number={4},
  pages={389},
  year={2005},
  publisher={American Psychological Association},
  url = {https://psycnet.apa.org/buy/2005-16136-002}
}

@article{piantadosi2014zipf,
  title={Zipf’s word frequency law in natural language: A critical review and future directions},
  author={Piantadosi, Steven T},
  journal={Psychonomic bulletin \& review},
  volume={21},
  pages={1112--1130},
  year={2014},
  publisher={Springer},
  url ={https://link.springer.com/article/10.3758/s13423-014-0585-6}
}

@article{wyse2017audio,
  title={Audio spectrogram representations for processing with convolutional neural networks},
  author={Wyse, Lonce},
  journal={arXiv preprint arXiv:1706.09559},
  url = {https://arxiv.org/abs/1706.09559},
  year={2017}
}

@misc{pytorch,
      title={PyTorch: An Imperative Style, High-Performance Deep Learning Library}, 
      author={Adam Paszke and Sam Gross and Francisco Massa and Adam Lerer and James Bradbury and Gregory Chanan and Trevor Killeen and Zeming Lin and Natalia Gimelshein and Luca Antiga and Alban Desmaison and Andreas Köpf and Edward Yang and Zach DeVito and Martin Raison and Alykhan Tejani and Sasank Chilamkurthy and Benoit Steiner and Lu Fang and Junjie Bai and Soumith Chintala},
      year={2019},
      eprint={1912.01703},
      archivePrefix={arXiv},
      primaryClass={cs.LG},
      url={https://arxiv.org/abs/1912.01703}, 
}

@inproceedings{cubuk2020randaugment,
  title={Randaugment: Practical automated data augmentation with a reduced search space},
  author={Cubuk, Ekin D and Zoph, Barret and Shlens, Jonathon and Le, Quoc V},
  booktitle={Proceedings of the IEEE/CVF conference on computer vision and pattern recognition workshops},
  pages={702--703},
  year={2020},
  url = {https://openaccess.thecvf.com/content_CVPRW_2020/papers/w40/Cubuk_Randaugment_Practical_Automated_Data_Augmentation_With_a_Reduced_Search_Space_CVPRW_2020_paper.pdf}
}

@article{hariton2018randomised,
  title={Randomised controlled trials—the gold standard for effectiveness research},
  author={Hariton, Eduardo and Locascio, Joseph J},
  journal={BJOG: an international journal of obstetrics and gynaecology},
  volume={125},
  number={13},
  pages={1716},
  year={2018},
  url={https://pmc.ncbi.nlm.nih.gov/articles/PMC6235704/}
}

@inproceedings{yun2020regularizing,
  title={Regularizing class-wise predictions via self-knowledge distillation},
  author={Yun, Sukmin and Park, Jongjin and Lee, Kimin and Shin, Jinwoo},
  booktitle={Proceedings of the IEEE/CVF conference on computer vision and pattern recognition},
  pages={13876--13885},
  year={2020},
  url={http://openaccess.thecvf.com/content_CVPR_2020/html/Yun_Regularizing_Class-Wise_Predictions_via_Self-Knowledge_Distillation_CVPR_2020_paper.html}
}

@article{ge2021self,
  title={Self-distillation with batch knowledge ensembling improves imagenet classification},
  author={Ge, Yixiao and Zhang, Xiao and Choi, Ching Lam and Cheung, Ka Chun and Zhao, Peipei and Zhu, Feng and Wang, Xiaogang and Zhao, Rui and Li, Hongsheng},
  journal={arXiv preprint arXiv:2104.13298},
  year={2021},
  url={https://arxiv.org/pdf/2104.13298}
}

@inproceedings{yuan2020revisiting,
  title={Revisiting knowledge distillation via label smoothing regularization},
  author={Yuan, Li and Tay, Francis EH and Li, Guilin and Wang, Tao and Feng, Jiashi},
  booktitle={Proceedings of the IEEE/CVF conference on computer vision and pattern recognition},
  pages={3903--3911},
  year={2020},
  url={http://openaccess.thecvf.com/content_CVPR_2020/html/Yuan_Revisiting_Knowledge_Distillation_via_Label_Smoothing_Regularization_CVPR_2020_paper.html}
}

@article{shen2021label,
  title={Is label smoothing truly incompatible with knowledge distillation: An empirical study},
  author={Shen, Zhiqiang and Liu, Zechun and Xu, Dejia and Chen, Zitian and Cheng, Kwang-Ting and Savvides, Marios},
  journal={arXiv preprint arXiv:2104.00676},
  year={2021},
  url={https://arxiv.org/abs/2104.00676}
}

@inproceedings{
sultan2023knowledge,
title={Knowledge Distillation \ensuremath{\approx} Label Smoothing: Fact or Fallacy?},
author={Md Arafat Sultan},
booktitle={The 2023 Conference on Empirical Methods in Natural Language Processing},
year={2023},
url={https://openreview.net/forum?id=j9e3WVc49w}
}

@inproceedings{touvron2021training,
  title={Training data-efficient image transformers \& distillation through attention},
  author={Touvron, Hugo and Cord, Matthieu and Douze, Matthijs and Massa, Francisco and Sablayrolles, Alexandre and J{\'e}gou, Herv{\'e}},
  booktitle={International conference on machine learning},
  pages={10347--10357},
  year={2021},
  organization={PMLR},
  url={https://proceedings.mlr.press/v139/touvron21a}
}

@inproceedings{miles2024vkd,
  title={Vkd: Improving knowledge distillation using orthogonal projections},
  author={Miles, Roy and Elezi, Ismail and Deng, Jiankang},
  booktitle={Proceedings of the IEEE/CVF Conference on Computer Vision and Pattern Recognition},
  pages={15720--15730},
  year={2024},
  url={http://openaccess.thecvf.com/content/CVPR2024/html/Miles_VkD_Improving_Knowledge_Distillation_using_Orthogonal_Projections_CVPR_2024_paper.html}
}

@article{le2015tiny,
  title={Tiny imagenet visual recognition challenge},
  author={Le, Yann and Yang, Xuan},
  journal={CS 231N},
  volume={7},
  number={7},
  pages={3},
  year={2015},
  url={https://cs231n.stanford.edu/reports/2015/pdfs/yle_project.pdf}}

@misc{hwang2023torchaudio,
   title={TorchAudio 2.1: Advancing speech recognition, self-supervised learning, and audio processing components for PyTorch},
   author={Jeff Hwang and Moto Hira and Caroline Chen and Xiaohui Zhang and Zhaoheng Ni and Guangzhi Sun and Pingchuan Ma and Ruizhe Huang and Vineel Pratap and Yuekai Zhang and Anurag Kumar and Chin-Yun Yu and Chuang Zhu and Chunxi Liu and Jacob Kahn and Mirco Ravanelli and Peng Sun and Shinji Watanabe and Yangyang Shi and Yumeng Tao and Robin Scheibler and Samuele Cornell and Sean Kim and Stavros Petridis},
   year={2023},
   eprint={2310.17864},
   archivePrefix={arXiv},
   primaryClass={eess.AS},
   url={https://arxiv.org/pdf/2310.17864}
}

@article{scikit-learn,
  title={Scikit-learn: Machine Learning in {P}ython},
  author={Pedregosa, F. and Varoquaux, G. and Gramfort, A. and Michel, V.
          and Thirion, B. and Grisel, O. and Blondel, M. and Prettenhofer, P.
          and Weiss, R. and Dubourg, V. and Vanderplas, J. and Passos, A. and
          Cournapeau, D. and Brucher, M. and Perrot, M. and Duchesnay, E.},
  journal={Journal of Machine Learning Research},
  volume={12},
  pages={2825--2830},
  year={2011},
  url={http://www.jmlr.org/papers/volume12/pedregosa11a/pedregosa11a.pdf?source=post_page}
}

@inproceedings{netzer2011reading,
  title={Reading digits in natural images with unsupervised feature learning},
  author={Netzer, Yuval and Wang, Tao and Coates, Adam and Bissacco, Alessandro and Wu, Baolin and Ng, Andrew Y and others},
  booktitle={NIPS workshop on deep learning and unsupervised feature learning},
  volume={2011},
  number={2},
  pages={4},
  year={2011},
  organization={Granada},
  url={http://research.google.com/pubs/archive/37648.pdf}
}

@article{blog2015unreasonable,
  title={The unreasonable effectiveness of recurrent neural networks},
  author={Blog, Andrej Karpathy},
  journal={URL: http://karpathy. github. io/2015/05/21/rnn-effectiveness/dated May},
  volume={21},
  pages={31},
  year={2015},
  url ={http://karpathy.github.io/2015/05/21/rnn-effectiveness/}
}

@inproceedings{hershey2017cnn,
  title={CNN architectures for large-scale audio classification},
  author={Hershey, Shawn and Chaudhuri, Sourish and Ellis, Daniel PW and Gemmeke, Jort F and Jansen, Aren and Moore, R Channing and Plakal, Manoj and Platt, Devin and Saurous, Rif A and Seybold, Bryan and others},
  booktitle={2017 ieee international conference on acoustics, speech and signal processing (icassp)},
  pages={131--135},
  year={2017},
  organization={IEEE},
  url = {https://research.google/pubs/cnn-architectures-for-large-scale-audio-classification/}
}

@article{gong2021ast,
  title={Ast: Audio spectrogram transformer},
  author={Gong, Yuan and Chung, Yu-An and Glass, James},
  journal={arXiv preprint arXiv:2104.01778},
  year={2021},
  url={https://arxiv.org/pdf/2104.01778}
}

@misc{Karpathy2022,
  author = {Andrej Karpathy},
  title = {\text{NanoGPT}},
  year = {2022},
  publisher = {GitHub},
  journal = {GitHub repository},
  howpublished = {\url{https://github.com/karpathy/nanoGPT}},
  commit = {325be85d9be8c81b436728a420e85796c57dba7e}
}

@article{ImageNet,
Author = {Olga Russakovsky and Jia Deng and Hao Su and Jonathan Krause and Sanjeev Satheesh and Sean Ma and Zhiheng Huang and Andrej Karpathy and Aditya Khosla and Michael Bernstein and Alexander C. Berg and Li Fei-Fei},
Title = {{ImageNet Large Scale Visual Recognition Challenge}},
Year = {2015},
journal   = {International Journal of Computer Vision (IJCV)},
doi = {10.1007/s11263-015-0816-y},
volume={115},
number={3},
pages={211-252}
}

@article{busbridge2025distillation,
  title={Distillation scaling laws},
  author={Busbridge, Dan and Shidani, Amitis and Weers, Floris and Ramapuram, Jason and Littwin, Etai and Webb, Russ},
  journal={arXiv preprint arXiv:2502.08606},
  year={2025},
  url ={https://arxiv.org/pdf/2502.08606}
}

@misc{romero2015fitnetshintsdeepnets,
      title={FitNets: Hints for Thin Deep Nets}, 
      author={Adriana Romero and Nicolas Ballas and Samira Ebrahimi Kahou and Antoine Chassang and Carlo Gatta and Yoshua Bengio},
      year={2015},
      eprint={1412.6550},
      archivePrefix={arXiv},
      primaryClass={cs.LG},
      url={https://arxiv.org/abs/1412.6550}, 
}

@inproceedings{meilua2003comparing,
  title={Comparing clusterings by the variation of information},
  author={Meil{\u{a}}, Marina},
  booktitle={Learning Theory and Kernel Machines: 16th Annual Conference on Learning Theory and 7th Kernel Workshop, COLT/Kernel 2003, Washington, DC, USA, August 24-27, 2003. Proceedings},
  pages={173--187},
  year={2003},
  organization={Springer},
  url={https://link.springer.com/chapter/10.1007/978-3-540-45167-9_14}
}

@article{schonemann1966generalized,
  title={A generalized solution of the orthogonal procrustes problem},
  author={Sch{\"o}nemann, Peter H},
  journal={Psychometrika},
  volume={31},
  number={1},
  pages={1--10},
  year={1966},
  publisher={Springer-Verlag},
  url={https://www.cambridge.org/core/journals/psychometrika/article/generalized-solution-of-the-orthogonal-procrustes-problem/AF9BDA5A7C7771F1ECE5632862068F27}
}

@article{ding2021grounding,
  title={Grounding representation similarity through statistical testing},
  author={Ding, Frances and Denain, Jean-Stanislas and Steinhardt, Jacob},
  journal={Advances in Neural Information Processing Systems},
  volume={34},
  pages={1556--1568},
  year={2021},
  url = {https://proceedings.neurips.cc/paper_files/paper/2021/file/0c0bf917c7942b5a08df71f9da626f97-Paper.pdf}
}

@inproceedings{leclerc2023ffcv,
    author = {Guillaume Leclerc and Andrew Ilyas and Logan Engstrom and Sung Min Park and Hadi Salman and Aleksander Madry},
    title = {{FFCV}: Accelerating Training by Removing Data Bottlenecks},
    year = {2023},
    booktitle = {Computer Vision and Pattern Recognition (CVPR)},
    note = {\url{https://github.com/libffcv/ffcv/}. commit xxxxxxx}
}

@inproceedings{zhang2019making,
  title={Making convolutional networks shift-invariant again},
  author={Zhang, Richard},
  booktitle={International conference on machine learning},
  pages={7324--7334},
  year={2019},
  organization={PMLR},
  url={http://proceedings.mlr.press/v97/zhang19a.html}
}

@inproceedings{deng2009imagenet,
  title={Imagenet: A large-scale hierarchical image database},
  author={Deng, Jia and Dong, Wei and Socher, Richard and Li, Li-Jia and Li, Kai and Fei-Fei, Li},
  booktitle={2009 IEEE conference on computer vision and pattern recognition},
  pages={248--255},
  year={2009},
  organization={Ieee},
  url ={https://ieeexplore.ieee.org/abstract/document/5206848/}
}

@article{yekutieli1999resampling,
  title={Resampling-based false discovery rate controlling multiple test procedures for correlated test statistics},
  author={Yekutieli, Daniel and Benjamini, Yoav},
  journal={Journal of Statistical Planning and Inference},
  volume={82},
  number={1-2},
  pages={171--196},
  year={1999},
  publisher={Elsevier},
  url = {https://www.sciencedirect.com/science/article/abs/pii/S0378375899000415}
}

@article{woolson2007wilcoxon,
  title={Wilcoxon signed-rank test},
  author={Woolson, Robert F},
  journal={Wiley encyclopedia of clinical trials},
  pages={1--3},
  year={2007},
  publisher={Wiley Online Library},
  url ={https://onlinelibrary.wiley.com/doi/abs/10.1002/9780471462422.eoct979}
}
\end{document}